\documentclass[runningheads]{llncs}
\usepackage{graphicx}
\usepackage{siunitx}
\usepackage{color} 
\usepackage{amssymb}
\usepackage{makecell}
\usepackage{multirow}
\usepackage{lipsum}
\usepackage{caption}
\usepackage{subcaption}
\usepackage{adjustbox}
\usepackage{comment}
\usepackage{float}
\usepackage{booktabs}
\usepackage{rotating}
\usepackage{setspace}
\usepackage{starfont}
\usepackage{svg}

\usepackage{pifont}
\definecolor{MyBlue}{rgb}{0,0.08,0.5}
\definecolor{MyRed}{rgb}{0.7,0.02,0.02}
\definecolor{MyOrange}{rgb}{1,0.5,0}
\definecolor{MyPurple}{rgb}{0.6,0.25,0.8}
\definecolor{MyGreen}{rgb}{0.1,0.8,0.1}

\usepackage[pagebackref=true,breaklinks=true,letterpaper=true,colorlinks=false,bookmarks=false]{hyperref}
\hypersetup{
     colorlinks = true,
     linkcolor = red,
     anchorcolor = black,
     citecolor = black,
     filecolor = black,
     urlcolor = blue
     }

\begin{document}
\title{Deep learning based domain adaptation for mitochondria segmentation on EM volumes}
\titlerunning{Deep learning based domain adaptation on EM volumes}

\newcommand*{\affaddr}[1]{#1} 
\newcommand*{\affmark}[1][*]{\textsuperscript{#1}}

\author{Daniel Franco-Barranco\affmark[1,2] \and Julio Pastor-Tronch\affmark[1] \and Aitor Gonzalez-Marfil\affmark[1] \and Arrate Muñoz-Barrutia\affmark[3,4] \and Ignacio Arganda-Carreras\affmark[1,2,5]  }

\authorrunning{Franco-Barranco {\em et al.}}
\institute{
\affaddr{\affmark[1] Dept. of Computer Science and Artificial Intelligence, University of the Basque Country (UPV/EHU)} \\
\affaddr{\affmark[2] Donostia International Physics Center (DIPC)} \\
\affaddr{\affmark[3] Universidad Carlos III de Madrid} \\
\affaddr{\affmark[4] Instituto de Investigación Sanitaria Gregorio Marañón} \\
\affaddr{\affmark[5] Ikerbasque, Basque Foundation for Science} \\
\email{daniel\_franco001@ehu.eus}
}

\maketitle   

\section*{Abstract}
\textbf{Background and Objective}: Accurate segmentation of electron microscopy (EM) volumes of the brain is essential to characterize neuronal structures at a cell or organelle level. While supervised deep learning methods have led to major breakthroughs in that direction during the past years, they usually require large amounts of annotated data to be trained, and perform poorly on other data acquired under similar experimental and imaging conditions. This is a problem known as domain adaptation, since models that learned from a sample distribution (or source domain) struggle to maintain their performance on samples extracted from a different distribution or target domain. In this work, we address the complex case of deep learning based domain adaptation for mitochondria segmentation across EM datasets from different tissues and species. \\
\textbf{Methods}: We present three unsupervised domain adaptation strategies to improve mitochondria segmentation in the target domain based on (1) state-of-the-art style transfer between images of both domains; (2) self-supervised learning to pre-train a model using unlabeled source and target images, and then fine-tune it only with the source labels; and (3) multi-task neural network architectures trained end-to-end with both labeled and unlabeled images. Additionally, to ensure good generalization in our models, we propose a new training stopping criterion based on morphological priors obtained exclusively in the source domain. The code
and its documentation are publicly available at \url{https://github.com/danifranco/EM_domain_adaptation}\\
\textbf{Results}: We carried out all possible cross-dataset experiments using three publicly available EM datasets. We evaluated our proposed strategies and those of others based on the mitochondria semantic labels predicted on the target datasets.\\
\textbf{Conclusions}: The methods introduced here outperform the baseline methods and compare favorably to the state of the art. In the absence of validation labels, monitoring our proposed morphology-based metric is an intuitive and effective way to stop the training process and select in average optimal models.

\section{Introduction}
\label{introduction}

Supervised learning has achieved great success in computer vision leading to the development of robust algorithms that have been successfully applied in diverse research areas. The generalization capability and reliability of these algorithms are based on the assumption that the data used to train them and the data used to test them are drawn from the same distribution or \textit{domain}. Thus, when the training data is not representative enough of the target population, there is a drop in the algorithm's performance~\cite{patel2015visual}. This performance gap is highly significant when the data acquisition changes (i.e., protocol, instrument) even for a similar target domain. In the particular case of biomedical imaging, data distributions are highly biased due to the variety of acquisition techniques and protocols. Therefore, a significant number of annotations is usually required to ensure a good representation of the population. 

Nevertheless, collecting and annotating these datasets is extremely expensive in both time and human resources~\cite{wang2018deep}. For that reason, the field of domain adaptation has emerged to tackle both issues: the reduction of the domain gap difference and the generation of annotated data. The purpose of domain adaptation is to learn from labeled data in a source domain to perform well on a different, but related target domain without any annotation~\cite{wilson2020survey}.  

Aiming to reduce source and target domain dissimilarity, many methods have been proposed to create synthetic source images, and therefore, increase the heterogeneity of the data~\cite{yi2019generative}. Some of these approaches generate new images from random noise without any other conditional information for Computed Tomography (CT) data~\cite{frid2018gan,bowles2018gan}, Magnetic Resonance (MR)~\cite{bowles2018gan,mondal2018few,bermudez2018learning} or chest X-rays~\cite{madani2018chest,madani2018semi}. Other methods of synthetic data generation aim to create new training samples using target domain samples and labeled source domain knowledge~\cite{wilson2020survey}. A large amount of this cross-modality synthesis work has been proposed for adapting MR data to CT~\cite{emami2018generating,nie2018medical}, CT to MR~\cite{jiang2018tumor,jin2019deep} and MR to Positron Emission Tomography (PET)~\cite{wei2018learning,pan2018synthesizing}. 

Additionally, image generation can be constrained by the appearance of the anatomical structures and segmentation maps. Many approaches have been presented in the literature that generate image-mask pairs, for instance, implementing domain adaptation from CT to MR~\cite{zhang2018translating}, generating synthetic samples to solve a segmentation task~\cite{guibas2017synthetic,costa2017towards,beers2018high,Ma_2021_ICCV} or for one-shot segmentation~\cite{zhao2019data,wang2020lt,tomar2022self}. 

In the particular case of Electron Microscopy (EM) volumes of the brain, its accurate segmentation is essential to characterize the neural structures present in the volume. Several recent works have been presented in the literature that use domain adaptation to segment neuronal structures~\cite{liu2014modular,fakhry2016residual,xiao2018deep}, vesicles~\cite{kaltdorf2017fiji}, mitochondria~\cite{oztel2017mitochondria,casser2020fast,khadangi2021net,franco2021stable} and whole-cell organelles~\cite{heinrich2021whole}. For the specific task of mitochondria segmentation, domain adaptation methods have been introduced to handle the limited availability of labeled data~\cite{bermudez2018domain,roels2019domain,peng2020unsupervised}.    

In this work, we address the complex case of domain adaptation for mitochondria segmentation across EM datasets from different tissues and species. We assume the absence of target domain annotations to simulate a real scenario. More specifically, we compare three deep learning based strategies to improve mitochondria segmentation in the target dataset based on 1) style transfer between domains, 2) self-supervised learning, and 3) multi-task neural network architectures. To demonstrate the potential of these three strategies, we employed a cross-domain thorough study between three publicly available datasets for mitochondria segmentation. The same initial conditions and basic architectural design choices are maintained across all strategies, which are also compared with the same supervised baseline methods.        

In brief, our main contributions are as follows:

\begin{itemize}
    \item We have presented state-of-the-art style transfer as a solution for domain adaptation for mitochondria segmentation in EM volumes.   
    \item We introduce a self-supervised approach based in a pre-training step using both datasets without annotations and a final fine-tuning with only source annotations. 
    \item We have performed a cross-dataset analysis of state-of-the-art deep multi-task networks for EM datasets in the context of domain adaptation and propose a novel architecture based on one of them.
    \item As a stopping criterion, we propose a new metric to ensure a good generalization towards the target domain based on the morphology of the resulting mitochondria segmentation.
\end{itemize}

\section{Related work}
\label{related}

The work presented here focuses on domain adaptation and style transfer methods for EM image analysis. By domain and style, we refer to the intrinsic feature space and characteristics of a particular dataset and the distribution from where it is drawn. Domain adaptation can be seen as a particular type of transfer learning where instead of trying to transfer the knowledge from task A in domain A to task B in domain B, the tasks are kept the same while the domains are different. On the other hand, style transfer is mainly focused on adapting the domain from one dataset to another.

Existing domain adaptation methods can be divided depending on the label availability during the training process. Thus, they can be supervised, if both source and target domain labels are available; semi-supervised, if source labels and some target labels are available; and unsupervised, if only source labels are available while target data is entirely unlabeled~\cite{Guan2021MedDAsurvey}. Moreover, methods can also be categorized based on the learning model used, i.e., either shallow (usually relying on predefined image features and traditional machine learning models) or deep (if they use deep learning architectures). In this paper, we focus on the strategy known as deep unsupervised domain adaptation.

One particular way of addressing this problem is by style transfer. For instance, the Cycle Generative Adversarial Networks (CycleGAN)~\cite{CycleGAN2017} approach is becoming an effective method in medical image synthesis. Many variations have been presented addressing cross-domain style transfer problems targeting different sources and target types of data, such as from MR to CT~\cite{yang2018unpaired,huo2018synseg,zhang2018translating,dou2018unsupervised}, transferring the stain style for histopathological images~\cite{cho2017neural,de2018stain,shaban2019staingan} or creating target-style data pairs, image and mask, without using any annotation~\cite{mahmood2018unsupervised,wang2019two,kim2021synthesis}.

More recent approaches to address style transfer exploit contrastive learning~\cite{chen2020simple}, where models are trained without labels to learn which data samples are similar or different. Similarity is defined in an unsupervised way, by using different data augmentation techniques to create similar examples to the original image and then maximizing a similarity function (e.g., mutual information) during training. Following this idea, Contrastive Unpaired Translation (CUT) \cite{park2020contrastive} compares unpaired image patches and associates similar patches to each other while disassociating them from others. This way, the model learns to pay attention to the commonalities between domains. For instance, a patch containing a mitochondrion will have a high similarity with a patch in a different tissue containing mitochondria, or at least a higher value than if it is compared with a patch showing other organelles. Thus, a generator learns to change the style of input images to match a target style.

Another way to address this domain problem is by using self-supervised learning (SSL), which consists in establishing a \textit{pretext} task using unlabeled related images that do not require to be annotated by an expert to initially train the model. Then, the model is used as the starting training point for the \textit{downstream} (segmentation) task. The main advantage is that the pretext examples (or pseudo-labels) are automatically generated from existing raw data, not being conditioned to the number of available expert-reviewed images. Therefore, during the pre-training step, models can leverage from all available images to learn useful feature representations.

In the computer vision literature, related to natural images, the usefulness of this self-supervised pre-training step has been widely explored for several tasks. Namely, the coloring of a grayscale image~\cite{chen2020simple,Katircioglu2020,Jenni2020}, the restoration of a distorted or deteriorated image~\cite{Lee2020,Dewil2020,Laine2019,li2021reinforcing}, the prediction of the transformation performed in an image~\cite{Noroozi2016} or even, the re-ordering of pieces or frames of images~\cite{Li2020,taleb2021} and videos~\cite{jiao2020}. However, there is hardly any work applying this methodology to microscopy images. The published works mostly focus on reducing the number of annotated images required for training thanks to a good network initialization achieved by pre-training with denoising~\cite{Krull2020,Buchholz,Prakash2020,Alex2017}, jigsaw solving~\cite{Taleb,Taleba} and image restoration~\cite{Chen2019}. 

Finally, another approach is based on multi-task deep neural network architectures that receive both source and target samples as input. In this case, apart from solving the downstream task for the source (labeled) data, the model aims to exploit the features of the target domain to learn the feature shift between domains. Among these types of unsupervised and semi-supervised domain adaptation methods, we find the Y-Net~\cite{roels2019domain}, used for the segmentation of EM images. Its architecture consists of an encoder-decoder such as a U-Net~\cite{ronneberger2015u}, coupled with a second decoder in an autoencoder strategy.  While one decoder is trained for segmentation, using the images with available labels, the second decoder is trained to reconstruct all available images, including the unlabeled ones, in an unsupervised manner. Since both decoders share the same encoder, the features learned by the autoencoder are used for segmentation too. Consequently, the model works with unlabeled (target domain) data features. Following this idea, in combination with adversarial losses, similar models such as Domain Adaptive Multi-Task Learning network (DAMT-Net)~\cite{peng2020unsupervised} have been proposed. This network builds on top of the Y-Net architecture and adds two discriminators during training, following a Generative Adversarial Network (GAN) approach. The first discriminator uses the predicted segmentation, while the second discriminator uses the final feature maps of the network.

\section{Methods}
\label{methods}

To address the problem of domain adaptation between different EM datasets, we present different approaches that reduce the domain shift. Firstly, a cross-domain baseline is introduced using stable state-of-the-art models~\cite{franco2021stable} trained only on source domains. Next, a simple histogram matching between domains is added as pre-processing prior to the use of the baseline models. Finally, more sophisticated domain adaptation approaches are presented based on (1) a modern style-transfer technique, (2) self-supervised pretext tasks, and (3) state-of-the-art domain adaptation multi-task deep neural networks.

\subsection{Cross-dataset baseline}\label{cross-baseline}
\begin{figure}[htp!]
\centering
 \includegraphics[width=\textwidth]{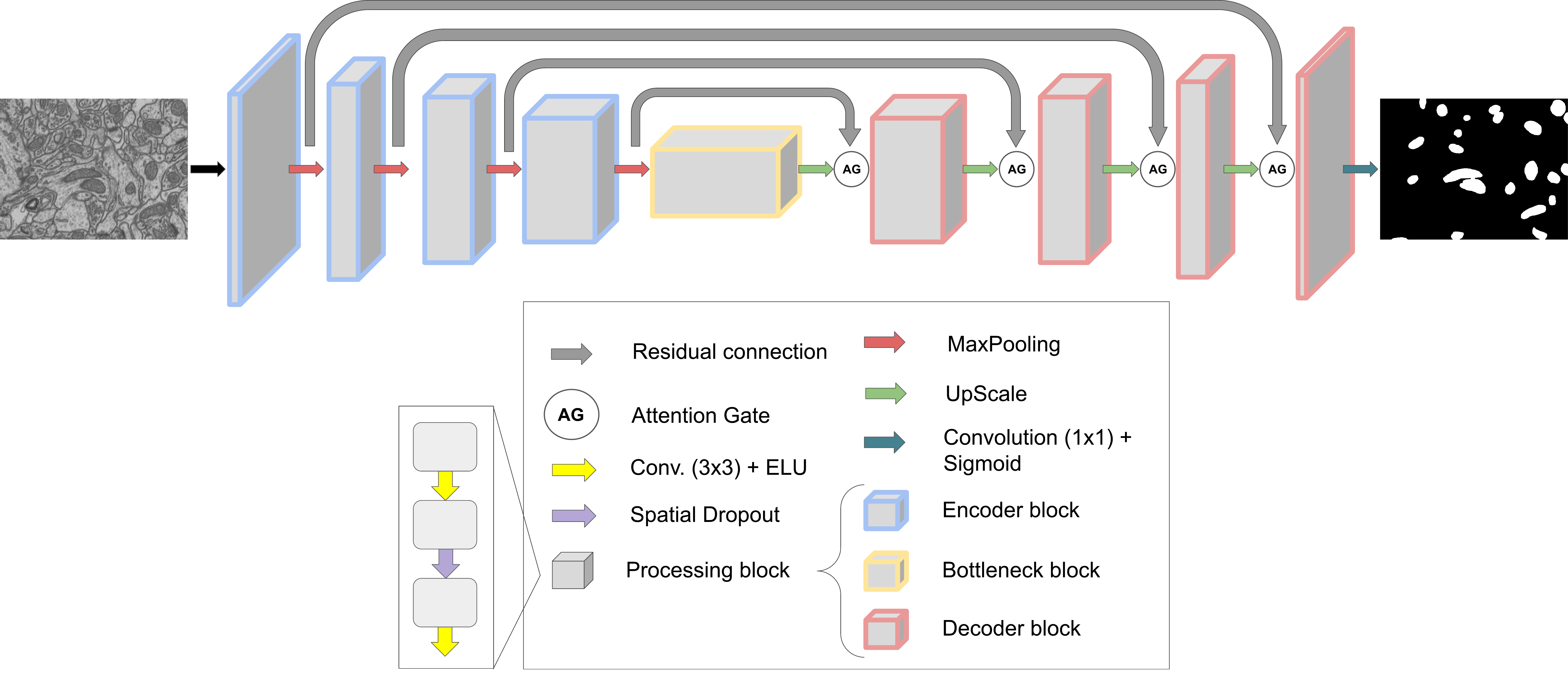}
\caption{Architecture of our Attention U-Net~\cite{franco2021stable} used for mitochondria semantic segmentation.}
\label{fig:attention_unet}
\end{figure}

As a reference method to compare our results with, we use our recent stable 2D Attention U-Net model~\cite{franco2021stable} trained on the labeled source domain and tested directly on the target domain (without any adaptation method). This network is a modified version of the U-Net~\cite{ronneberger2015u} including attention gates~\cite{SCHLEMPER2019197} in the skip connections that has proven to produce consistently robust results in the segmentation of mitochondria on EM volumes~\cite{franco2021stable}. Its architecture is shown in Figure \ref{fig:attention_unet}. 
 
\subsection{Histogram matching}\label{hist-methods}
A straightforward approach to make the images of one domain look closer to the images of another domain is histogram matching. Most commonly, this technique is applied to one source image so that its histogram matches the histogram of a target image~\cite{Gonzalez2007}. Here instead, we use as target histogram the mean histogram of the target domain images, so the histogram of all source images are transformed to match it.

\begin{figure}[htp!]
\centering
 \includegraphics[width=\textwidth]{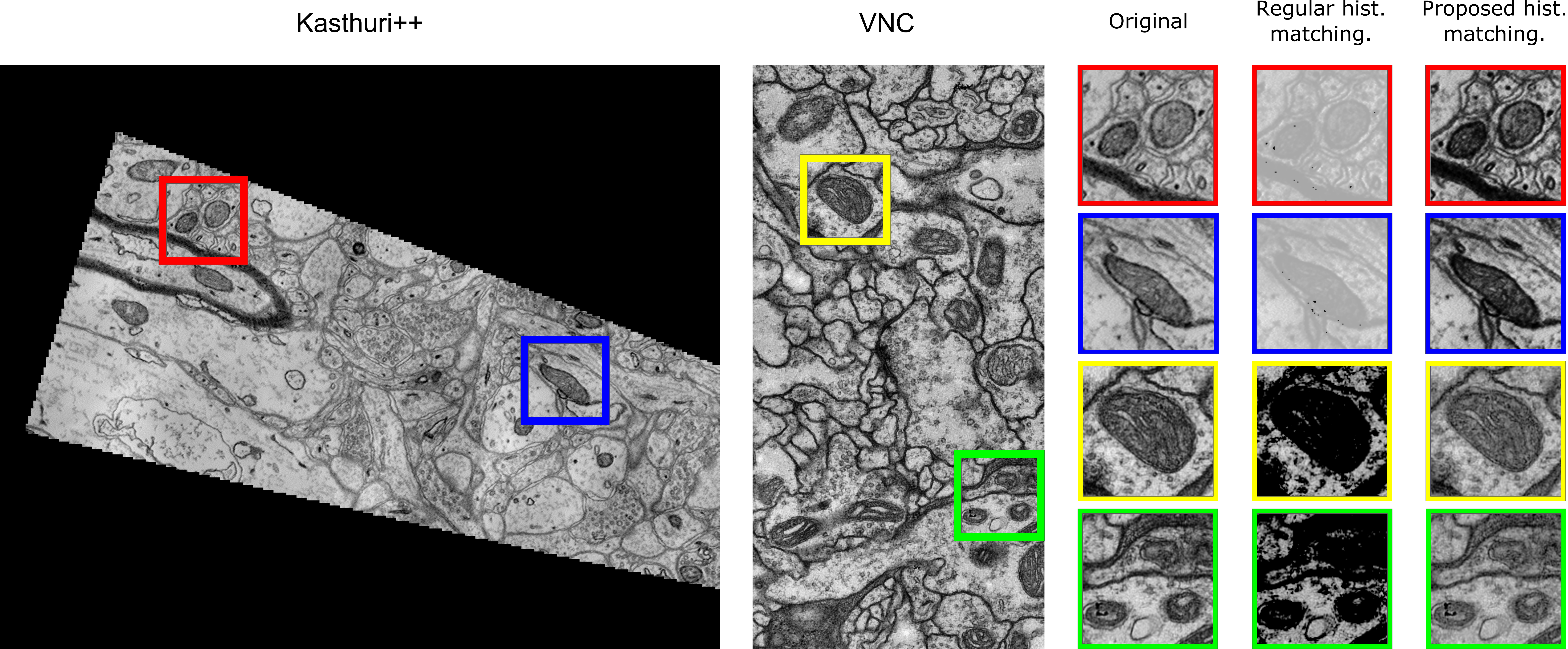}
\caption{Examples of histogram matching between source and target domain images. When using a dataset such as Kasthuri++ containing padding (non-tissue) pixels, regular histogram matching methods fail and need to be corrected to focus only on tissue intensities. From left to right: original full-size images from the Kasthuri++ and VNC datasets; four zoomed areas of both images (in red, blue, yellow and green), with their corresponding (Original) pixel values, followed by their histogram-matched versions using the full histogram (Regular hist. matching) and our proposed method to predict the zero values and avoid using padding pixels (Proposed hist. matching). For the red and blue examples, Kasthuri++ is the source domain and VNC is the target domain, while the opposite occurs for the yellow and green examples.}
\label{fig:histogram_matching}
\end{figure}

Some images of our datasets present zero-padding surrounding the tissue, which provokes an artificial high pick at the zero value in their histograms. Since we are only interested in matching the histogram of the tissue part of the images, we modified the actual number of zeros with linear regression using the first bins of the original histogram. We set the value to zero in the absence of initial values or when predicting a negative number. This process is done for both target and source histograms. Some example images processed with this histogram matching method can be seen in Figure~\ref{fig:histogram_matching}.

\subsection{Style transfer approach}
\label{st-methods}

As described in the previous section, domain adaptation can be considered a style-transfer problem. In particular, we were motivated by the success of the recent Contrastive Unpaired Translation (CUT) method~\cite{park2020contrastive} for the problem of unpaired image-to-image translation. Therefore, we tested it on our EM datasets for mitochondria segmentation and re-analyzed the cross-domain performance of our supervised baseline networks on the translated target datasets.

\begin{figure}[htp!]
\centering
 \includegraphics[width=\textwidth]{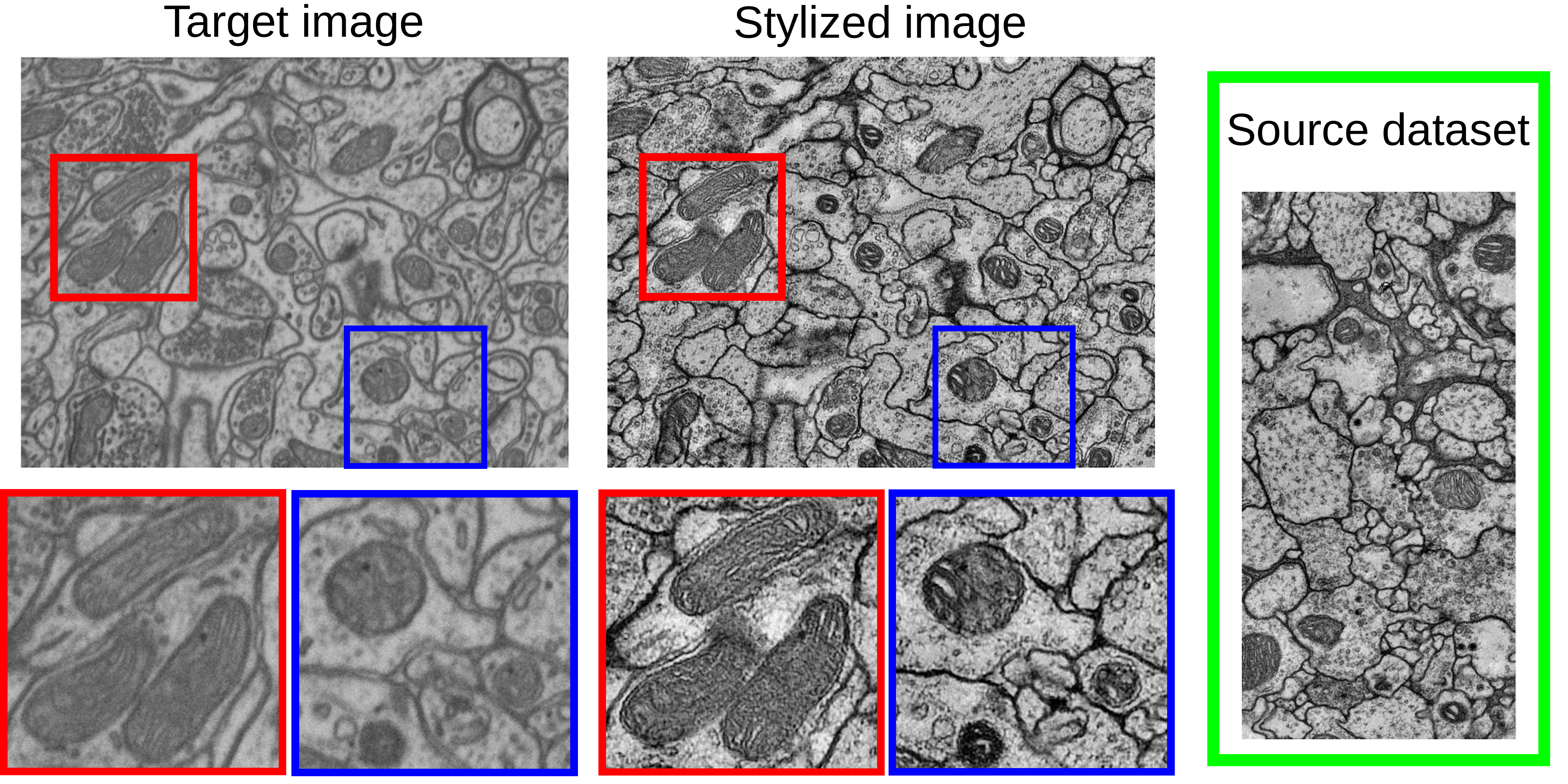}
\caption{Stylization made by the CUT~\cite{park2020contrastive} method using Lucchi++ images as source and the VNC dataset as reference (target) style. From left to right: Original Lucchi++ sample; its stylized result with the appearance of VNC; and a VNC image sample (green box). Blue and red boxes show zoomed areas from the source and stylized images.}
\label{fig:cut}
\end{figure}

In order to learn the translation between source and target images, this method randomly crops the images to patches of $512\times512$ pixels and maximizes the mutual information between the input and output patches using a contrastive learning framework. This way, corresponding patches (positives) are mapped together in feature space and far from other patches (negatives). Results of this method are shown in Figure~\ref{fig:cut}. All cross-dataset stylization results can be found in Section S1.

Following the recommendations of the original paper, we used the default hyperparameter setting as provided in their public implementation, which corresponds with training the method for $400$ epochs, with Adam as optimizer and a learning rate of $2e-4$.

\subsection{Self-supervised approach}
\label{ssl-methods}
\par
\begin{figure}[htp!]
 \includegraphics[width=\textwidth]{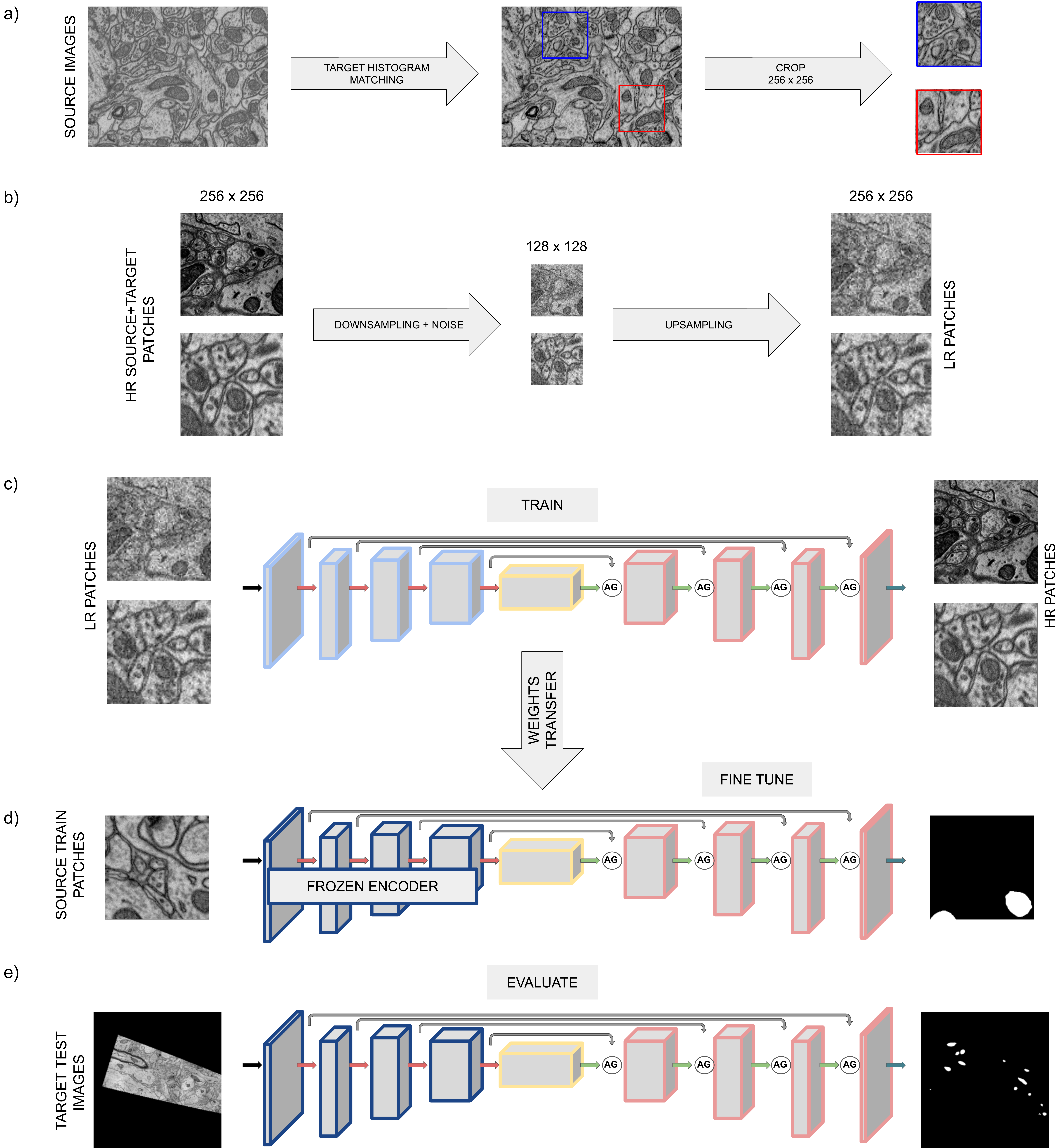}
\caption{Diagram of our self-supervised workflow for domain adaptation. From top to bottom: a) The source dataset is adjusted to the target image histogram and cropped into patches of $256\times256$ pixels; b) crops from both datasets are used to generate low-resolution samples by undersampling them and adding Gaussian noise; c) our Attention U-Net is pre-trained by learning to super-resolve the generated patches to their original versions; d) the encoder of the model is frozen and the rest of the network is fine-tuned for the mitochondria segmentation task using only source training patches and their corresponding binary masks; e) the model is evaluated on the target test dataset.}
\label{fig:ssl_flow}
\end{figure}

 As an alternative approach, we propose a self-supervised framework where we leverage from the use of two sequential training steps: (1) an initial generative self-supervised step including both (source and target) datasets without annotations, and (2) a fully-supervised fine-tuning step using only the source images and their labels. A summary of our self-supervised workflow is depicted in Figure~\ref{fig:ssl_flow}.

\textbf{Super-resolution pretext task}. In this pretext task, our Attention U-Net is trained to enhance the resolution of images from both the source and target datasets. This first step aims to reach a good starting point to solve the downstream task (i.e., supervised mitochondria segmentation). The input images are synthetically generated low-resolution images, while the ground truth is formed by the (high-resolution) original ones. To generate the synthetic input images, the original images are distorted with normally distributed Gaussian noise with \textbf{$\mu=0$} and \textbf{$\sigma=0.1$} as a fraction of the dynamic range of the image. Next, the images are downsampled by a factor of two in both axes and then upsampled by the same factor to simulate a process where the original resolution is worsened. For both downsampling and upsampling, bilinear interpolation is used. 

\textbf{Source supervised training}. Once the model has been pre-trained, the encoder gets frozen. Then, the rest of the network (bottleneck and decoder) are fine-tuned with the available source image annotations to perform semantic segmentation.  The source images are pre-processed so their histogram matches that of the target domain. The idea behind freezing the encoder is to enforce the model to remember features learnt during the previous super-resolution step from the target dataset. Thus, allowing for a better generalization and performance in the unlabeled target dataset. 

\par It is worth noting that during the super-resolution step, all available source and target images are used to train the model. That is because the input-label pairs are automatically generated from the raw data but no annotations are used. In the second step, only the training subset from the source dataset and its annotations are used to fine-tune the model.

During the pre-training step, the network is run for $200$ epochs, following a one-cycle learning rate policy~\cite{smith2019super} with a maximum learning rate of $5e-4$, and Adam optimizer. Next, the fine-tuning step is carried out for $60$ epochs, using as well a one-cycle learning rate scheduler with a maximum learning rate of $1e-4$ and Adam optimizer. In both cases, the optimal batch size was found to be $1$. All training images were randomly cropped to patches of $256\times256$ pixels, from which $10\%$ was used for validation. A more detailed description of the hyperparameters can be found in Table S3.1 as well as all combinations tested.
\newpage

\subsection{Multi-task neural networks}
\label{da-methods}

\begin{figure}[htp!]
    \centering
    \includegraphics[width=\textwidth]{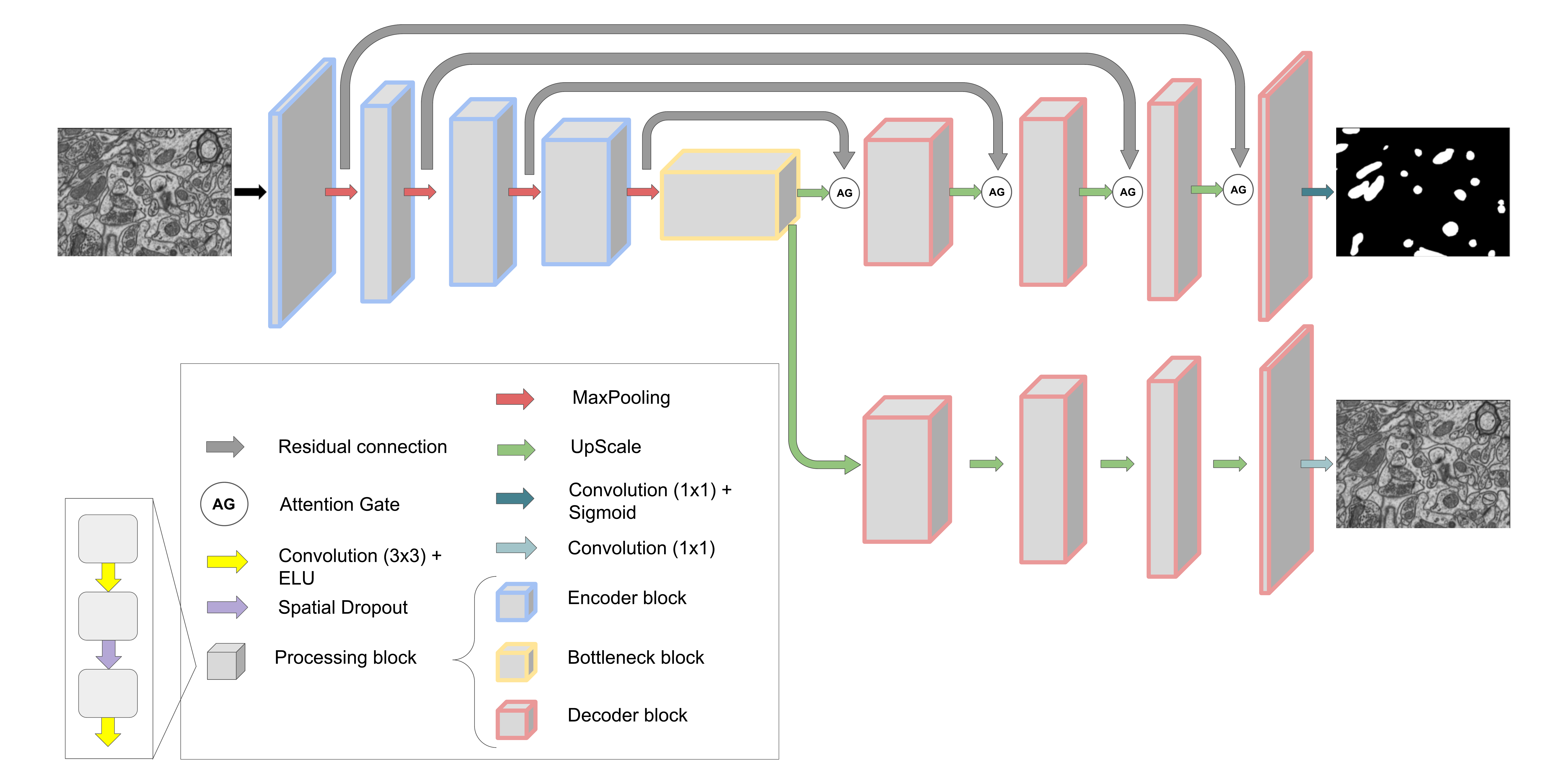}
    \caption{Architecture of the proposed Attention Y-Net used for domain adaptation. The architecture is formed by one encoder and two decoders: one for image reconstruction (without skip connections) and one for segmentation (with skip connections and attention gates).}
    \label{fig:att_ynet}
\end{figure}

Following the idea behind Y-Net~\cite{roels2019domain}, we have built a similar architecture taking as a base model the previously mentioned Attention U-Net~\cite{franco2021stable}. We refer to this network as Attention Y-Net. In short, the architecture consists of the classical encoder-decoder setup, where a new second decoder is placed. We can see the architecture as the combination of the Attention U-Net and an autoencoder, where both parts share the same encoder. The architecture is illustrated in Figure~\ref{fig:att_ynet}.

The network is trained using a loss function ($\mathcal{L}$) made of two terms: a segmentation term based on the binary cross-entropy between the predicted and ground truth masks ($\mathcal{L}_{BCE}$), and a reconstruction term based on the mean squared error between the predicted and the original grayscale images ($\mathcal{L}_{MSE}$), as given by

\begin{equation}
    \centering
    \mathcal{L} = \alpha \mathcal{L}_{MSE} + (1-\alpha) \mathcal{L}_{BCE},
    \label{eq:ynet_loss}
\end{equation}

where the weight $\alpha$ is a numeric value between $0$ and $1$. For those images without available labels (binary masks), the $\mathcal{L}_{BCE}$ value will be $0$.

In its original work, the training of the Y-Net~\cite{roels2019domain} was proposed in two sequential steps. First, the network is trained unsupervised to perform only reconstruction ($\alpha = 1$). Then, the model is fine-tuned to perform segmentation with the available labels ($\alpha = 0$). However, we have experienced instability in this step. Namely, quite often, the predicted reconstruction of the network was a flat grey-value image. Therefore, we propose a new additional step before the unsupervised pre-training, which combines both tasks using all the available data. We set $\alpha = 0.98$, which was experimentally found to help balancing both loss terms.

With our additional pre-training step, the network consistently outputs improved results, out of the local minimum achieved with the flat grey-value image. Next, we freeze the network encoder (blue blocks in Figure~\ref{fig:att_ynet}). Otherwise, the network forgets the target domain features in the next step. Experimentally, we observed that the network performs better if we let the bottleneck and the two decoders unfrozen. Remarkably, as observed with the self-supervised approach, the performance of the whole process was greatly enhanced thanks to the use of histogram matching after the first step.

The first step was carried out for $50$ epochs. We used an initial learning rate of $1e-3$ that got reduced when reaching plateaus, stochastic gradient descent (SGD) as optimizer and a patience of $7$ epochs over the monitored validation loss. In the second training step, we train for $40$ epochs (with a patience of 6). We use a learning rate of $2e-4$, and a “reduce on plateau” scheduler once again, but this time with Adam optimizer. Finally, in the last training step, we train for $100$ epochs (the different stop criteria will be analysed later). We follow a one-cycle learning rate policy~\cite{smith2019super} with a maximum learning rate of $2e-4$, and use Adam as optimizer. For all training steps,  the optimal batch size was found to be $1$. The input to the model consists of $1000$ random cropped patches of $256\times256$ pixels, from which $10\%$ is used for validation. This training configuration was empirically found. A more detailed description of the hyperparameters as well as all combinations tested can be found in Table S3.2.

\section{Experimental Results}
\label{experiments}

\subsection{EM Datasets}
\label{sec:datasets}

All the experiments performed in this work are based on the following publicly available datasets:

\textbf{EPFL Hippocampus or Lucchi dataset~\cite{lucchi2011supervoxel}}. 
The original volume represents a $5\times5\times5$  $(\mu\mbox{m})^3$ section of the CA1 hippocampus region of a mouse brain, with an isotropic resolution of $5\times5\times5$~nm per voxel. The volume of $2048\times1536\times1065$~voxels was acquired using scanning electron microscopes (SEM), specifically with focused ion beam scanning electron microscopy (FIB-SEM). The mitochondria of two sub-volumes formed by $165$ slices of $1024\times768$~pixels were manually labeled by experts, and are used as the official training and test partitions. In particular, we used a more recent version of the labels~\cite{casser2020fast} after two neuroscientists and a senior biologist re-labeled mitochondria by fixing misclassifications and boundary inconsistencies.

\textbf{Kasthuri++ dataset~\cite{casser2020fast}}. This is a re-labeling of the dataset by \cite{kasthuri2015saturated}. The volume corresponds to a part of the somatosensory cortex of an adult mouse and was acquired using scanning electron microscopes (SEM) as Lucchi++, but specifically with serial section electron microscopy (ssEM). The train and test volume dimensions are  $1463\times1613\times{85}$~voxels and  $1334\times1553\times75$~voxels, respectively, with an anisotropic resolution of $3\times3\times30$~nm per voxel. 

\textbf{VNC dataset~\cite{gerhard2013segmented}}. This dataset represents a $4.7\times4.7\times1$~$(\mu\mbox{m})^3$ serial section transmission electron microscopy (ssTEM), acquired using transmission electron microscopy (TEM), of the \textit{Drosophila} melanogaster third instar larva ventral nerve cord, with an  an isotropic resolution of $4.6\times4.6\times45-50$~nm per voxel. Two volumes of  $1024\times1024\times20$~voxels were acquired, but only one of them was labeled. For that reason and following common practice, we use only the later and split the data volume along the x axis into two subsets with equal size ($20\times512\times1024$~voxels) that constitute our training and test partitions.

For fair comparison with other published work, only the training set labels of the source datasets are used during the supervised or fine-tuning steps of our approaches, while the quantitative evaluation is performed only on the test set of the target datasets.

\subsection{Evaluation metrics}
\label{sec:eval_metrics}
Since our downstream task is semantic segmentation, we evaluate all methods using the Jaccard index of the positive class or foreground intersection over union ($IoU_{F}$), defined as
\begin{equation}
    \centering
    IoU_{F} = \frac{TP}{TP+FP+FN}
\end{equation}

where TP are the true positives, FP the false positives and FN the false negatives. As a convention, the positive class is foreground and the negative class, background. This way, $IoU_{F}$ values range from $0$ to $1$, where $0$ represents no overlap at all between the ground truth and the predicted mitochondria masks, and $1$ means a perfect overlap. 

\subsection{Stopping criterion}\label{sec:stop-criterion}

An intrinsic issue of unsupervised domain adaptation methods is blindly deciding when to stop their respective optimization processes since no labels are available from the target domain samples to guide us in such optimization. This problem is common to all our proposed approaches, either to select the number of stylization iterations or to fix the number of epochs to train our self-supervised or multi-task models. For that reason, we have selected a stopping criterion using morphological priors extracted from the source labels. More specifically, we calculate the average \textit{solidity} $\overline{S}$ of each mitochondrion in the image as:

\begin{equation}
   \overline{S} = \frac{1}{N} \sum_{n=1}^{N} solidity(n)
\end{equation}

where $N$ is the total number of objects (in our case mitochondria instances) in the image and $solidity(n)$ is the ratio of pixels in the $nth$ object to pixels of the convex hull of that  object. In practice, each instance is found by the connected components algorithm on the binarized outputs of the models.

The main advantage of the average solidity is that it is agnostic of the dataset resolution and easy to implement. As a criterion, we can monitor the $\overline{S}$ value of the predictions in the target dataset and stop optimizing our domain adaptation methods when it moves away from the objective $\overline{S}$ value (measured in the source domain). To select the best model, one can simply take the model producing test masks with the $\overline{S}$ value that is closest to the objective one. Moreover, to increase the robustness of this criterion, we discard very tiny objects (with less than ten pixels) for all datasets. 

\begin{figure*}[t!]
    \centering
    \begin{subfigure}[t]{0.5\textwidth}
        \centering
        \includegraphics[height=1.6in]{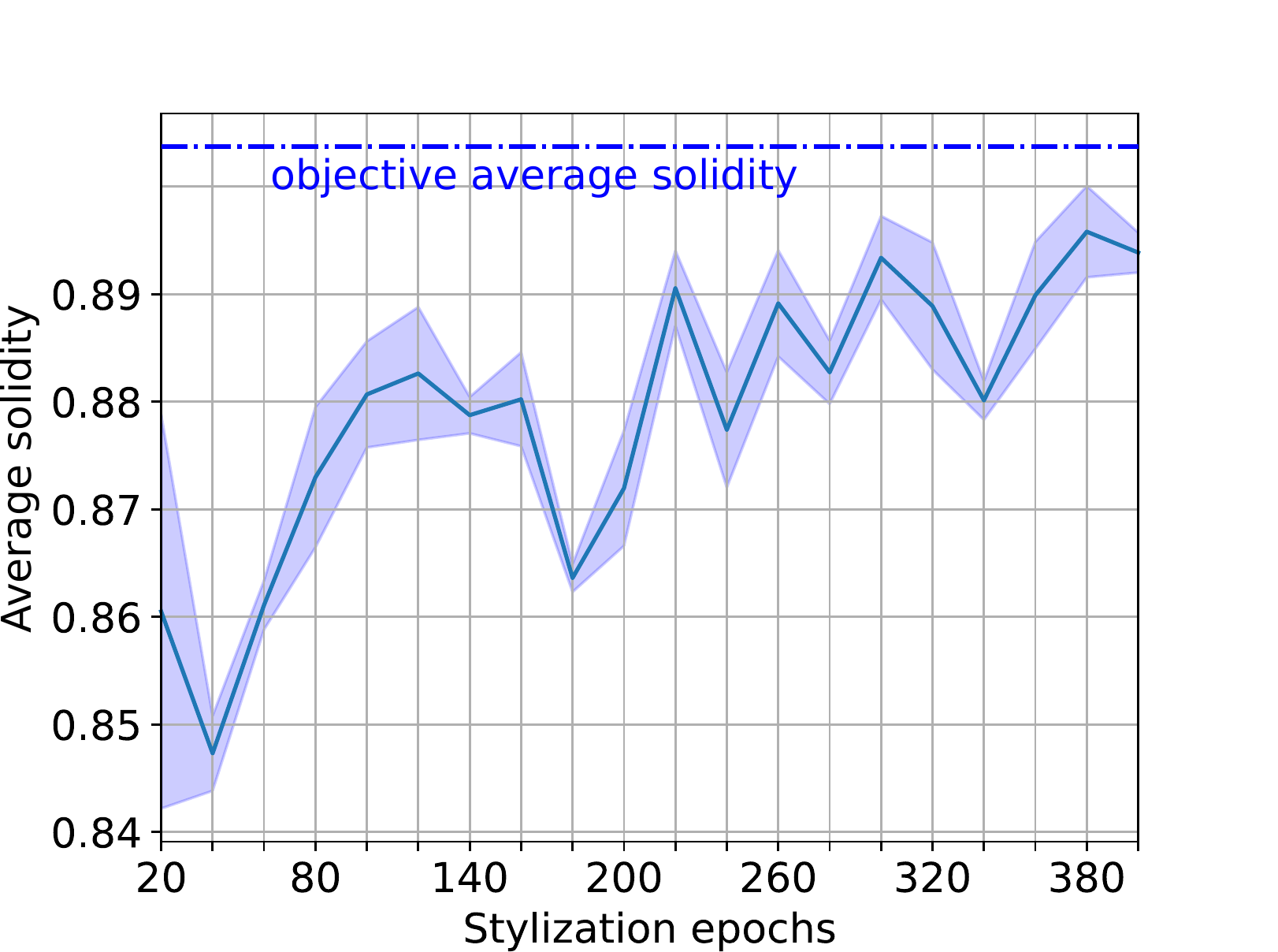}
        \caption{}
        \label{fig:solidity_example:solidity}
    \end{subfigure}%
    ~ 
    \begin{subfigure}[t]{0.5\textwidth}
        \centering
        \includegraphics[height=1.6in]{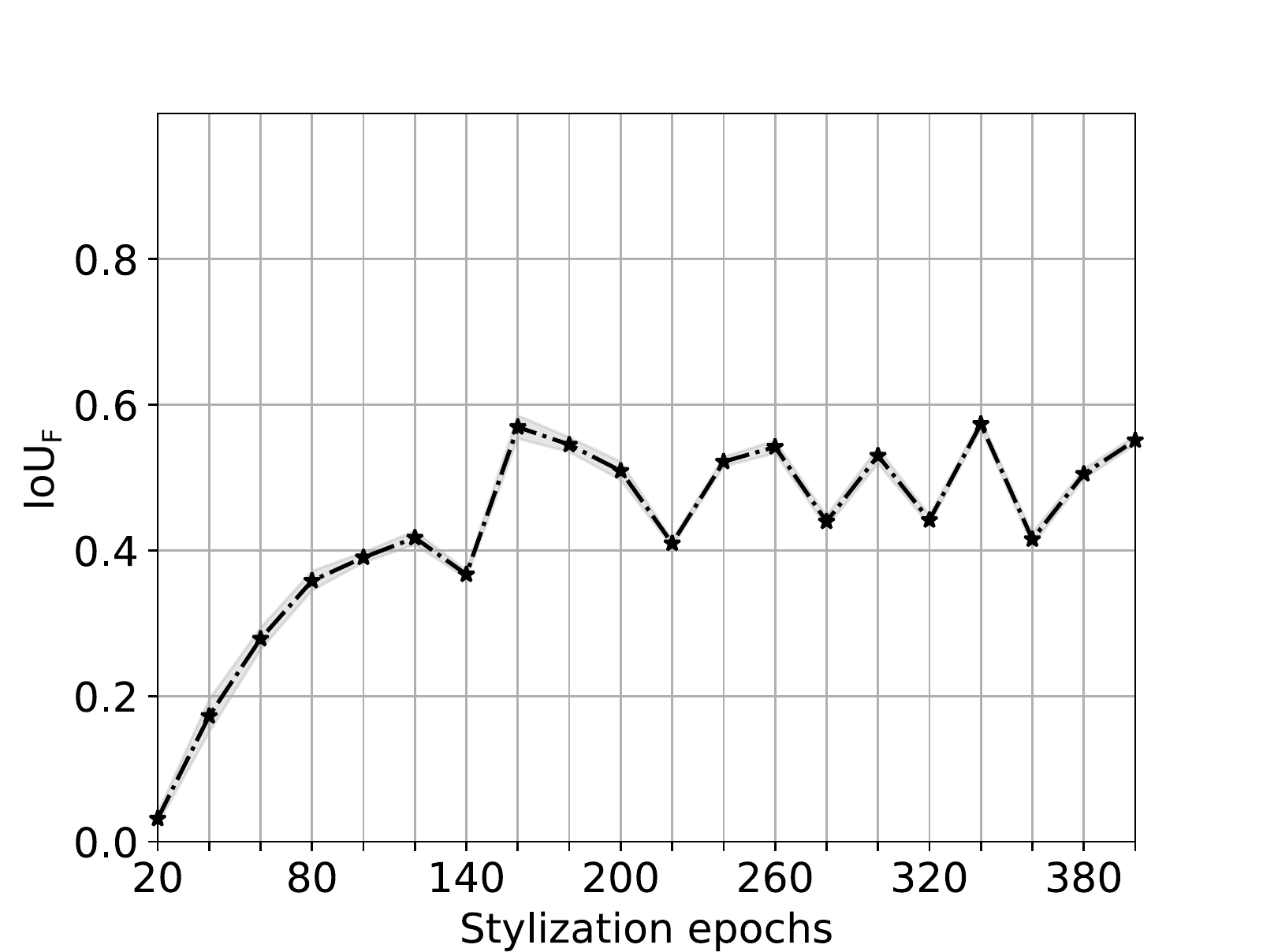}
        \caption{}
        \label{fig:solidity_example:IoU}
    \end{subfigure}
    \caption{Example of connection between average solidity $\overline{S}$ and $IoU_F$ values: (a) Average value of $\overline{S}$ (shaded area represents its standard deviation) as function of the epochs run for the style-transfer method. The source $\overline{S}$ value is depicted with a dashed line. (b) Average $IoU_F$ value (shaded area represents its standard deviation) as function of the same epochs. Values are calculated over the output of ten style-transfer model executions using Kasthuri++ and VNC as source and target domains, respectively.}
    \label{fig:solidity_example}
\end{figure*}

An example of the connection between the $\overline{S}$ values of test predictions and their respective segmentation results expressed in terms of $IoU_F$ is shown in Figure~\ref{fig:solidity_example}. One can observe that the range of epochs where the test $\overline{S}$ values are closer to the objective $\overline{S}$ (calculated in the source domain) in Figure~\ref{fig:solidity_example:solidity} correspond, overall, to the epochs with higher $IoU_F$ values in Figure~\ref{fig:solidity_example:IoU}. The same plots for all methods and cross-dataset experiments can be found in Section S2.

\subsection{Cross-dataset results}
All the methods proposed here were applied to all the possible source-target combinations of the three EM datasets introduced in Section~\ref{sec:datasets}. Moreover, for a more detailed evaluation and comparison with the state of the art, we executed as well the same experiments using the publicly available implementation of DAMT-Net~\cite{peng2020unsupervised}. As it is an extended practice on EM image processing, we also tested all methods on the same image data after preprocessing them using contrast limited adaptive histogram equalization (CLAHE)~\cite{zuiderveld1994contrast}. Notice CLAHE is a contrast equalization method, thus not intended to match two intensity distributions. However, its effect on the image contrast may bring the histogram of our datasets closer to each other.

To ensure the robustness of the proposed training configurations and hyperparameters, each experiment was repeated ten times using exactly the same setup. A full description of the search of hyperparameters for each approach can be found in Section S3.

The best results based on the average $IoU_F$ of the predicted mitochondria in the corresponding target test images for each method are shown in Table~\ref{table:cross_dataset_results}. Furthermore, we explored the impact of stopping the model training by each of the following criteria: (1) monitoring the $IoU_F$ value of the source validation set (and also selecting the best model based on that value); (2) leaving the model train for a fixed number of epochs; and (3) monitoring the average solidity values of the target test set (and selecting the model that better approaches the known source average solidity value).

First, although expected, it is worth mentioning that all tested methods outperform the baseline in all cases, demonstrating the need for a domain adaptation strategy that allows addressing the domain shift problem.
Secondly, we can observe an evident boost in performance by simply applying either our histogram matching method to the target images or CLAHE as preprocessing for all images, and re-using the baseline models for inference. Interestingly, on one of the source-target combinations (Lucchi++ as source and Kasthuri++ as target) these strategies provide very good segmentation results ($IoU_F=0.679$ and $0.620$ respectively), but they perform poorly ($IoU_F=0.268$ and $0.249$) on the opposite experiment (Kasthuri++ as source and Lucchi++ as target). This reflects an asymmetric aspect of the problem and the need for solutions that learn more than just simple histogram image features.
Moreover, these results show our proposed methods generally perform favourably to the state of the art, represented by DAMT-Net~\cite{peng2020unsupervised}. In particular, our style-transfer based approach provides consistent results across all datasets, followed by our proposed multi-task Attention Y-Net.

Finally, the choice of the stopping criterion seems to play an important role improving the segmentation results depending on the dataset combination. Although the monitoring of the source validation results is a good indicator of the performance in the target domain by the multi-task networks (DAMT-Net and Attention Y-Net), we observe their segmentation can be improved by either leaving the training converge (with a maximum number of epochs) or by monitoring the target average solidity instead.

\begin{table}[ht]
\begin{adjustbox}{max width=\textwidth}

\begin{tabular}{clllllllllll}
\multicolumn{1}{l}{}                                                               &               &                                                            &               & \multicolumn{2}{c}{\textbf{Source: Lucchi++}}                              &               & \multicolumn{2}{c}{\textbf{Source: Kasthuri++}}                              &               & \multicolumn{2}{c}{\textbf{Source: VNC}}                                            \\ \cline{5-6} \cline{8-9} \cline{11-12} 
\textbf{Stop criteria}                                                             & \phantom{abc} & \textbf{Method}                                            & \phantom{abc} & \multicolumn{1}{c}{\textbf{Kasthuri++}} & \multicolumn{1}{c}{\textbf{VNC}} & \phantom{abc} & \multicolumn{1}{c}{\textbf{Lucchi++~~~~}} & \multicolumn{1}{c}{\textbf{VNC}} & \phantom{abc} & \multicolumn{1}{c}{\textbf{Lucchi++~~~~}} & \multicolumn{1}{c}{\textbf{Kasthuri++}} \\ \hline
\multirow{7}{*}{\textbf{\begin{tabular}[c]{@{}c@{}}Source\\ val set\end{tabular}}} &               & Baseline~\cite{franco2021stable}                           &               & 0.017$\pm$0.008                         & 0.009$\pm$0.010                  &               & 0.000$\pm$0.000                           & 0.095$\pm$0.013                  &               & 0.351$\pm$0.101                           & 0.288$\pm$0.050                         \\
                                                                                   &               & Baseline~\cite{franco2021stable} + CLAHE                   &               & 0.620$\pm$0.051                         & 0.249$\pm$0.021                  &               & 0.433$\pm$0.085                           & 0.121$\pm$0.045                  &               & 0.586$\pm$0.016                           & 0.534$\pm$0.065                         \\
                                                                                   &               & Baseline~\cite{franco2021stable} + HM (ours)               &               & 0.679$\pm$0.043                         & 0.265$\pm$0.028                  &               & 0.268$\pm$0.048                           & 0.111$\pm$0.011                  &               & 0.531$\pm$0.019                           & 0.454$\pm$0.035                         \\
                                                                                   &               & Attention Y-Net + HM (ours)                                &               & 0.668$\pm$0.020                         & 0.402$\pm$0.040                  &               & 0.704$\pm$0.045                           & 0.252$\pm$0.048                  &               & 0.536$\pm$0.022                           & 0.389$\pm$0.041                         \\
                                                                                   &               & DAMT-Net~\cite{peng2020unsupervised}                       &               & 0.279$\pm$0.078                         & 0.469$\pm$0.054                  &               & 0.569$\pm$0.088                           & 0.324$\pm$0.038                  &               & 0.491$\pm$0.102                           & 0.162$\pm$0.042                         \\
                                                                                   &               & DAMT-Net~\cite{peng2020unsupervised} + HM                  &               & 0.226$\pm$0.037                         & 0.489$\pm$0.040                  &               & 0.438$\pm$0.094                           & 0.274$\pm$0.080                  &               & 0.371$\pm$0.123                           & 0.170$\pm$0.049                         \\
                                                                                   &               & DAMT-Net~\cite{peng2020unsupervised} + CLAHE               &               & 0.299$\pm$0.099                         & 0.497$\pm$0.029                  &               & 0.547$\pm$0.088                           & 0.346$\pm$0.047                  &               & 0.545$\pm$0.039                           & 0.221$\pm$0.085                         \\ \hline
\multirow{8}{*}{\textbf{\begin{tabular}[c]{@{}c@{}}Last \\ epoch\end{tabular}}}    &               & Style transfer (ours, ~\cite{park2020contrastive})         &               & 0.515$\pm$0.011                         & 0.586$\pm$0.009                  &               & 0.569$\pm$0.003                           & \textbf{0.551$\pm$0.006}         &               & 0.638$\pm$0.014                           & \textbf{0.654$\pm$0.026}                \\
                                                                                   &               & SSL + HM (ours)                                            &               & 0.568$\pm$0.165                         & 0.327$\pm$0.135                  &               & 0.511$\pm$0.145                           & 0.138$\pm$0.043                  &               & 0.582$\pm$0.237                           & 0.237$\pm$0.191                         \\
                                                                                   &               & SSL + CLAHE (ours)                                         &               & 0.254$\pm$0.159                         & 0.149$\pm$0.113                  &               & 0.456$\pm$0.189                           & 0.153$\pm$0.077                  &               & 0.205$\pm$0.156                           & 0.162$\pm$0.111                         \\
                                                                                   &               & SSL + HM + CLAHE (ours)                                    &               & 0.578$\pm$0.160                          & 0.187$\pm$0.084                  &               & 0.421$\pm$0.177                           & 0.166$\pm$0.061                  &               & 0.270$\pm$0.139                           & 0.116$\pm$0.091                         \\
                                                                                   &               & Attention Y-Net + HM (ours)                                &               & 0.669$\pm$0.019                         & 0.388$\pm$0.026                  &               & 0.719$\pm$0.024                           & 0.232$\pm$0.024                  &               & 0.540$\pm$0.014                           & 0.404$\pm$0.016                         \\
                                                                                   &               & DAMT-Net~\cite{peng2020unsupervised}                       &               & 0.261$\pm$0.039                         & 0.455$\pm$0.066                  &               & 0.581$\pm$0.057                           & 0.295$\pm$0.040                  &               & 0.449$\pm$0.082                           & 0.169$\pm$0.055                         \\
                                                                                   &               & DAMT-Net~\cite{peng2020unsupervised} + HM                  &               & 0.258$\pm$0.062                         & 0.422$\pm$0.169                  &               & 0.416$\pm$0.078                           & 0.276$\pm$0.072                  &               & 0.380$\pm$0.077                           & 0.187$\pm$0.118                         \\
                                                                                   &               & DAMT-Net~\cite{peng2020unsupervised} + CLAHE               &               & 0.284$\pm$0.047                         & 0.482$\pm$0.050                  &               & 0.440$\pm$0.135                           & 0.319$\pm$0.100                  &               & 0.488$\pm$0.061                           & 0.233$\pm$0.100                         \\ \hline
\multirow{11}{*}{\textbf{Solidity}}                                                &               & Style transfer (ours, ~\cite{park2020contrastive})         &               & 0.703$\pm$0.009                         & 0.605$\pm$0.032                  &               & 0.572$\pm$0.044                           & 0.509$\pm$0.034                  &               & 0.608$\pm$0.017                           & 0.560$\pm$0.032                         \\
                                                                                   &               & Style transfer (ours, ~\cite{park2020contrastive}) + CLAHE &               & \textbf{0.768$\pm$0.020}                & \textbf{0.671$\pm$0.009}         &               & 0.529$\pm$0.051                           & 0.146$\pm$0.165                  &               & 0.581$\pm$0.017                           & 0.572$\pm$0.078                         \\
                                                                                   &               & SSL + HM (ours)                                            &               & 0.685$\pm$0.092                         & 0.394$\pm$0.102                  &               & 0.572$\pm$0.109                           & 0.136$\pm$0.027                  &               & \textbf{0.694$\pm$0.022}                  & 0.278$\pm$0.171                         \\
                                                                                   &               & SSL + CLAHE (ours)                                         &               & 0.204$\pm$0.164                         & 0.165$\pm$0.164                  &               & 0.477$\pm$0.168                           & 0.170$\pm$0.073                   &               & 0.253$\pm$0.150                            & 0.186$\pm$0.088                         \\
                                                                                   &               & SSL + HM + CLAHE (ours)                                    &               & 0.649$\pm$0.093                         & 0.152$\pm$0.069                  &               & 0.510$\pm$0.177                           & 0.177$\pm$0.043                  &               & 0.249$\pm$0.105                           & 0.083$\pm$0.084                         \\
                                                                                   &               & Attention Y-Net + HM (ours)                                &               & 0.713$\pm$0.029                         & 0.397$\pm$0.013                  &               & \textbf{0.728$\pm$0.031}                  & 0.310$\pm$0.067                  &               & 0.508$\pm$0.030                           & 0.416$\pm$0.026                         \\
                                                                                   &               & Attention Y-Net + CLAHE (ours)                             &               & 0.729$\pm$0.017                         & 0.420$\pm$0.041                  &               & 0.635$\pm$0.049                           & 0.350$\pm$0.025                  &               & 0.545$\pm$0.031                           & 0.516$\pm$0.094                         \\
                                                                                   &               & Attention Y-Net + HM + CLAHE (ours)                        &               & 0.731$\pm$0.047                         & 0.396$\pm$0.023                  &               & 0.678$\pm$0.029                           & 0.360$\pm$0.017                  &               & 0.551$\pm$0.030                           & 0.565$\pm$0.020                         \\
                                                                                   &               & DAMT-Net~\cite{peng2020unsupervised}                       &               & 0.223$\pm$0.091                         & 0.441$\pm$0.140                  &               & 0.608$\pm$0.070                           & 0.224$\pm$0.073                  &               & 0.502$\pm$0.048                           & 0.180$\pm$0.063                         \\
                                                                                   &               & DAMT-Net~\cite{peng2020unsupervised} + HM                  &               & 0.230$\pm$0.038                         & 0.497$\pm$0.059                  &               & 0.655$\pm$0.037                           & 0.308$\pm$0.035                  &               & 0.551$\pm$0.045                           & 0.172$\pm$0.114                         \\
                                                                                   &               & DAMT-Net~\cite{peng2020unsupervised} + CLAHE               &               & 0.244$\pm$0.043                         & 0.506$\pm$0.037                  &               & 0.625$\pm$0.066                           & 0.291$\pm$0.100                  &               & 0.554$\pm$0.053                           & 0.288$\pm$0.097                         \\ \hline
\end{tabular}

\end{adjustbox}
\vspace{0.1cm}
\caption{Cross-dataset domain adaptation methods evaluation. Results are shown based on the mean $IoU_F$ value ($\pm$ standard deviation) obtained in the test partition of the target datasets under the three possible stopping criteria: (1) performance on the validation partition of the source dataset, (2) maximum number of epochs (experimentally found for each method), and (3) the proposed average solidity metric. The best results of each column are shown in bold. CLAHE and HM refer to the use of contrast limited adaptive histogram equalization~\cite{zuiderveld1994contrast} and histogram matching as pre-processing methods, respectively. }
\label{table:cross_dataset_results}
\end{table}

\begin{figure}[ht]
\centering
\includegraphics[width=\textwidth]{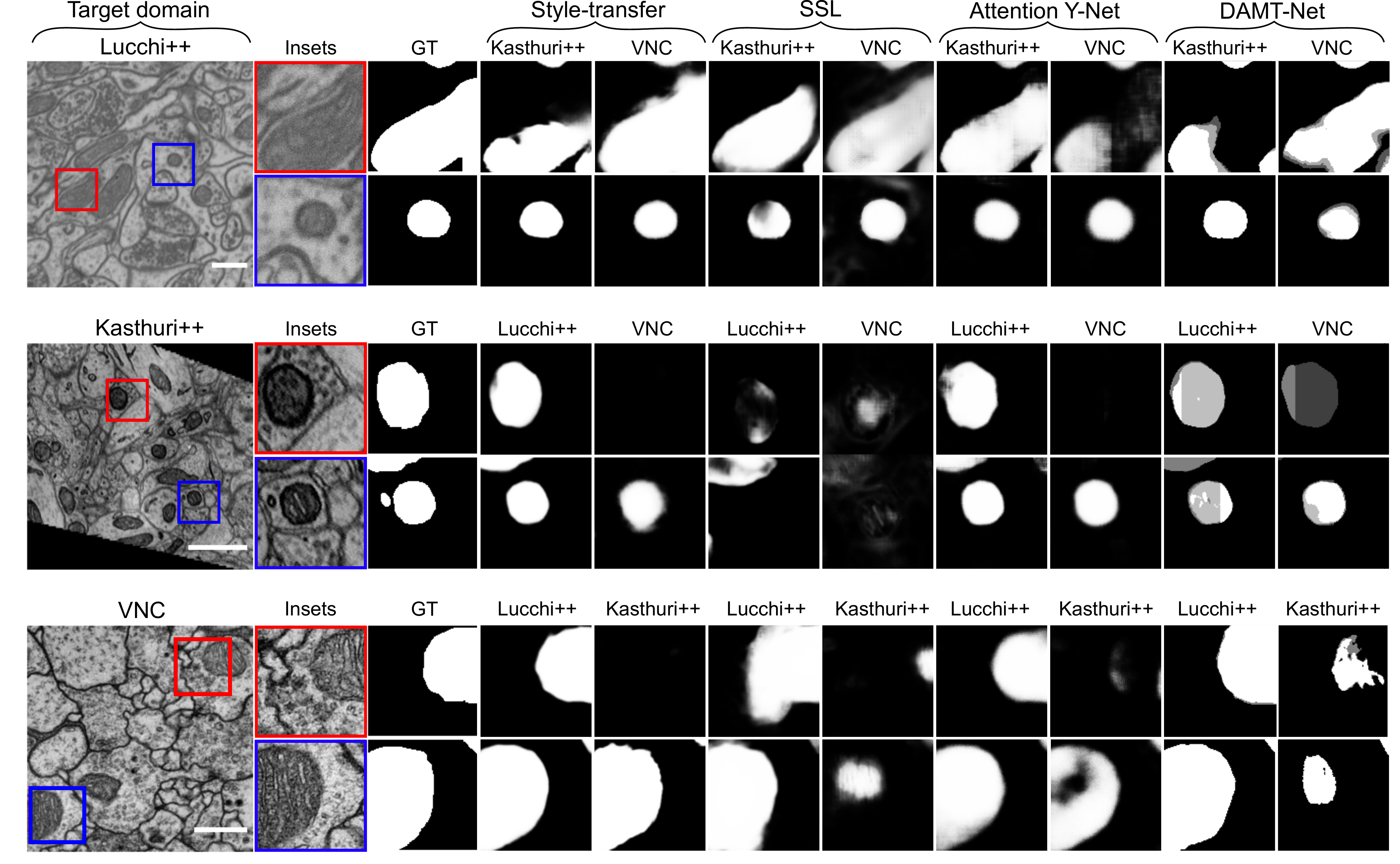}
\caption{Cross-dataset segmentation results using average solidity as stopping criterion for all learning based methods. From top to bottom: segmentation results when using Lucchi++, Kasthuri++ and VNC as target datasets. From left to right: image sample from the target dataset, two crops of that sample (in red and blue), their corresponding ground-truth (GT) binary masks, and probability maps produced by each method (Style-transfer, SSL, Attention Y-Net and DAMT-Net). The white scale bar represents $500$~nm.}
\label{fig:cross_dataset_results}
\end{figure}

Some qualitative results of the learning-based methods are shown in Figure~\ref{fig:cross_dataset_results}, where the probability maps of mitochondria masks produced by each method are displayed side by side for the same sample images. More specifically, the predictions shown were obtained using average solidity as stopping criterion. In agreement with the quantitative results of Table~\ref{table:cross_dataset_results}, we can observe most methods predict reasonable masks when Lucchi++ is used as the target dataset (where the $IoU_F$ values are in the range of $\sim0.5-0.7$), but present different levels of performance when predicting the mitochondria of the two other datasets used as target. Remarkably, all methods except our style-transfer approach struggle with the VNC/Kasthuri++ and Kasthuri++/VNC combinations, suggesting a larger domain shift between those two datasets.

\section{Conclusions and Discussion}
\label{conclusions}

In this paper, we address the problem of domain adaptation for the challenging task of semantic segmentation of EM volumes. More specifically, we propose three novel solutions that built on top the deep-learning based state of the art by means of (1) unsupervised style transfer to transform the target domain images into the "style" of the source domain and then reuse robust models trained on annotated data; (2) self-supervised learning to pre-train our segmentation models without annotations and then fine-tune them using the source labels; and (3) a multi-task deep architecture able to learn from both labeled and unlabeled data. All methods have been evaluated under the same setups using three publicly available EM datasets of different modalities (FIB-SEM, ssEM and ssTEM) and each of their possible source-target combinations. In addition, we propose a novel unsupervised metric to avoid blindly selecting the best model during training.

First of all, quantitative and qualitative results prove that learning-based methods are needed to deal with the domain shift in five out of the six cross-dataset experiments. Only in one combination (Lucchi++ as source domain and Kasthuri++ as target domain) an ad-hoc histogram matching method has been able to reduce the shift at the level of the learning approaches.

Regarding the proposed approaches, the style-transfer based method produces segmentation results with consistently medium-high $IoU_F$ values ($\sim 0.5-0.6$), specially when the stylization is run for a large number of epochs ($>200$, see Section S2).
The performance of our SSL and Attention Y-Net methods also gets stabilized after a fixed number of training epochs ($60$ and $100$, respectively) as can be seen in Section S2.
However, their results are not as consistent as those of the style-transfer approach, oscillating between low ($0.1-0.2$) and high ($0.6-0.7$) values of $IoU_F$ depending on the specific source and target dataset combination. Nevertheless, we have been able to estimate the correct number of epochs to train the models thanks to the availability of target labels (although they are not used at all during training). In a real scenario, monitoring the proposed average solidity metric is an intuitive and effective way to stop the training process in the absence of validation labels, and select (in average) models of similar or better accuracy. Although other morphological and area measurements were initially tested, the average solidity correlates better with the $IoU_F$ value of the test labels. Nevertheless, the performance of this metric depends on how close its value is the source and target domains.

It is also interesting to note that TEM and SEM images are different, with TEM images usually having higher resolution. Consequently, Lucchi++ and Kasthuri++ datasets (SEM) are -in principle- in closer domains compared to VNC (TEM) as reflected by the baseline results in Table~\ref{table:cross_dataset_results}. When Lucchi++ or Kasthuri++ are used as sources, the results obtained with VNC are clearly lower than with Kasthuri++ and Lucchi++, respectively. However, when VNC is used as source, the results obtained with Lucchi++ or Kasthuri++ are similar. As similar discussion is applicable to the figures presented in Section S1: going to lower resolution (i.e., from VNC as a source, to Lucchi++ or Kasthuri++) in principle, could be easier than the opposite (from Lucchi++ or Kasthuri++ as a source, to VNC). Apart from the intrinsic variability due to the modality, we need to acknowledge also the variability due to the differences in the samples itself, their preparation and the acquisition protocol.

In summary, from a practical point of view, the style-transfer approach appears as both the safest and simplest way of addressing the domain shift in EM volumes for semantic segmentation. Nevertheless, using self-supervised or multi-task models may provide better results on specific datasets at the cost of more complex training setups and a larger set of hyperparameters.

The present work is an initial assessment of the three competing approaches running under the same conditions and compared with the same supervised baseline methods. In a future work, we plan to explore the performance of meaningful combinations of the proposed strategies. Namely, the outputs of the style transfer method could be used as inputs or the self-supervised learning and the multi-tasks neural network architectures. We expect the combined strategies to outperform the histogram matching approach.

Moreover, current initiatives (e.g., volume EM, \url{http://www.volumeem.org}) are developing massive databases of heterogeneous 3DEM data. These initiatives promise to facilitate deep-learning-based model building for automated segmentation~\cite{conrad2021cem500k}. In our view, the style-transfer strategies could be more effective when pre-trained in massive databases of heterogeneous 3DEM data than in a small dataset of well-defined characteristics.

Finally, it is important to highlight that even the best results among all our proposed domain adaptation strategies lie much lower than the fully supervised approaches. As a reference, the average $IoU_F$ values obtained by our baseline models trained on the target annotated images are $0.9066$ for Lucchi++, $0.9154$ for Kasthurhi++, and $0.8041$ for VNC. This leaves plenty of room for improvement and future lines of research. In particular, we will explore the use of massive databases of heterogeneous 3D EM data, with the combination of some of our proposed strategies and the exploitation of segmentation-specific pretext tasks.

\section*{Code Availability}
The developed software that support the findings of this study are publicly available at \url{https://github.com/danifranco/EM_domain_adaptation}.

\section*{Data availability}
The Lucchi++ and Kasthuri++ datasets can be downloaded from \url{https://sites.google.com/view/connectomics/}. The VNC dataset can be downloaded from \url{https://github.com/unidesigner/groundtruth-drosophila-vnc}. 

\section*{Acknowledgments}
I. Arganda-Carreras would like to acknowledge the support of the 2020 Leonardo Grant for Researchers and Cultural Creators, BBVA Foundation. This work is supported in part by the University of the Basque Country UPV/EHU grant GIU19/027 and by Ministerio de Ciencia, Innovación y Universidades, Agencia Estatal de Investigación, under grant PID2019-109820RB-I00, MCIN/AEI /10.13039/501100011033/, cofinanced by European Regional
Development Fund (ERDF), "A way of making Europe."

\bibliographystyle{ieeetr}
\bibliography{main}

\end{document}


\title{Supplementary material\\Deep learning based domain adaptation for mitochondria segmentation on EM volumes}
\titlerunning{Deep learning based domain adaptation on EM volumes}

\newcommand*{\affaddr}[1]{#1} 
\newcommand*{\affmark}[1][*]{\textsuperscript{#1}}

\author{Daniel Franco-Barranco\affmark[1,2] \and Julio Pastor-Tronch\affmark[1] \and Aitor Gonzalez-Marfil\affmark[1] \and Arrate Muñoz-Barrutia\affmark[3,4] \and Ignacio Arganda-Carreras\affmark[1,2,5]  }

\authorrunning{Franco-Barranco {\em et al.}}
\institute{
\affaddr{\affmark[1] Dept. of Computer Science and Artificial Intelligence, University of the Basque Country (UPV/EHU)} \\
\affaddr{\affmark[2] Donostia International Physics Center (DIPC)} \\
\affaddr{\affmark[3] Universidad Carlos III de Madrid} \\
\affaddr{\affmark[4] Instituto de Investigación Sanitaria Gregorio Marañón} \\
\affaddr{\affmark[5] Ikerbasque, Basque Foundation for Science} \\
\email{daniel\_franco001@ehu.eus}
}

\maketitle   

\renewcommand{\thetable}{S\arabic{section}.\arabic{table}}
\renewcommand{\thefigure}{S\arabic{section}.\arabic{figure}}
\setcounter{table}{0}
\renewcommand{\thesection}{S\arabic{section}}

\section{Cross-dataset results}
\label{appendix:style_results}
To complete the overview of results shown in the manuscript, in this section, we show examples of histogram-matching and style-transfer image transformations, together with segmentation results of all cross-dataset experiments. Full-size images are shown for qualitative evaluation purposes.
\subsection{Source: Lucchi++ - Target: Kasthuri++}
\begin{figure}[ht]
\begin{minipage}{.5\linewidth}
\centering
\subfloat[]{
    \includegraphics[scale=.16]{img_sup_material/lucchi++/lucchi++_test_original}
    \label{fig:appendix-hm-st-Lucc-Kasth:lucchi}
    }

\end{minipage}%
\begin{minipage}{.5\linewidth}
\centering
\subfloat[]{\includegraphics[scale=.10]{img_sup_material/kasthuri++/kasthuri++_test_original}\label{fig:appendix-hm-st-Lucc-Kasth:kasthuri}}
\end{minipage}\par\medskip
\begin{minipage}{.5\linewidth}
\centering
\subfloat[]{\includegraphics[scale=.10]{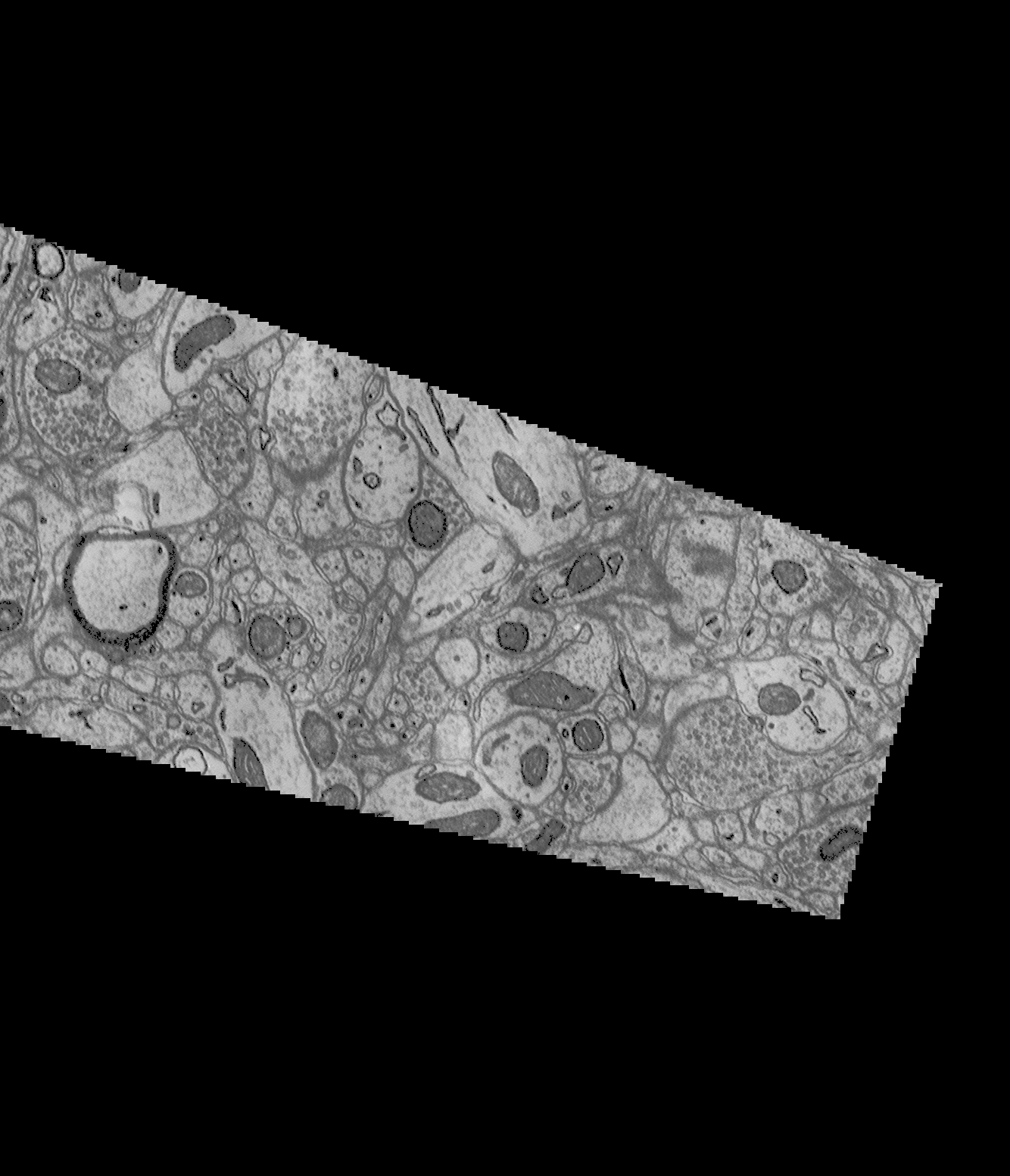}\label{fig:appendix-hm-st-Lucc-Kasth:hm}}
\end{minipage}
\begin{minipage}{.5\linewidth}
\centering
\subfloat[]{\includegraphics[scale=.10]{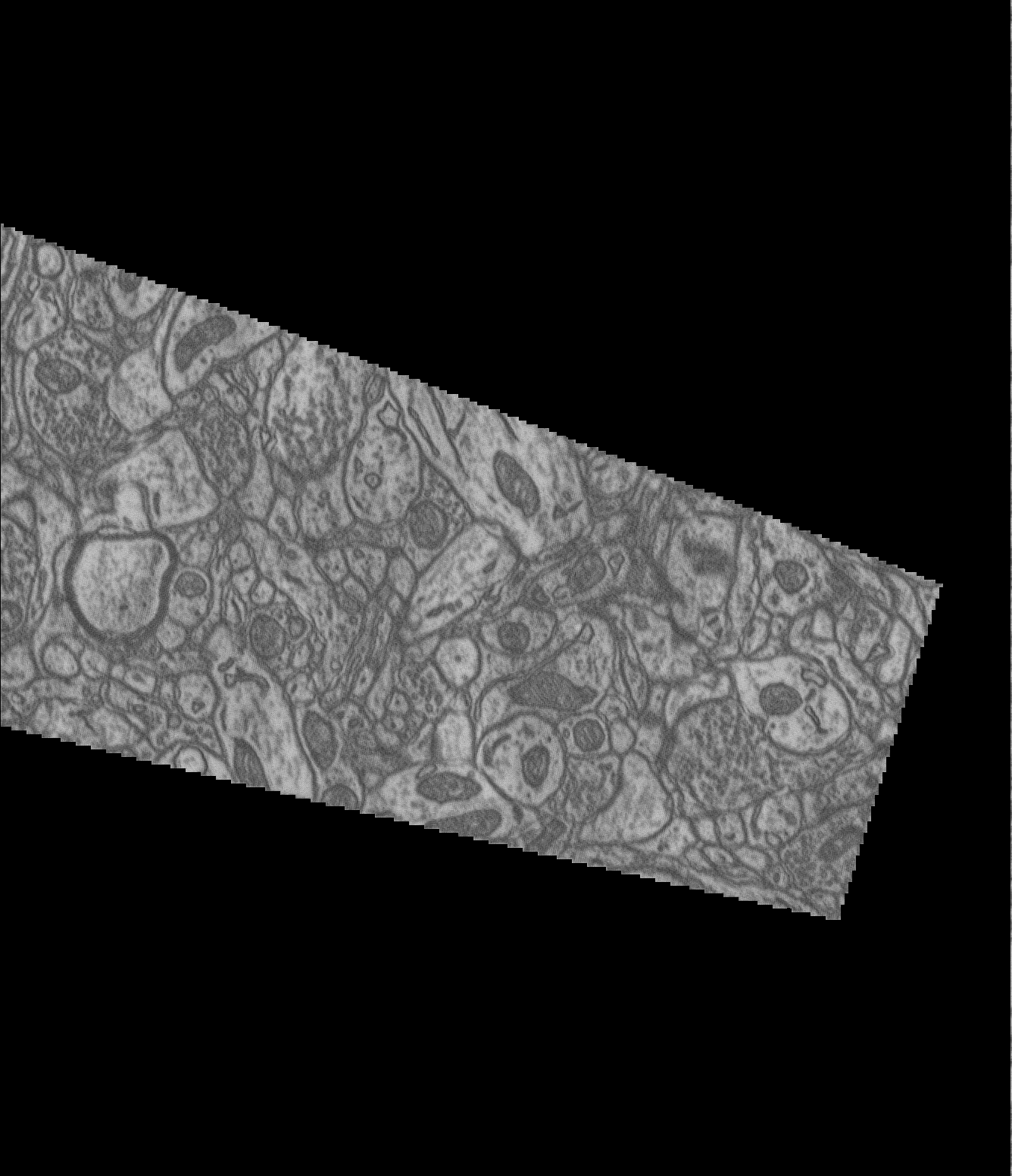}\label{fig:appendix-hm-st-Lucc-Kasth:style}}
\end{minipage}\par\medskip
\caption{Examples of histogram-matching and style-transfer results using Lucchi++ as reference histogram/style to transform Kasthuri++ images: (a) Lucchi++ dataset sample; (b) original Kasthuri++ test image; (c) histogram-matched version of (b); and (d) stylized version of (b).}
\label{fig:appendix-hm-st-Lucc-Kasth}
\end{figure}

The effect of our histogram-matching and style-transfer methods on an image from the Kasthuri++ dataset is shown in Figure~\ref{fig:appendix-hm-st-Lucc-Kasth} using the Lucchi++ dataset as the source domain. Remarkably, the domain shift in this source-target combination seems to be the smallest of all cases, and the histogram-matched images (see Figures~\ref{fig:appendix-hm-st-Lucc-Kasth:lucchi},~\ref{fig:appendix-hm-st-Lucc-Kasth:kasthuri}) appear to be very close to the source domain images.

\begin{figure}[ht]
\centering
\begin{minipage}{.4\linewidth}
\centering
\subfloat[]{\includegraphics[scale=.12]{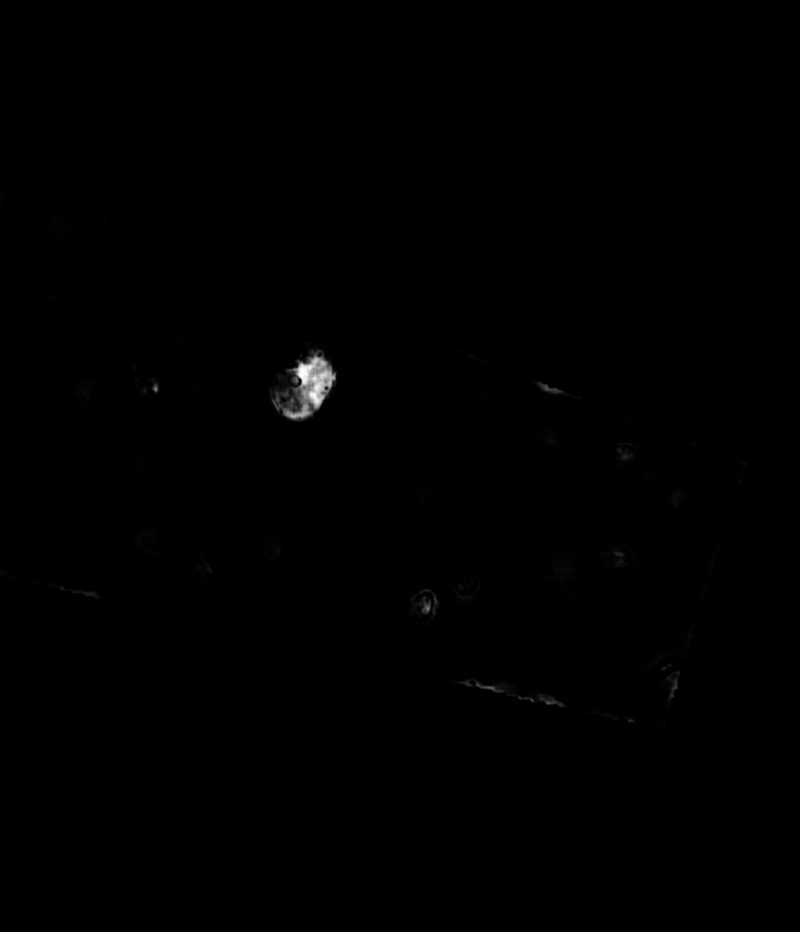}\label{fig:appendix-seg-Lucc-Kasth:baseline}}
\end{minipage}
\begin{minipage}{.4\linewidth}
\centering
\subfloat[]{\includegraphics[scale=.07]{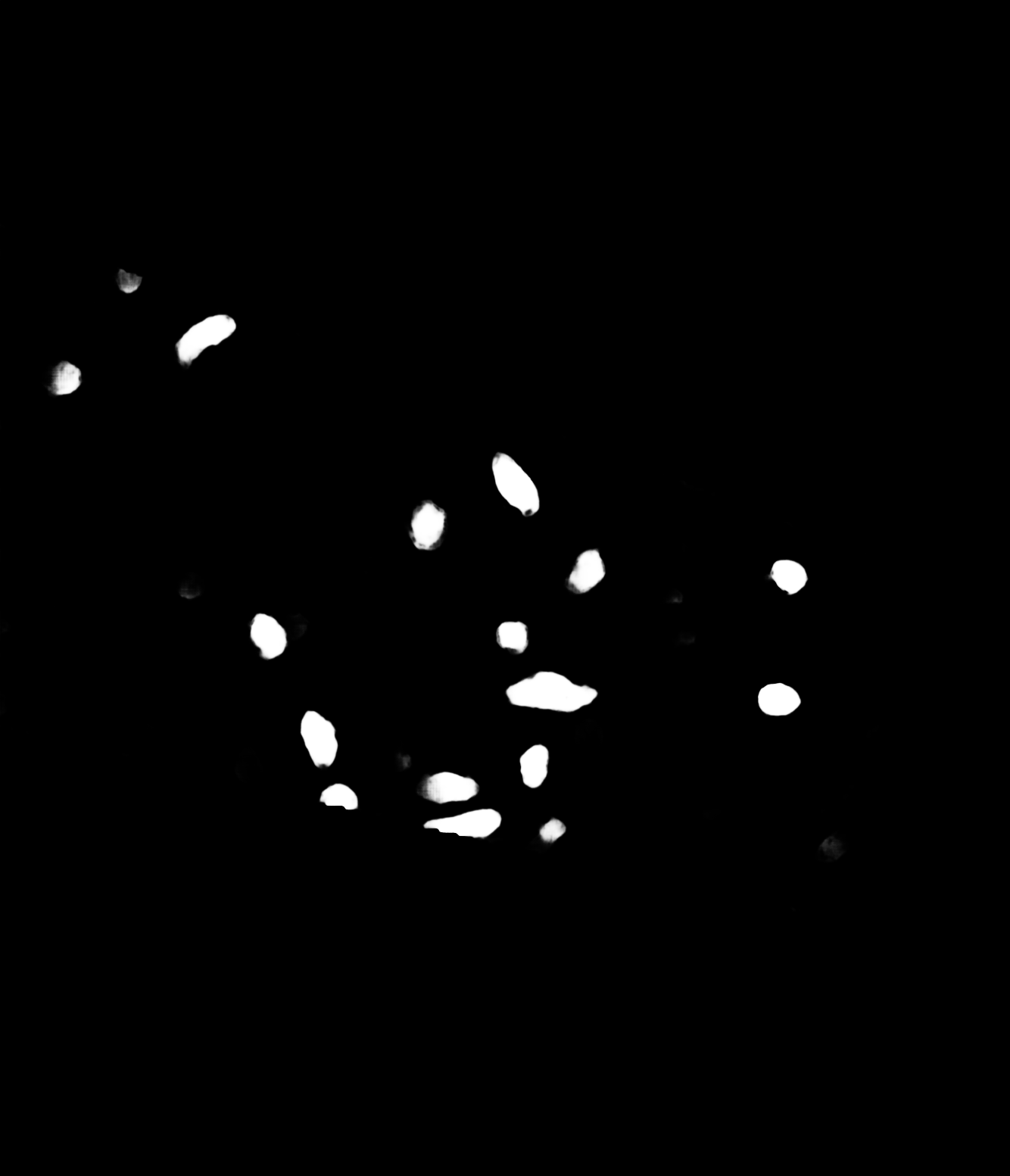}\label{fig:appendix-seg-Lucc-Kasth:hm_baseline}}
\end{minipage}\par\medskip
\begin{minipage}{.4\linewidth}
\centering
\subfloat[]{\includegraphics[scale=.07]{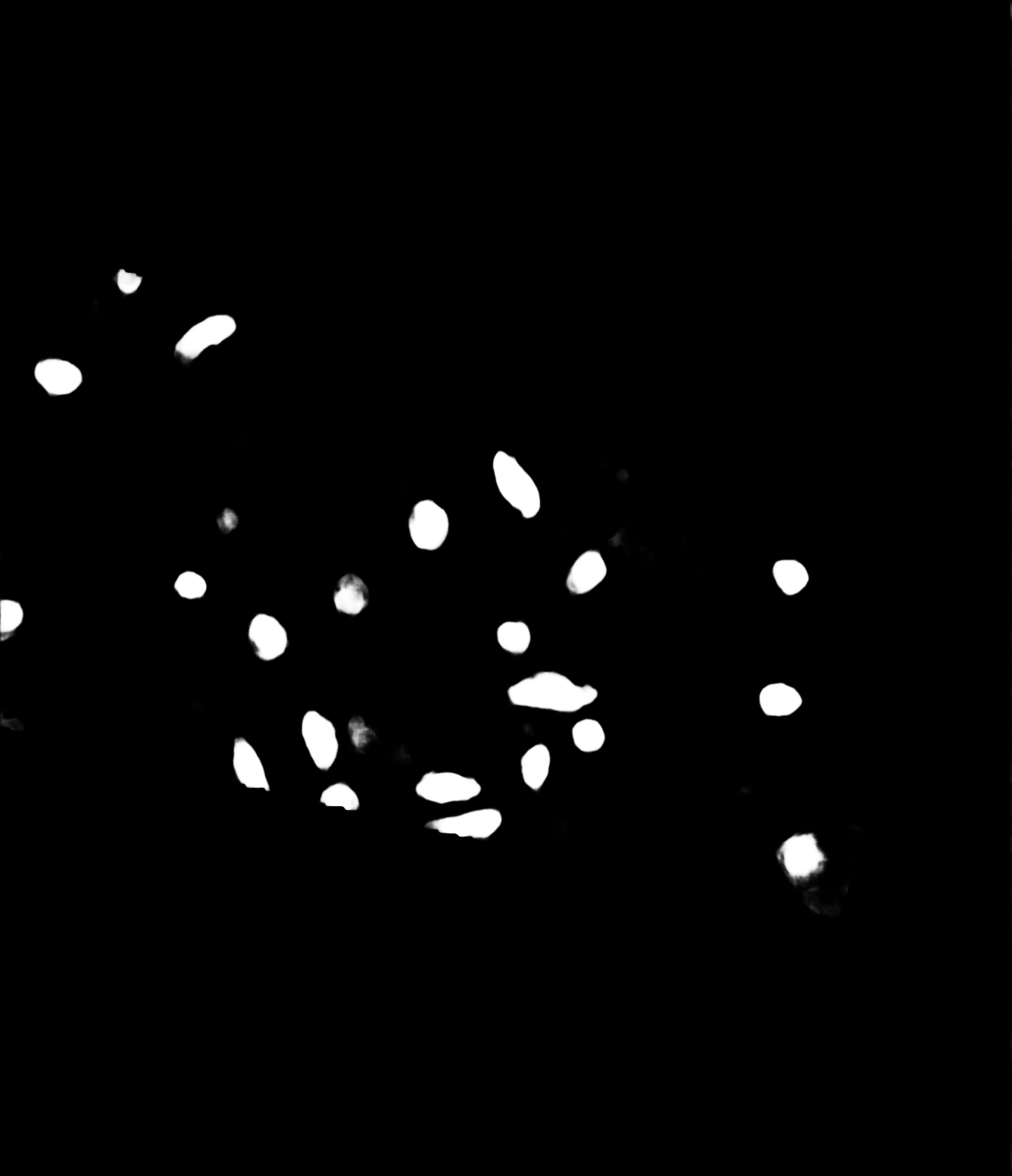}\label{fig:appendix-seg-Lucc-Kasth:style}}
\end{minipage}
\begin{minipage}{.4\linewidth}
\centering
\subfloat[]{\includegraphics[scale=.07]{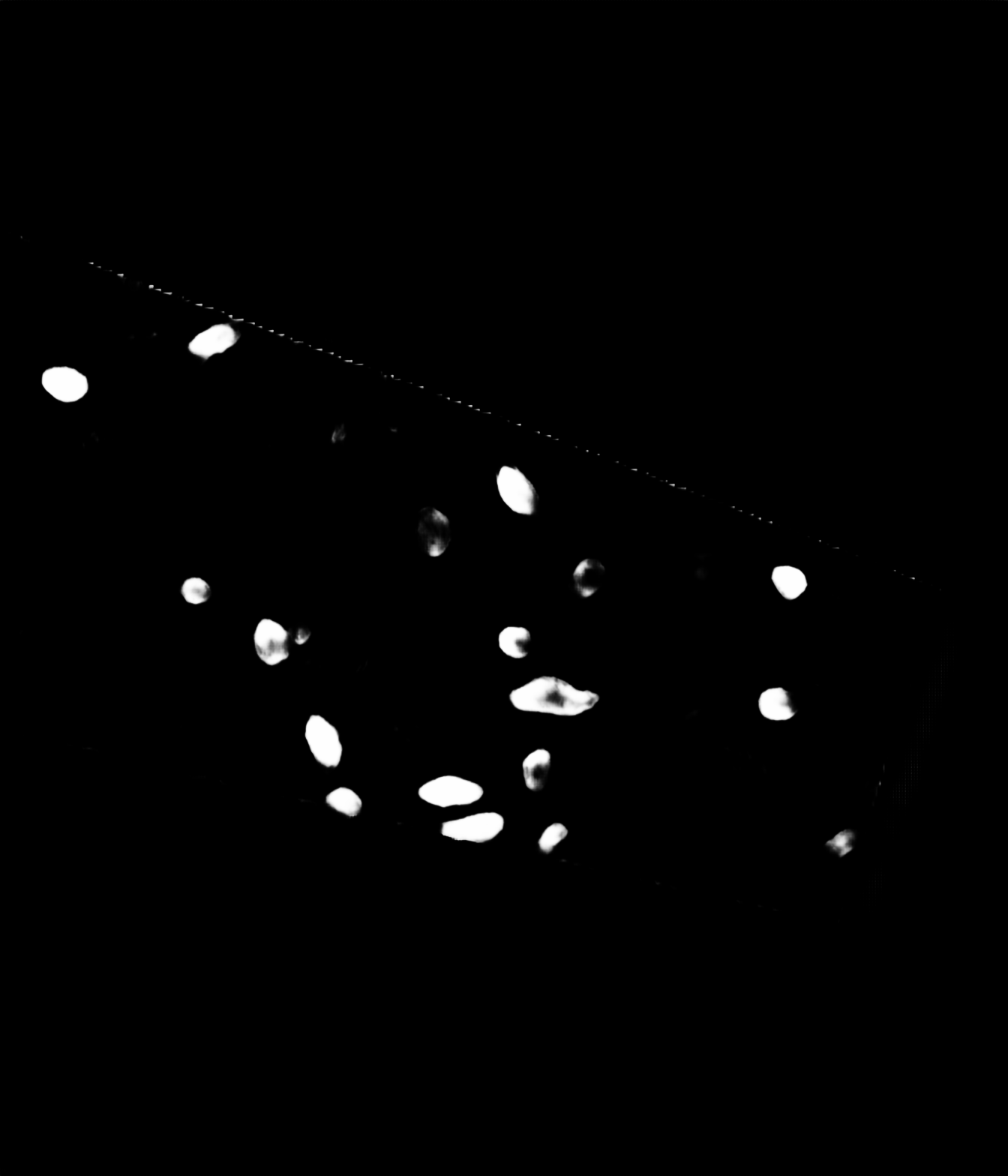}\label{fig:appendix-seg-Lucc-Kasth:ssl}}
\end{minipage}\par\medskip
\begin{minipage}{.4\linewidth}
\centering
\subfloat[]{\includegraphics[scale=.07]{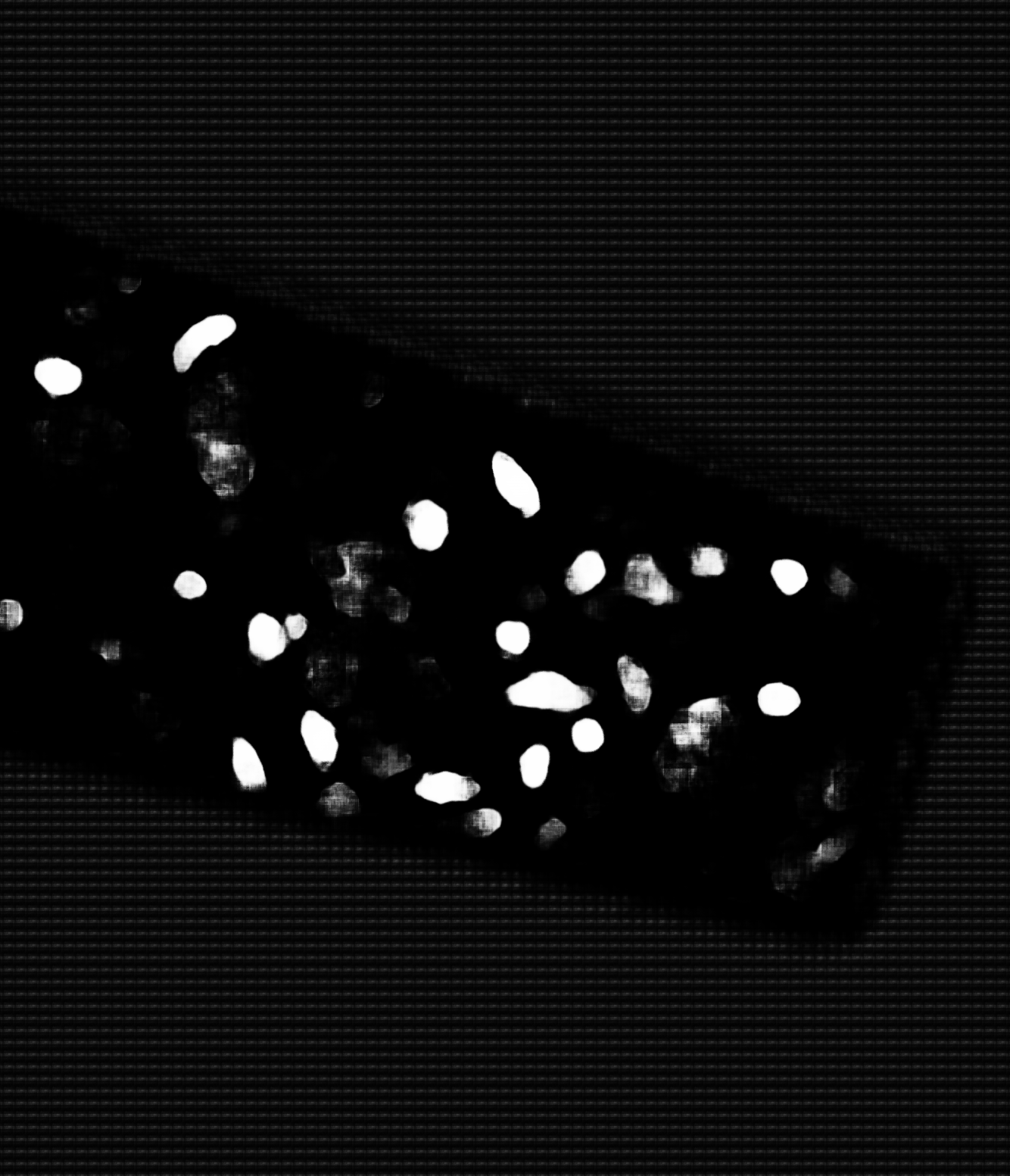}\label{fig:appendix-seg-Lucc-Kasth:attYNet}}
\end{minipage}
\begin{minipage}{.4\linewidth}
\centering
\subfloat[]{\includegraphics[scale=.07]{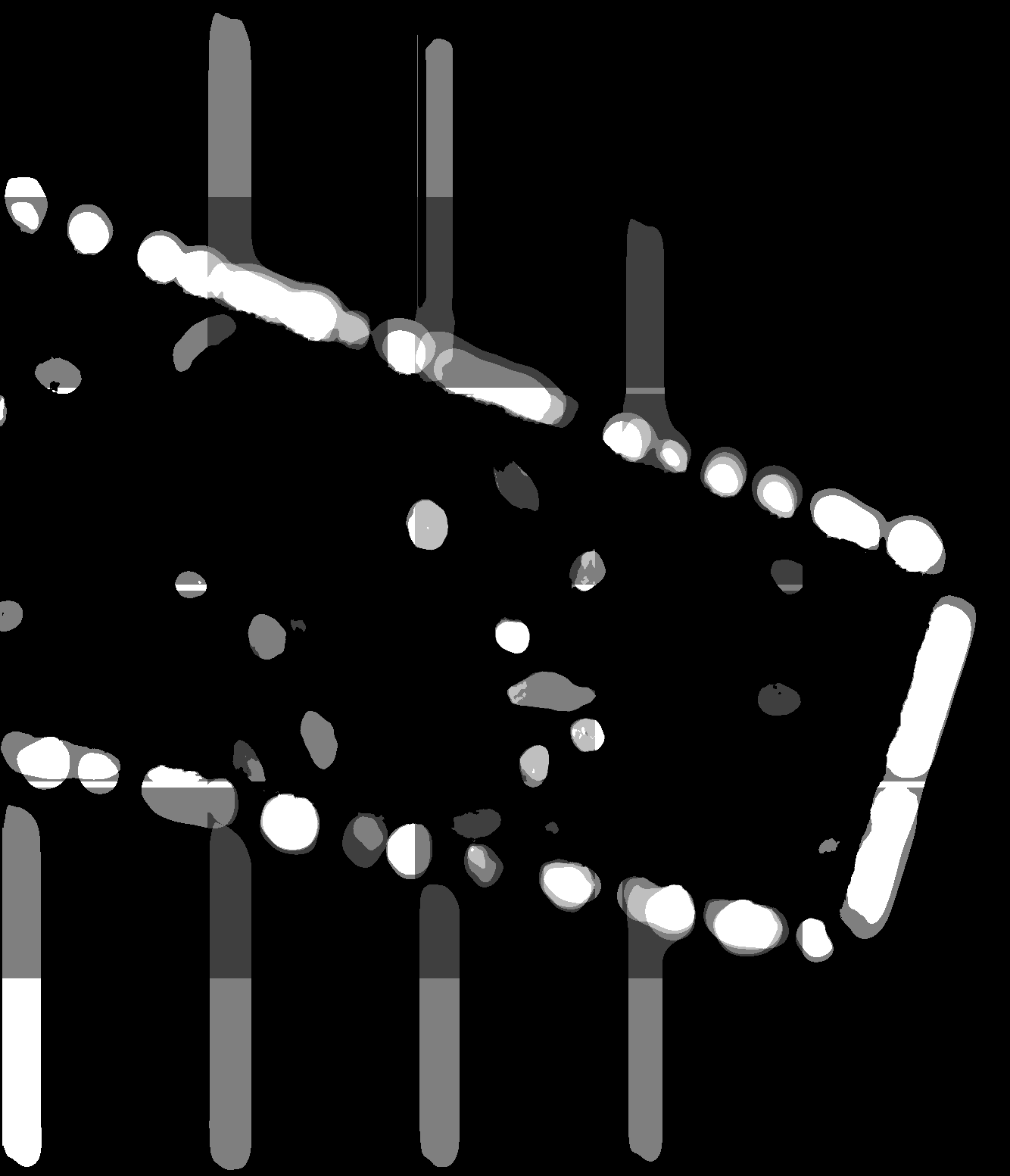}\label{fig:appendix-seg-Lucc-Kasth:damtnet}}
\end{minipage}\par\medskip
\begin{minipage}{.4\linewidth}
\centering
\subfloat[]{\includegraphics[scale=.07]{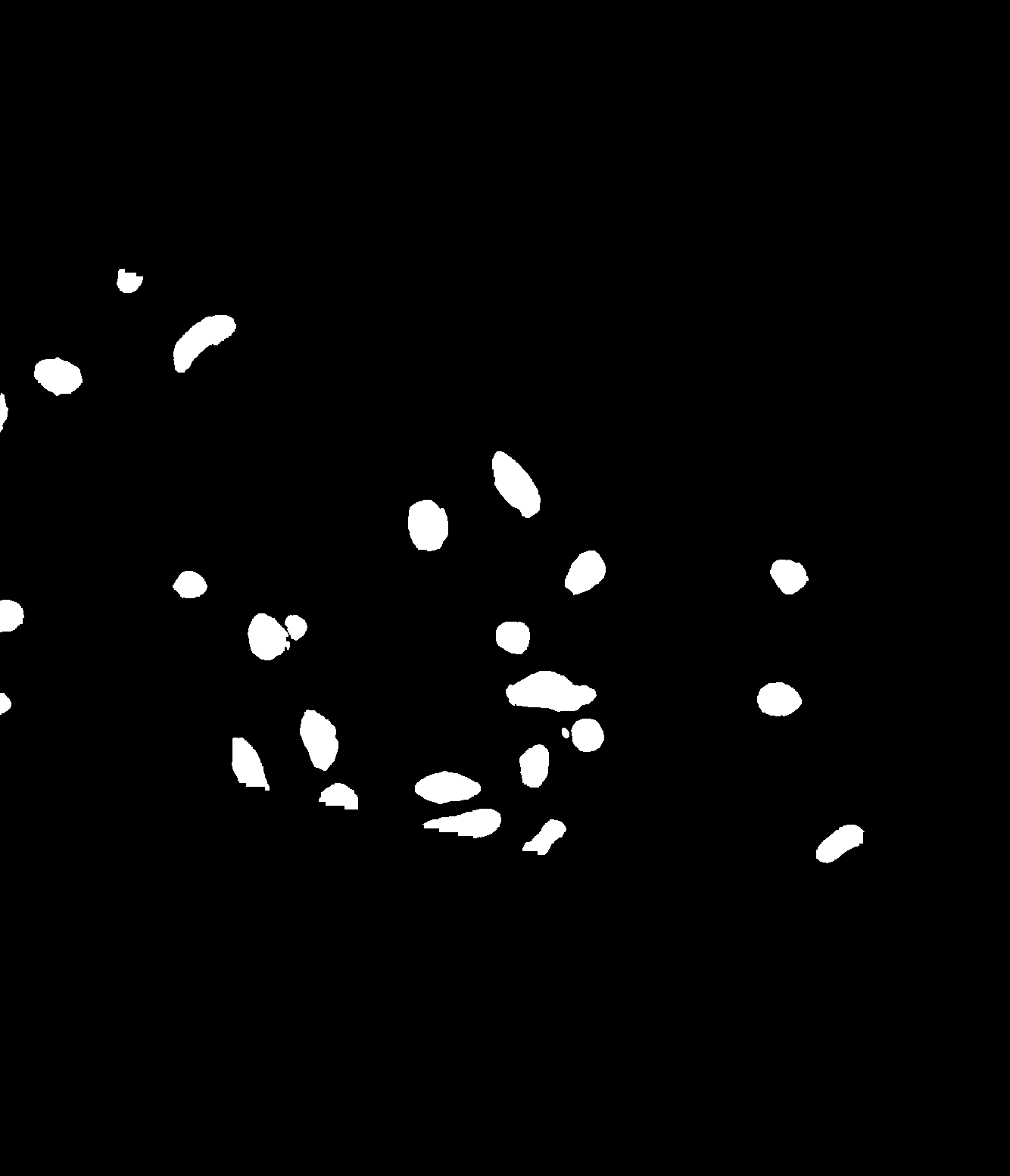}\label{fig:appendix-seg-Lucc-Kasth:gt}}
\end{minipage}
\begin{minipage}{.4\linewidth}
\centering
\subfloat[]{\includegraphics[scale=.07]{img_sup_material/kasthuri++/kasthuri++_test_original}\label{fig:appendix-seg-Lucc-Kasth:original}}
\end{minipage}\par\medskip
\caption{Examples of semantic segmentation results using Lucchi++ as source and Kasthuri++ as the target. The resulting mitochondria probability maps are shown for: (a) the baseline method (no adaptation); (b) the baseline method applied to the histogram-matched images; our (c) style-transfer, (d) self-supervised learning, and (e) Attention Y-Net approaches; and (f) the DAMT-Net method; together with the corresponding (g) ground truth and (h) original test sample from Kasthuri++.}
\label{fig:appendix-seg-Lucc-Kasth}
\end{figure}

The mitochondria probability maps produced by all our tested methods on the first test image from Kasthuri++ are shown in Figure~\ref{fig:appendix-seg-Lucc-Kasth} together with its corresponding ground-truth binary labels and original EM image. The best qualitative results seem to be produced by the histogram-matching and style-transfer approaches (see Figures~\ref{fig:appendix-seg-Lucc-Kasth:hm_baseline},~\ref{fig:appendix-seg-Lucc-Kasth:style}), while the state-of-the-art DAMT-Net method struggles to produce compact mitochondria masks and presents border artifacts due to the zero-padding of the Kasthuri++ dataset (see Figure~\ref{fig:appendix-seg-Lucc-Kasth:damtnet}). Notice that the displayed results for the style-transfer, SSL, Attention Y-Net, and DAMT-Net approaches correspond to executions using our proposed stop criterion (solidity, see Section 4.3).

\subsection{Source: Lucchi++ - Target: VNC}
\begin{figure}[ht]
\centering
\begin{minipage}{.5\linewidth}
\centering
\subfloat[]{\includegraphics[scale=.14]{img_sup_material/lucchi++/lucchi++_test_original}\label{fig:appendix-hm-st-Lucc-VNC:lucchi}}
\end{minipage}%
\begin{minipage}{.5\linewidth}
\centering
\subfloat[]{\includegraphics[scale=.14]{img_sup_material/vnc/vnc_test_original}\label{fig:appendix-hm-st-Lucc-VNC:vnc}}
\end{minipage}\par\medskip
\begin{minipage}{.49\linewidth}
\centering
\subfloat[]{\includegraphics[scale=.14]{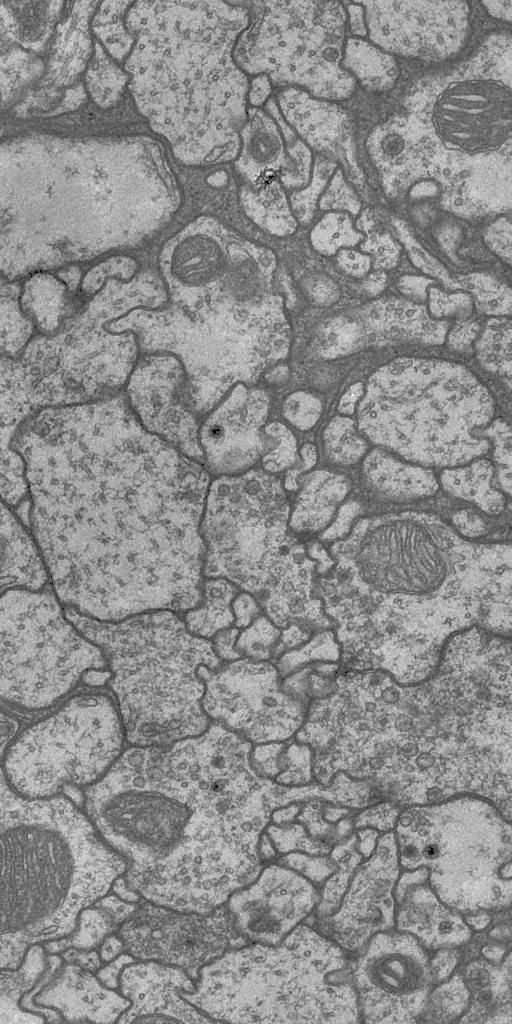}\label{fig:appendix-hm-st-Lucc-VNC:hm}}
\end{minipage}
\begin{minipage}{.49\linewidth}
\centering
\subfloat[]{\includegraphics[scale=.14]{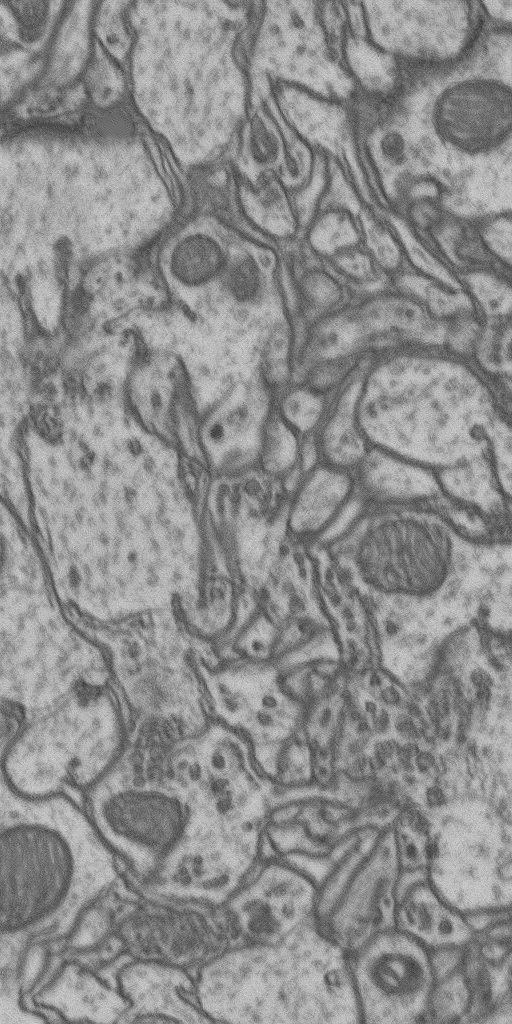}\label{fig:appendix-hm-st-Lucc-VNC:style}}
\end{minipage}\par\medskip
\caption{Examples of histogram-matching and style-transfer results using Lucchi++ as reference histogram/style to transform VNC images: (a) Lucchi++ dataset sample; (b) original VNC test image; (c) histogram-matched version of (b); and (d) stylized version of (b).}
\label{fig:appendix-hm-st-Lucc-VNC}
\end{figure}

The effect of our histogram-matching and style-transfer methods on an image from the VNC dataset is shown in Figure~\ref{fig:appendix-hm-st-Lucc-VNC} using the Lucchi++ dataset as the source domain. The domain shift in this source-target combination seems much larger than in the previous case, and the histogram-matched images (see Figures~\ref{fig:appendix-hm-st-Lucc-VNC:lucchi},~\ref{fig:appendix-hm-st-Lucc-VNC:hm}) appear to be further away from the source domain images than the stylized ones (see Figure~\ref{fig:appendix-hm-st-Lucc-VNC:style}). In particular, the style-transfer method successfully adapted the texture of the neural process from one domain (ssTEM) to the other (FIB-SEM).

\begin{figure}[ht]
\centering
\begin{minipage}{.4\linewidth}
\centering
\subfloat[]{\includegraphics[scale=.12]{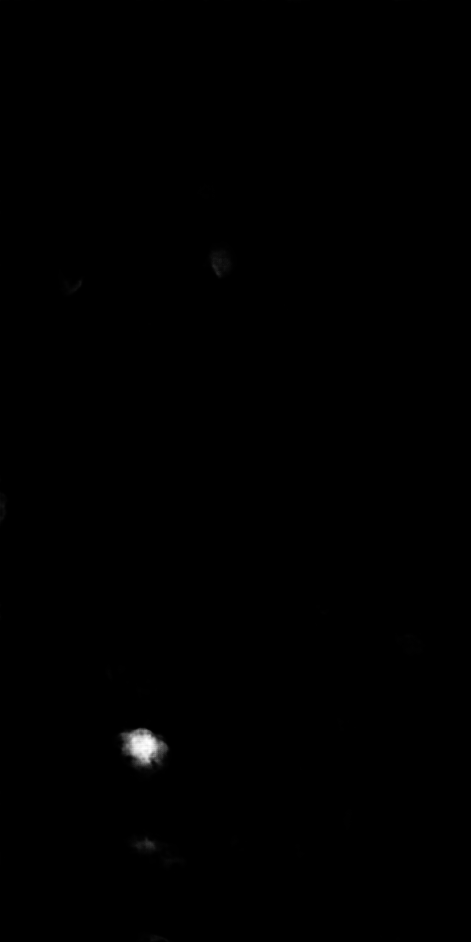}\label{fig:appendix-seg-Lucc-VNC:baseline}}
\end{minipage}%
\begin{minipage}{.4\linewidth}
\centering
\subfloat[]{\includegraphics[scale=.11]{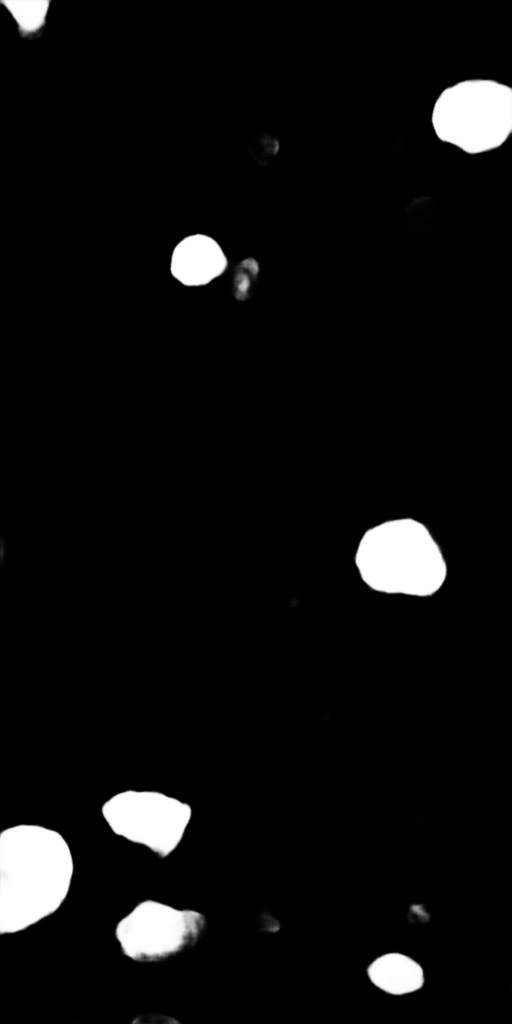}\label{fig:appendix-seg-Lucc-VNC:hm_baseline}}
\end{minipage}\par\medskip
\begin{minipage}{.4\linewidth}
\centering
\subfloat[]{\includegraphics[scale=.11]{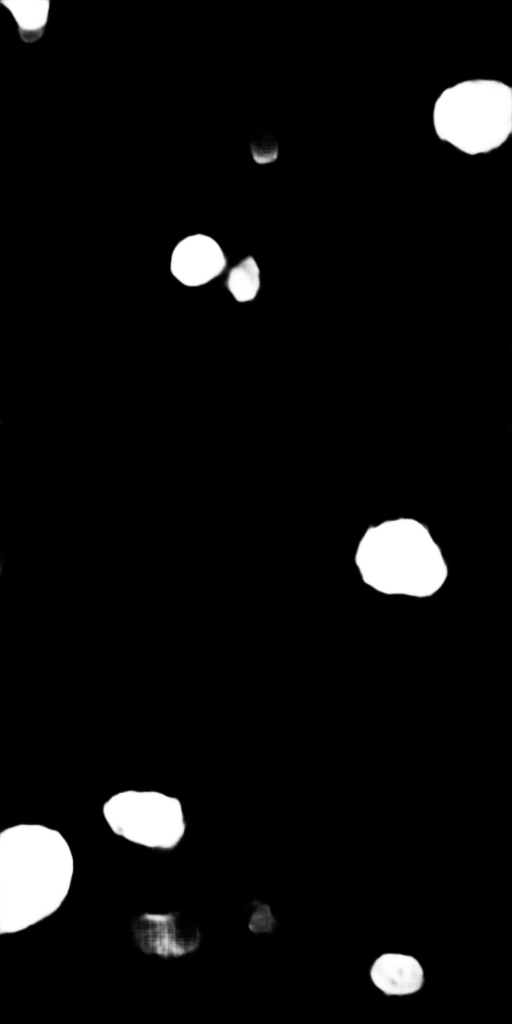}\label{fig:appendix-seg-Lucc-VNC:style}}
\end{minipage}%
\begin{minipage}{.4\linewidth}
\centering
\subfloat[]{\includegraphics[scale=.11]{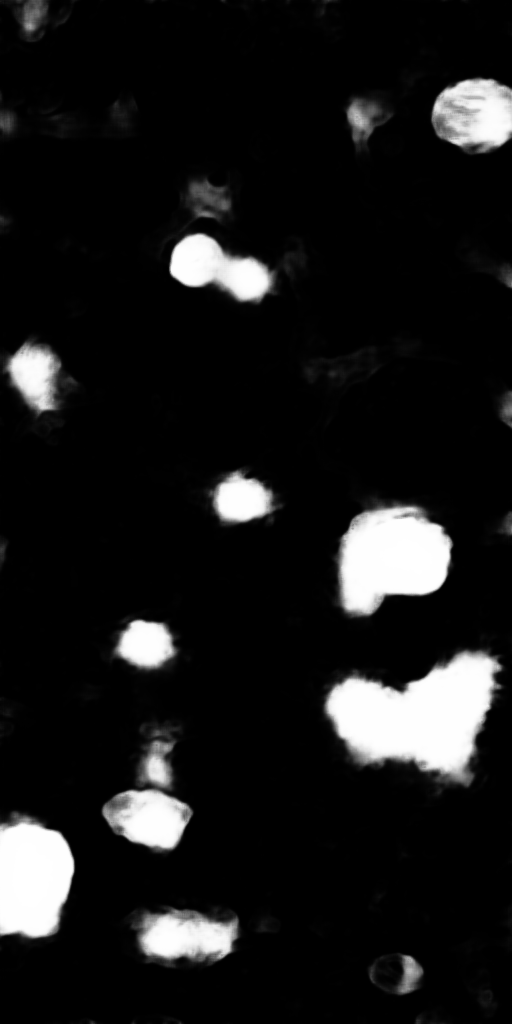}\label{fig:appendix-seg-Lucc-VNC:ssl}}
\end{minipage}\par\medskip
\begin{minipage}{.4\linewidth}
\centering
\subfloat[]{\includegraphics[scale=.11]{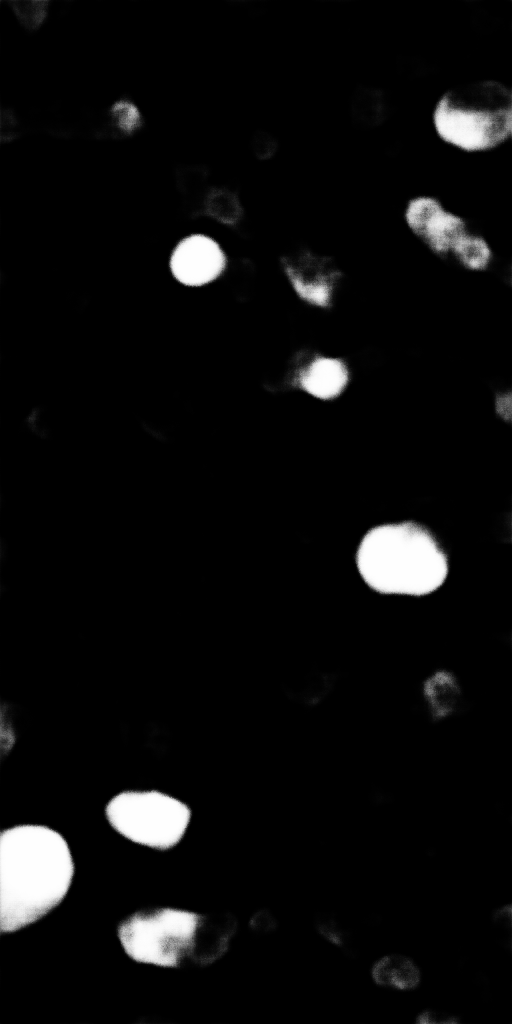}\label{fig:appendix-seg-Lucc-VNC:attynet}}
\end{minipage}%
\begin{minipage}{.4\linewidth}
\centering
\subfloat[]{\includegraphics[scale=.11]{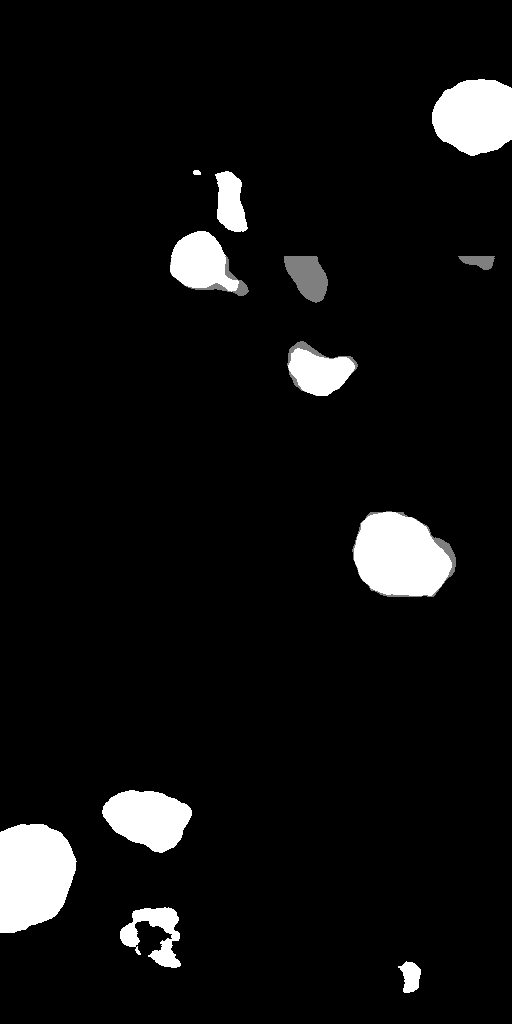}\label{fig:appendix-seg-Lucc-VNC:damtnet}}
\end{minipage}\par\medskip
\begin{minipage}{.4\linewidth}
\centering
\subfloat[]{\includegraphics[scale=.11]{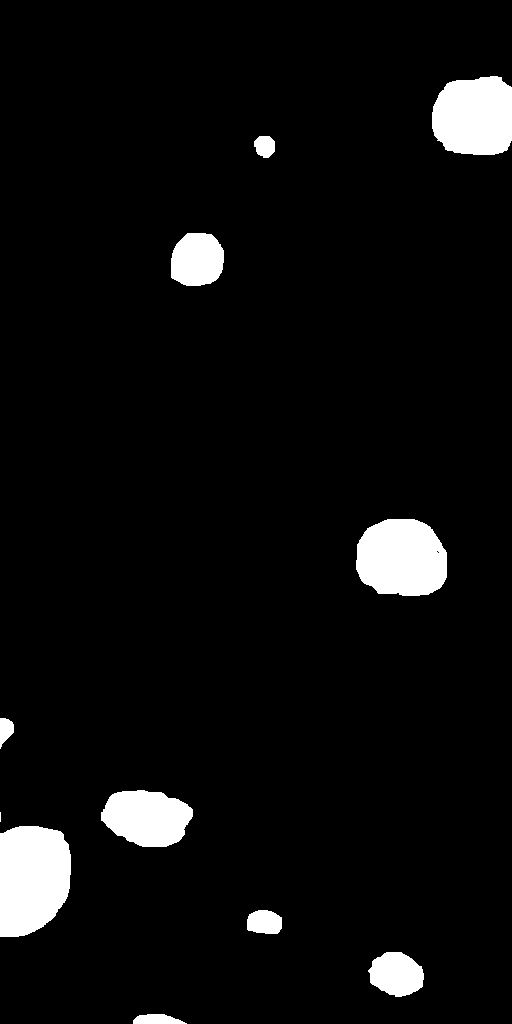}\label{fig:appendix-seg-Lucc-VNC:gt}}
\end{minipage}
\begin{minipage}{.4\linewidth}
\centering
\subfloat[]{\includegraphics[scale=.11]{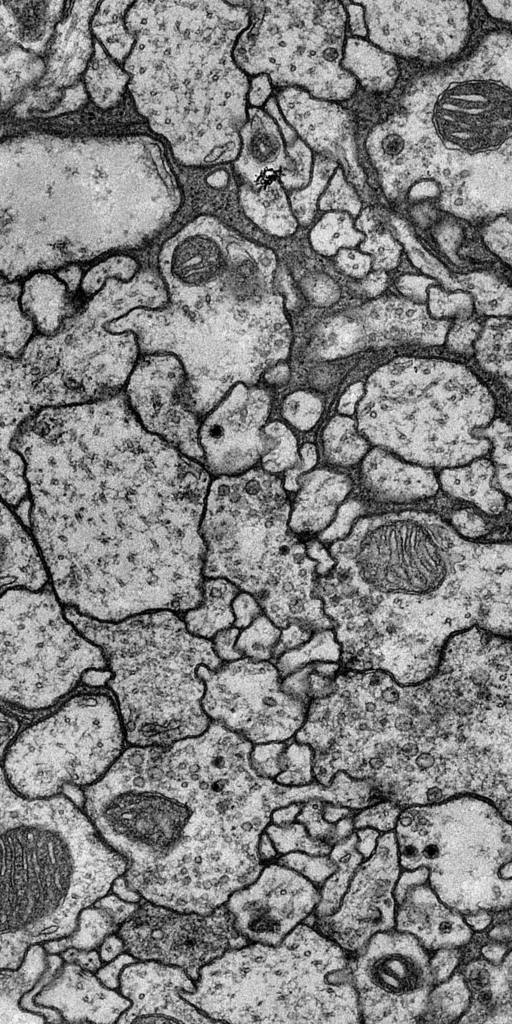}\label{fig:appendix-seg-Lucc-VNC:original}}
\end{minipage}\par\medskip

\caption{Examples of semantic segmentation results using Lucchi++ as the source and VNC as the target. The resulting mitochondria probability maps are shown for: (a) the baseline method (no adaptation); (b) the baseline method applied to the histogram-matched images; our (c) style-transfer, (d) self-supervised learning, and (e) Attention Y-Net approaches; and (f) the DAMT-Net method; together with the corresponding (g) ground truth and (h) original test sample from VNC.}
\label{fig:appendix-seg-Lucc-VNC}
\end{figure}

The mitochondria probability maps produced by all our tested methods on the first test image from VNC are shown in Figure~\ref{fig:appendix-seg-Lucc-VNC} together with its corresponding ground-truth binary labels and original EM image. Although all methods seem to approximate the location of mitochondria correctly, the best qualitative results seem to be produced by our style-transfer approach (see Figure~\ref{fig:appendix-seg-Lucc-VNC:style}). In this case, the state-of-the-art DAMT-Net method seems to produce under-segmented results (see Figure~\ref{fig:appendix-seg-Lucc-VNC:damtnet}) while our SSL and Attention Y-Net (see Figures~\ref{fig:appendix-seg-Lucc-VNC:ssl},~\ref{fig:appendix-seg-Lucc-VNC:attynet}) methods output over-segmented masks. Notice that the displayed results for the style-transfer, SSL, Attention Y-Net, and DAMT-Net approaches correspond to executions using our proposed stop criterion (solidity, see Section 4.3).

\subsection{Source: Kasthuri++ - Target: Lucchi++}

\begin{figure}[ht]
\begin{minipage}{.5\linewidth}
\centering
\subfloat[]{\includegraphics[scale=.12]{img_sup_material/kasthuri++/kasthuri++_test_original}\label{fig:appendix-ht-st-Kasth-Lucc:kasthuri}}
\end{minipage}%
\begin{minipage}{.5\linewidth}
\centering
\subfloat[]{\includegraphics[scale=.16]{img_sup_material/lucchi++/lucchi++_test_original}\label{fig:appendix-ht-st-Kasth-Lucc:lucchi}}
\end{minipage}\par\medskip
\begin{minipage}{.5\linewidth}
\centering
\subfloat[]{\includegraphics[scale=.16]{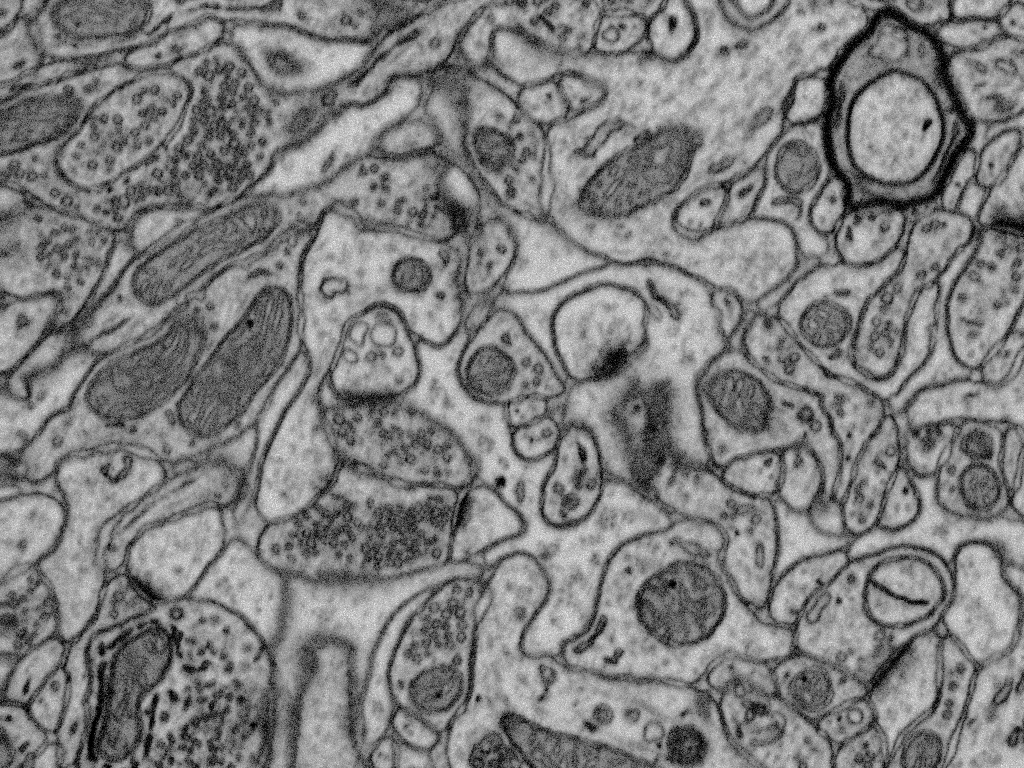}\label{fig:appendix-ht-st-Kasth-Lucc:hm}}
\end{minipage}%
\begin{minipage}{.5\linewidth}
\centering
\subfloat[]{\includegraphics[scale=.16]{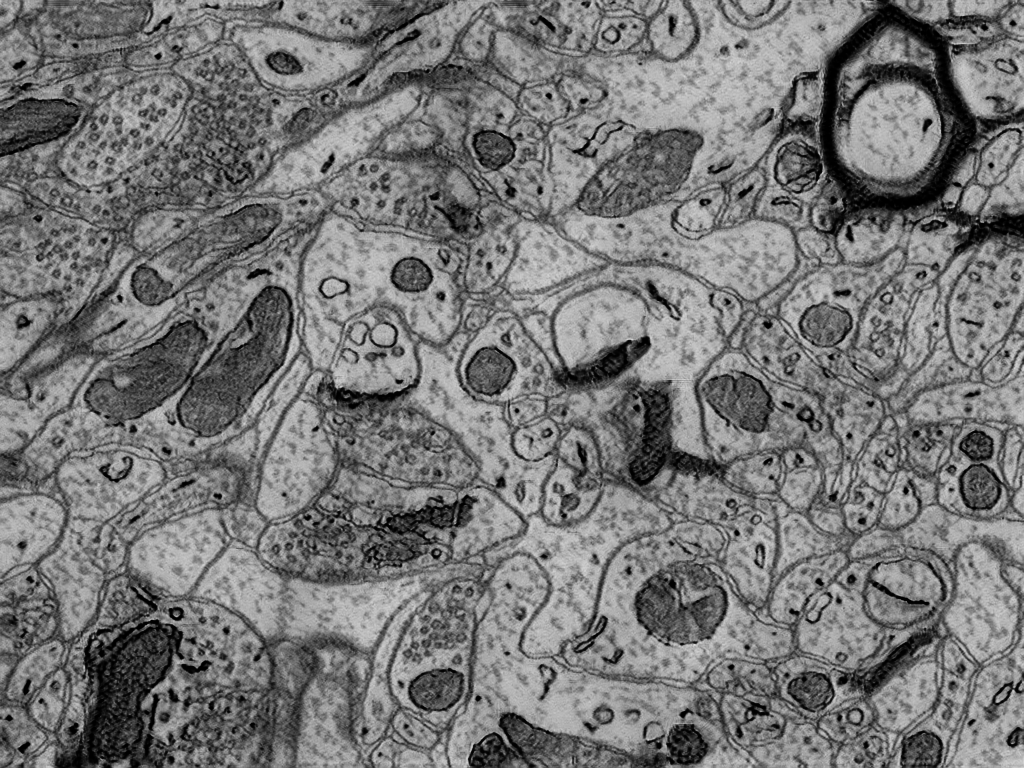}\label{fig:appendix-ht-st-Kasth-Lucc:style}}
\end{minipage}%
\caption{Examples of histogram-matching and style-transfer results using Kasthuri++ as reference histogram/style to transform Lucchi++ images: (a) Kasthuri++ dataset sample; (b) original Lucchi++ test image; (c) histogram-matched version of (b); and (d) stylized version of (b).}
\label{fig:appendix-ht-st-Kasth-Lucc}
\end{figure}

The effect of our histogram-matching and style-transfer methods on an image from the Lucchi++ dataset is shown in Figure~\ref{fig:appendix-ht-st-Kasth-Lucc} using the Kasthuri++ dataset as the source domain. The domain shift in this source-target combination seems larger than in the opposite combination, where the histogram matching obtained excellent results. As in the previous case, here, the histogram-matched images (see Figures~\ref{fig:appendix-ht-st-Kasth-Lucc:kasthuri},~\ref{fig:appendix-ht-st-Kasth-Lucc:hm}) appear to be far away from the source domain images. However, the style-transfer method (see Figure~\ref{fig:appendix-ht-st-Kasth-Lucc:style}) has managed to successfully capture the texture of both the neural process and organelles from one domain (FIB-SEM) to the other (ssEM).

\begin{figure}[ht]
\centering
\begin{minipage}{.5\linewidth}
\centering
\subfloat[]{\includegraphics[scale=.085]{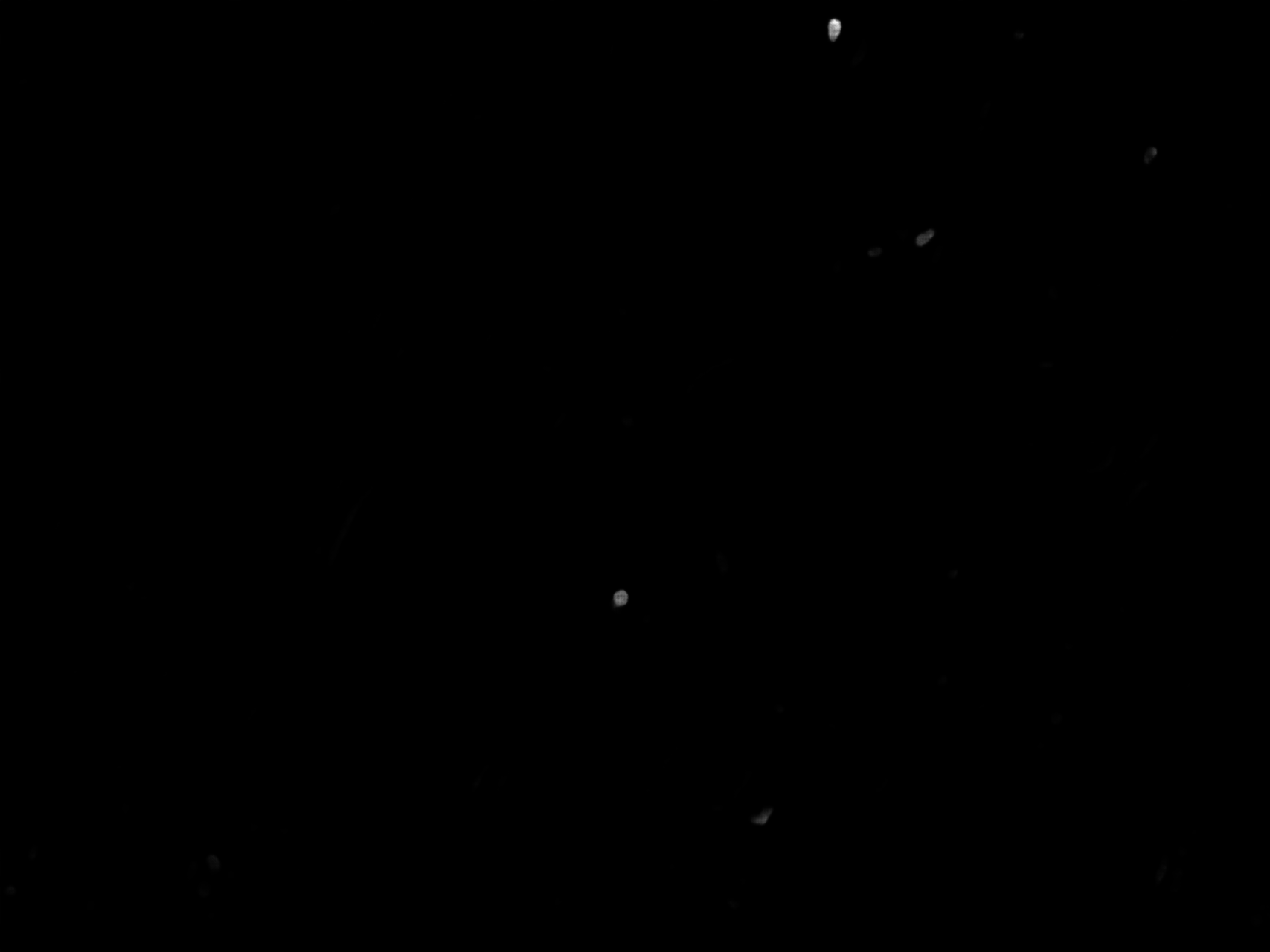}\label{fig:appendix-seg-Kasth-Lucc:baseline}}
\end{minipage}%
\begin{minipage}{.5\linewidth}
\centering
\subfloat[]{\includegraphics[scale=.14]{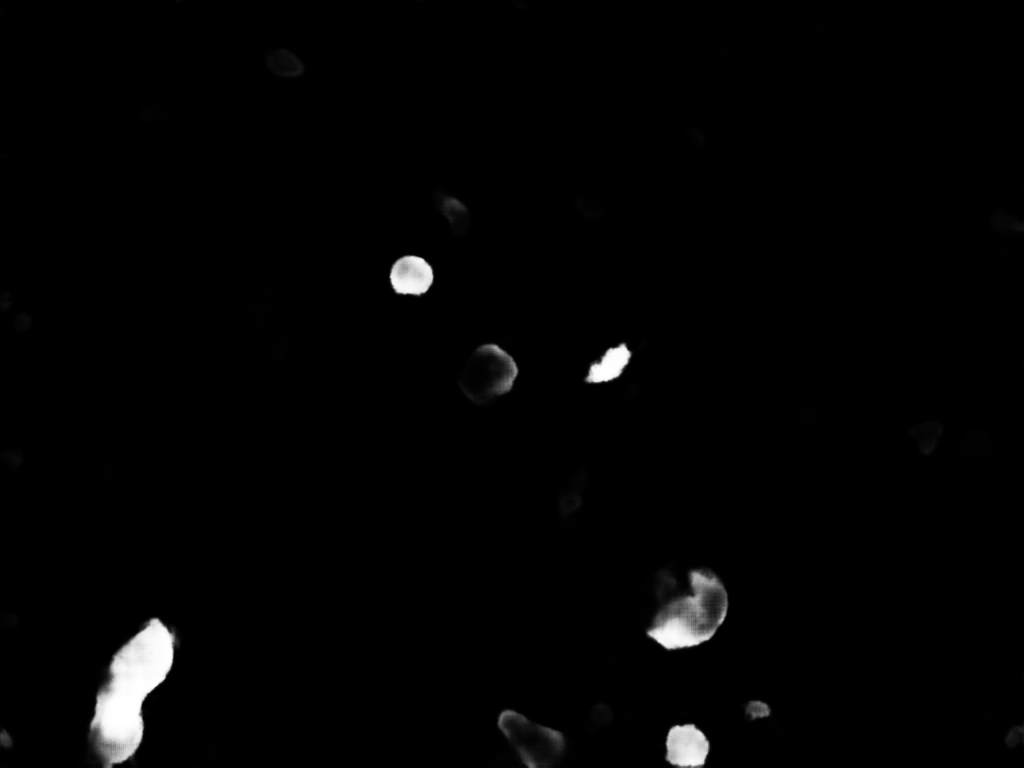}\label{fig:appendix-seg-Kasth-Lucc:hm_baseline}}
\end{minipage}\par\medskip
\begin{minipage}{.5\linewidth}
\centering
\subfloat[]{\includegraphics[scale=.14]{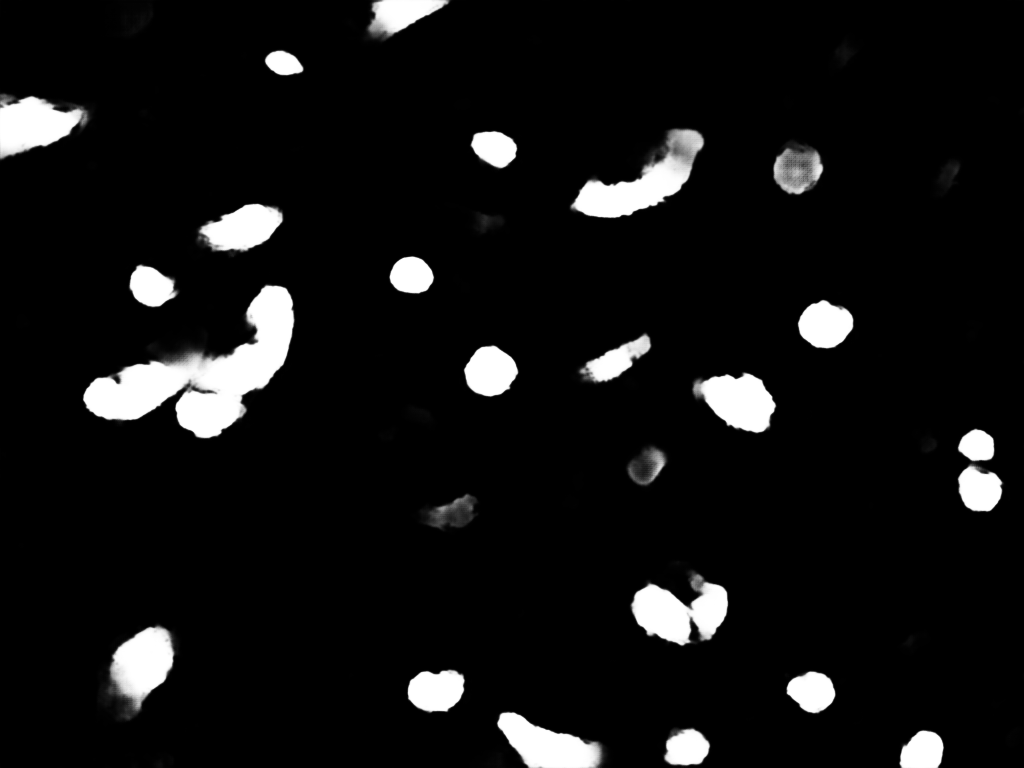}\label{fig:appendix-seg-Kasth-Lucc:style}}
\end{minipage}%
\begin{minipage}{.5\linewidth}
\centering
\subfloat[]{\includegraphics[scale=.14]{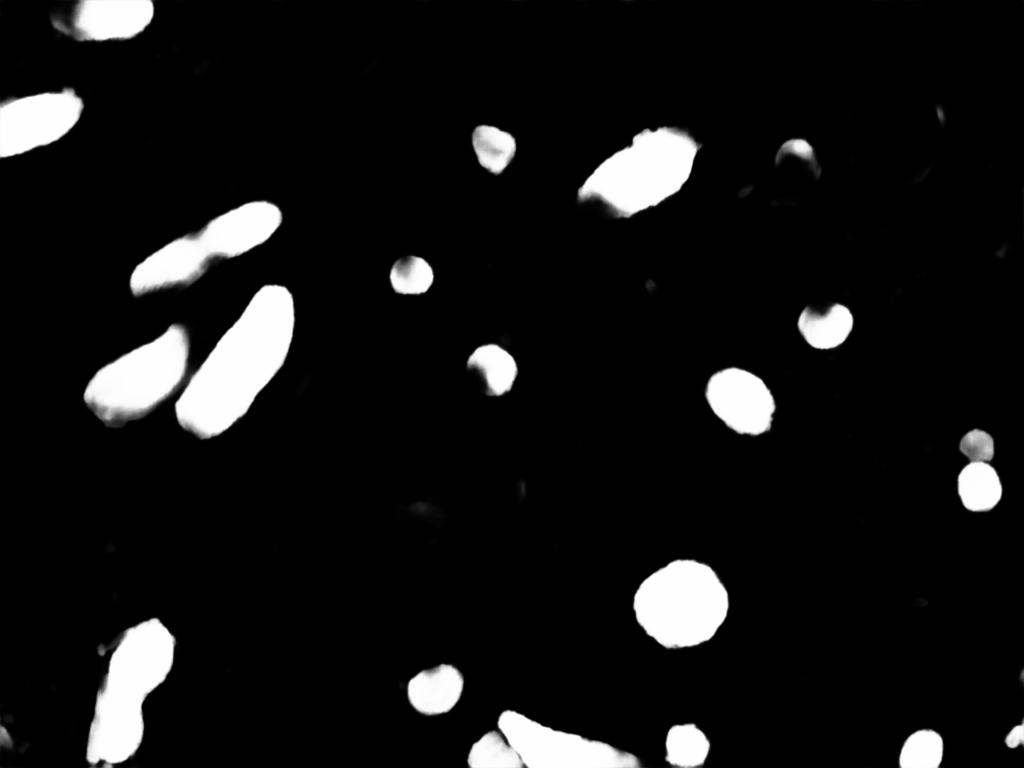}\label{fig:appendix-seg-Kasth-Lucc:ssl}}
\end{minipage}\par\medskip
\begin{minipage}{.5\linewidth}
\centering
\subfloat[]{\includegraphics[scale=.14]{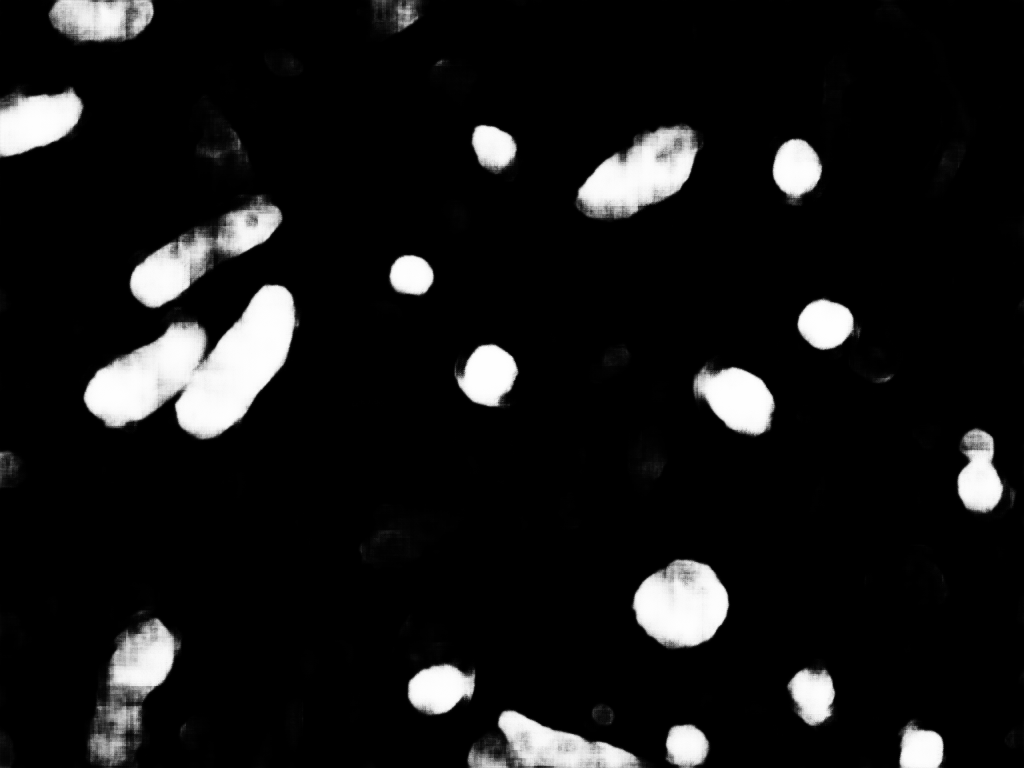}\label{fig:appendix-seg-Kasth-Lucc:attynet}}
\end{minipage}%
\begin{minipage}{.5\linewidth}
\centering
\subfloat[]{\includegraphics[scale=.14]{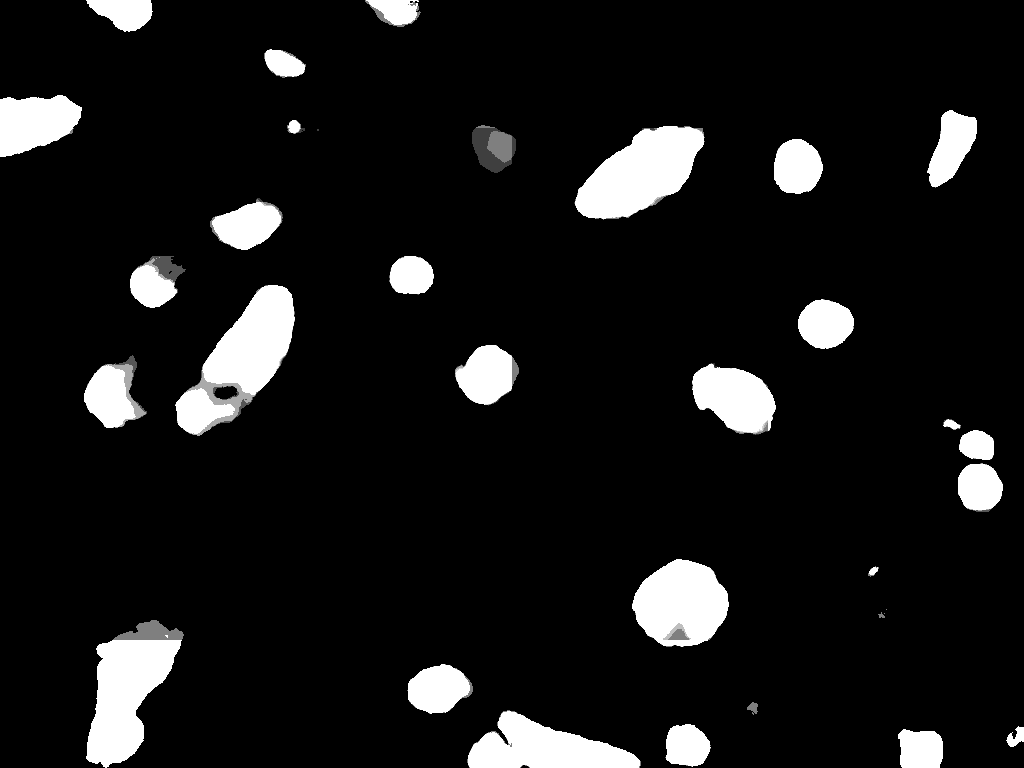}\label{fig:appendix-seg-Kasth-Lucc:damtnet}}
\end{minipage}\par\medskip
\begin{minipage}{.49\linewidth}
\centering
\subfloat[]{\includegraphics[scale=.14]{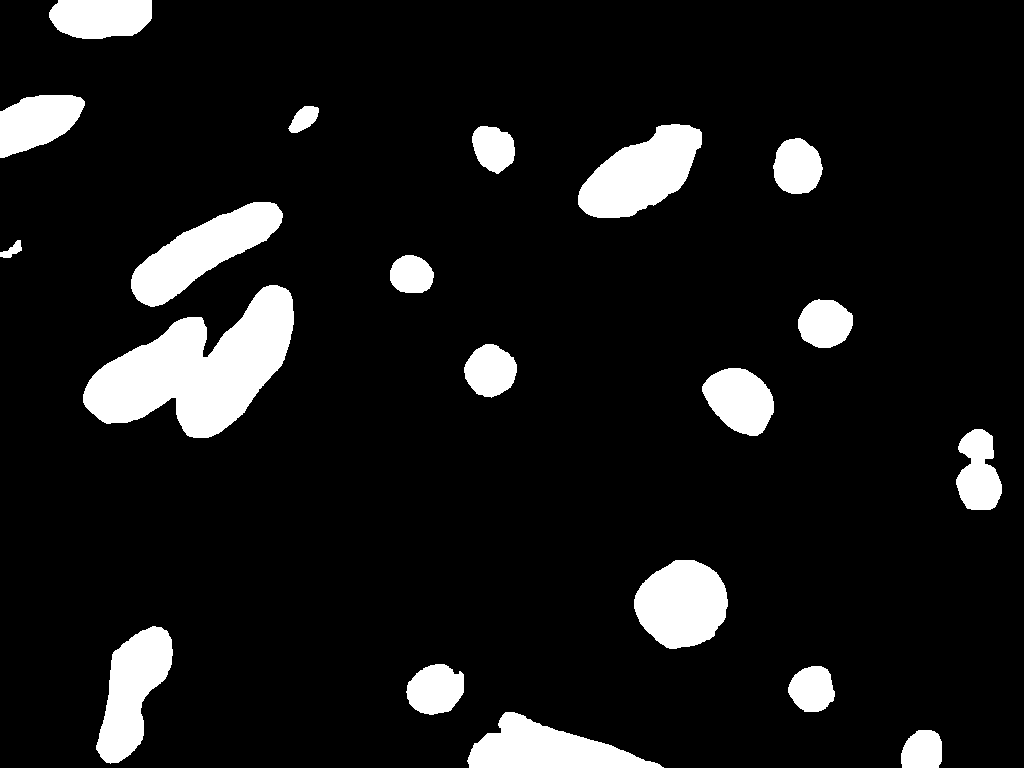}\label{fig:appendix-seg-Kasth-Lucc:gt}}
\end{minipage}
\begin{minipage}{.49\linewidth}
\centering
\subfloat[]{\includegraphics[scale=.14]{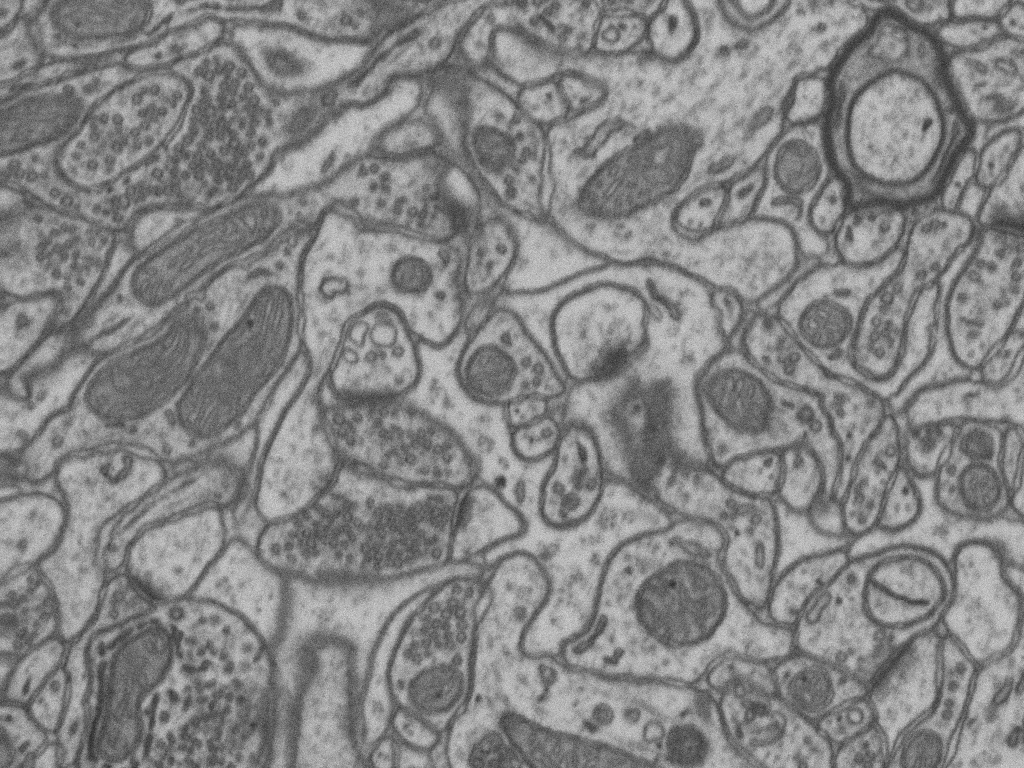}\label{fig:appendix-seg-Kasth-Lucc:original}}
\end{minipage}\par\medskip

\caption{Examples of semantic segmentation results using Kasthuri++ as the source and Lucchi++ as the target. The resulting mitochondria probability maps are shown for: (a) the baseline method (no adaptation); (b) the baseline method applied to the histogram-matched images; our (c) style-transfer, (d) self-supervised learning, and (e) Attention Y-Net approaches; and (f) the DAMT-Net method; together with the corresponding (g) ground truth and (h) original test sample from Lucchi++.}
\label{fig:appendix-seg-Kasth-Lucc}
\end{figure}

The mitochondria probability maps produced by all our tested methods on the first test image from Lucchi++  are shown in Figure~\ref{fig:appendix-seg-Kasth-Lucc} together with its corresponding ground-truth binary labels and original EM image. In these experiments, all learning-based methods perform notably well (see Figures~\ref{fig:appendix-seg-Kasth-Lucc:style}-~\ref{fig:appendix-seg-Kasth-Lucc:damtnet}). Nevertheless, the best qualitative results appear to be those produced by our Attention Y-Net approach (see Figure~\ref{fig:appendix-seg-Kasth-Lucc:attynet}), which are very close to the desired ground truth output (Figure~\ref{fig:appendix-seg-Kasth-Lucc:gt}). Notice that the displayed results for the style-transfer, SSL, Attention Y-Net, and DAMT-Net approaches correspond to executions using our proposed stop criterion (solidity, see Section 4.3).

\subsection{Source: Kasthuri++ - Target: VNC}
\begin{figure}[ht]
\begin{minipage}{.5\linewidth}
\centering
\subfloat[]{\includegraphics[scale=.12]{img_sup_material/kasthuri++/kasthuri++_test_original}\label{fig:appendix-ht-st-Kasth-VNC:kasthuri}}
\end{minipage}%
\begin{minipage}{.5\linewidth}
\centering
\subfloat[]{\includegraphics[scale=.15]{img_sup_material/vnc/vnc_test_original}\label{fig:appendix-ht-st-Kasth-VNC:vnc}}
\end{minipage}\par\medskip
\begin{minipage}{.5\linewidth}
\centering
\subfloat[]{\includegraphics[scale=.15]{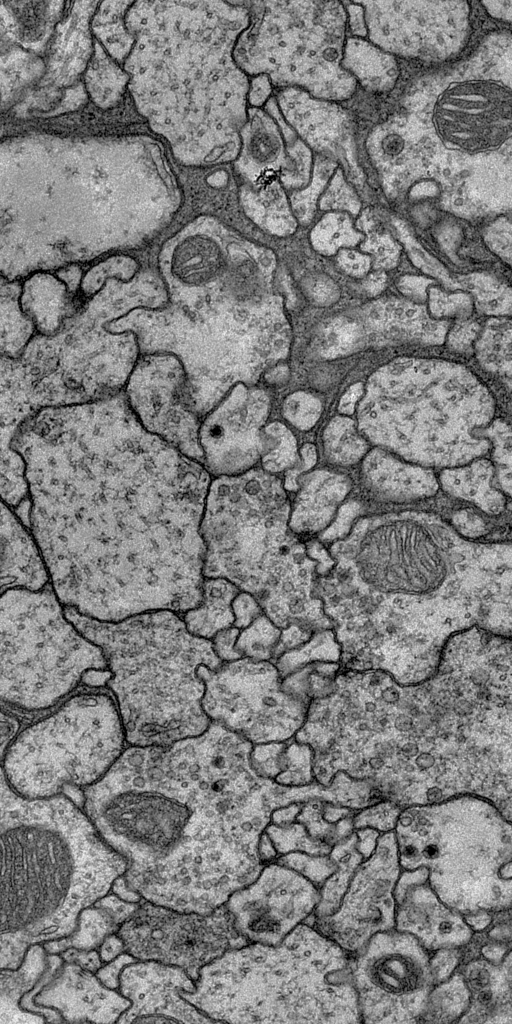}\label{fig:appendix-ht-st-Kasth-VNC:hm}}
\end{minipage}
\begin{minipage}{.5\linewidth}
\centering
\subfloat[]{\includegraphics[scale=.15]{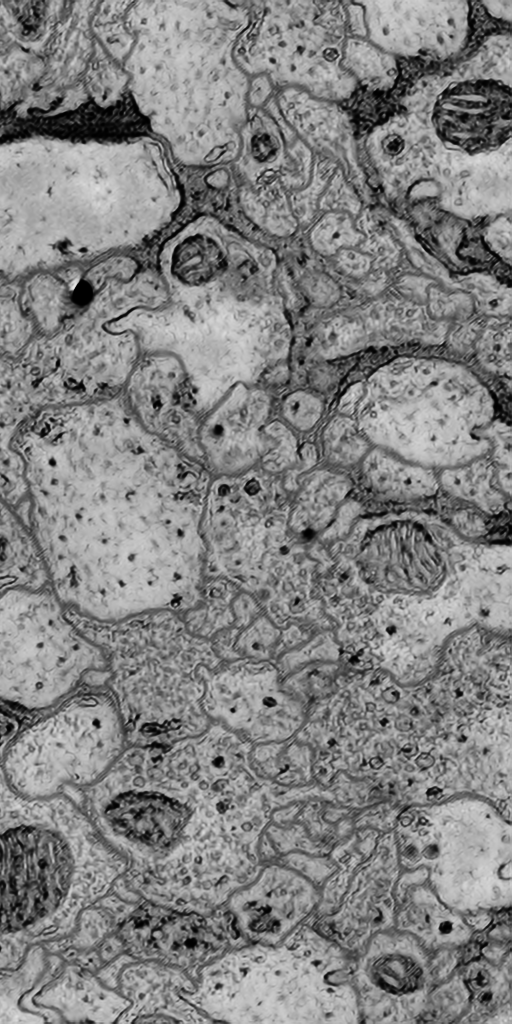}\label{fig:appendix-ht-st-Kasth-VNC:style}}
\end{minipage}\par\medskip
\caption{Examples of histogram-matching and style-transfer results using Kasthuri++ as reference histogram/style to transform VNC images: (a) Kasthuri++ dataset sample; (b) original VNC test image; (c) histogram-matched version of (b); and (d) stylized version of (b).}
\label{fig:appendix-ht-st-Kasth-VNC}
\end{figure}

The effect of our histogram-matching and style-transfer methods on an image from the VNC dataset is shown in Figure~\ref{fig:appendix-ht-st-Kasth-VNC} using the Kasthuri++ dataset as the source domain. The domain shift seems quite large in this case, and the histogram-matched images (Figure~\ref{fig:appendix-ht-st-Kasth-VNC:hm}) appear to be far away from the source domain images (Figure~\ref{fig:appendix-ht-st-Kasth-VNC:kasthuri}). In appearance, the style-transfer results (Figure~\ref{fig:appendix-ht-st-Kasth-VNC:style}) do not look much better either, but the results from Table 1 indicate the method was quite successful at transferring the style from the ssEM to the ssTEM dataset.

\begin{figure}[ht]
\centering
\begin{minipage}{.5\linewidth}
\centering
\subfloat[]{\includegraphics[scale=.07]{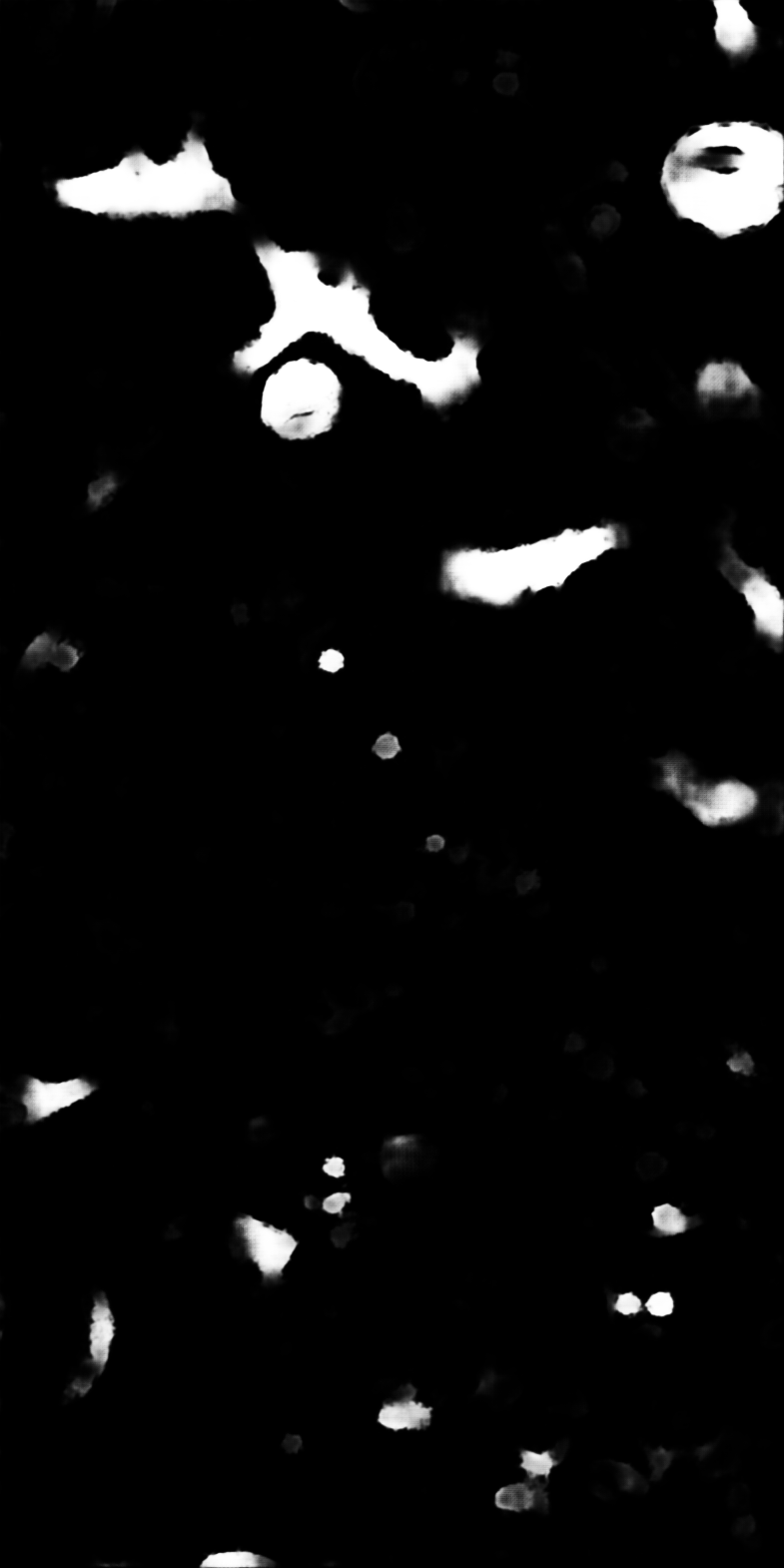}\label{fig:appendix-seg-Kasth-VNC:baseline}}
\end{minipage}%
\begin{minipage}{.5\linewidth}
\centering
\subfloat[]{\includegraphics[scale=.11]{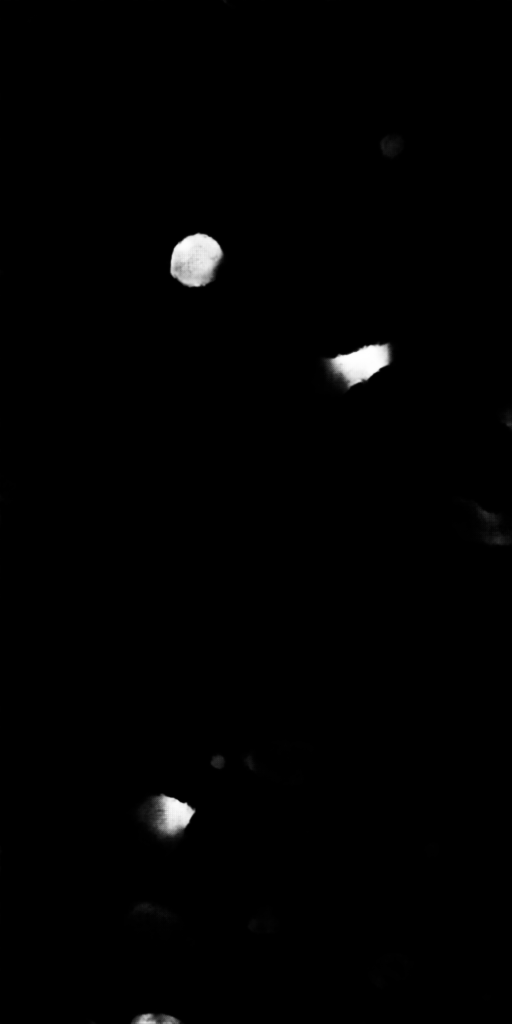}\label{fig:appendix-seg-Kasth-VNC:hm_baseline}}
\end{minipage}\par\medskip
\begin{minipage}{.5\linewidth}
\centering
\subfloat[]{\includegraphics[scale=.11]{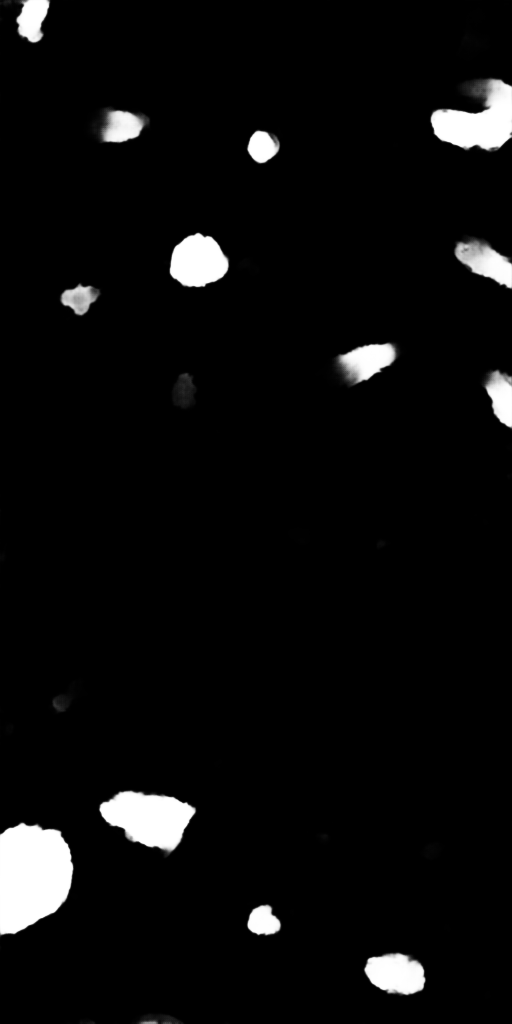}\label{fig:appendix-seg-Kasth-VNC:style}}
\end{minipage}%
\begin{minipage}{.5\linewidth}
\centering
\subfloat[]{\includegraphics[scale=.11]{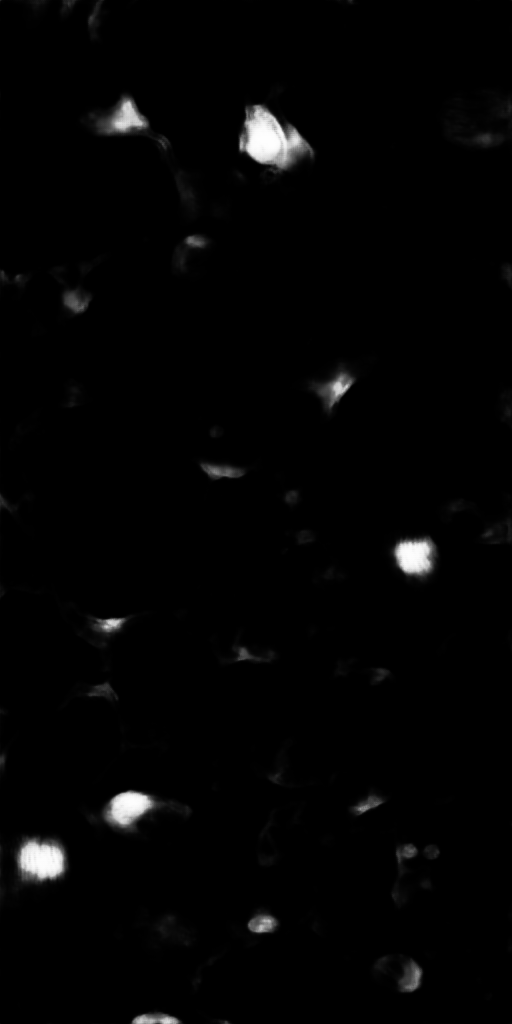}\label{fig:appendix-seg-Kasth-VNC:ssl}}
\end{minipage}\par\medskip
\begin{minipage}{.5\linewidth}
\centering
\subfloat[]{\includegraphics[scale=.11]{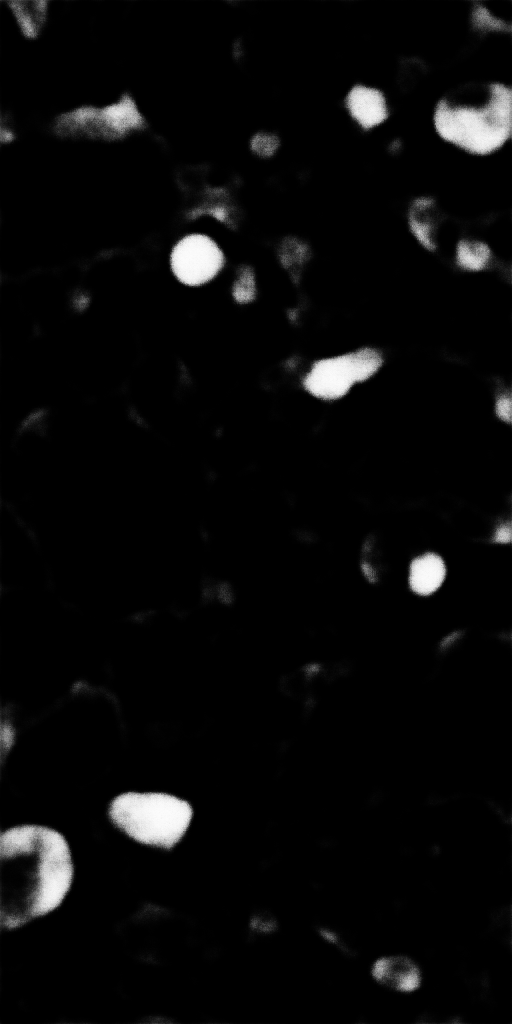}\label{fig:appendix-seg-Kasth-VNC:attynet}}
\end{minipage}%
\begin{minipage}{.5\linewidth}
\centering
\subfloat[]{\includegraphics[scale=.11]{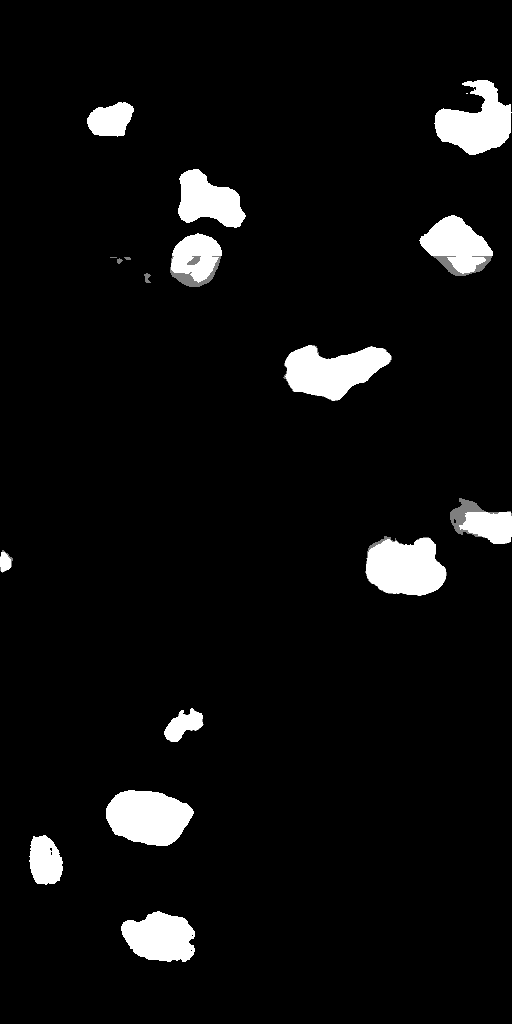}\label{fig:appendix-seg-Kasth-VNC:damtnet}}
\end{minipage}\par\medskip
\begin{minipage}{.49\linewidth}
\centering
\subfloat[]{\includegraphics[scale=.11]{img_sup_material/vnc/vnc_test_gt}\label{fig:appendix-seg-Kasth-VNC:gt}}
\end{minipage}
\begin{minipage}{.49\linewidth}
\centering
\subfloat[]{\includegraphics[scale=.11]{img_sup_material/vnc/vnc_test_original.png}\label{fig:appendix-seg-Kasth-VNC:original}}
\end{minipage}\par\medskip
\caption{Examples of semantic segmentation results using Kasthuri++ as the source and VNC as the target. The resulting mitochondria probability maps are shown for: (a) the baseline method (no adaptation); (b) the baseline method applied to the histogram-matched images; our (c) style-transfer, (d) self-supervised learning, and (e) Attention Y-Net approaches; and (f) the DAMT-Net method; together with the corresponding (g) ground truth and (h) original test sample from VNC.}
\label{fig:appendix-seg-Kasth-VNC}
\end{figure}

The mitochondria probability maps produced by all our tested methods on the first test image from VNC are shown in Figure~\ref{fig:appendix-seg-Kasth-VNC} together with its corresponding ground-truth binary labels and original EM image. In these experiments, most methods struggle to produce proper mitochondria masks. The exception is our style-transfer approach (Figure~\ref{fig:appendix-seg-Kasth-VNC:style}), which correctly finds all mitochondria present in the ground truth (Figure~\ref{fig:appendix-seg-Kasth-VNC:gt}) but also produces a couple of large mitochondria-like artifacts. Notice that the displayed results for the style-transfer, SSL, Attention Y-Net, and DAMT-Net approaches correspond to executions using our proposed stop criterion (solidity, see Section 4.3).

\subsection{Source: VNC - Target: Lucchi++}

\begin{figure}[ht]
\begin{minipage}{.5\linewidth}
\centering
\subfloat[]{\includegraphics[scale=.14]{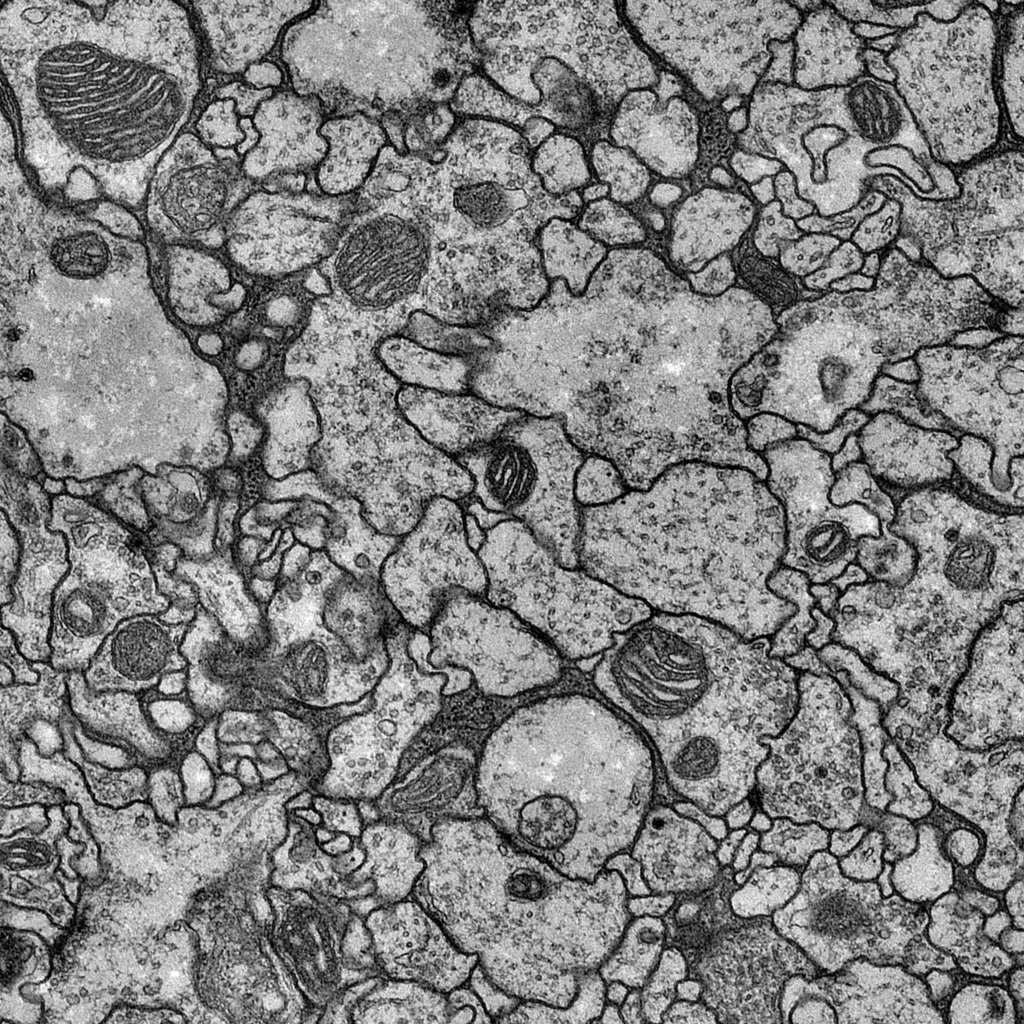}\label{fig:appendix-ht-st-VNC-Lucc:vnc}}
\end{minipage}%
\begin{minipage}{.5\linewidth}
\centering
\subfloat[]{\includegraphics[scale=.16]{img_sup_material/lucchi++/lucchi++_test_original}\label{fig:appendix-ht-st-VNC-Lucc:lucchi}}
\end{minipage}\par\medskip
\begin{minipage}{.5\linewidth}
\centering
\subfloat[]{\includegraphics[scale=.16]{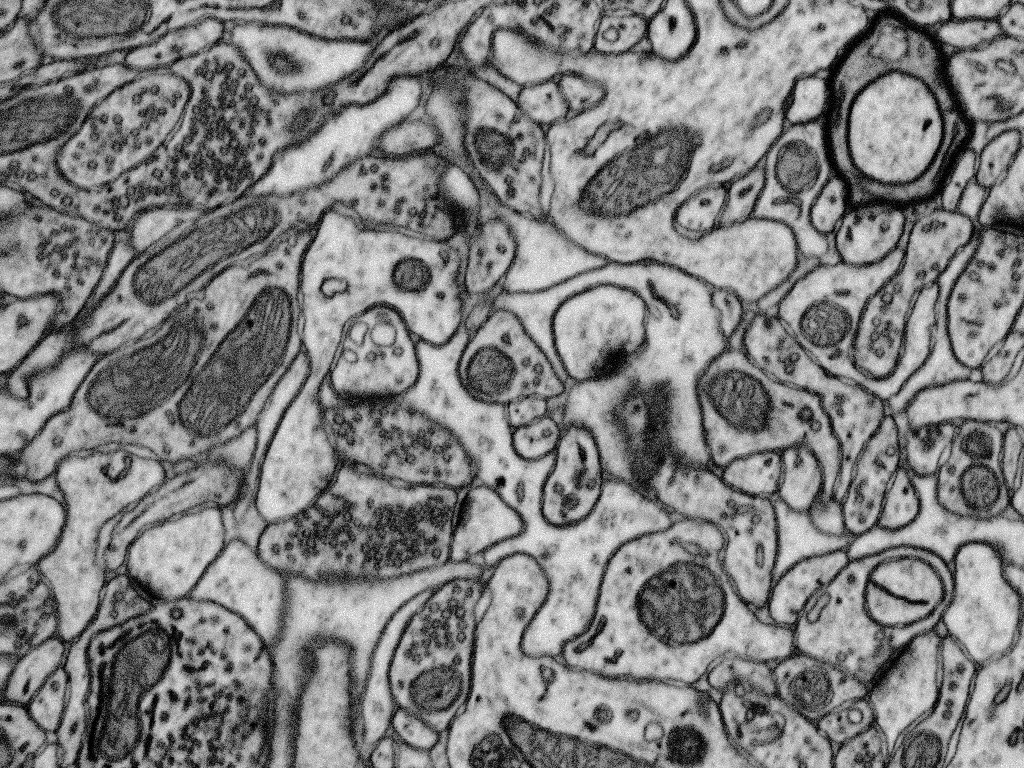}\label{fig:appendix-ht-st-VNC-Lucc:hm}}
\end{minipage}
\begin{minipage}{.5\linewidth}
\centering
\subfloat[]{\includegraphics[scale=.16]{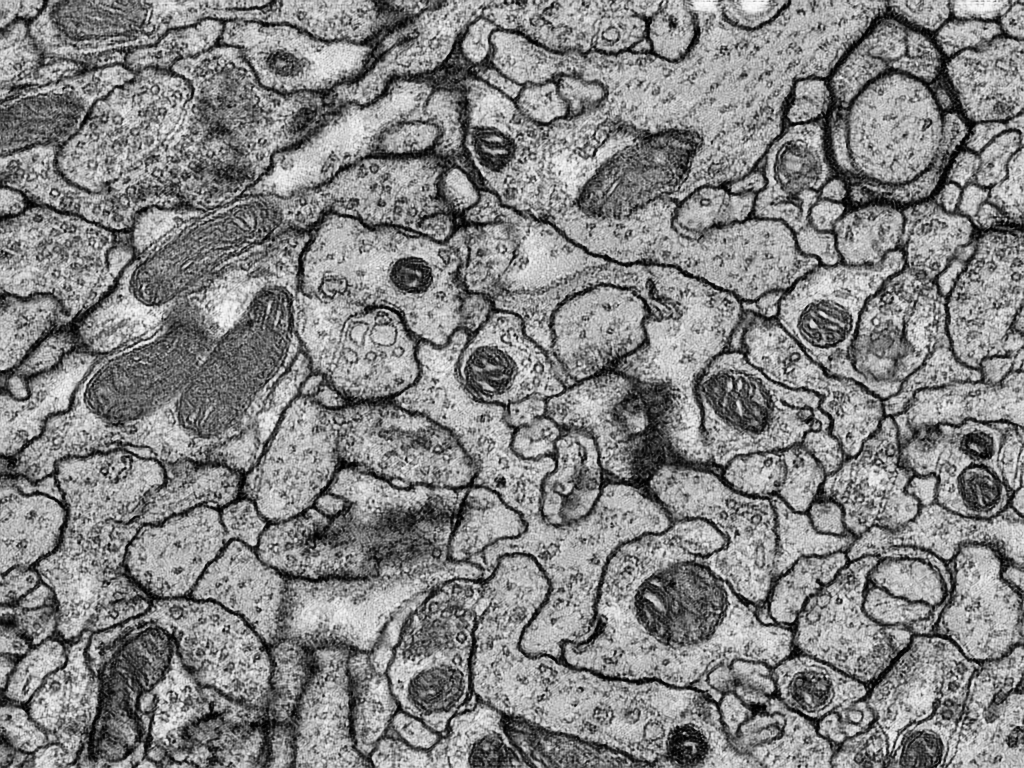}\label{fig:appendix-ht-st-VNC-Lucc:style}}
\end{minipage}\par\medskip
\caption{Examples of histogram-matching and style-transfer results using VNC as reference histogram/style to transform Lucchi++ images. From top to bottom and from left to right: (a) VNC dataset sample; (b) original Lucchi++ test image; (c) histogram-matched version of (b); and (d) stylized version of (b).}
\label{fig:appendix-ht-st-VNC-Lucc}
\end{figure}

The effect of our histogram-matching and style-transfer methods on an image from the Lucchi++ dataset is shown in Figure~\ref{fig:appendix-ht-st-VNC-Lucc} using the VNC dataset as the source domain. Both the histogram-matched image (Figure~\ref{fig:appendix-ht-st-VNC-Lucc:hm}) and the stylized image (Figure~\ref{fig:appendix-ht-st-VNC-Lucc:style}) seem to reproduce the appearance of the source domain image (Figure~\ref{fig:appendix-ht-st-VNC-Lucc:vnc}). In particular, the style-transfer results (Figure~\ref{fig:appendix-ht-st-VNC-Lucc:style}) are able to not only reproduce the source intensities but also correctly replicate the textures inside the neural processes.

\begin{figure}[ht]
\centering
\begin{minipage}{.5\linewidth}
\centering
\subfloat[]{\includegraphics[scale=.13]{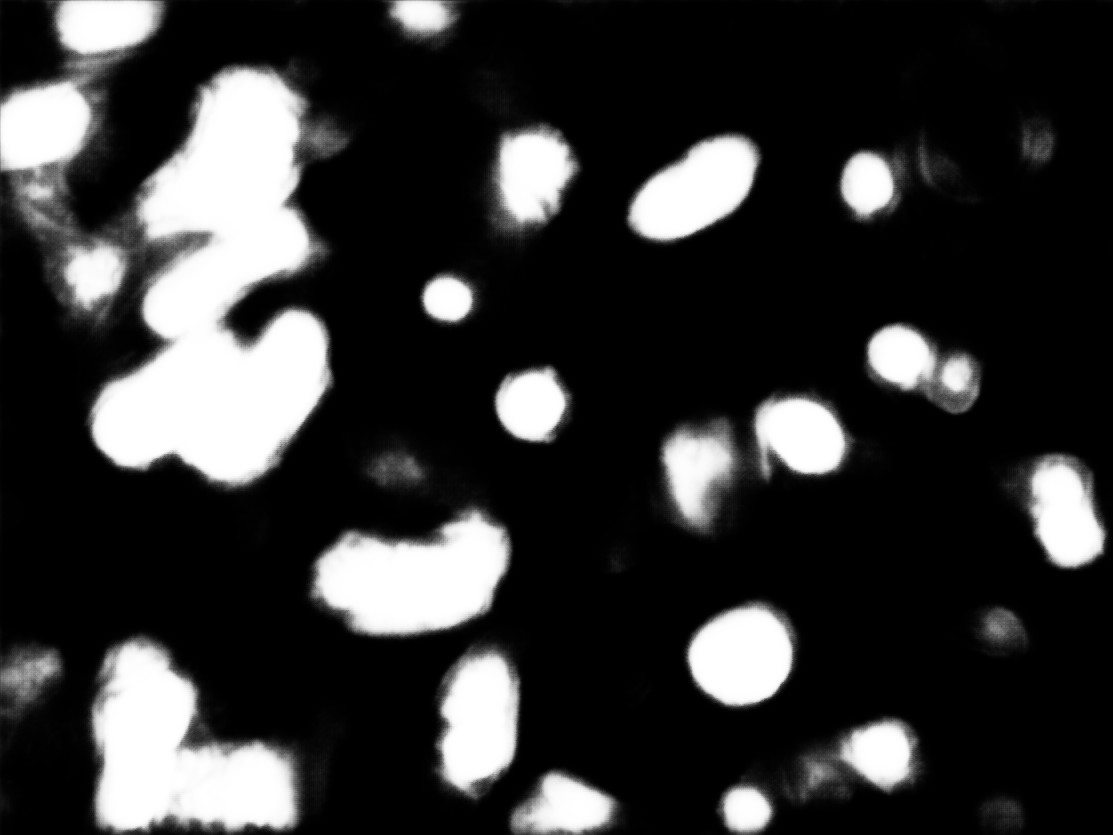}\label{fig:appendix-seg-VNC-Lucc:baseline}}
\end{minipage}%
\begin{minipage}{.5\linewidth}
\centering
\subfloat[]{\includegraphics[scale=.14]{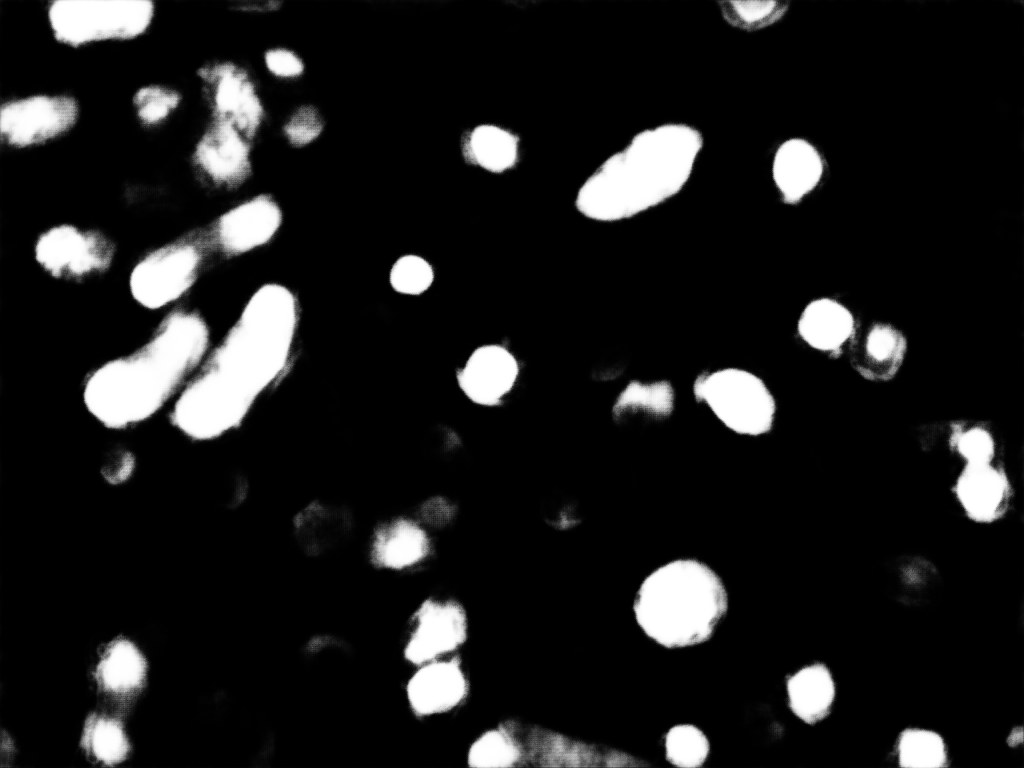}\label{fig:appendix-seg-VNC-Lucc:hm_baseline}}
\end{minipage}\par\medskip
\begin{minipage}{.5\linewidth}
\centering
\subfloat[]{\includegraphics[scale=.14]{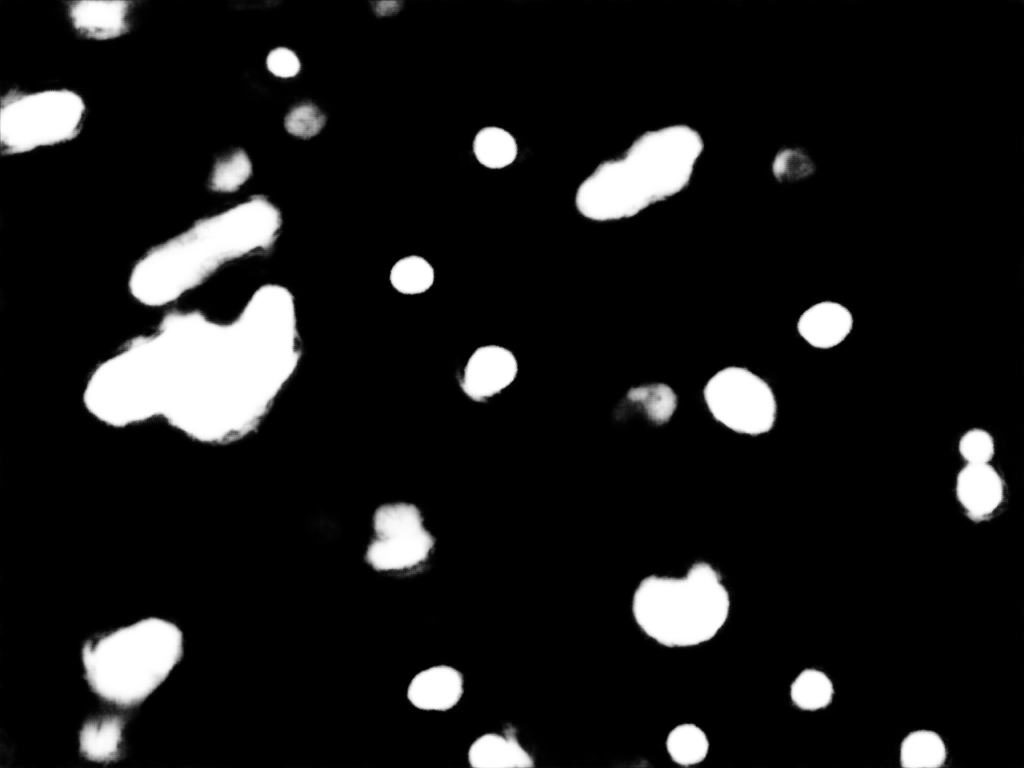}\label{fig:appendix-seg-VNC-Lucc:style}}
\end{minipage}%
\begin{minipage}{.5\linewidth}
\centering
\subfloat[]{\includegraphics[scale=.14]{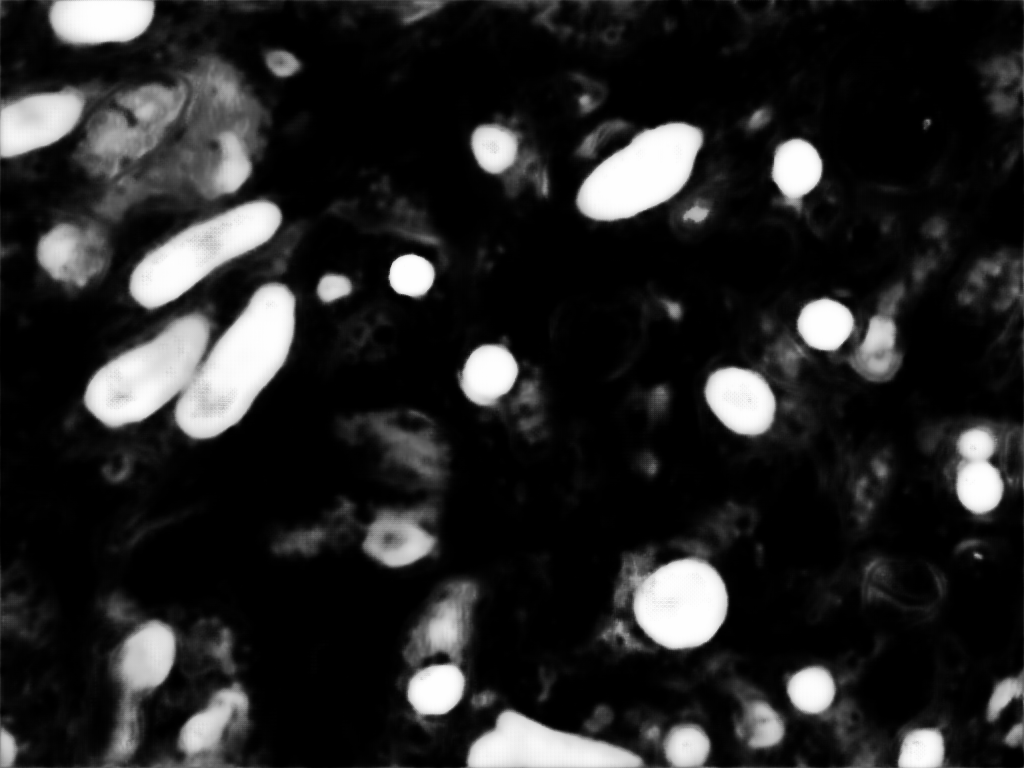}\label{fig:appendix-seg-VNC-Lucc:ssl}}
\end{minipage}\par\medskip
\begin{minipage}{.5\linewidth}
\centering
\subfloat[]{\includegraphics[scale=.14]{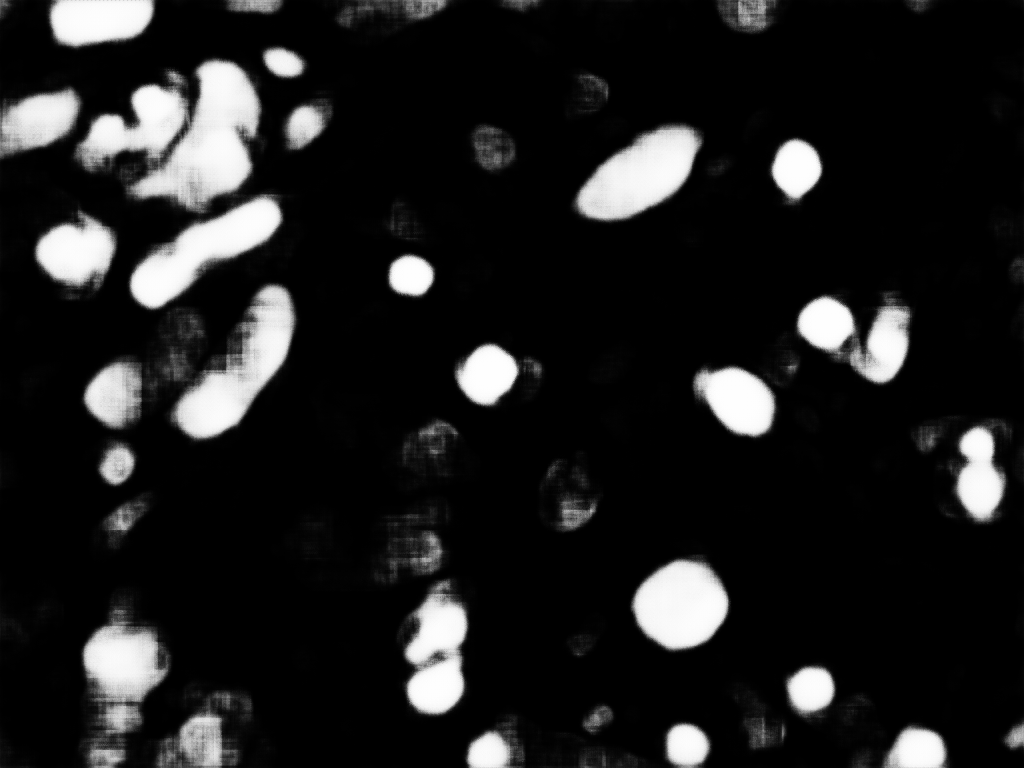}\label{fig:appendix-seg-VNC-Lucc:attynet}}
\end{minipage}%
\begin{minipage}{.5\linewidth}
\centering
\subfloat[]{\includegraphics[scale=.14]{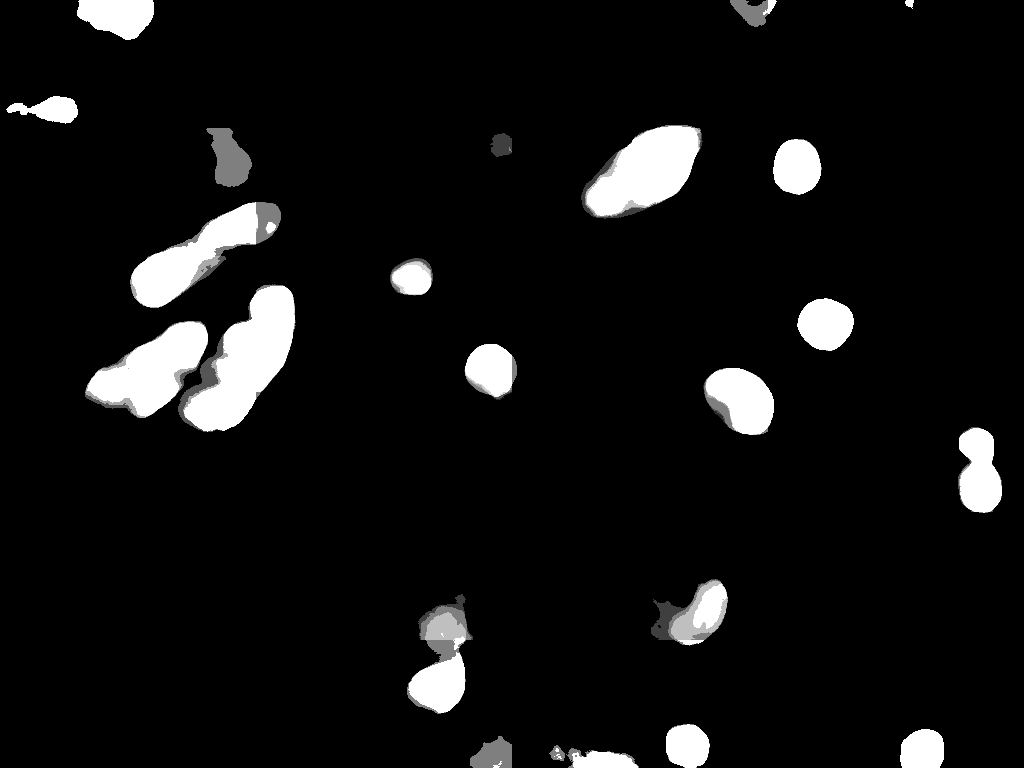}\label{fig:appendix-seg-VNC-Lucc:damtnet}}
\end{minipage}\par\medskip
\begin{minipage}{.49\linewidth}
\centering
\subfloat[]{\includegraphics[scale=.14]{img_sup_material/lucchi++/lucchi++_test_gt}\label{fig:appendix-seg-VNC-Lucc:gt}}
\end{minipage}
\begin{minipage}{.49\linewidth}
\centering
\subfloat[]{\includegraphics[scale=.14]{img_sup_material/lucchi++/lucchi++_test_original.png}\label{fig:appendix-seg-VNC-Lucc:original}}
\end{minipage}\par\medskip
\caption{Examples of semantic segmentation results using VNC as the source and Lucchi++ as the target. The resulting mitochondria probability maps are shown for: (a) the baseline method (no adaptation); (b) the baseline method applied to the histogram-matched images; our (c) style-transfer, (d) self-supervised learning, and (e) Attention Y-Net approaches; and (f) the DAMT-Net method; together with the corresponding (g) ground truth and (h) original test sample from Lucchi++.}
\label{fig:appendix-seg-VNC-Lucc}
\end{figure}

The mitochondria probability maps produced by all our tested methods on the first test image from Lucchi++ are shown in Figure~\ref{fig:appendix-seg-VNC-Lucc} together with its corresponding ground-truth binary labels and original EM image. While all methods identify all the mitochondria present in the ground truth correctly (Figure~\ref{fig:appendix-seg-VNC-Lucc:gt}), most of them produce an over-segmentation, except for DAMT-Net, which is under-segmenting  (Figure~\ref{fig:appendix-seg-VNC-Lucc:damtnet}). Although some extra low-probability maps are created by the SSL method (Figure~\ref{fig:appendix-seg-VNC-Lucc:ssl}), its medium-high probability maps nicely capture the real mitochondria. Notice that the displayed results for the style-transfer, SSL, Attention Y-Net, and DAMT-Net approaches correspond to executions using our proposed stop criterion (solidity, see Section 4.3).

\subsection{Source: VNC - Target: Kasthuri++}

\begin{figure}[ht]
\begin{minipage}{.5\linewidth}
\centering
\subfloat[]{\includegraphics[scale=.15]{img_sup_material/vnc/vnc_test_original}\label{fig:appendix-ht-st-VNC-Kasth:vnc}}
\end{minipage}%
\begin{minipage}{.5\linewidth}
\centering
\subfloat[]{\includegraphics[scale=.12]{img_sup_material/kasthuri++/kasthuri++_test_original}\label{fig:appendix-ht-st-VNC-Kasth:kasthuri}}
\end{minipage}\par\medskip
\centering
\begin{minipage}{.5\linewidth}
\centering
\subfloat[]{\includegraphics[scale=.12]{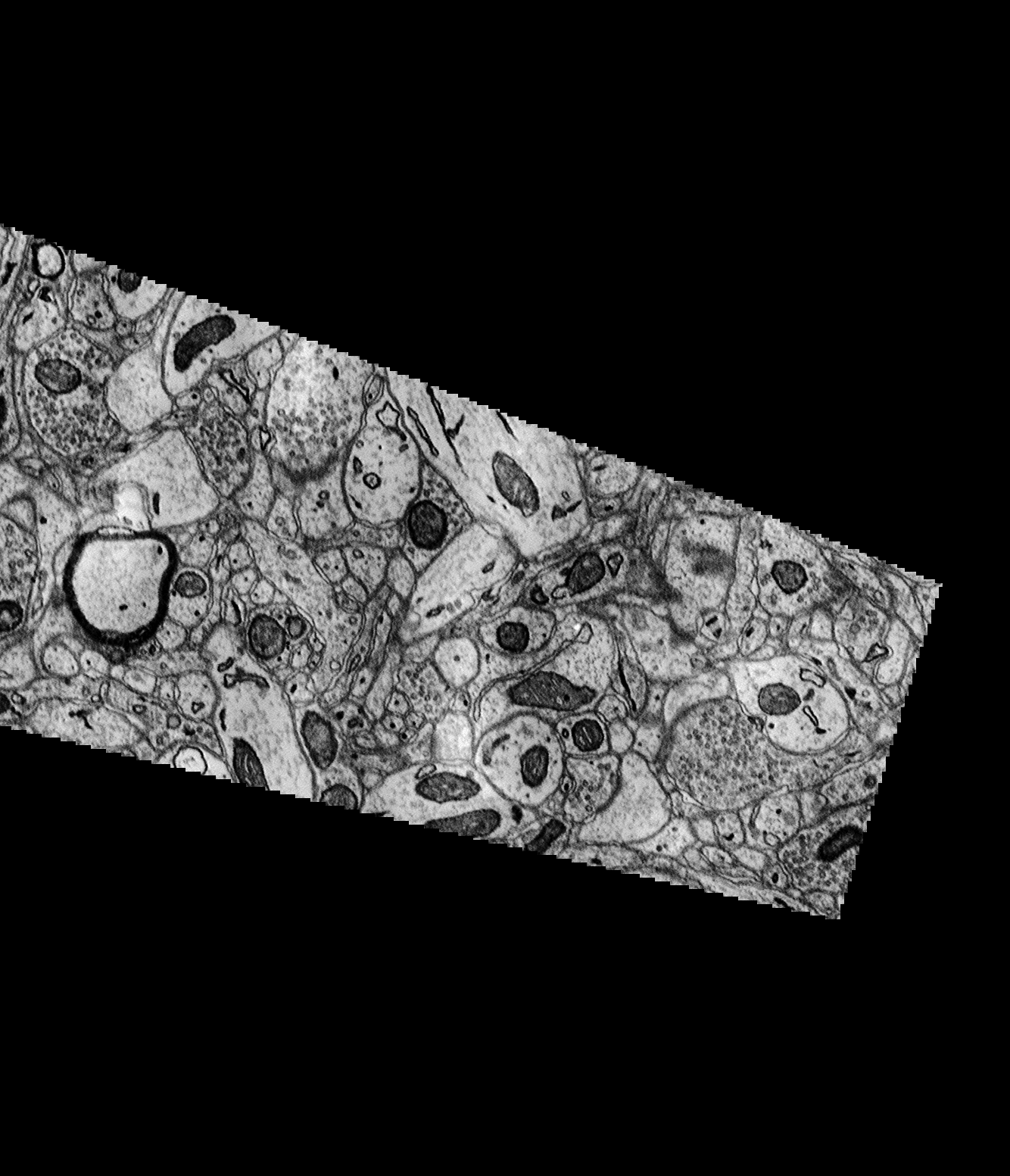}\label{fig:appendix-ht-st-VNC-Kasth:hm}}
\end{minipage}%
\begin{minipage}{.5\linewidth}
\centering
\subfloat[]{\includegraphics[scale=.12]{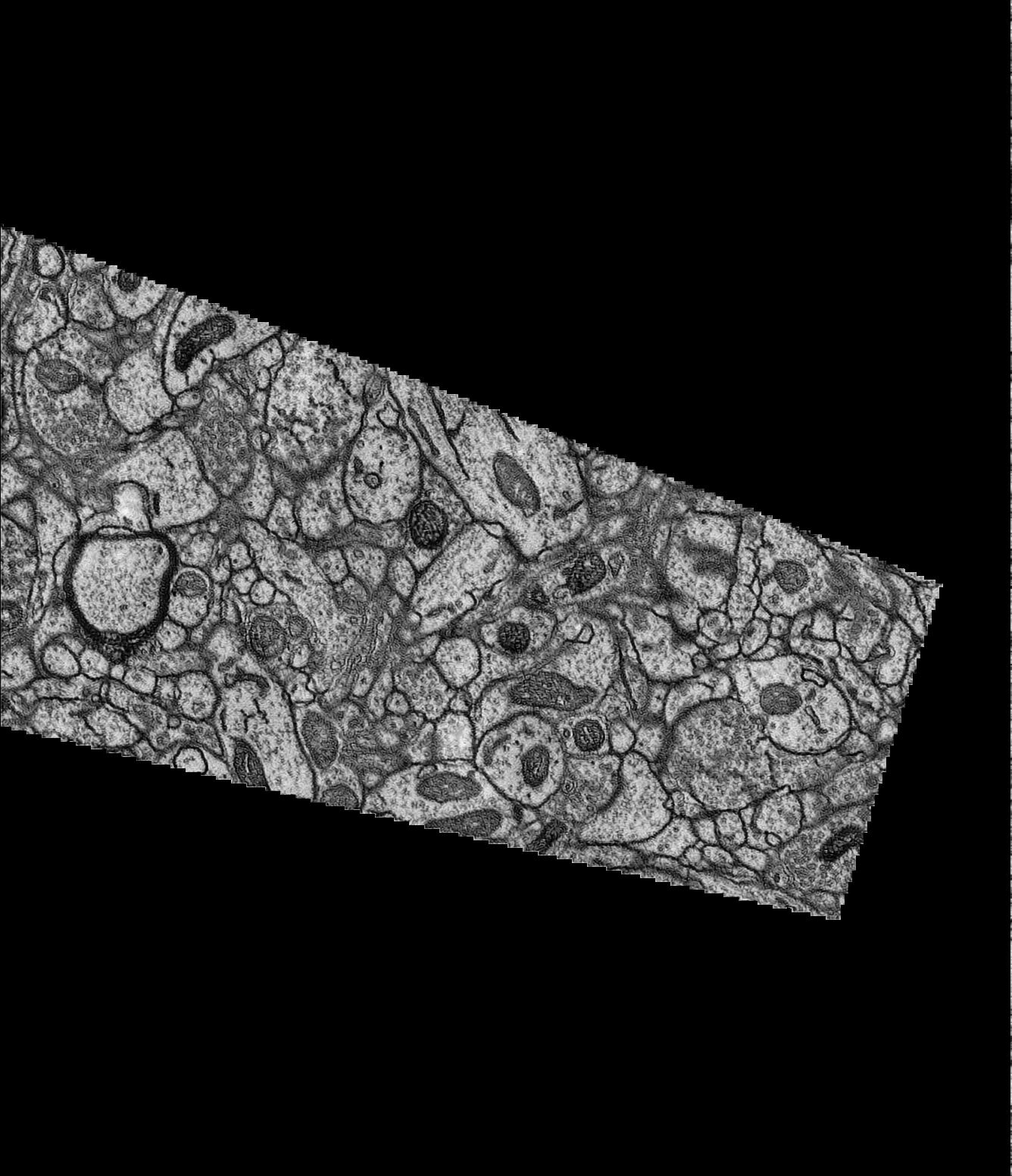}\label{fig:appendix-ht-st-VNC-Kasth:style}}
\end{minipage}\par\medskip
\caption{Examples of histogram-matching and style-transfer results using VNC as reference histogram/style to transform Kasthuri++ images. From top to bottom and from left to right: (a) VNC dataset sample; (b) original Kasthuri++ test image; (c) histogram-matched version of (b); and (d) stylized version of (b).}
\label{fig:appendix-ht-st-VNC-Kasth}
\end{figure}

The effect of our histogram-matching and style-transfer methods on an image from the Kasthuri++ dataset is shown in Figure~\ref{fig:appendix-ht-st-VNC-Kasth} using the VNC dataset as the source domain. As in the previous case, both the histogram-matched image (Figure~\ref{fig:appendix-ht-st-VNC-Kasth:hm}) and the stylized image (Figure~\ref{fig:appendix-ht-st-VNC-Kasth:style}) seem to reproduce the appearance of the source domain image (Figure~\ref{fig:appendix-ht-st-VNC-Kasth:vnc}). Again, the style-transfer results (Figure~\ref{fig:appendix-ht-st-VNC-Kasth:style}) seem to not only reproduce the source intensities but also correctly replicate the textures inside the neural processes.

\begin{figure}[ht]
\centering
\begin{minipage}{.5\linewidth}
\centering
\subfloat[]{\includegraphics[scale=.115]{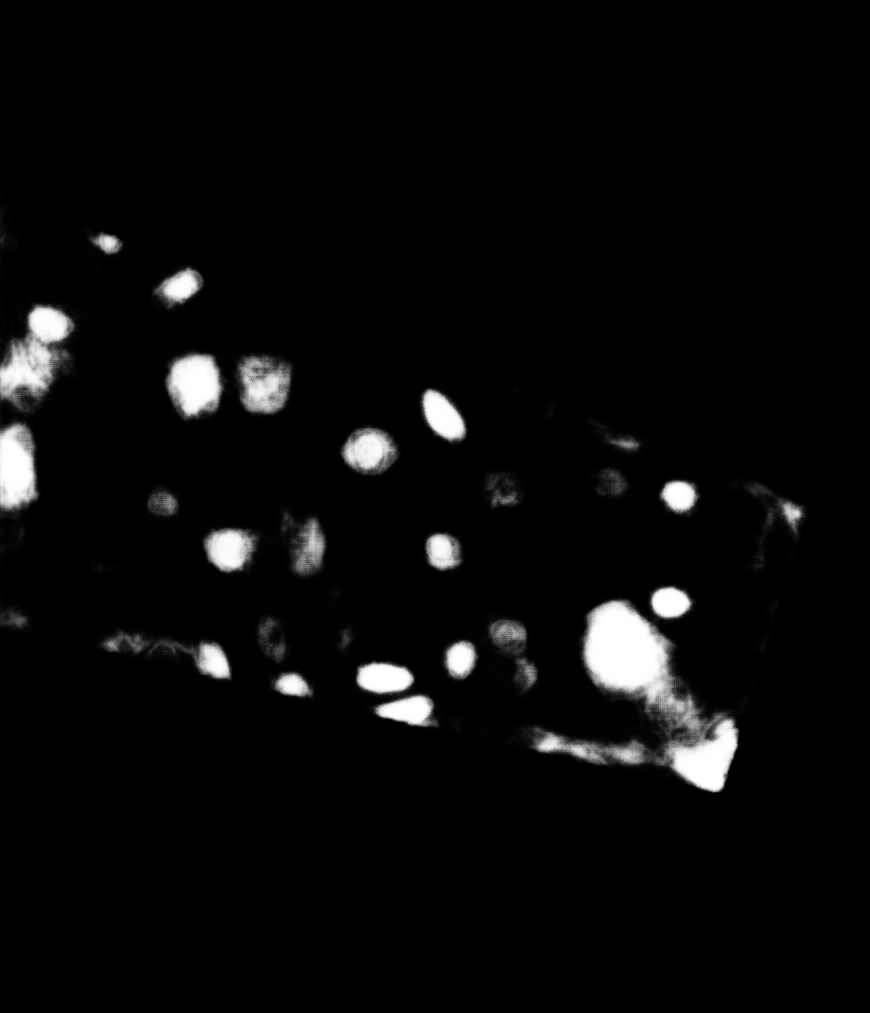}\label{fig:appendix-seg-VNC-Kasth:baseline}}
\end{minipage}%
\begin{minipage}{.5\linewidth}
\centering
\subfloat[]{\includegraphics[scale=.07]{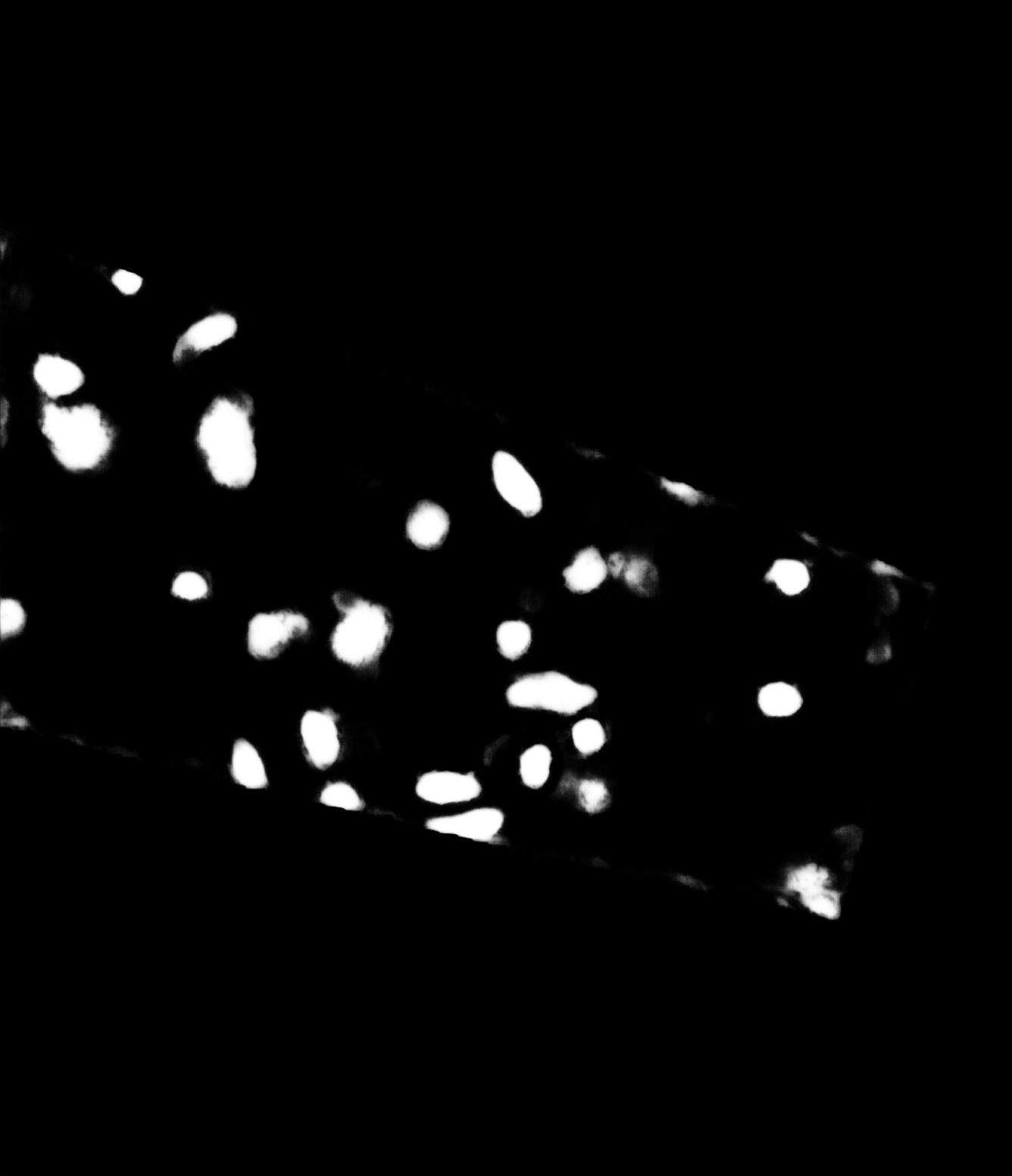}\label{fig:appendix-seg-VNC-Kasth:hm_baseline}}
\end{minipage}\par\medskip
\begin{minipage}{.5\linewidth}
\centering
\subfloat[]{\includegraphics[scale=.07]{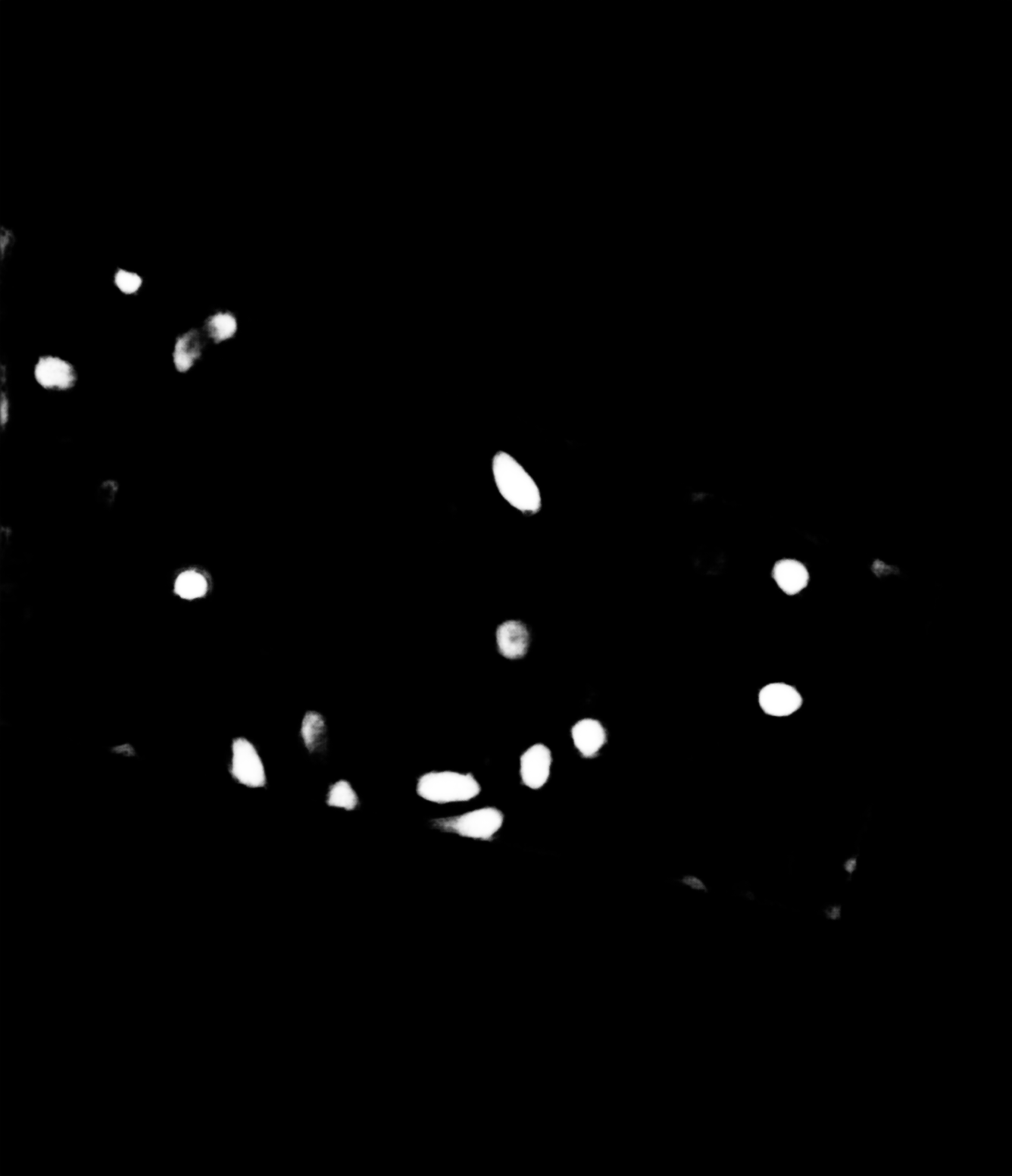}\label{fig:appendix-seg-VNC-Kasth:style}}
\end{minipage}%
\begin{minipage}{.5\linewidth}
\centering
\subfloat[]{\includegraphics[scale=.07]{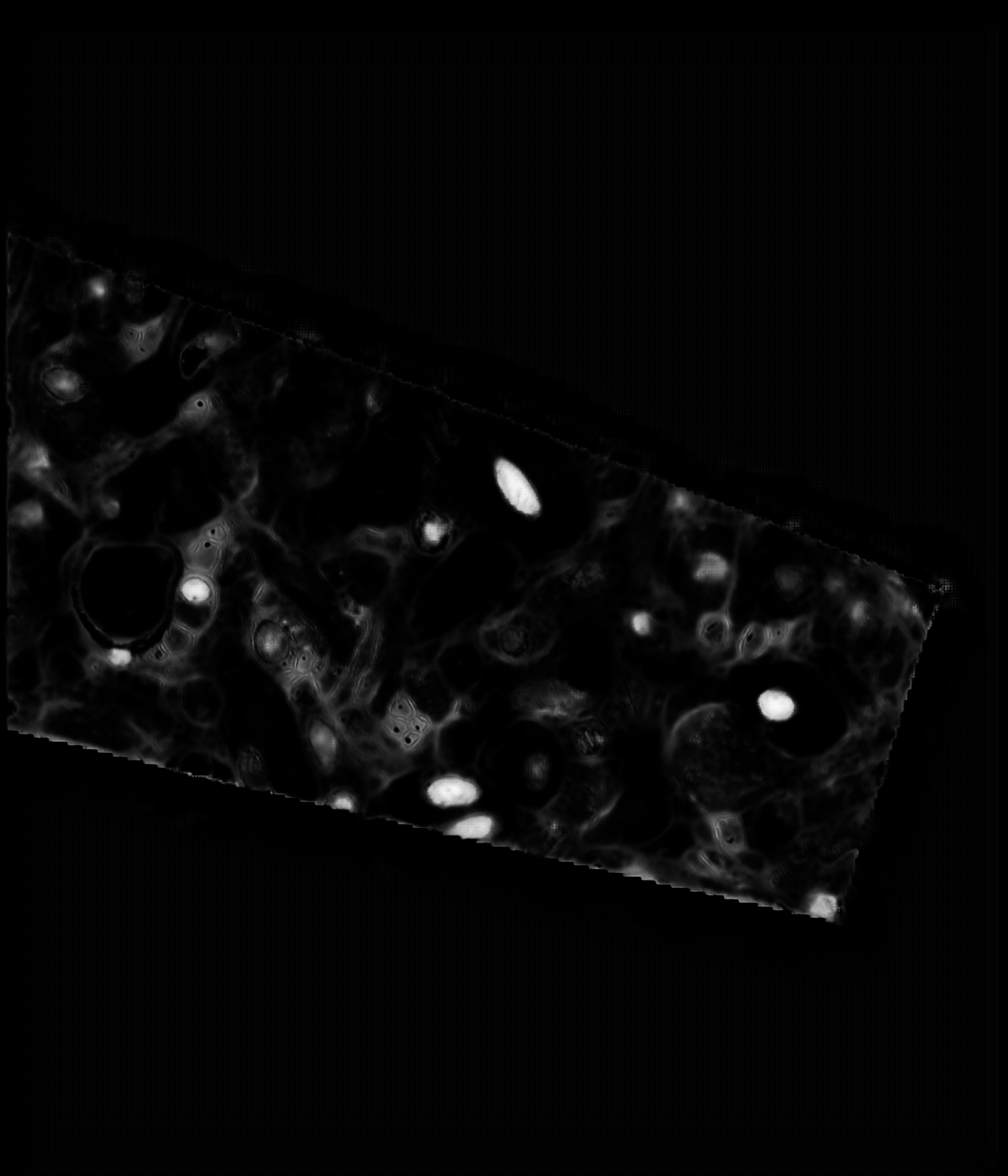}\label{fig:appendix-seg-VNC-Kasth:ssl}}
\end{minipage}\par\medskip
\begin{minipage}{.5\linewidth}
\centering
\subfloat[]{\includegraphics[scale=.07]{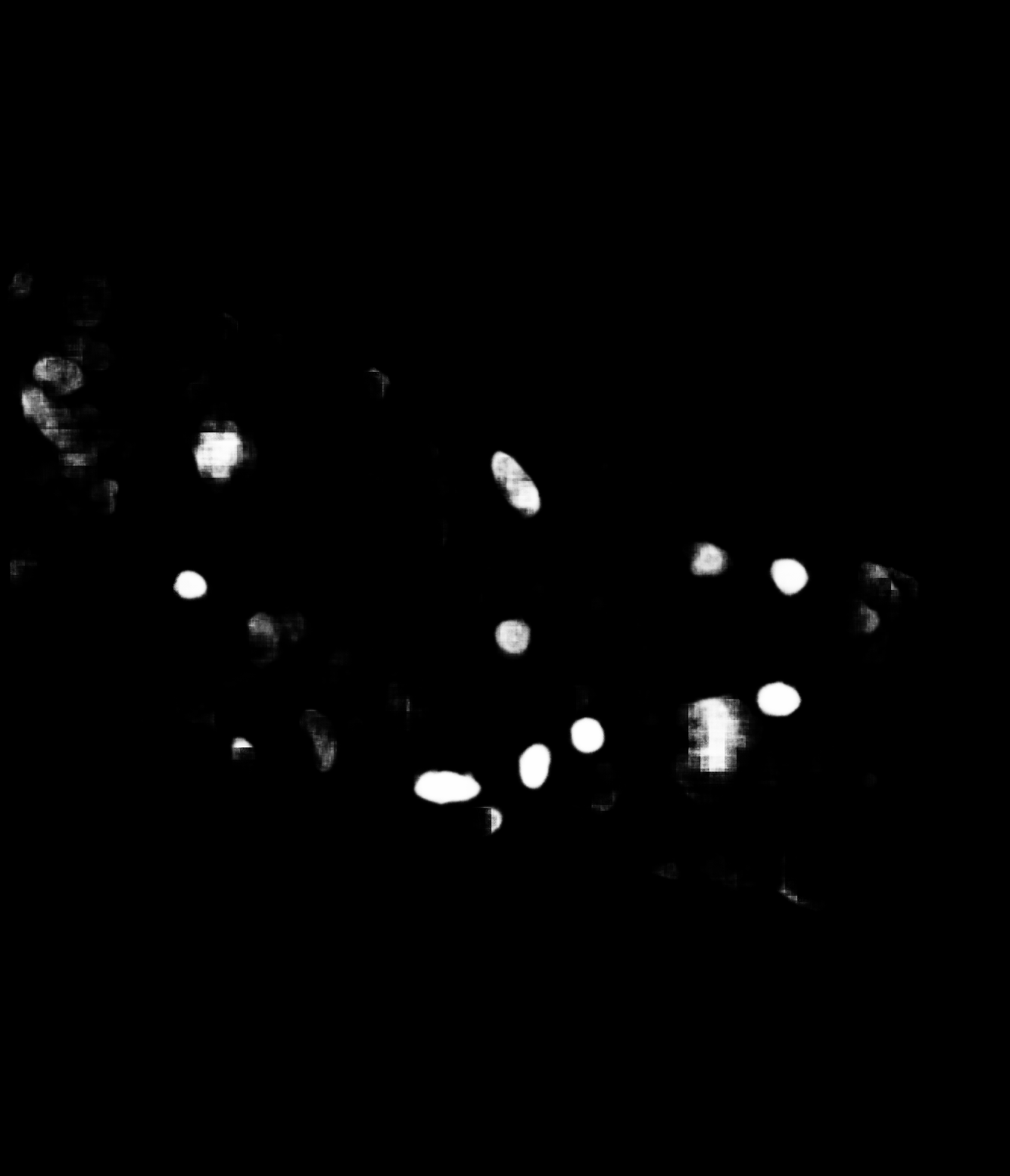}\label{fig:appendix-seg-VNC-Kasth:attynet}}
\end{minipage}%
\begin{minipage}{.5\linewidth}
\centering
\subfloat[]{\includegraphics[scale=.07]{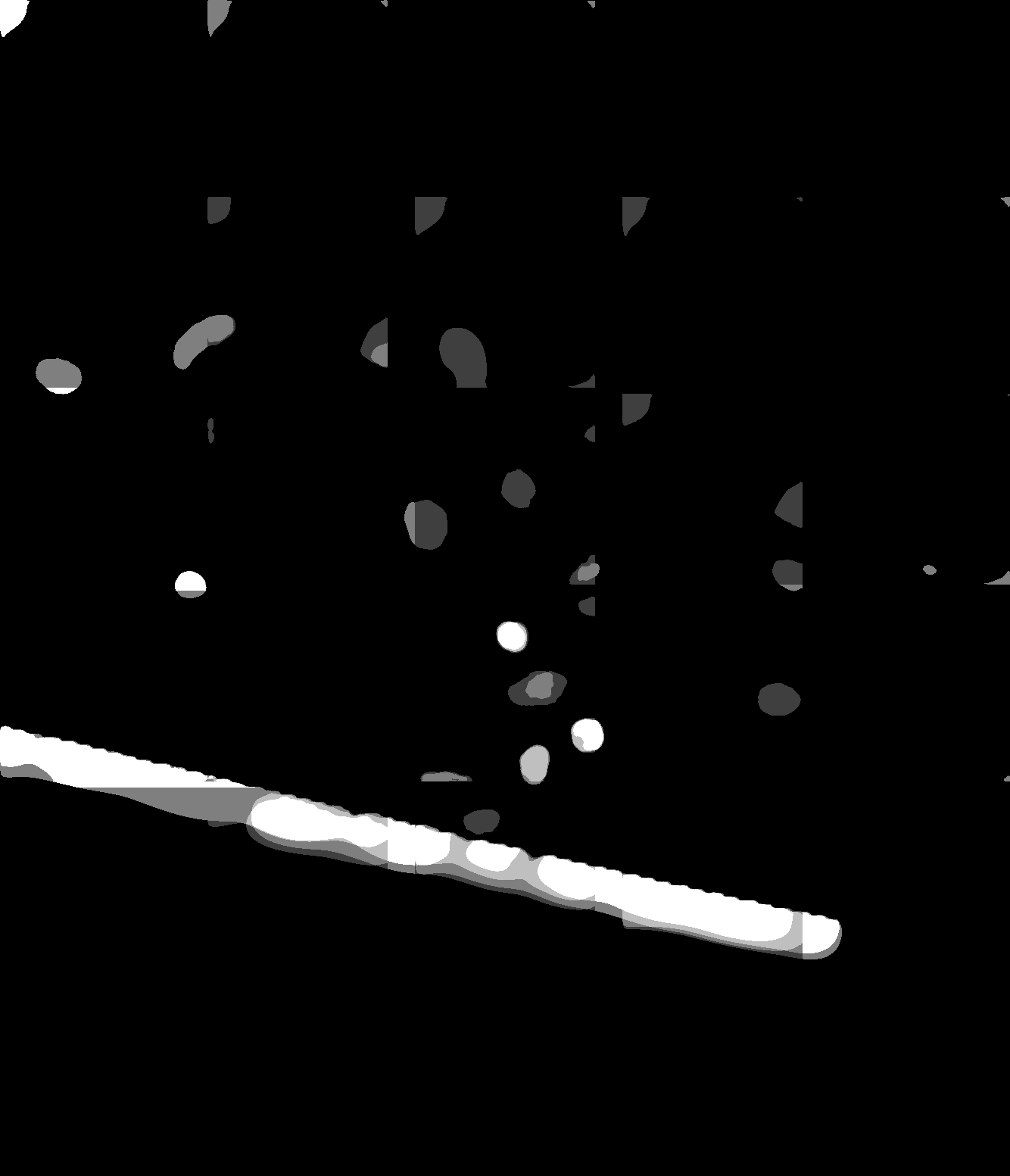}\label{fig:appendix-seg-VNC-Kasth:damtnet}}
\end{minipage}\par\medskip
\begin{minipage}{.49\linewidth}
\centering
\subfloat[]{\includegraphics[scale=.07]{img_sup_material/kasthuri++/kasthuri++_test_gt}\label{fig:appendix-seg-VNC-Kasth:gt}}
\end{minipage}
\begin{minipage}{.49\linewidth}
\centering
\subfloat[]{\includegraphics[scale=.07]{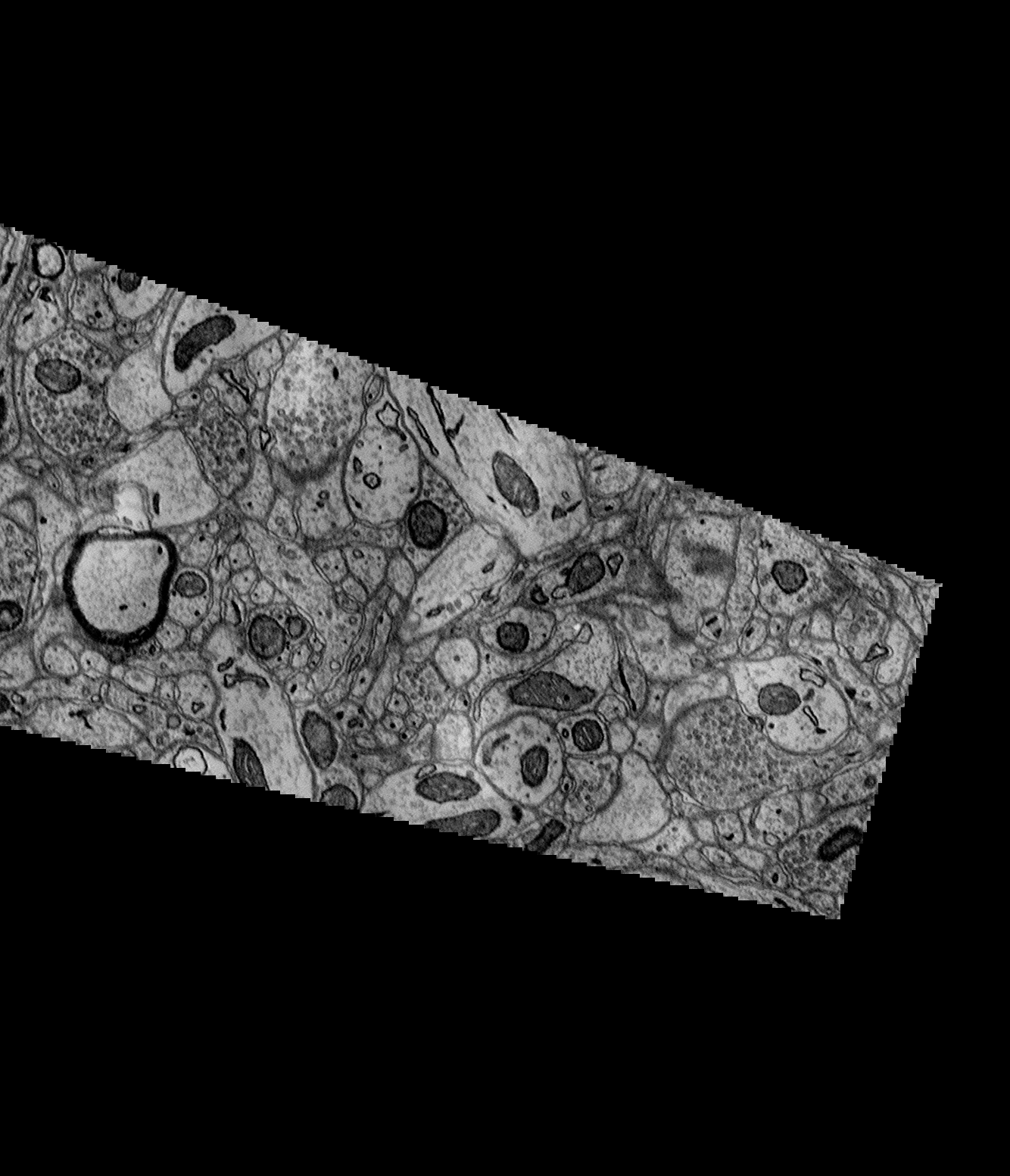}\label{fig:appendix-seg-VNC-Kasth:original}}
\end{minipage}\par\medskip

\caption{Examples of semantic segmentation results using VNC as the source and Kasthuri++ as the target. The resulting mitochondria probability maps are shown for: (a) the baseline method (no adaptation); (b) the baseline method applied to the histogram-matched images; our (c) style-transfer, (d) self-supervised learning, and (e) Attention Y-Net approaches; and (f) the DAMT-Net method; together with the corresponding (g) ground truth and (h) original test sample from Kasthuri++.}
\label{fig:appendix-seg-VNC-Kasth}
\end{figure}

The mitochondria probability maps produced by all our tested methods on the first test image from Kasthuri++ are shown in Figure~\ref{fig:appendix-seg-VNC-Kasth} together with its corresponding ground-truth binary labels and original EM image. Here all methods struggle to correctly identify the mitochondria present in the ground truth (Figure~\ref{fig:appendix-seg-VNC-Kasth:gt}). Some of them produce an over-segmentation (Figures~\ref{fig:appendix-seg-VNC-Kasth:baseline}, \ref{fig:appendix-seg-VNC-Kasth:style}, \ref{fig:appendix-seg-VNC-Kasth:attynet}), while others are under-segmenting  (Figures~\ref{fig:appendix-seg-VNC-Kasth:ssl},~\ref{fig:appendix-seg-VNC-Kasth:damtnet}). As observed before, the DAMT-Net method produces artifacts in the border of the tissue areas due to the padding (Figure~\ref{fig:appendix-seg-VNC-Kasth:damtnet}). Notice that the displayed results for the style-transfer, SSL, Attention Y-Net, and DAMT-Net approaches correspond to executions using our proposed stop criterion (solidity, see Section 4.3).

\clearpage
\newpage
\section{Analysis of solidity as stop condition}
\label{appendix:stop-condition}

\setcounter{figure}{0}  

In this section, we analyze the effect of using the solidity of the predicted masks (see Section 4.3) as a stop condition in all the tested learning methods. With that aim, we plot the solidity values at each epoch of every cross-dataset experiment and, on a complementary plot, the IoU values produced in the test partition of the target dataset at the same epochs.

\begin{figure}[ht]

\begin{minipage}{.5\linewidth}
\centering
\subfloat[]{\includegraphics[scale=.35]{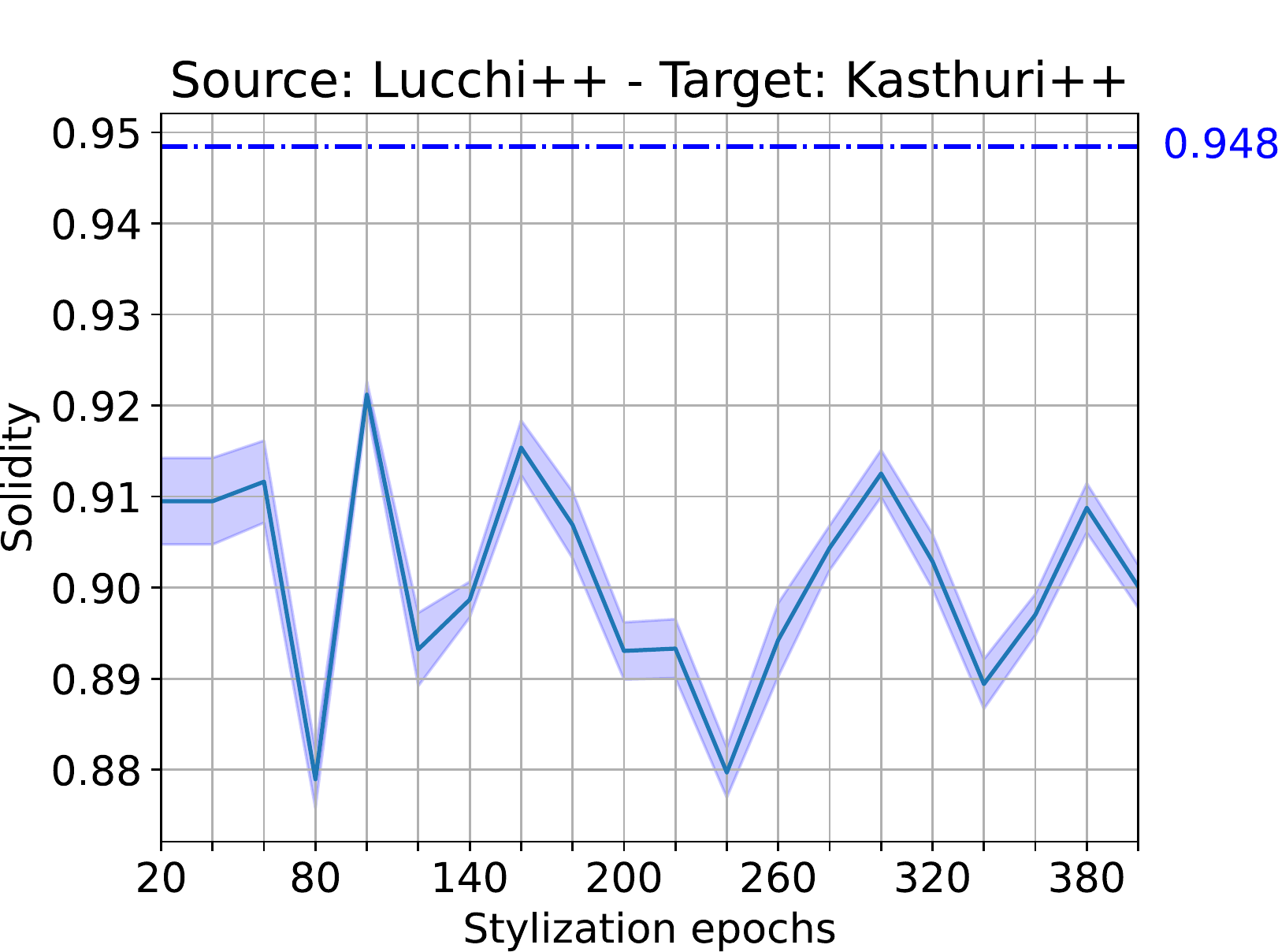}}
\end{minipage}
\begin{minipage}{.5\linewidth}
\centering
\subfloat[]{\includegraphics[scale=.35]{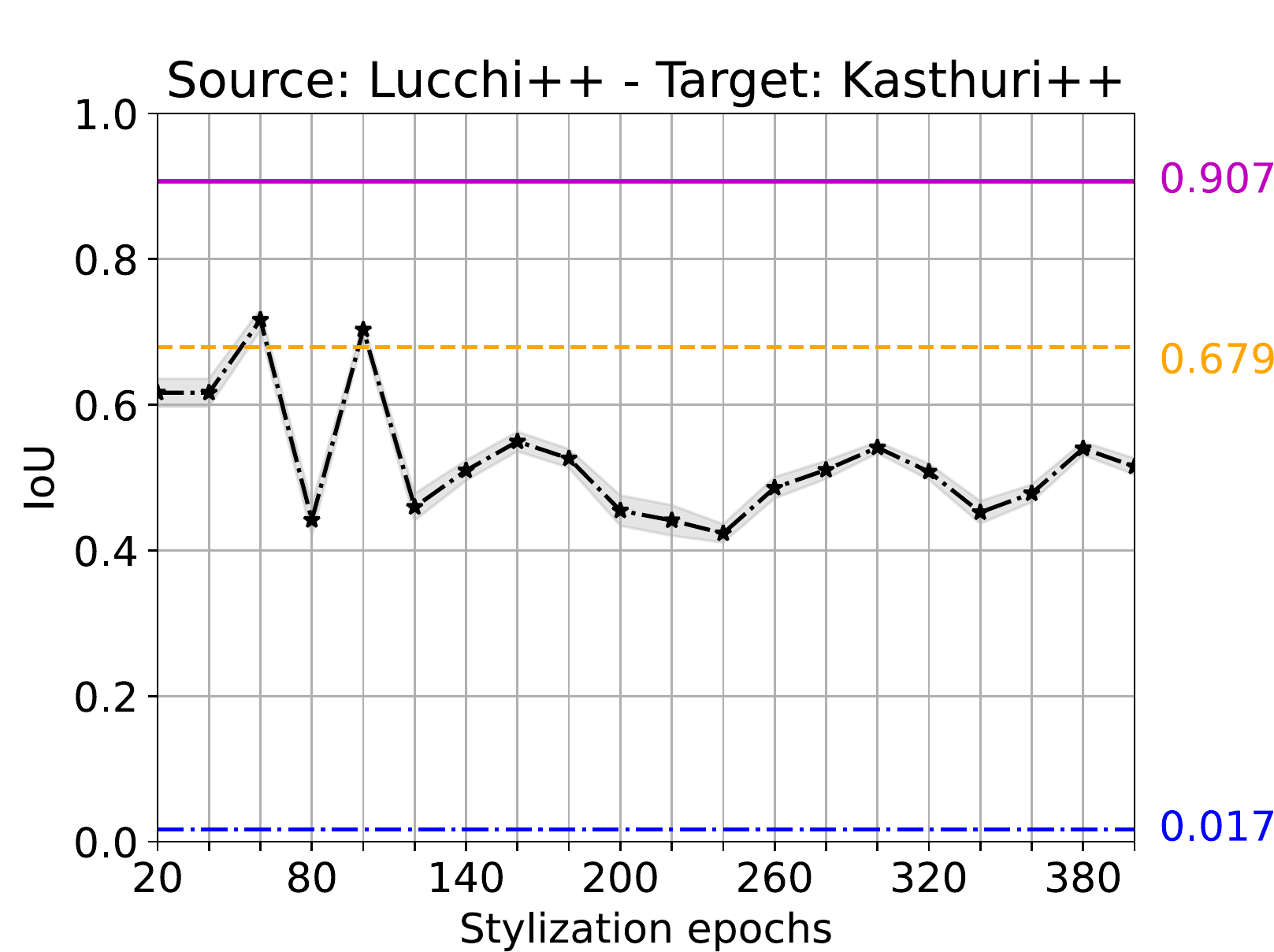}}
\end{minipage}\par\medskip
\begin{minipage}{.5\linewidth}
\centering
\subfloat[]{\includegraphics[scale=.35]{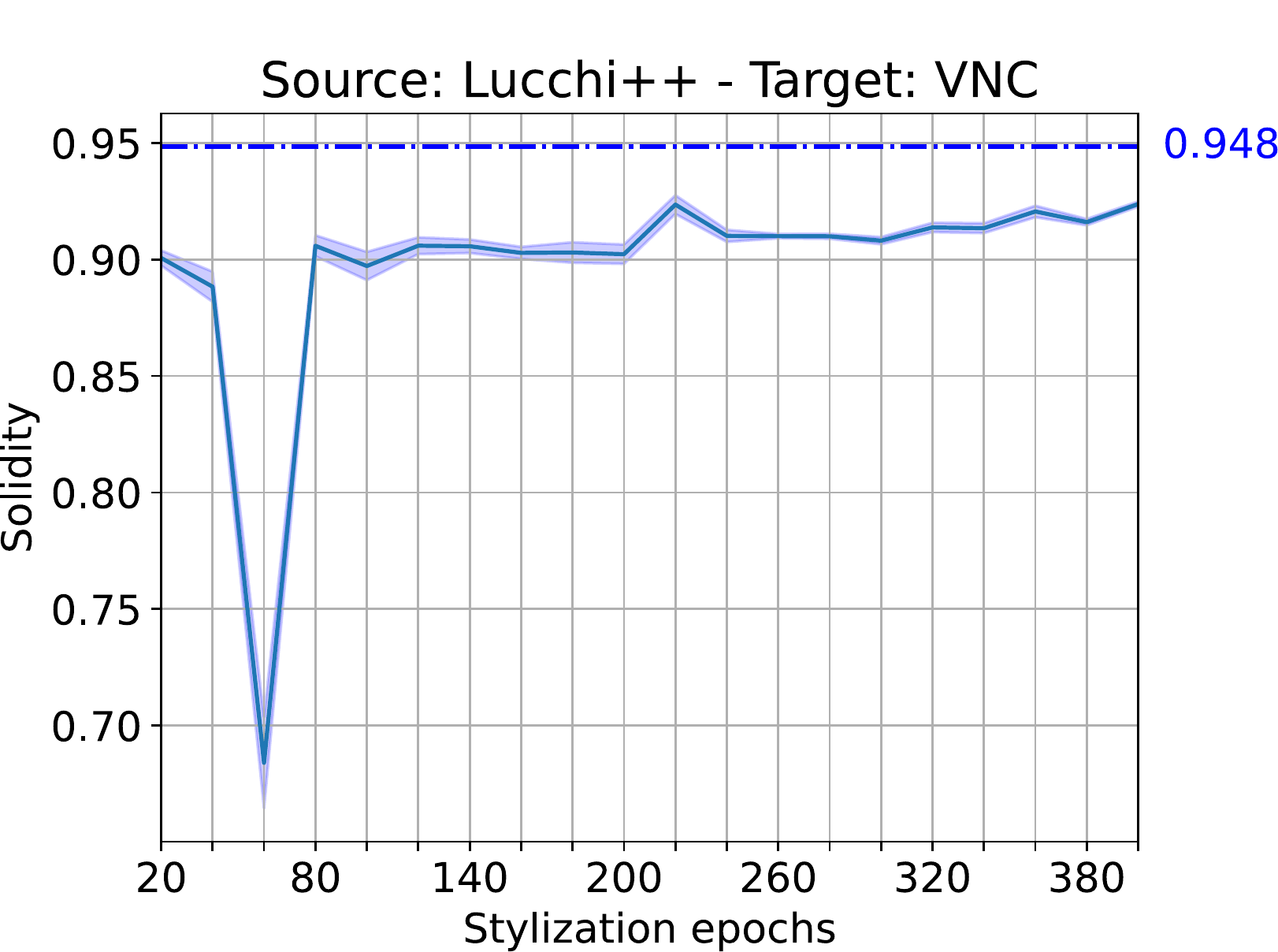}}
\end{minipage}
\begin{minipage}{.5\linewidth}
\centering
\subfloat[]{\includegraphics[scale=.35]{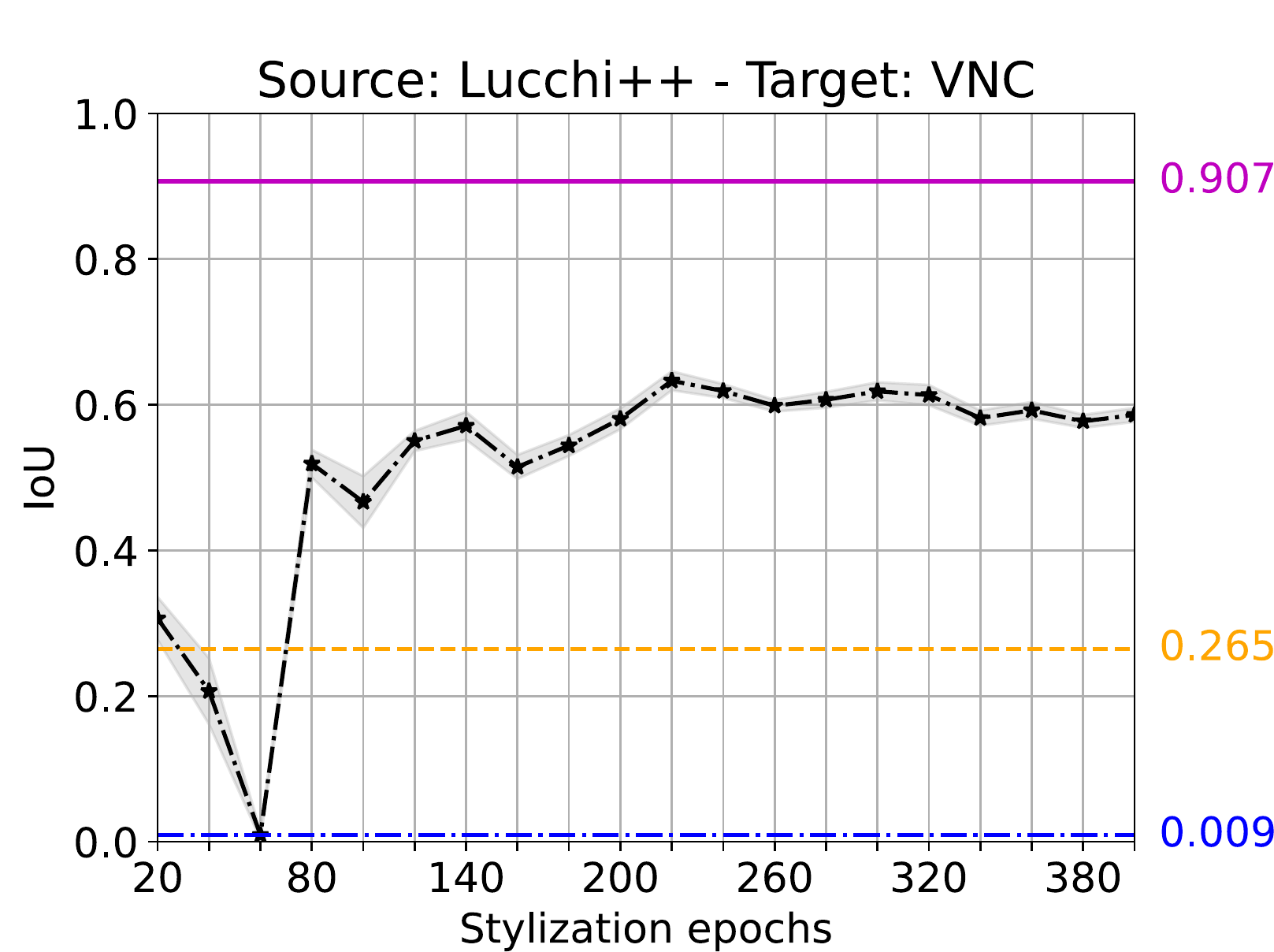}}
\end{minipage}

\caption{Relation between solidity and IoU in the style-transfer approach with Lucchi++ as source domain. On the left, the evolution of the solidity value (averaged for ten executions) as a function of the stylization epochs with (a) Kasthuri++ and (c) VNC as target domains (dashed lines represent the source solidity value). On the right, the evolution of the test IoU (also averaged over ten executions) as a function of the epochs with (b) Kasthuri++ and (d) VNC as target domains. The magenta lines represent the maximum IoU value obtained by the fully supervised baseline models. In contrast, the blue and orange lines represent the IoU values obtained by the baseline methods applied without adaptation and after histogram matching to the target datasets, respectively.}
\end{figure}

\begin{figure}[ht]
\begin{minipage}{.5\linewidth}
\centering
\subfloat[]{\includegraphics[scale=.35]{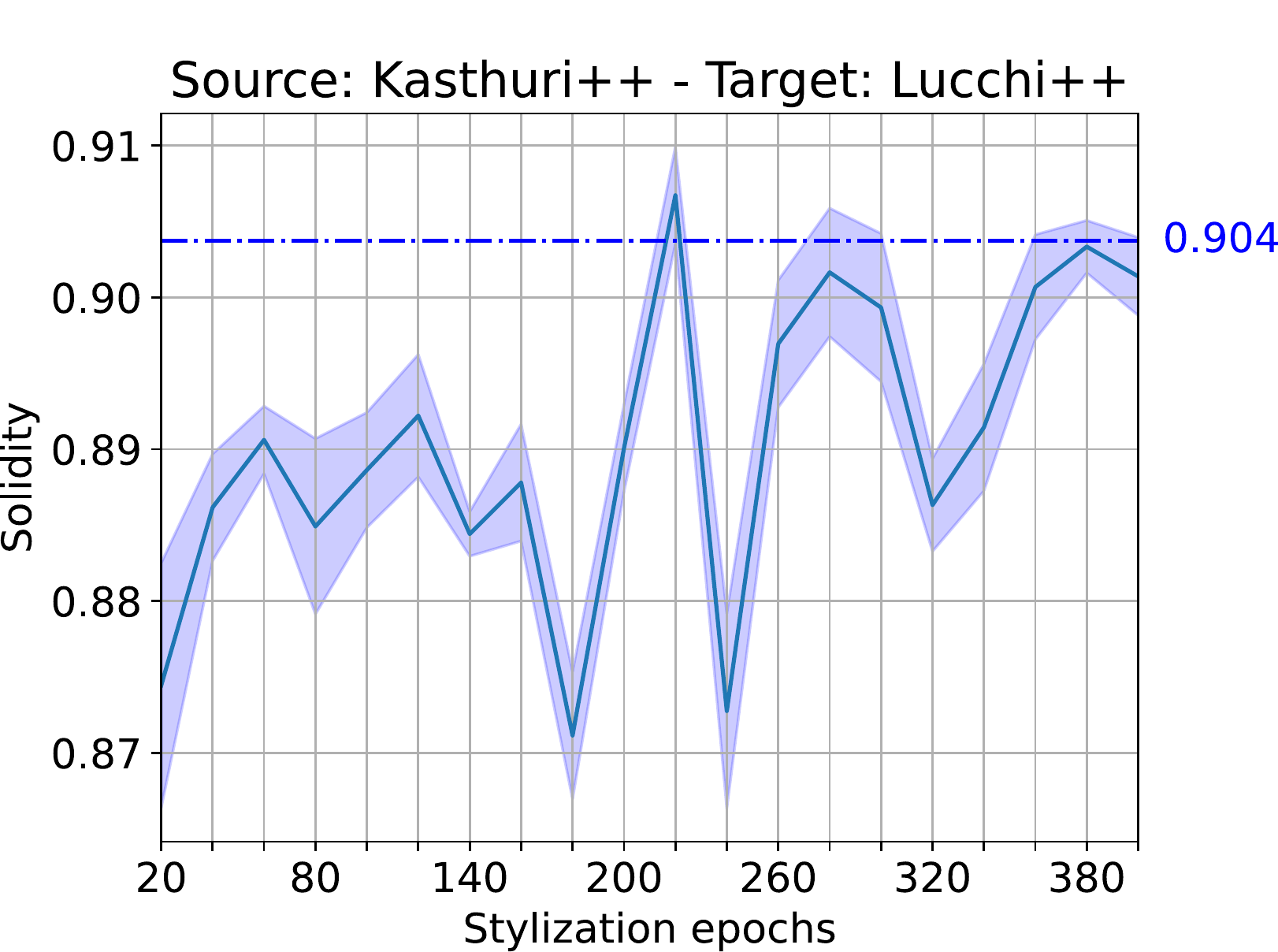}}
\end{minipage}
\begin{minipage}{.5\linewidth}
\centering
\subfloat[]{\includegraphics[scale=.35]{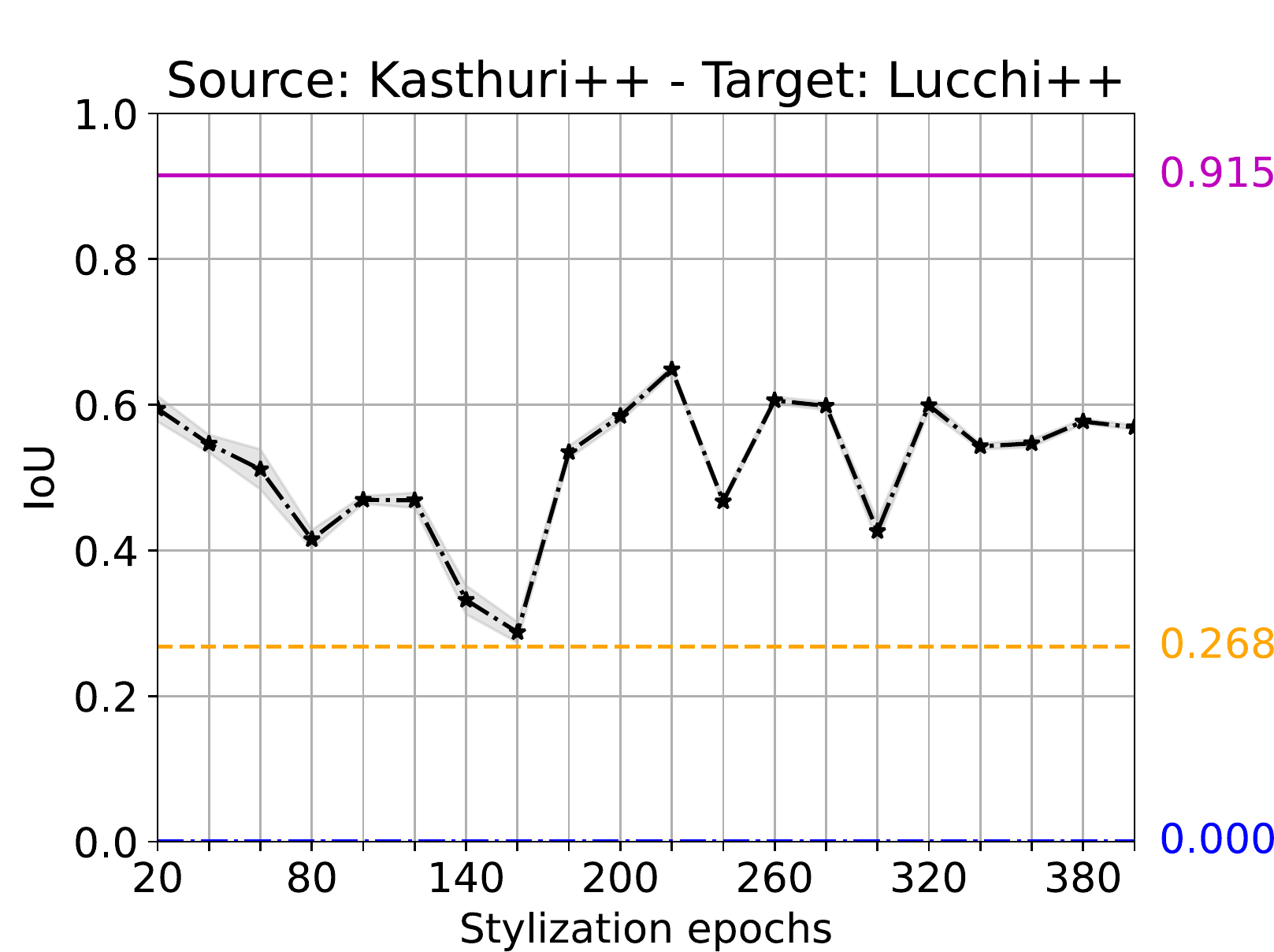}}
\end{minipage}\par\medskip
\begin{minipage}{.5\linewidth}
\centering
\subfloat[]{\includegraphics[scale=.35]{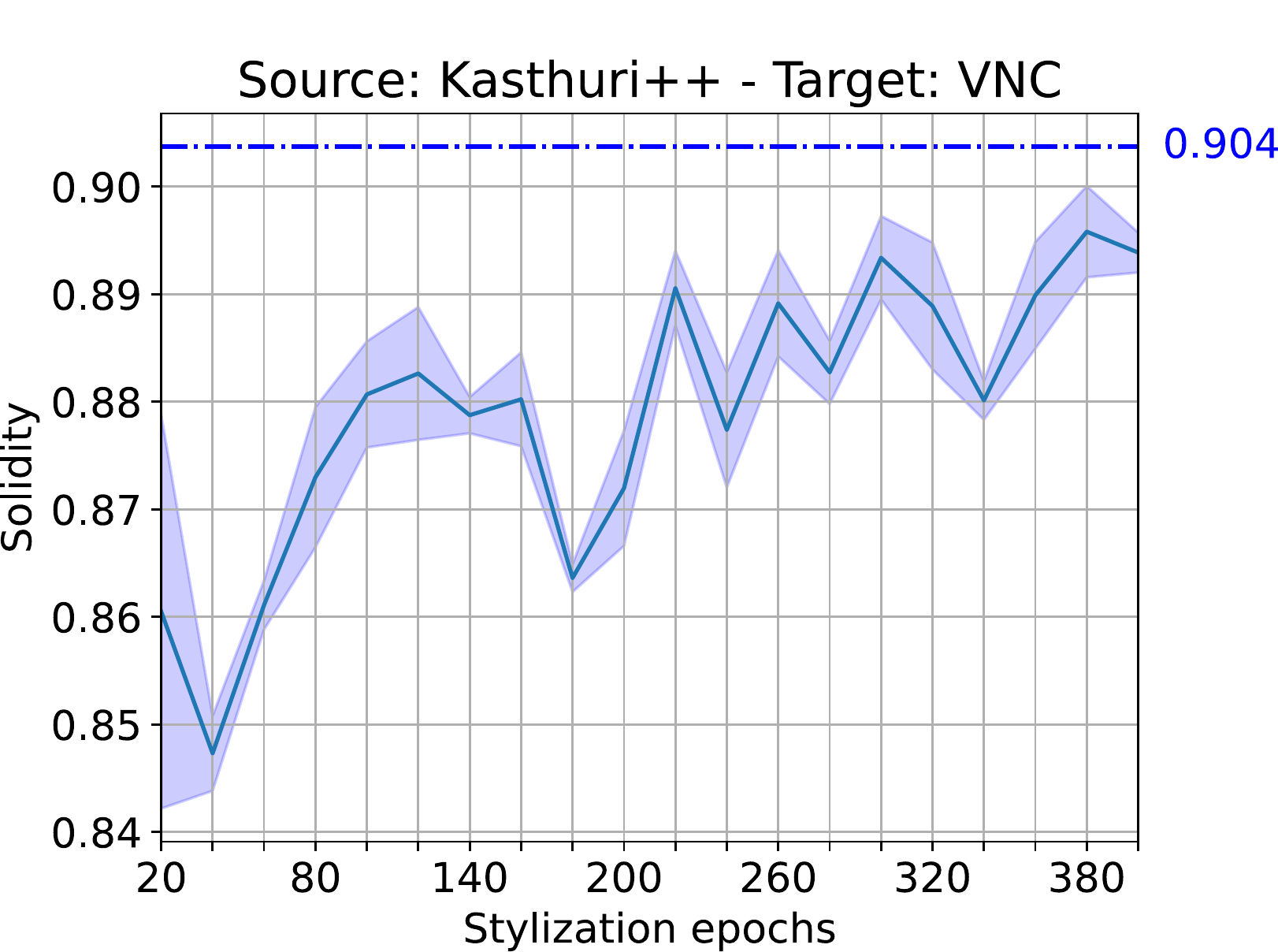}}
\end{minipage}
\begin{minipage}{.5\linewidth}
\centering
\subfloat[]{\includegraphics[scale=.35]{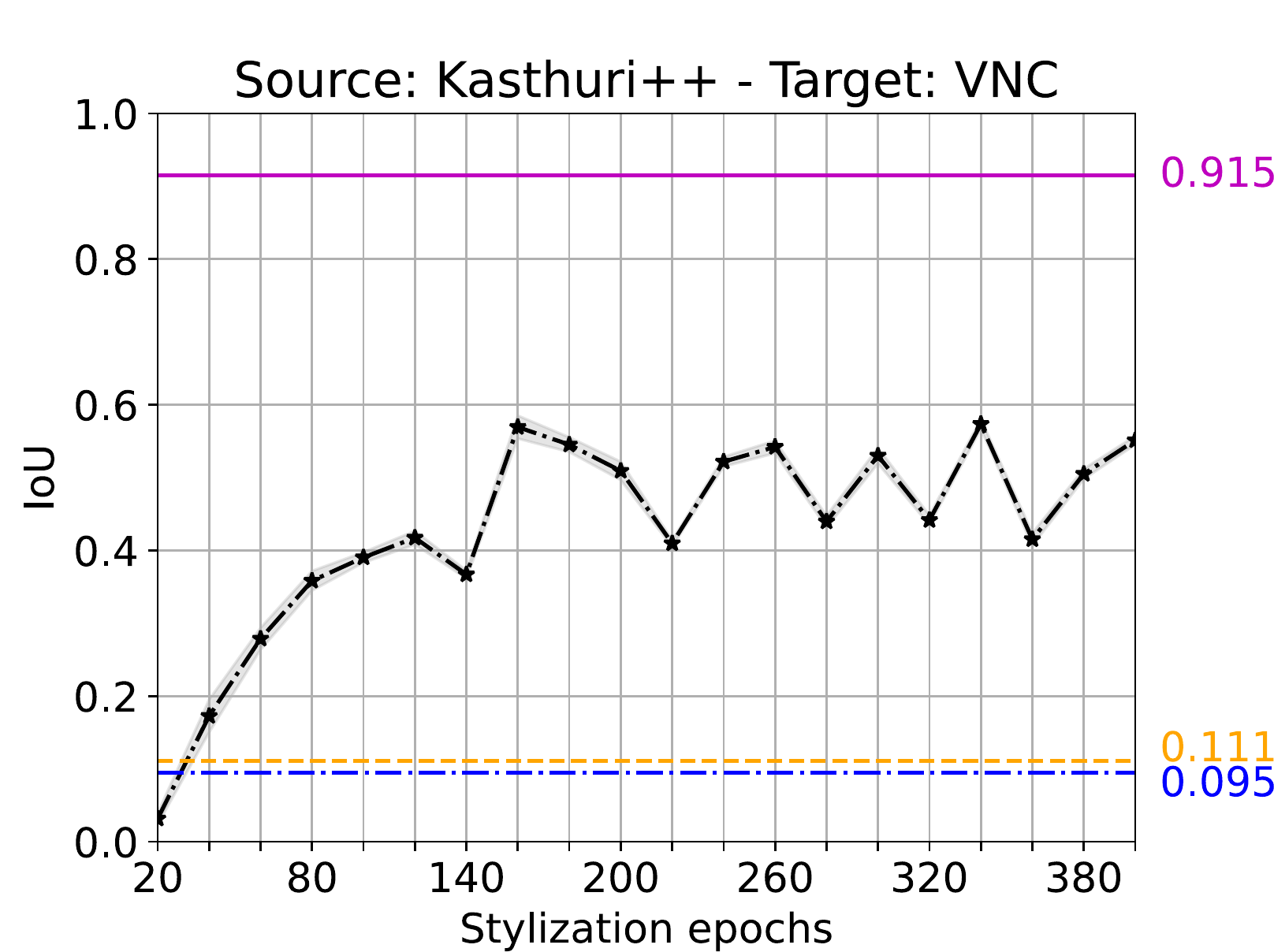}}
\end{minipage}\par\medskip

\caption{Relation between solidity and IoU in the style-transfer approach with Kasthuri++ as source domain. On the left, the evolution of the solidity value (averaged for ten executions) as a function of the stylization epochs with (a) Lucchi++ and (c) VNC as target domains (dashed lines represent the source solidity value). On the right, the evolution of the test IoU (also averaged over ten executions) as a function of the epochs with (b) Lucchi++ and (d) VNC as target domains. The magenta lines represent the maximum IoU value obtained by the fully supervised baseline models. In contrast, the blue and orange lines represent the IoU values obtained by the baseline methods applied without adaptation and after histogram matching to the target datasets, respectively.}
\end{figure}

\begin{figure}[ht]
\begin{minipage}{.5\linewidth}
\centering
\subfloat[]{\includegraphics[scale=.35]{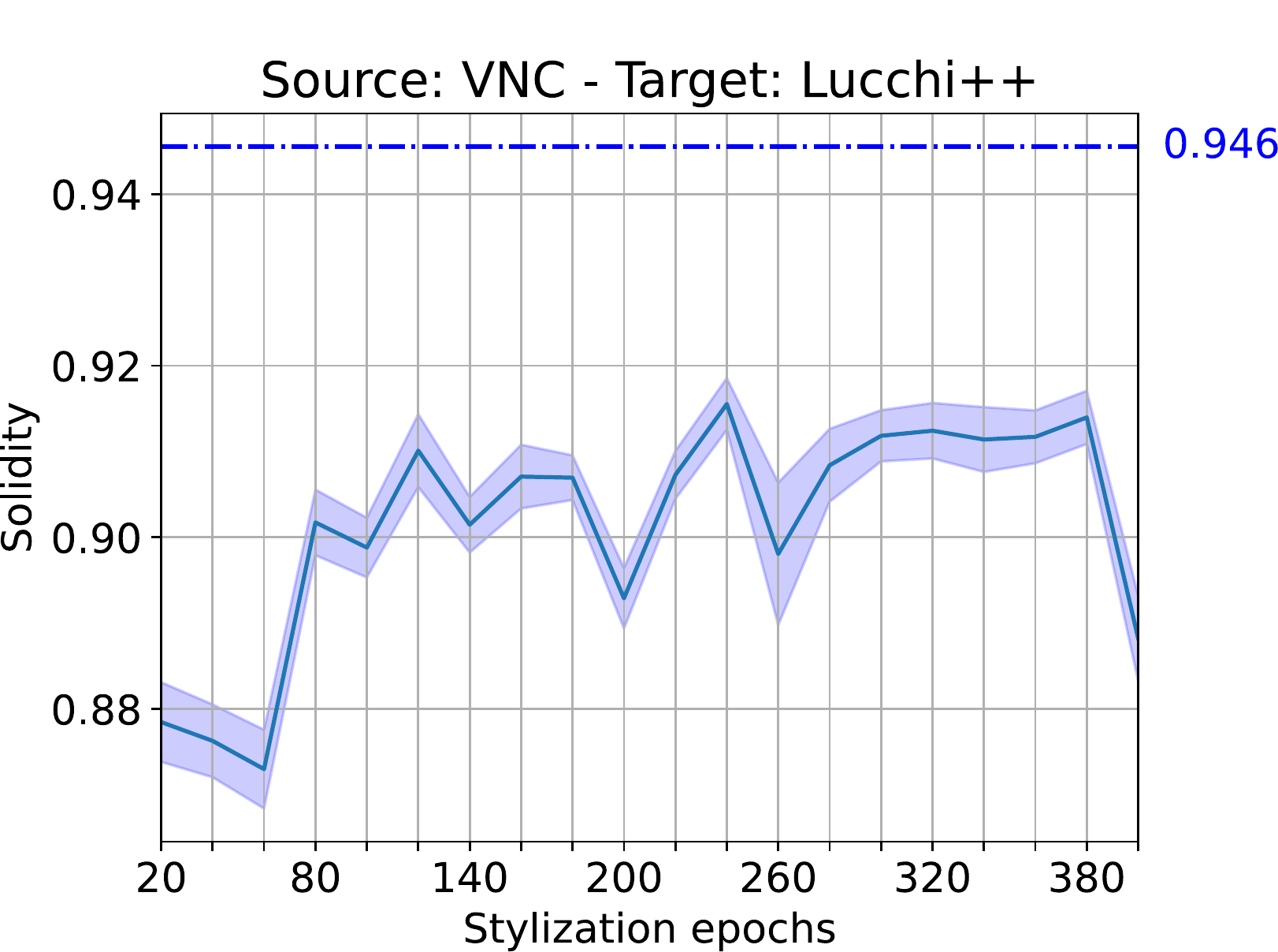}}
\end{minipage}
\begin{minipage}{.5\linewidth}
\centering
\subfloat[]{\includegraphics[scale=.35]{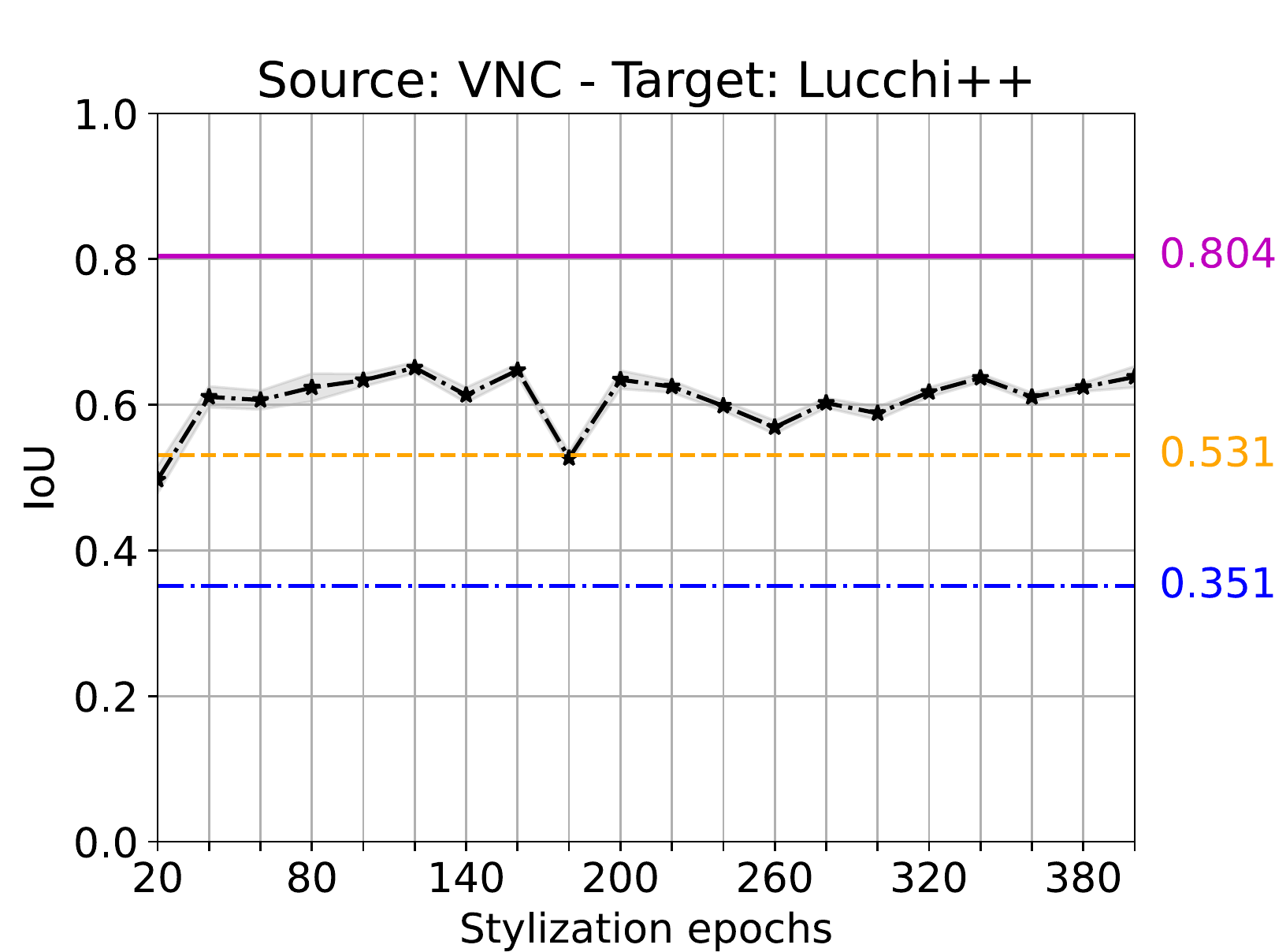}}
\end{minipage}\par\medskip
\begin{minipage}{.5\linewidth}
\centering
\subfloat[]{\includegraphics[scale=.35]{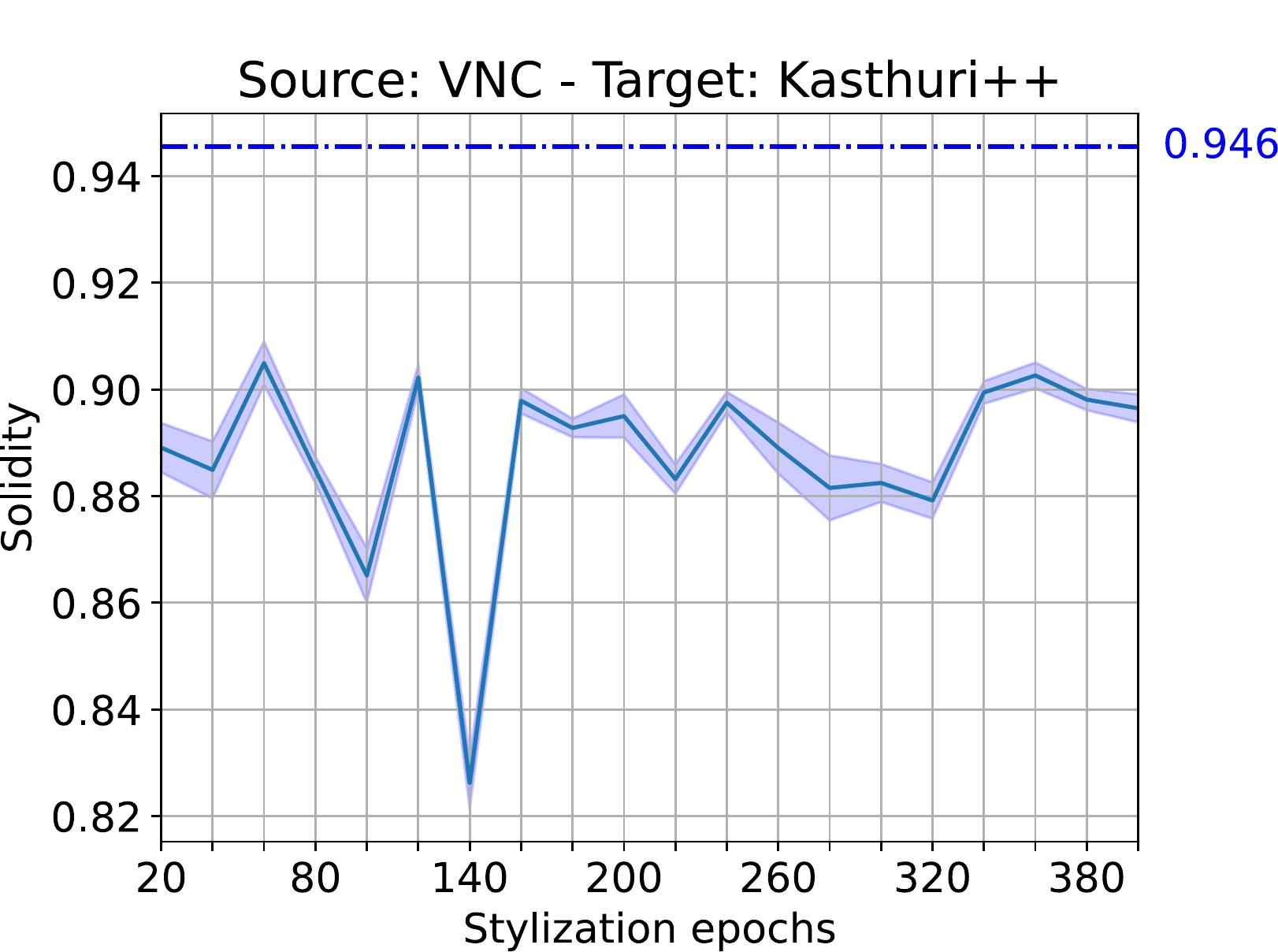}}
\end{minipage}%
\begin{minipage}{.5\linewidth}
\centering
\subfloat[]{\includegraphics[scale=.35]{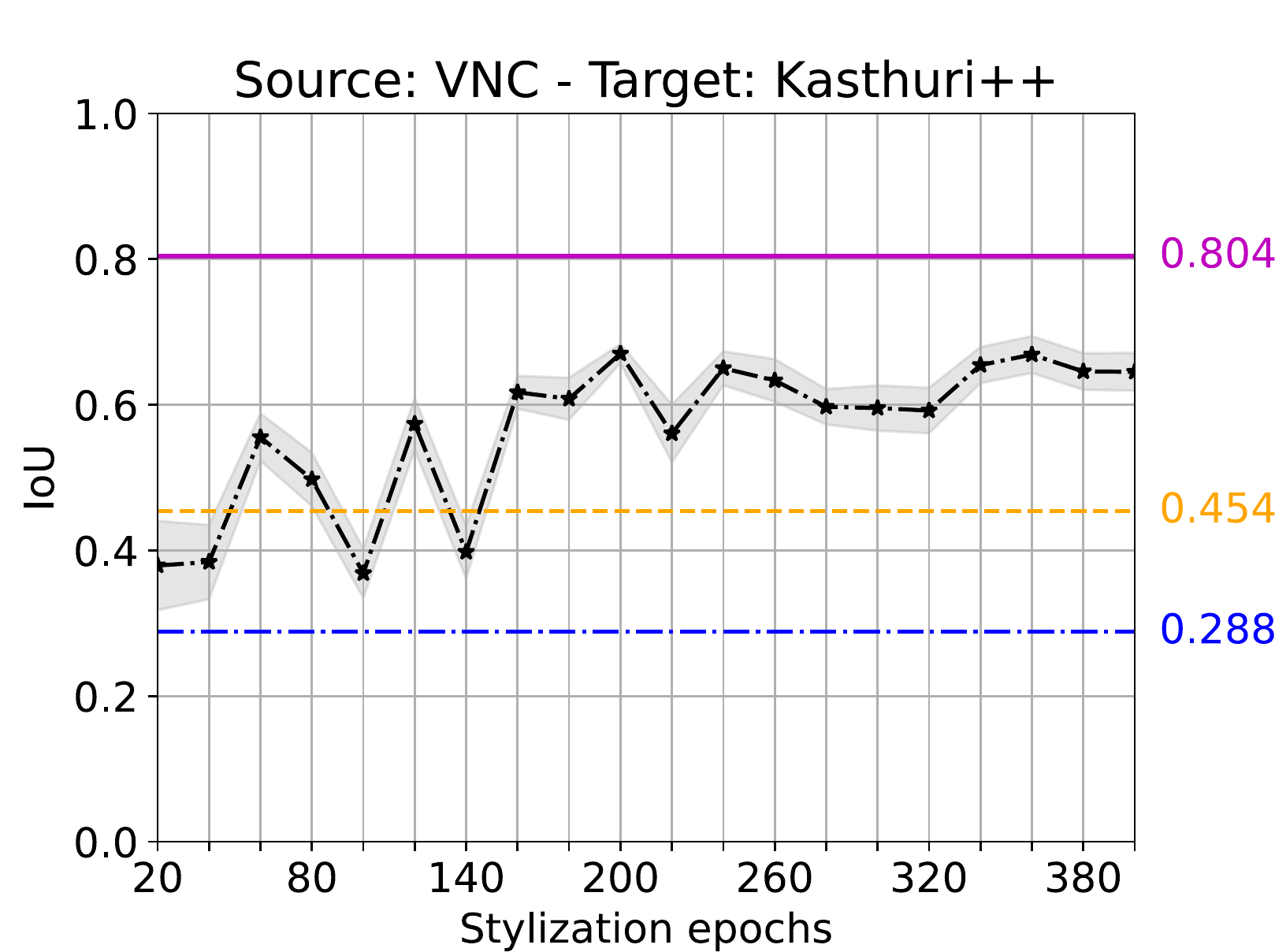}}
\end{minipage}\par\medskip
\caption{Relation between solidity and IoU in the style-transfer approach with VNC as source domain. On the left, the evolution of the solidity value (averaged for ten executions) as a function of the stylization epochs with (a) Lucchi++ and (c) Kasthuri++ as target domains (dashed lines represent the source solidity value). On the right, the the test IoU evolution ( averaged over ten executions) as a function of the epochs with (b) Lucchi++ and (d) Kasthuri++ as target domains. The magenta lines represent the maximum IoU value obtained by the fully supervised baseline models. In contrast, the blue and orange lines represent the IoU values obtained by the baseline methods applied without adaptation and after histogram matching to the target datasets, respectively.}
\end{figure}

\begin{figure}[ht]
\begin{minipage}{.5\linewidth}
\centering
\subfloat[]{\includegraphics[scale=.29]{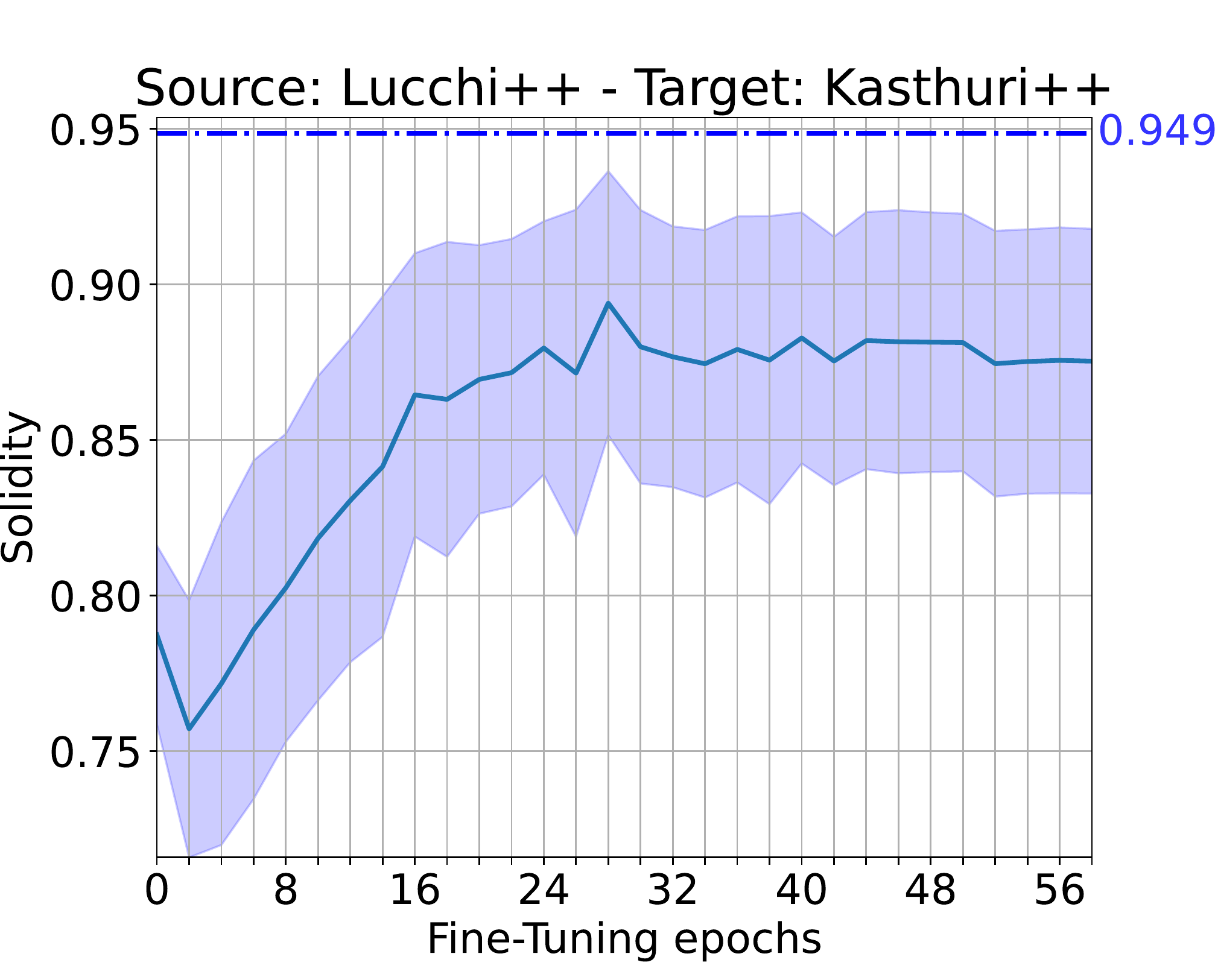}}
\end{minipage}
\begin{minipage}{.5\linewidth}
\centering
\subfloat[]{\includegraphics[scale=.29]{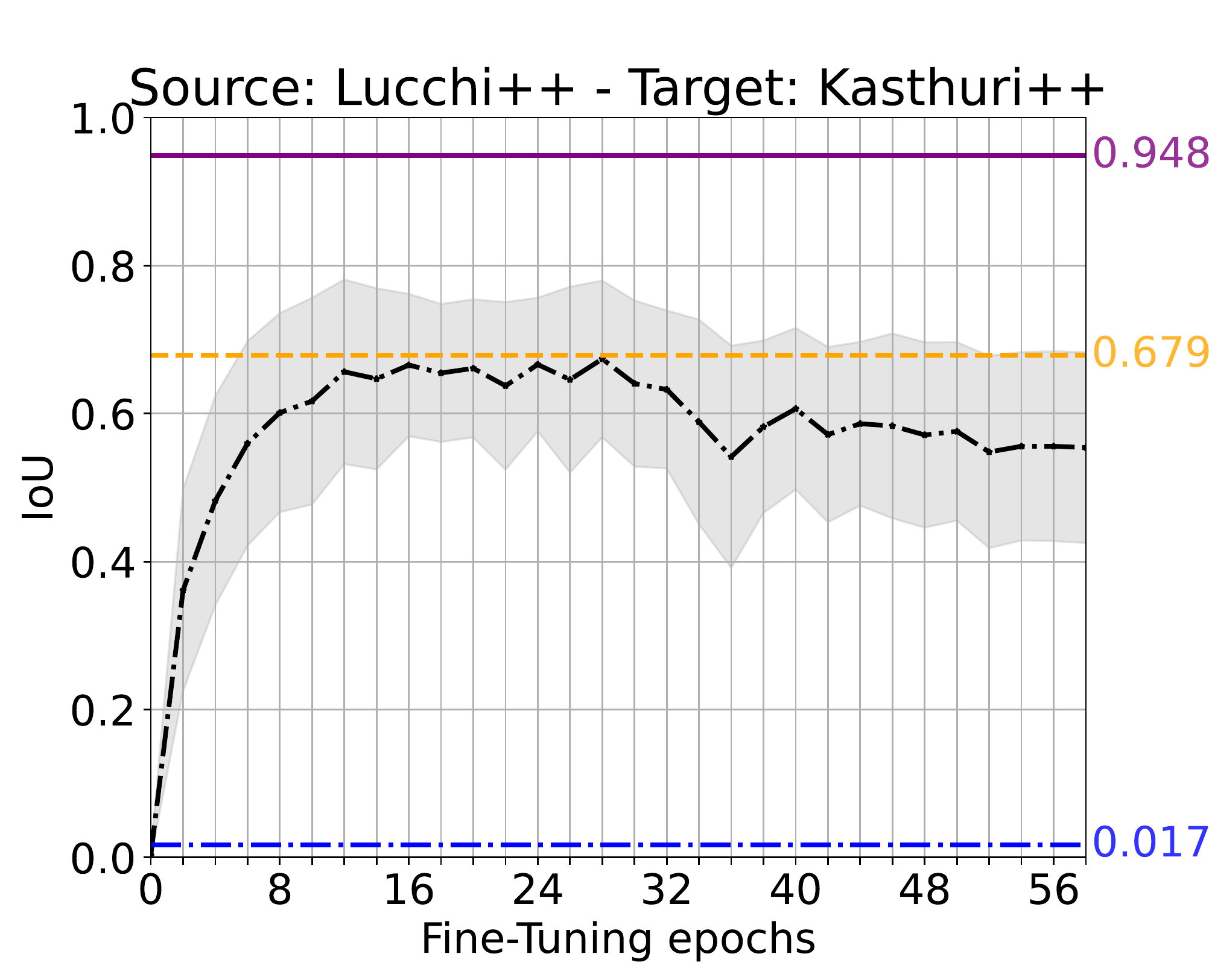}}
\end{minipage}\par\medskip
\begin{minipage}{.5\linewidth}
\centering
\subfloat[]{\includegraphics[scale=.29]{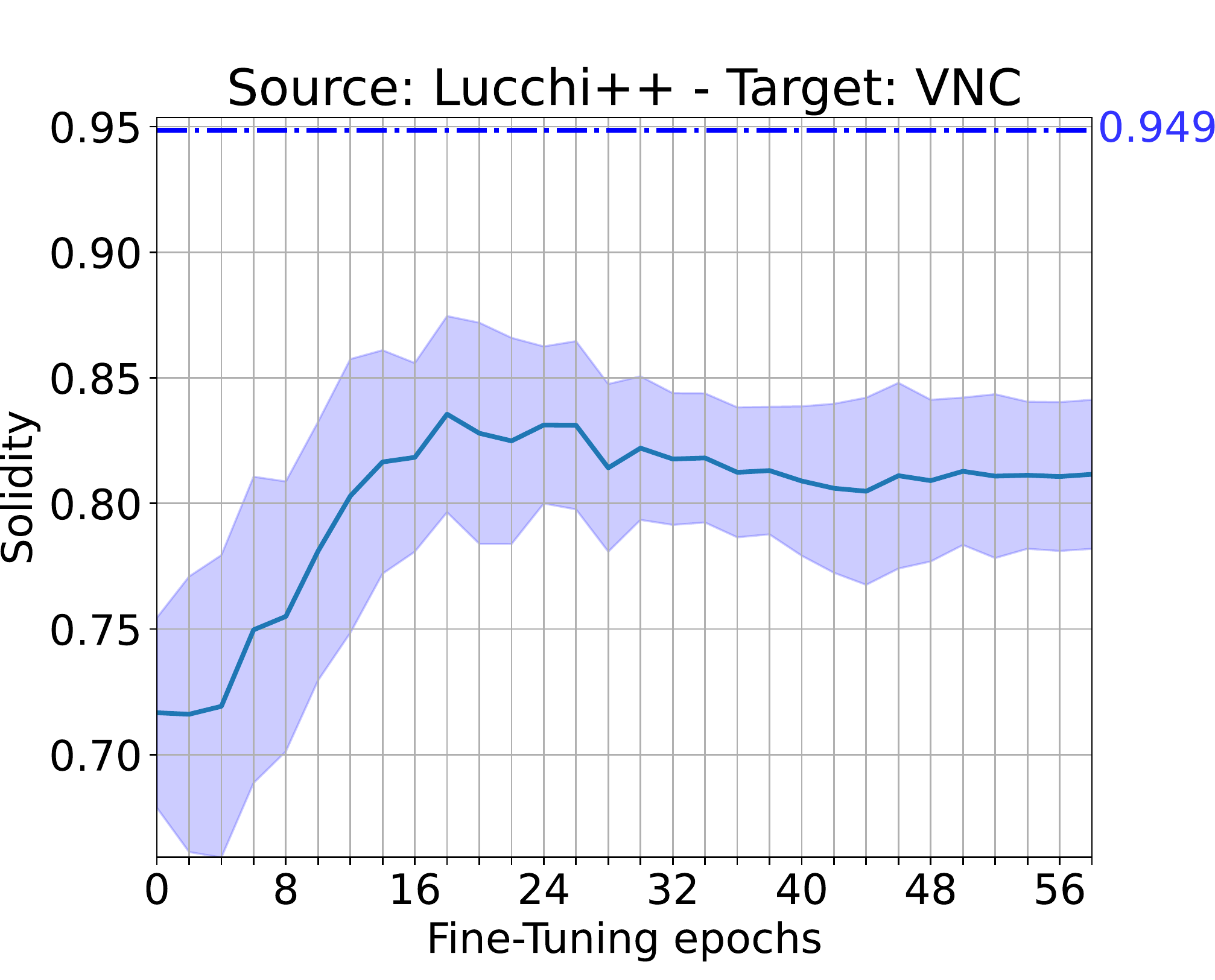}}
\end{minipage}
\begin{minipage}{.5\linewidth}
\centering
\subfloat[]{\includegraphics[scale=.29]{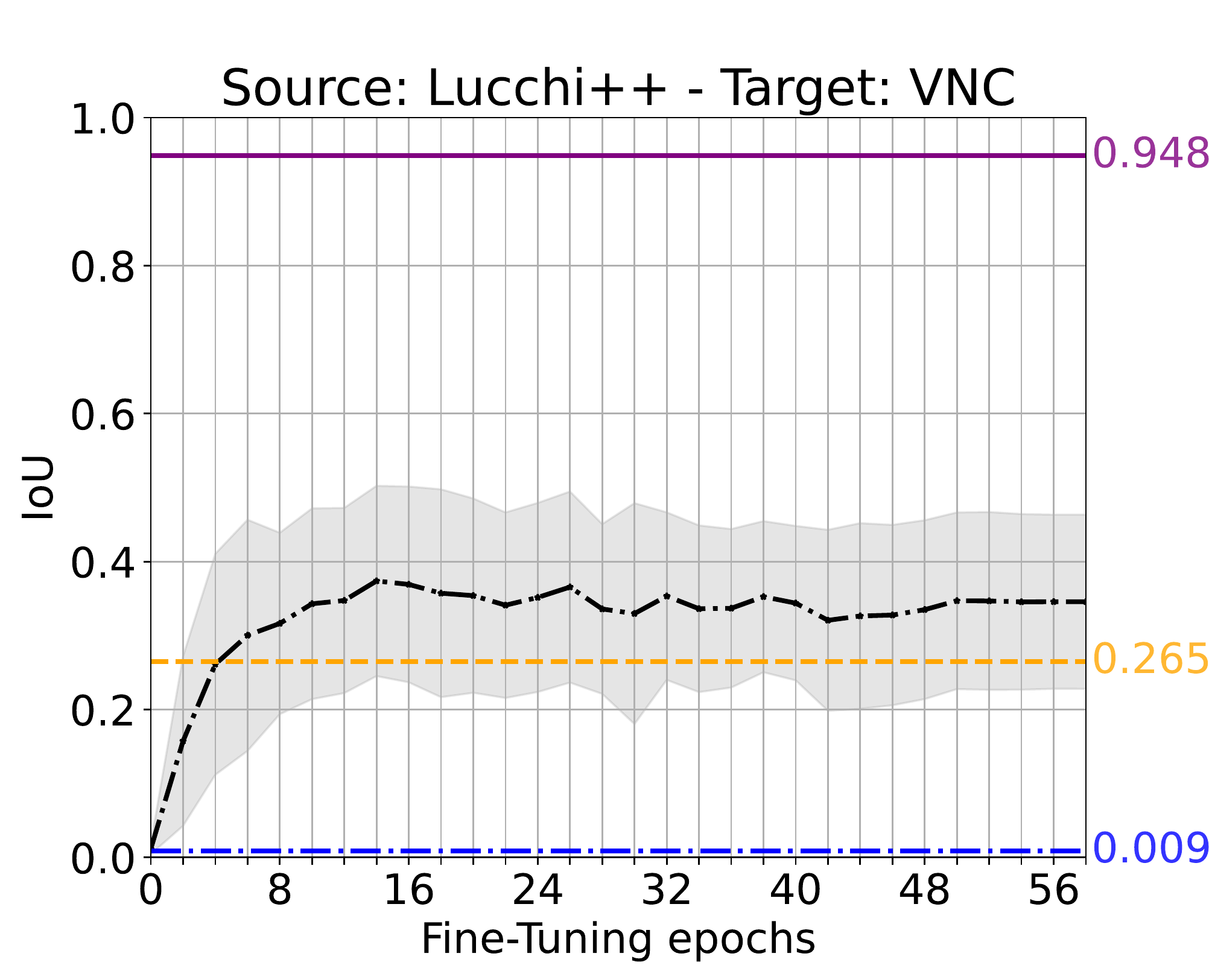}}
\end{minipage}

\caption{Relation between solidity and IoU in the self-supervised approach with Lucchi++ as source domain. On the left, the evolution of the solidity value (averaged for ten executions) as a function of the stylization epochs with (a) Kasthuri++ and (c) VNC as target domains (dashed lines represent the source solidity value). On the right, the evolution of the test IoU (also averaged over ten executions) as a function of the epochs with (b) Kasthuri++ and (d) VNC as target domains. The magenta lines represent the maximum IoU value obtained by the fully supervised baseline models. In contrast, the blue and orange lines represent the IoU values obtained by the baseline methods applied without adaptation and after histogram matching to the target datasets, respectively.}
\end{figure}

\begin{figure}[ht]
\begin{minipage}{.5\linewidth}
\centering
\subfloat[]{\includegraphics[scale=.29]{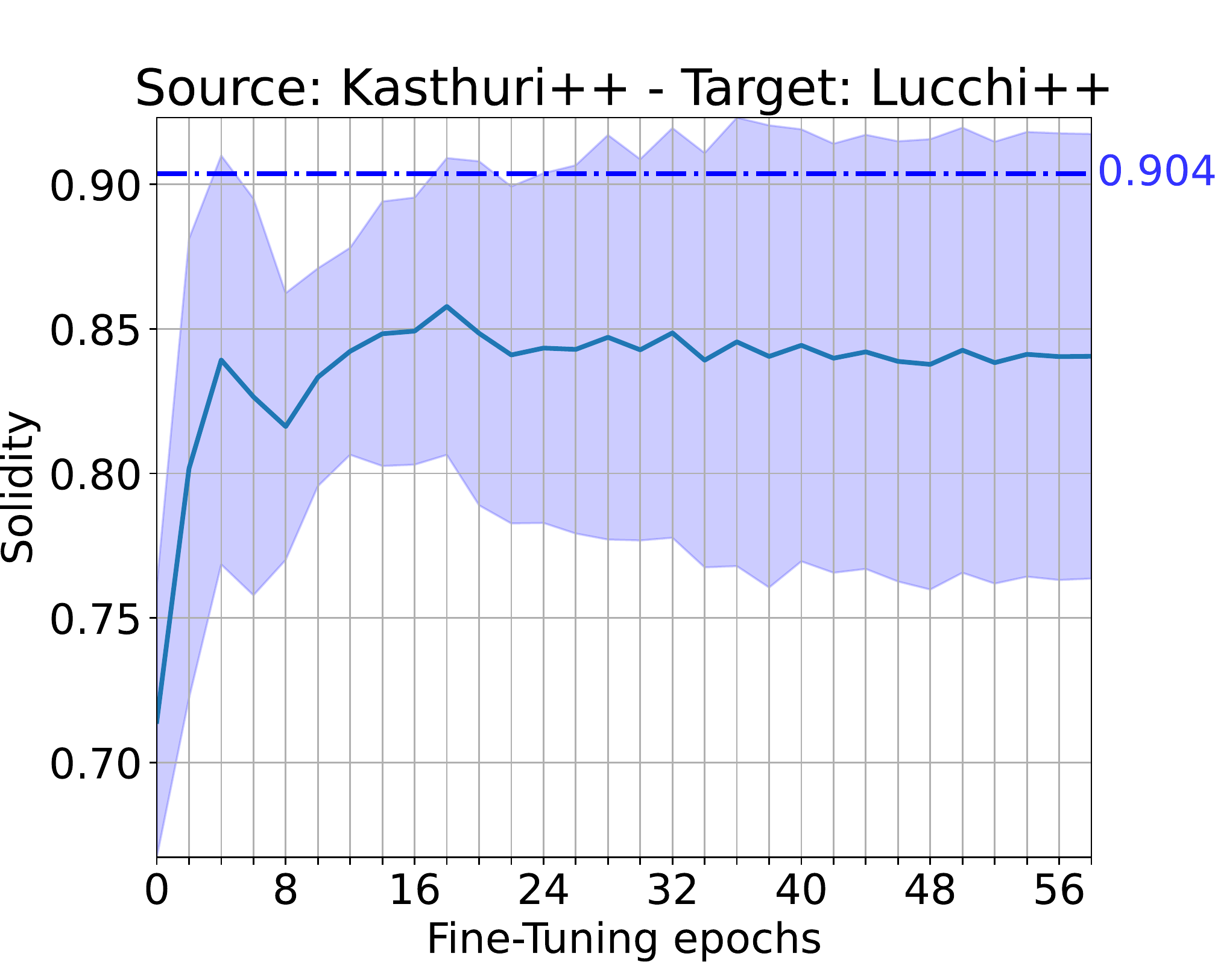}}
\end{minipage}
\begin{minipage}{.5\linewidth}
\centering
\subfloat[]{\includegraphics[scale=.29]{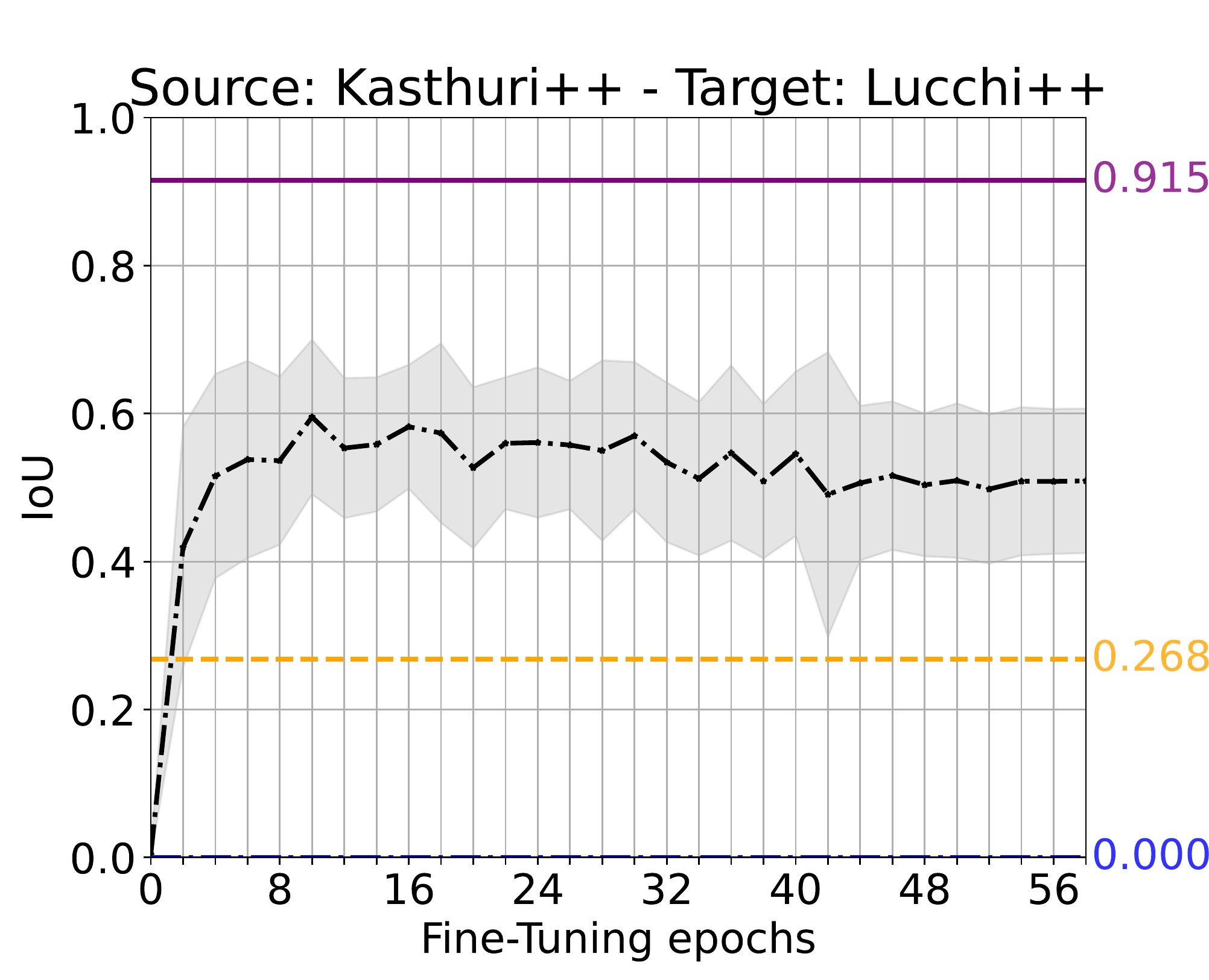}}
\end{minipage}\par\medskip
\begin{minipage}{.5\linewidth}
\centering
\subfloat[]{\includegraphics[scale=.29]{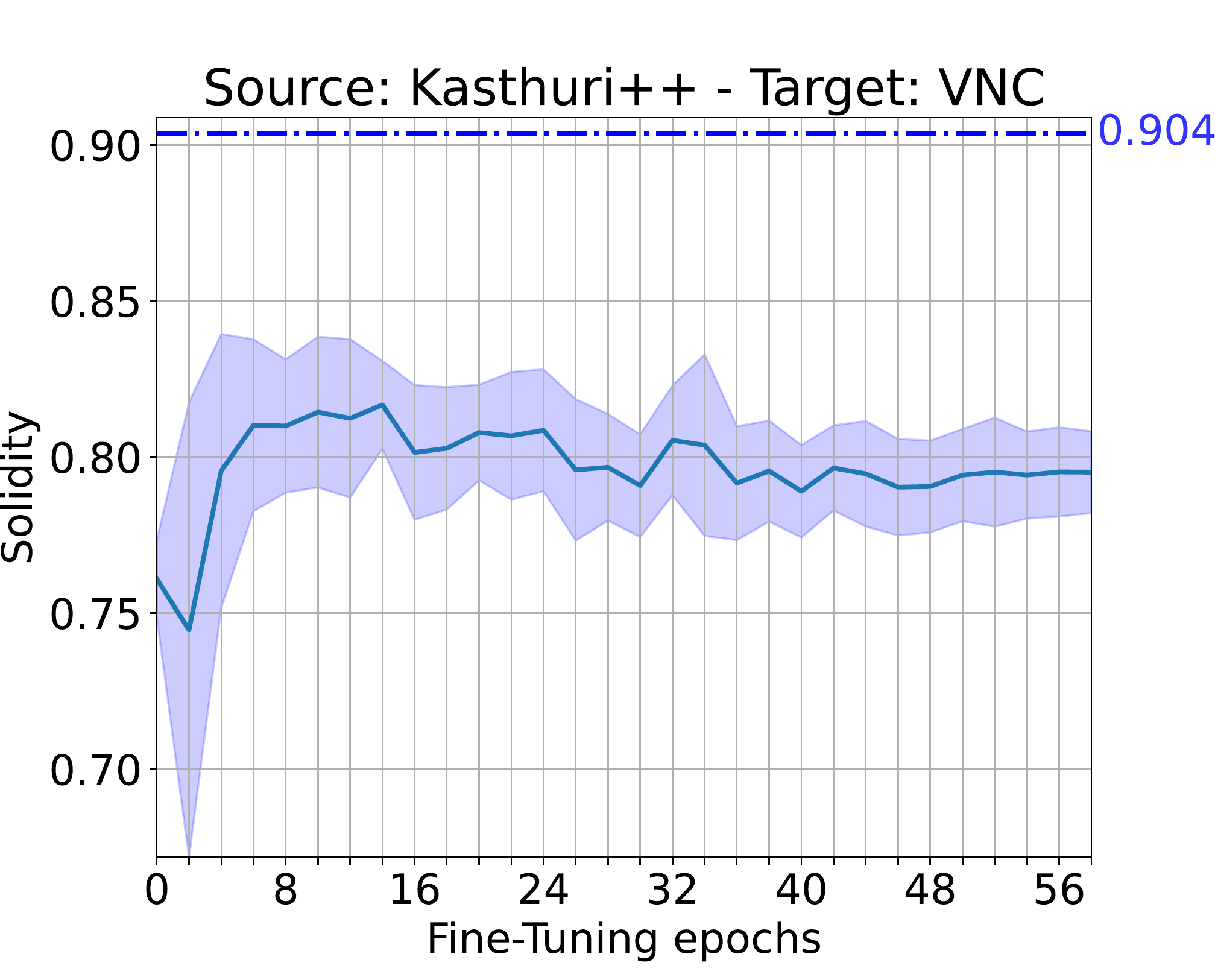}}
\end{minipage}
\begin{minipage}{.5\linewidth}
\centering
\subfloat[]{\includegraphics[scale=.29]{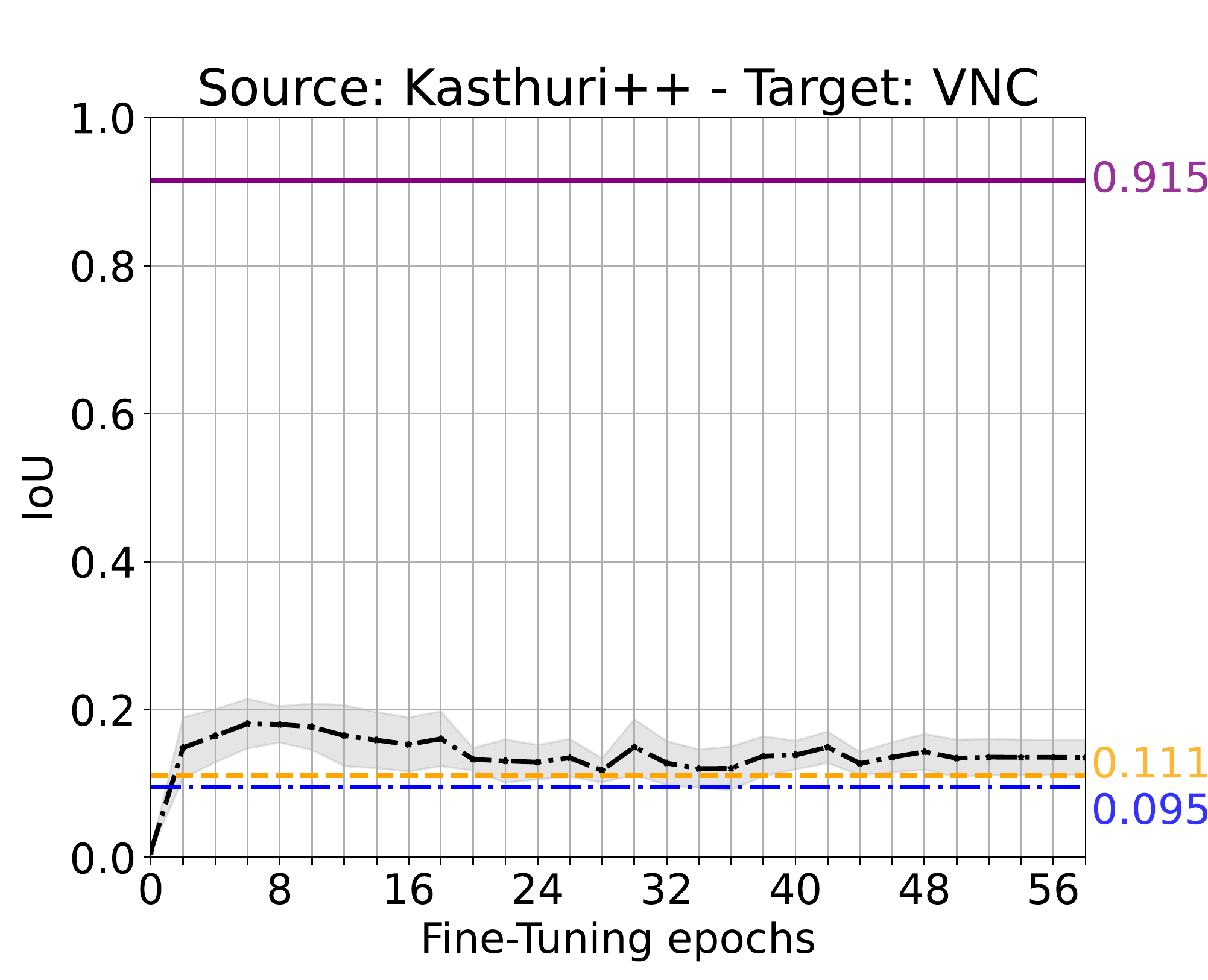}}
\end{minipage}\par\medskip

\caption{Relation between solidity and IoU in the self-supervised approach with Kasthuri++ as source domain. On the left, the evolution of the solidity value (averaged for ten executions) as a function of the stylization epochs with (a) Kasthuri++ and (c) VNC as target domains (dashed lines represent the source solidity value). On the right, the evolution of the test IoU (also averaged over ten executions) as a function of the epochs with (b) Kasthuri++ and (d) VNC as target domains. The magenta lines represent the maximum IoU value obtained by the fully supervised baseline models. In contrast, the blue and orange lines represent the IoU values obtained by the baseline methods applied without adaptation and after histogram matching to the target datasets, respectively.}
\end{figure}

\begin{figure}[ht]
\begin{minipage}{.5\linewidth}
\centering
\subfloat[]{\includegraphics[scale=.29]{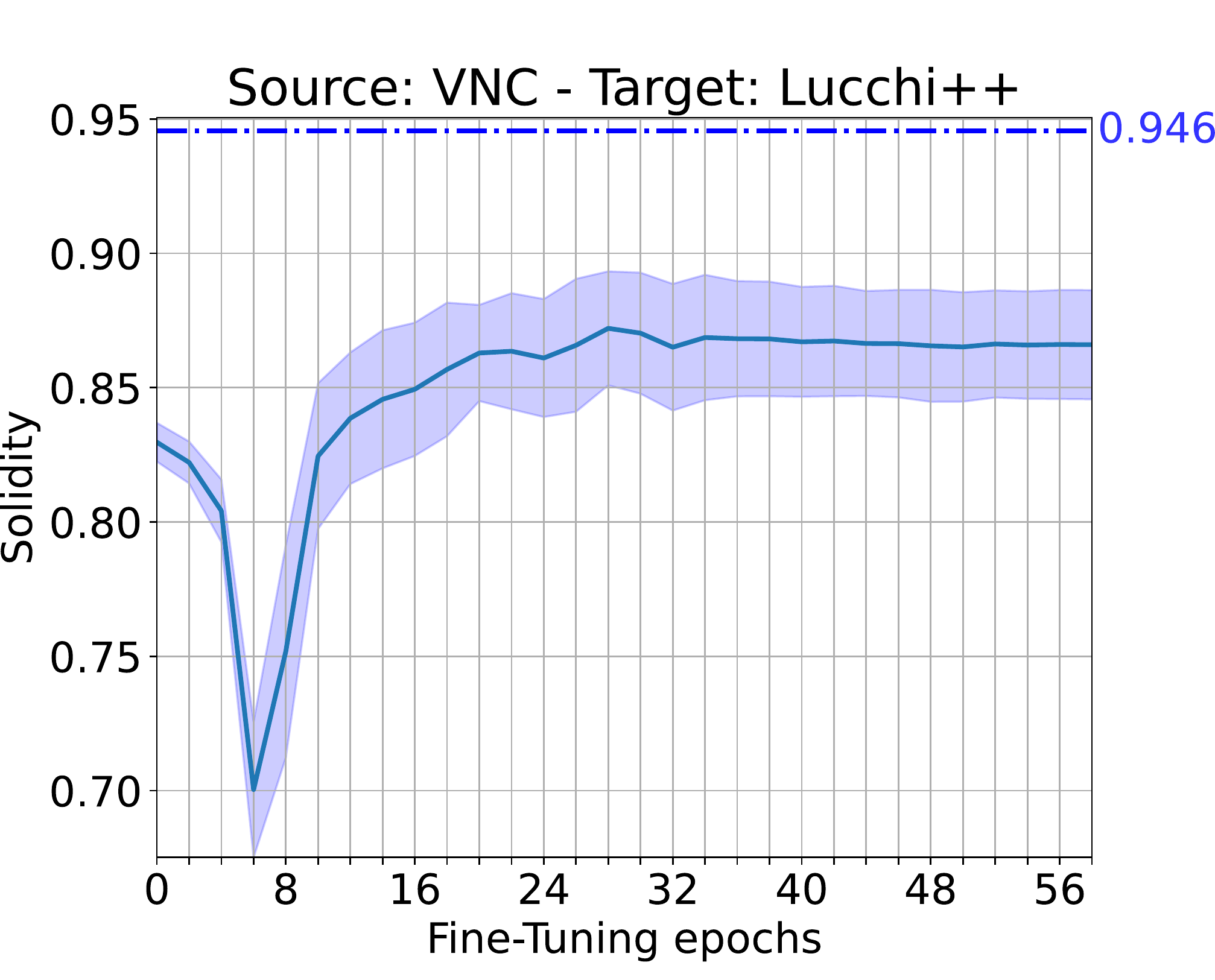}}
\end{minipage}
\begin{minipage}{.5\linewidth}
\centering
\subfloat[]{\includegraphics[scale=.29]{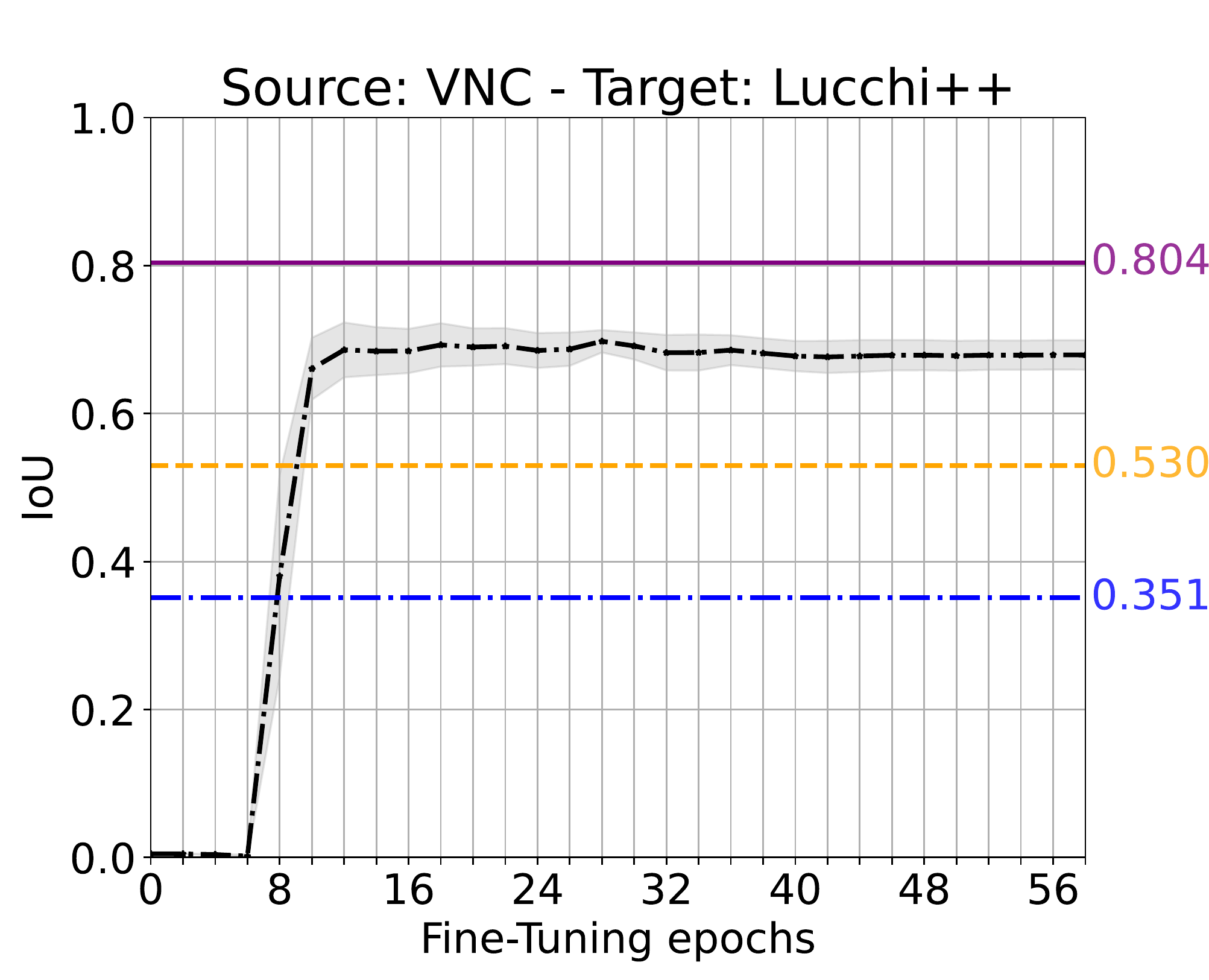}}
\end{minipage}\par\medskip
\begin{minipage}{.5\linewidth}
\centering
\subfloat[]{\includegraphics[scale=.29]{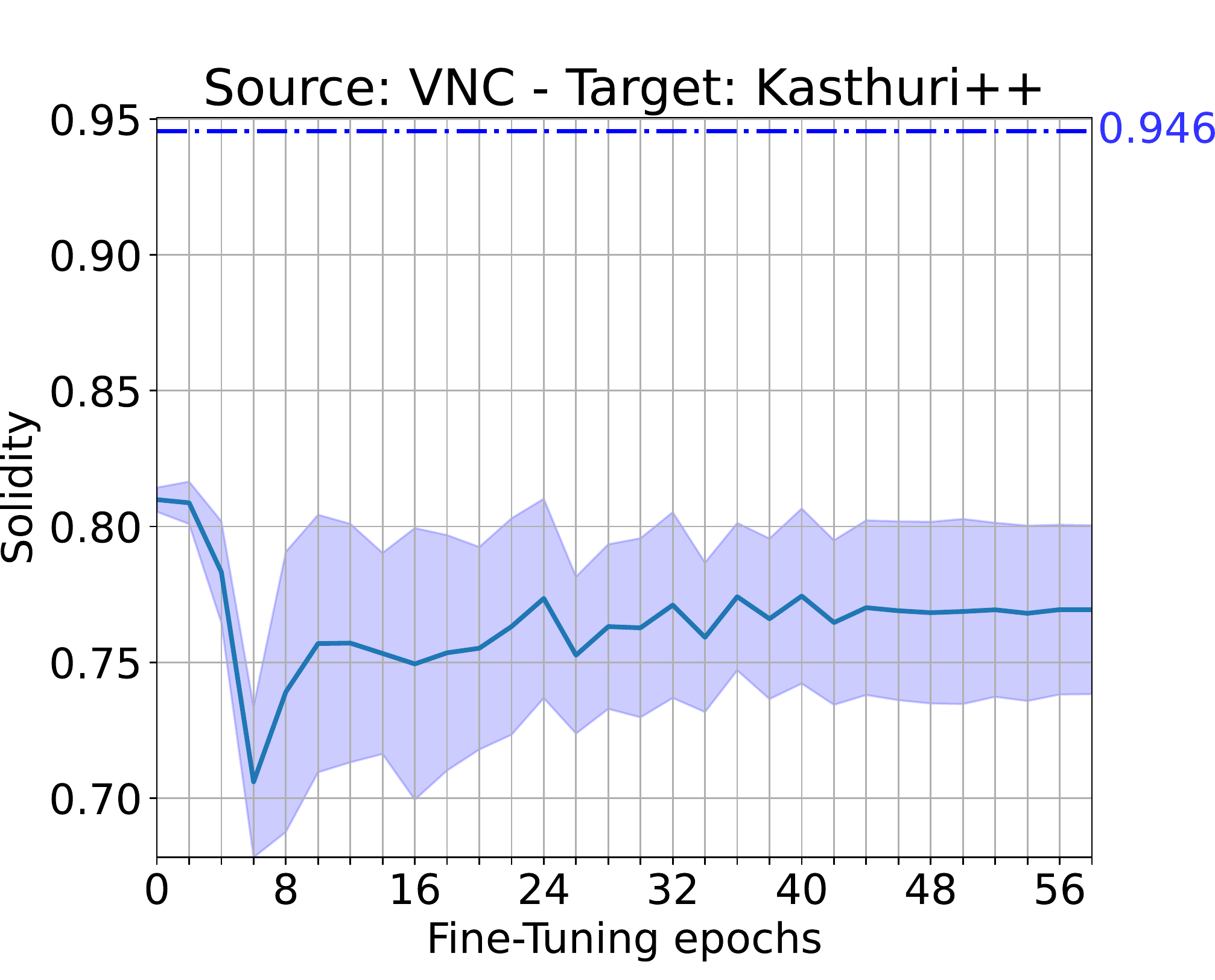}}
\end{minipage}%
\begin{minipage}{.5\linewidth}
\centering
\subfloat[]{\includegraphics[scale=.29]{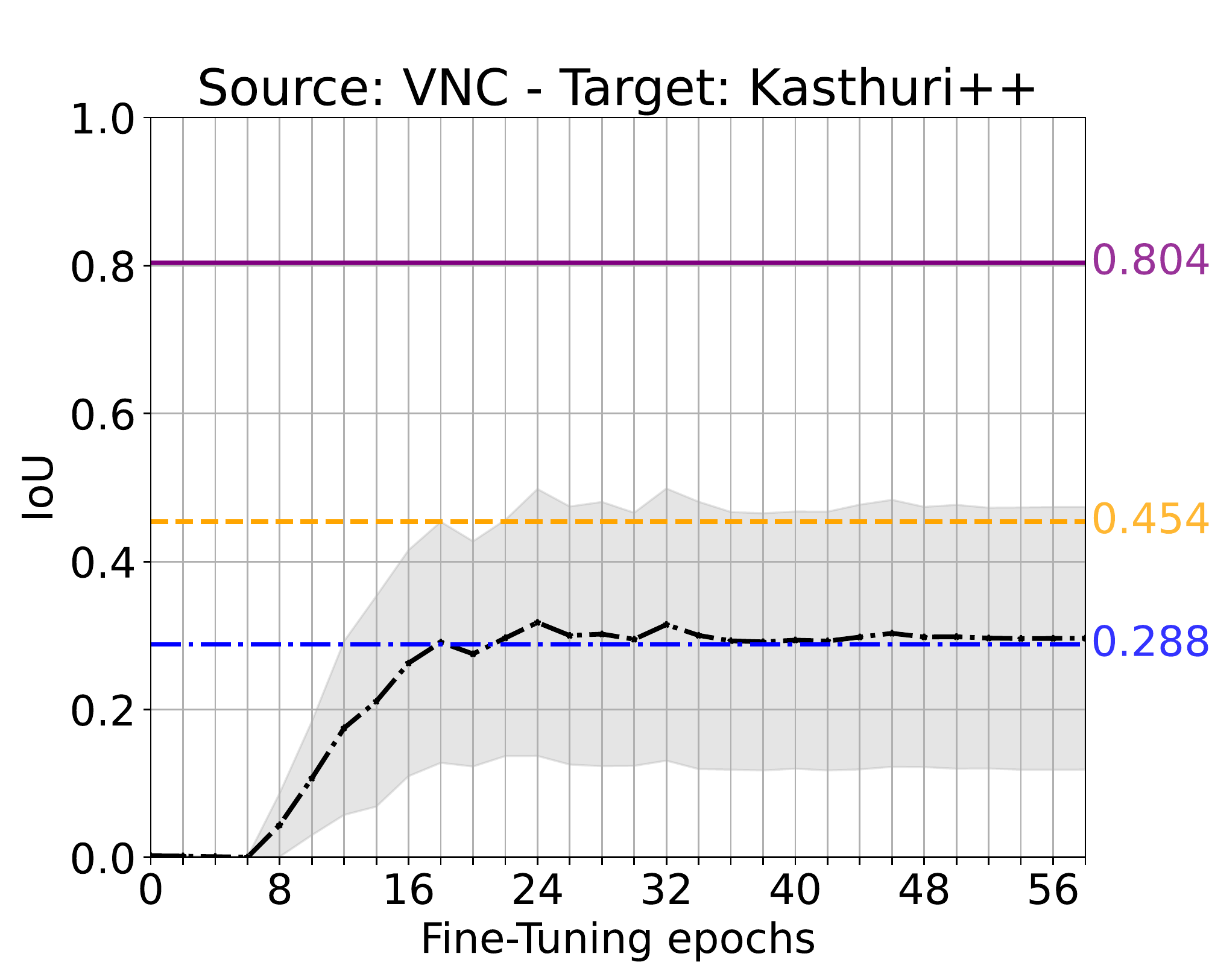}}
\end{minipage}\par\medskip
\caption{Relation between solidity and IoU in the self-supervised approach with VNC as source domain. On the left, the evolution of the solidity value (averaged for ten executions) as a function of the stylization epochs with (a) Lucchi++ and (c) Kasthuri++ as target domains (dashed lines represent the source solidity value). On the right, the test IoU evolution (averaged over ten executions) as a function of the epochs with (b) Lucchi++ and (d) Kasthuri++ as target domains. The magenta lines represent the maximum IoU value obtained by the fully supervised baseline models. In contrast, the blue and orange lines represent the IoU values obtained by the baseline methods applied without adaptation and after histogram matching to the target datasets, respectively.}
\end{figure}

\begin{figure}[ht]
    \begin{minipage}{.5\linewidth}
        \centering
        \subfloat[]{\includegraphics[scale=.39]{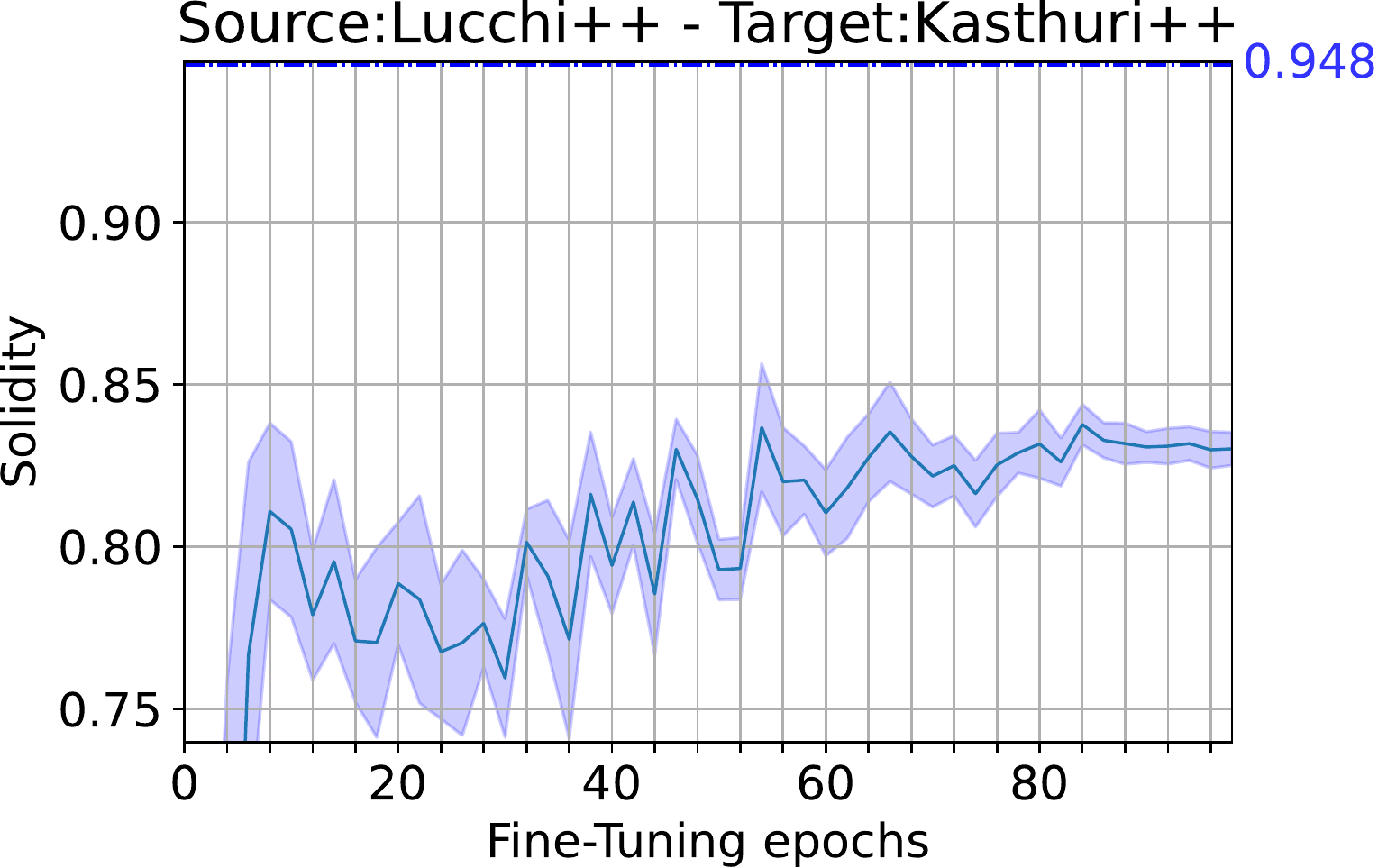}}
    \end{minipage}
    \begin{minipage}{.5\linewidth}
        \centering
        \subfloat[]{\includegraphics[scale=.39]{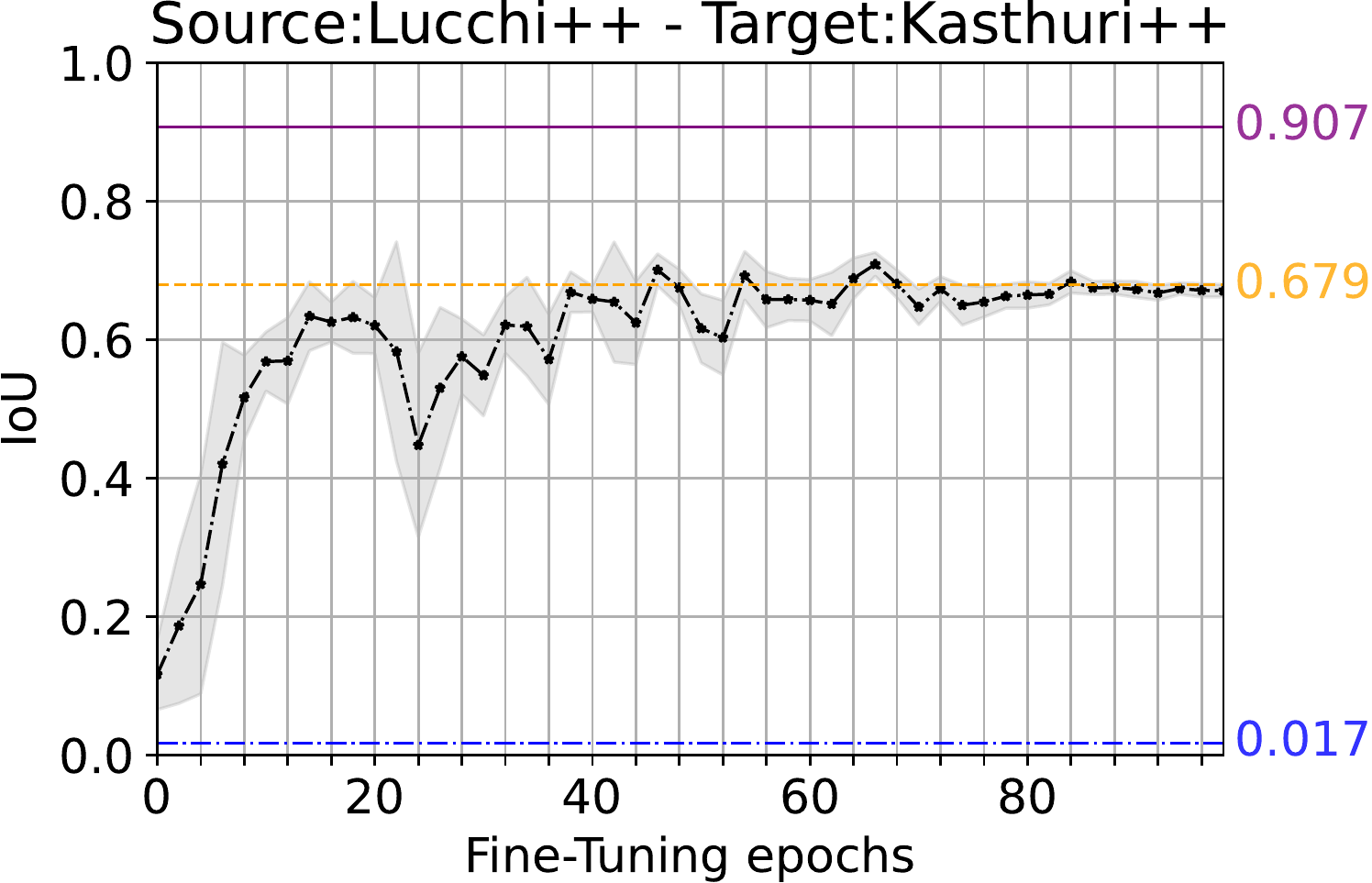}}
    \end{minipage}\par\medskip
    \begin{minipage}{.5\linewidth}
        \centering
        \subfloat[]{\includegraphics[scale=.39]{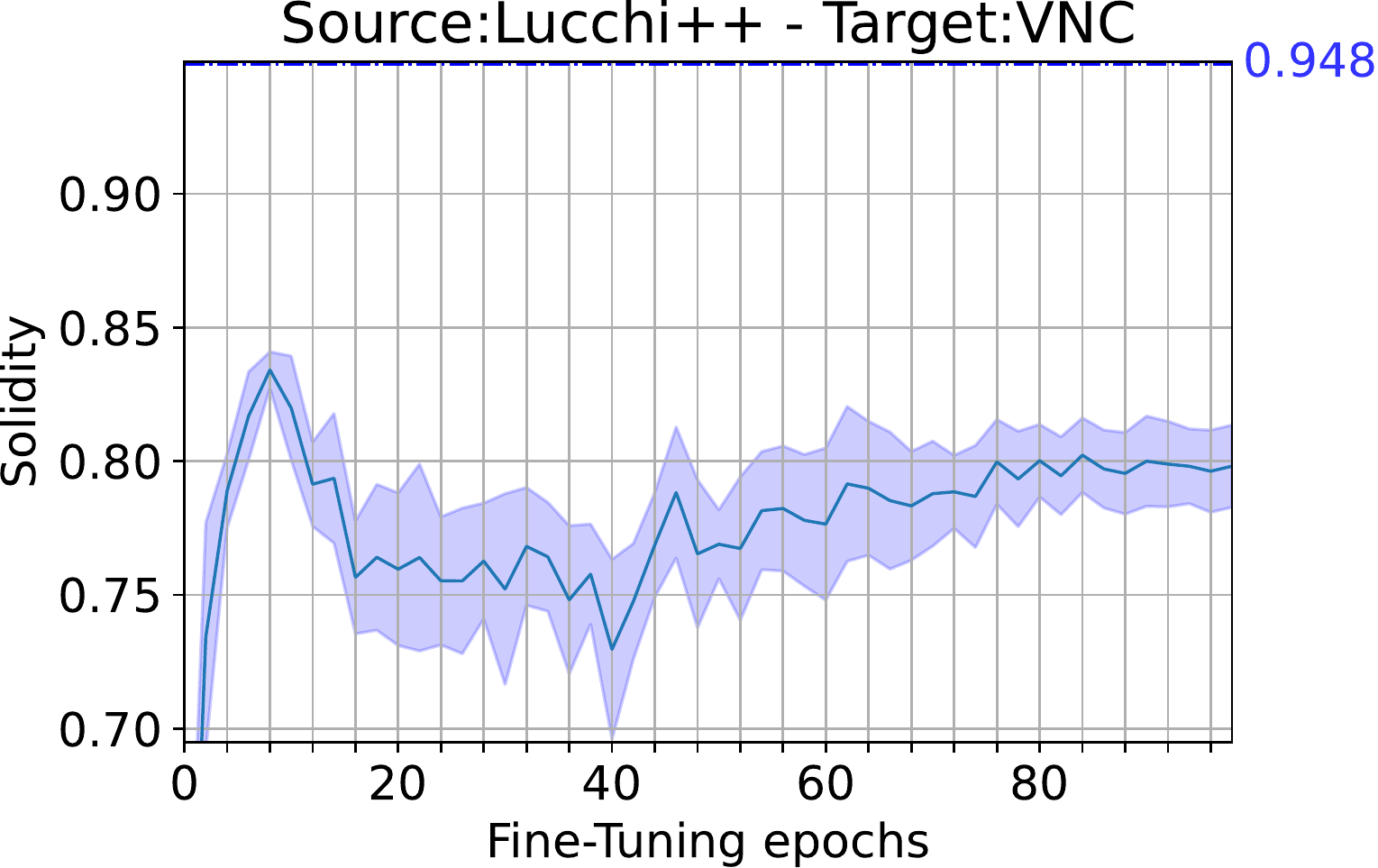}}
    \end{minipage}
    \begin{minipage}{.5\linewidth}
        \centering
        \subfloat[]{\includegraphics[scale=.39]{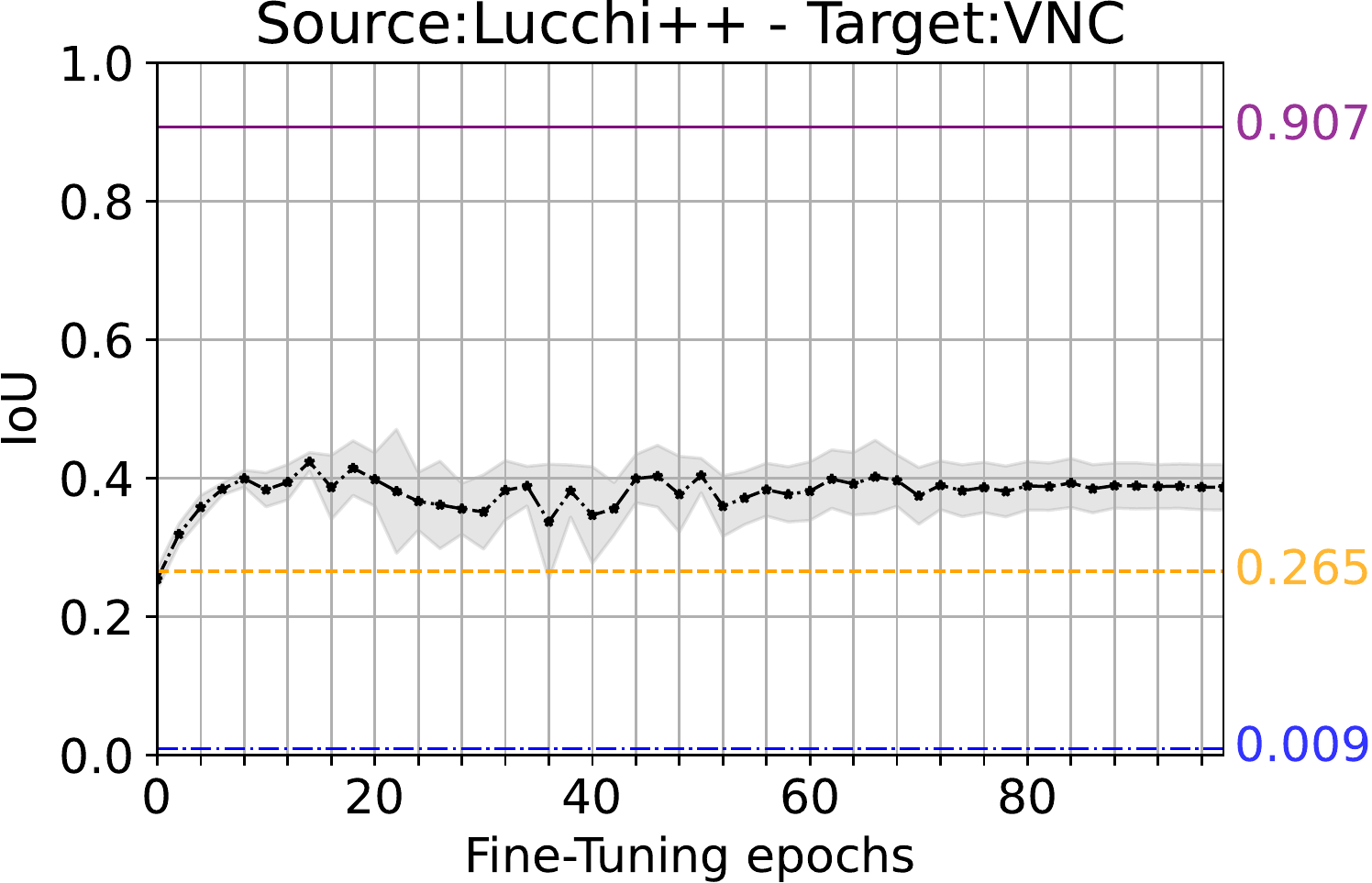}}
    \end{minipage}\par\medskip
    
    \caption{Relation between solidity and IoU in the Attention Y-Net approach with Lucchi++ as source domain. On the left, the evolution of the solidity value (averaged for ten executions) as a function of the stylization epochs with (a) Kasthuri++ and (c) VNC as target domains (dashed lines represent the source solidity value). On the right, the evolution of the test IoU (also averaged over ten executions) as a function of the epochs with (b) Kasthuri++ and (d) VNC as target domains. The magenta lines represent the maximum IoU value obtained by the fully supervised baseline models. In contrast, the blue and orange lines represent the IoU values obtained by the baseline methods applied without adaptation and after histogram matching to the target datasets, respectively.}
\end{figure}

\begin{figure}[ht]
    \begin{minipage}{.5\linewidth}
        \centering
        \subfloat[]{\includegraphics[scale=.39]{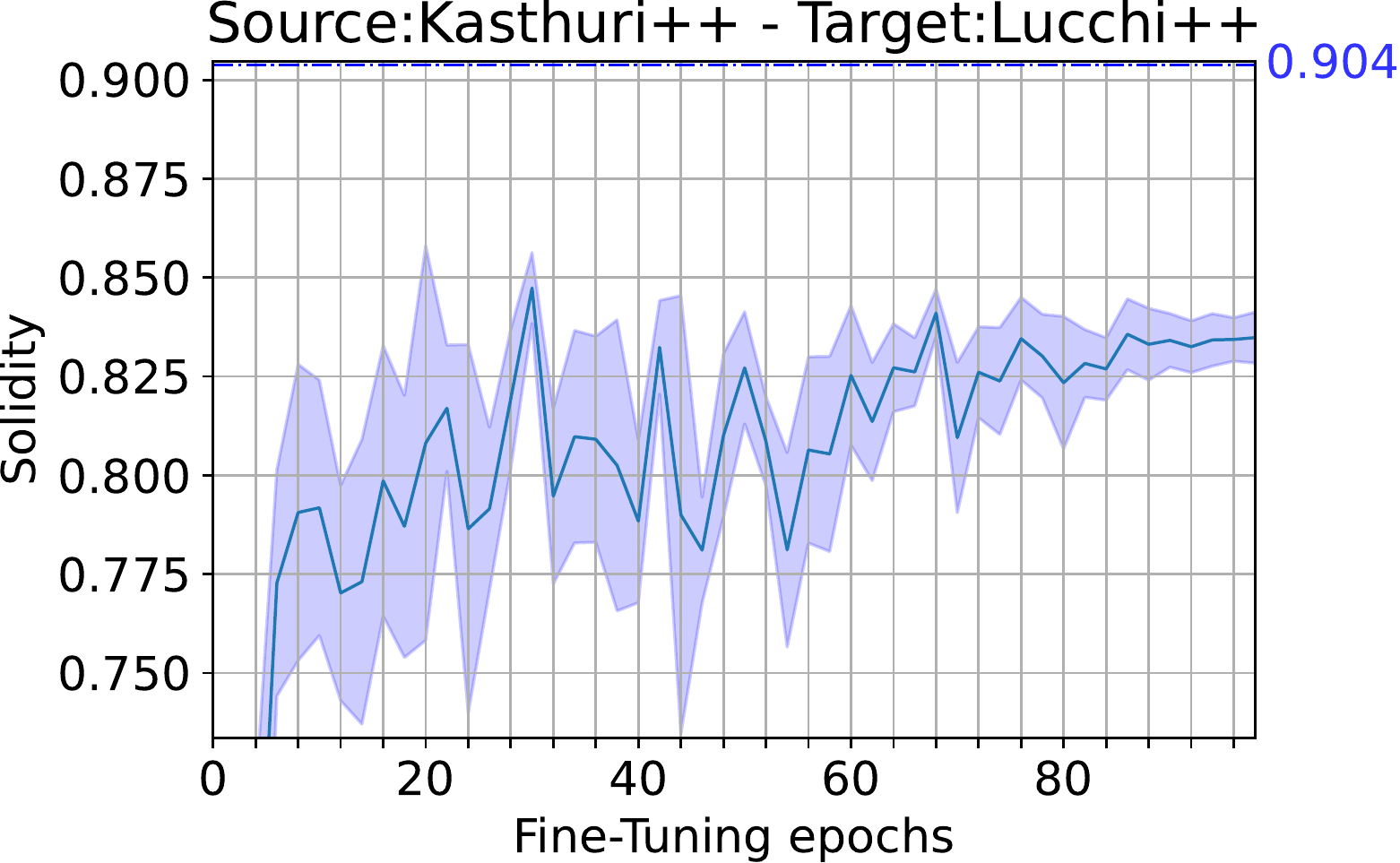}}
    \end{minipage}
    \begin{minipage}{.5\linewidth}
        \centering
        \subfloat[]{\includegraphics[scale=.39]{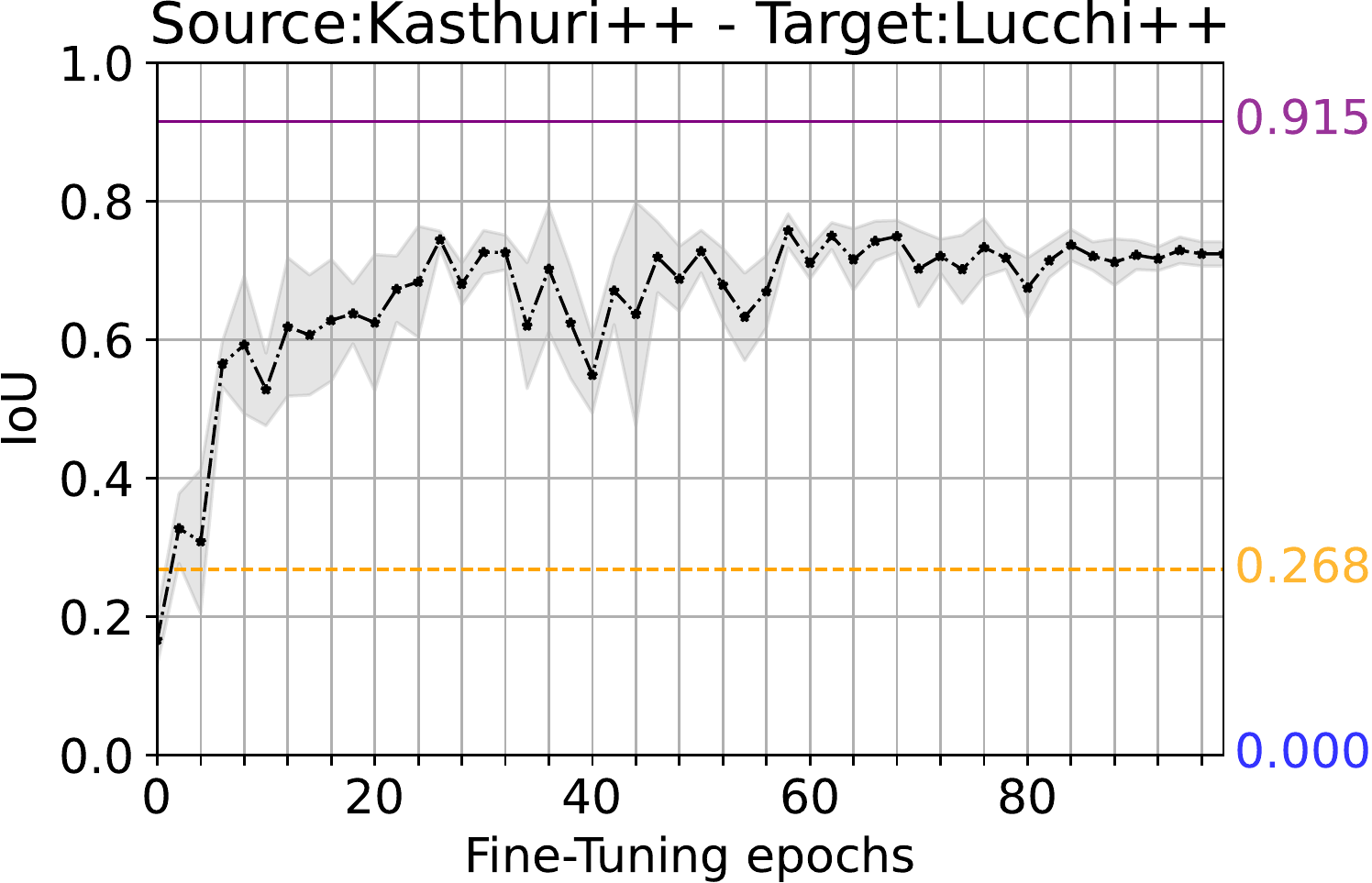}}
    \end{minipage}\par\medskip
    \begin{minipage}{.5\linewidth}
        \centering
        \subfloat[]{\includegraphics[scale=.39]{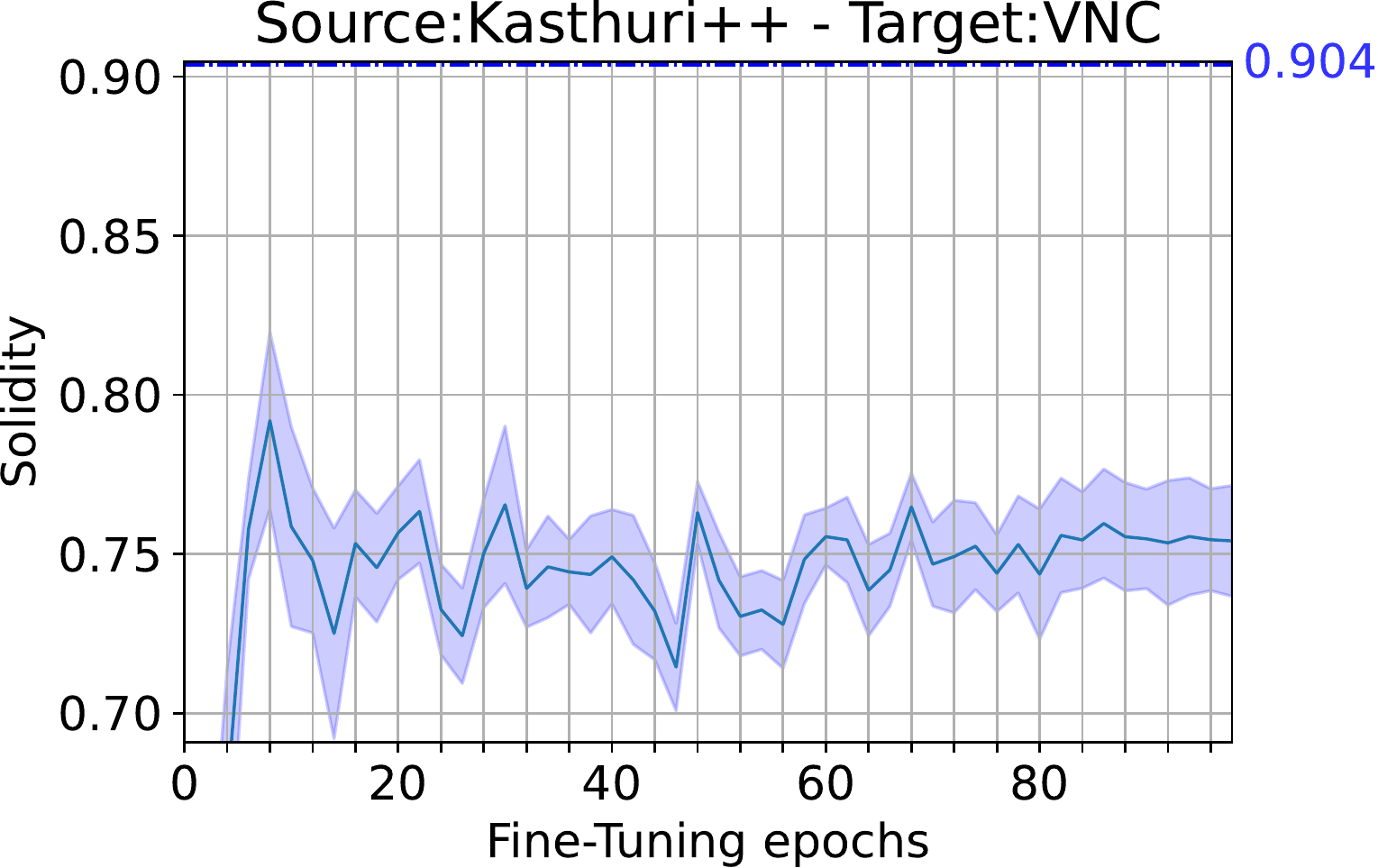}}
    \end{minipage}
    \begin{minipage}{.5\linewidth}
        \centering
        \subfloat[]{\includegraphics[scale=.39]{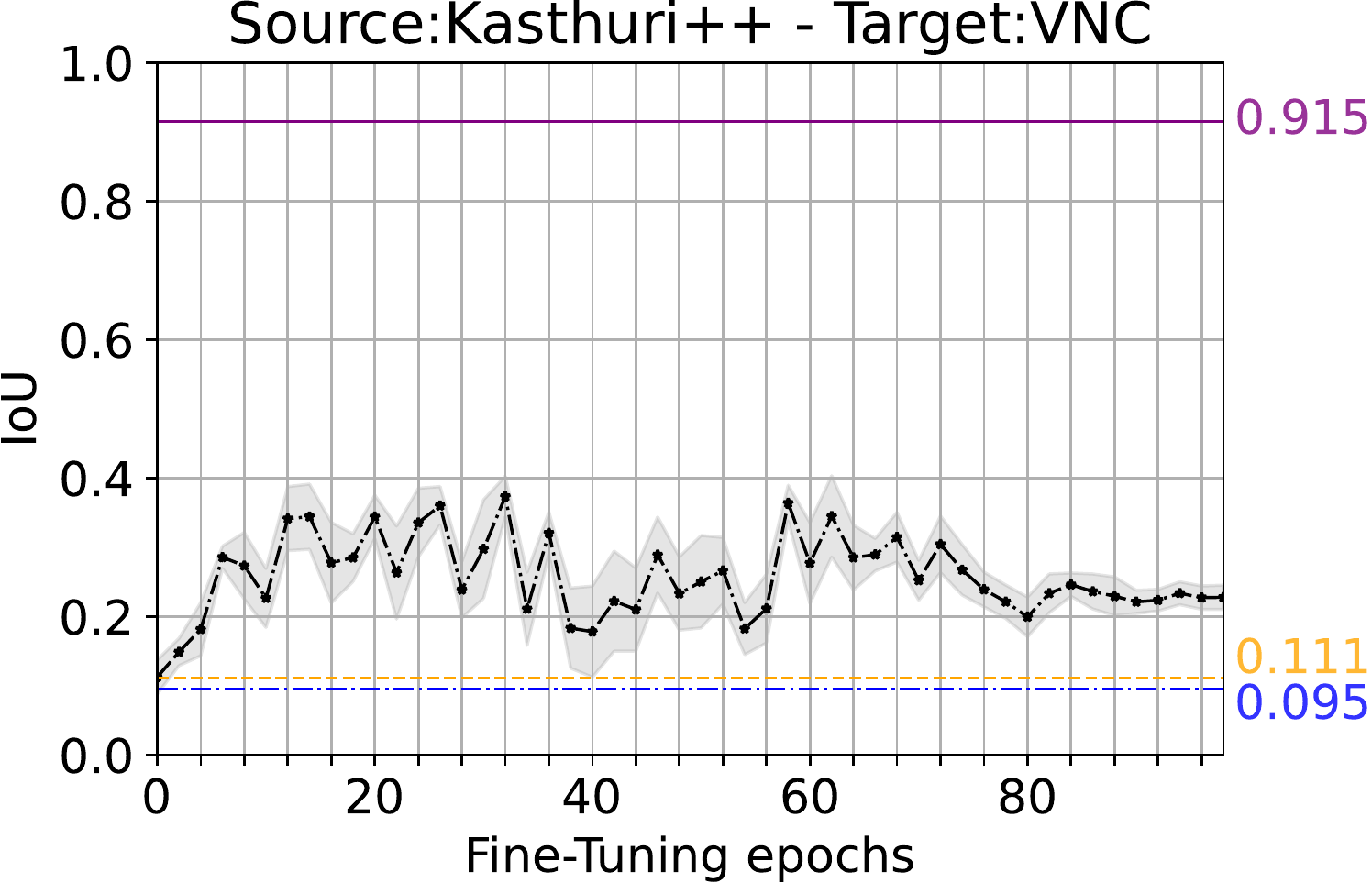}}
    \end{minipage}\par\medskip
    
    \caption{Relation between solidity and IoU in the Attention Y-Net approach with Kasthuri++ as source domain. On the left, the evolution of the solidity value (averaged for ten executions) as a function of the stylization epochs with (a) Lucchi++ and (c) VNC as target domains (dashed lines represent the source solidity value). On the right, the evolution of the test IoU (also averaged over ten executions) as a function of the epochs with (b) Lucchi++ and (d) VNC as target domains. The magenta lines represent the maximum IoU value obtained by the fully supervised baseline models. In contrast, the blue and orange lines represent the IoU values obtained by the baseline methods applied without adaptation and after histogram matching to the target datasets, respectively.}
\end{figure}

\begin{figure}[ht]
    \begin{minipage}{.5\linewidth}
        \centering
        \subfloat[]{\includegraphics[scale=.39]{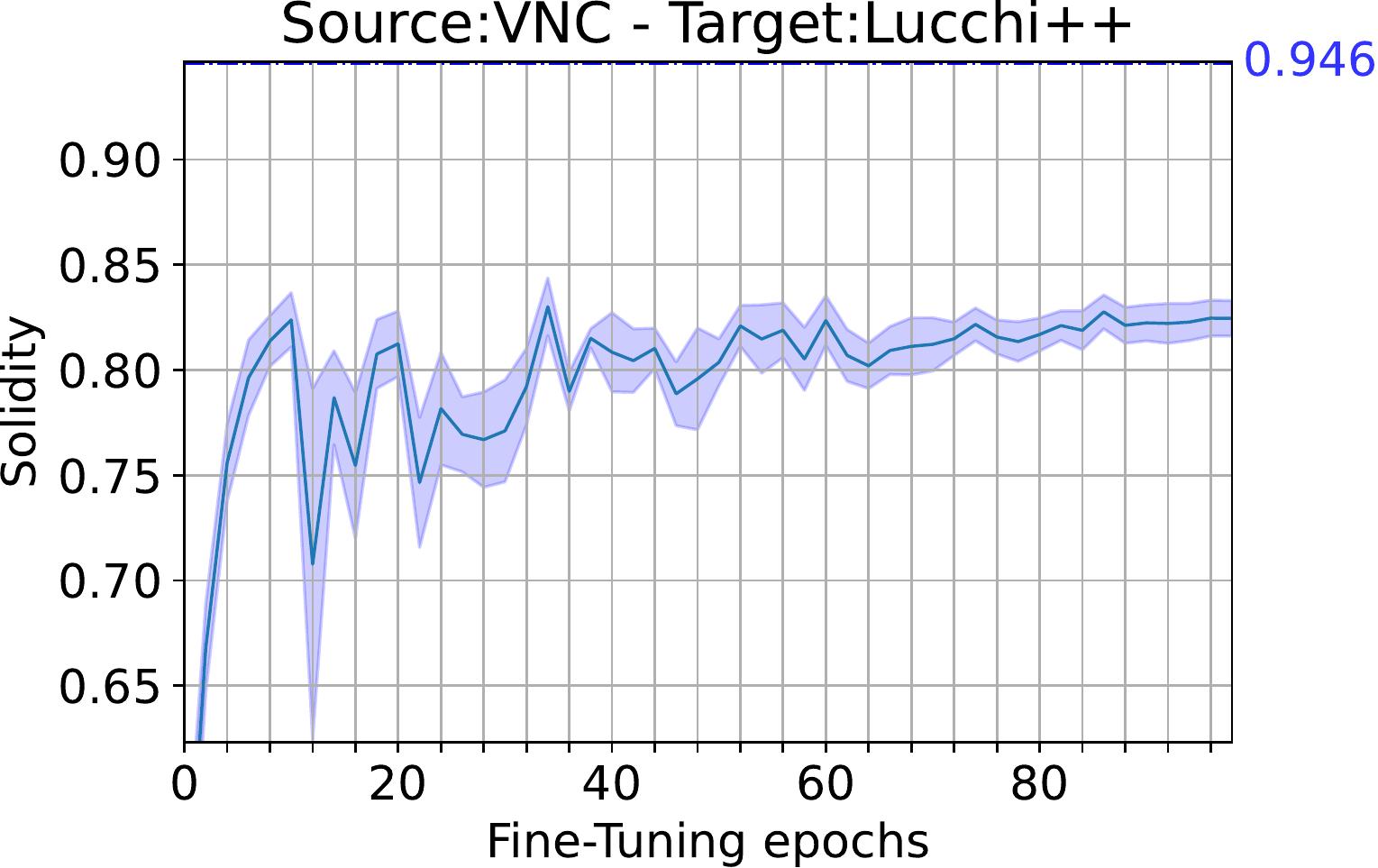}}
    \end{minipage}
    \begin{minipage}{.5\linewidth}
        \centering
        \subfloat[]{\includegraphics[scale=.39]{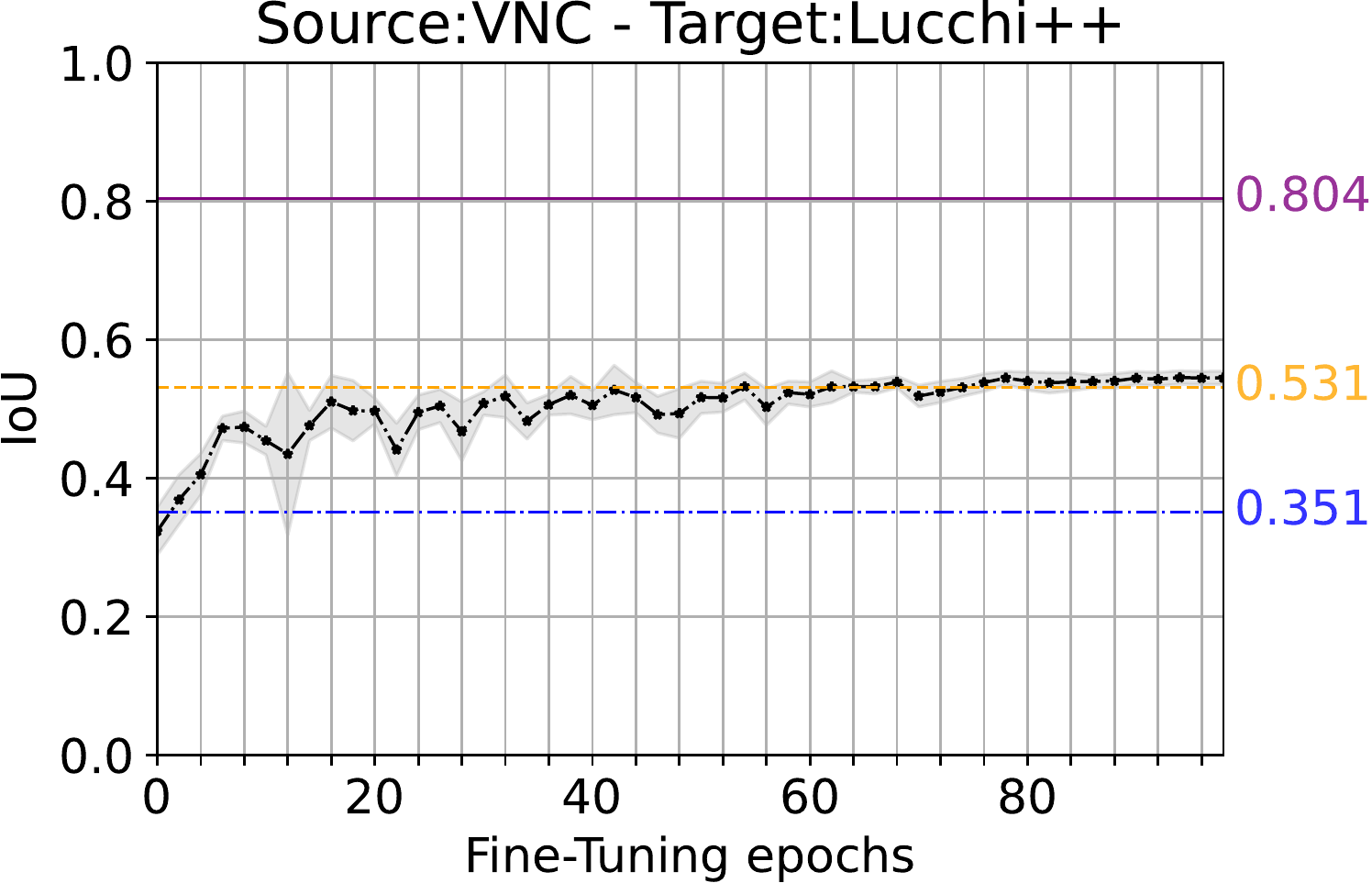}}
    \end{minipage}\par\medskip
    \begin{minipage}{.5\linewidth}
        \centering
        \subfloat[]{\includegraphics[scale=.39]{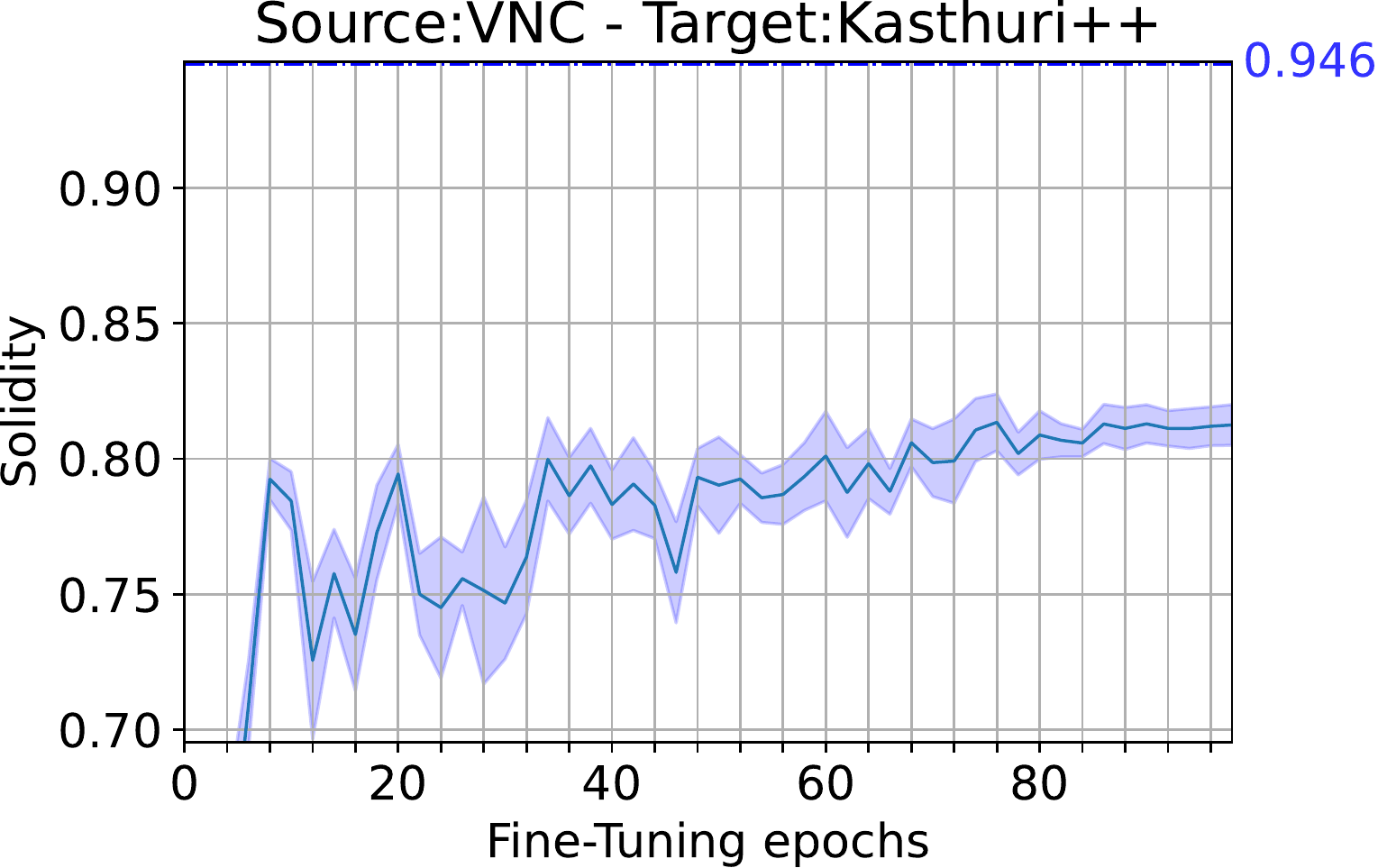}}
    \end{minipage}
    \begin{minipage}{.5\linewidth}
        \centering
        \subfloat[]{\includegraphics[scale=.39]{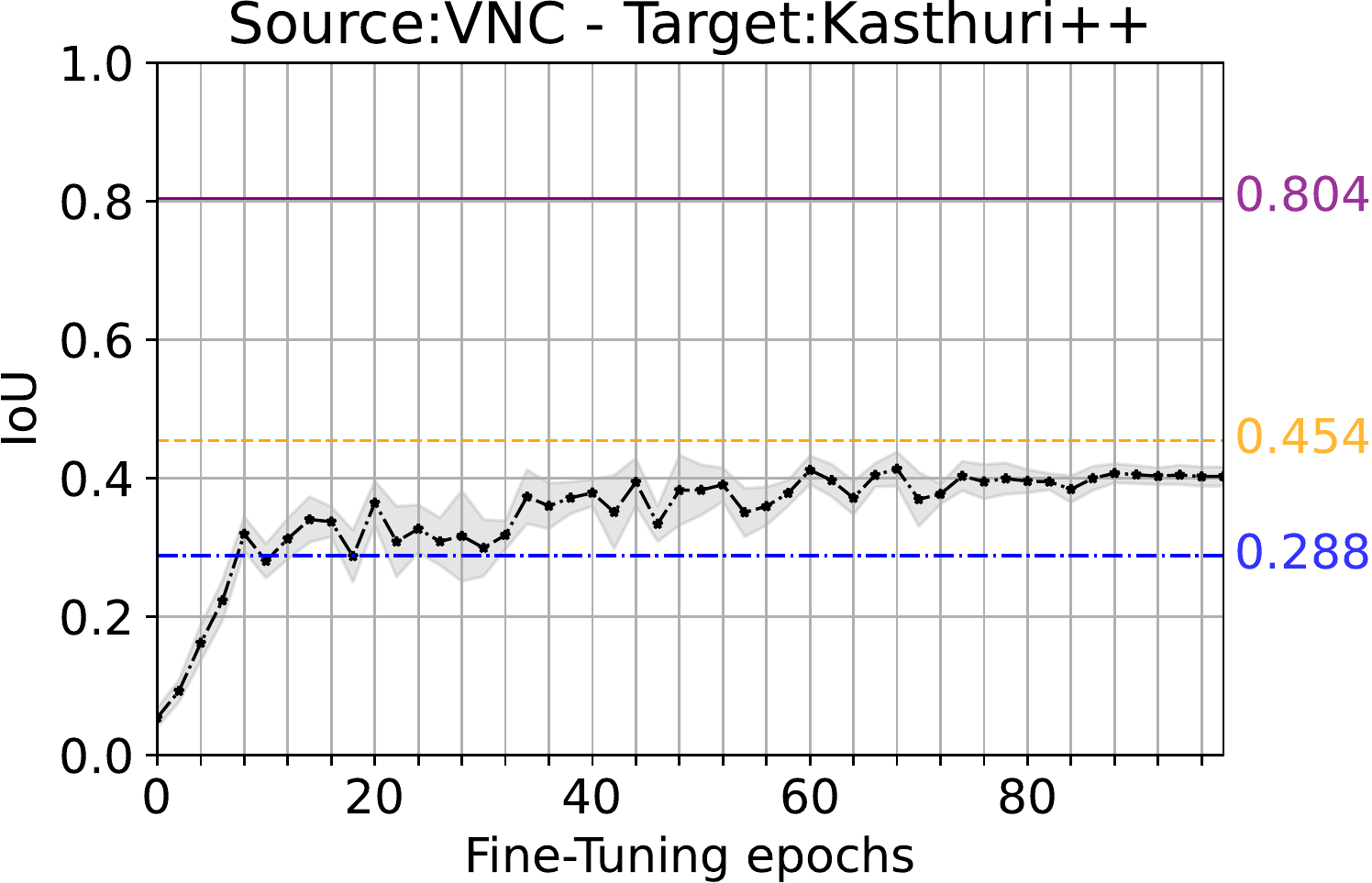}}
    \end{minipage}\par\medskip
    
    \caption{Relation between solidity and IoU in the Attention Y-Net approach with VNC as source domain. On the left, the evolution of the solidity value (averaged for ten executions) as a function of the stylization epochs with (a) Lucchi++ and (c) Kasthuri++ as target domains (dashed lines represent the source solidity value). On the right, the test IoU evolution (averaged over ten executions) as a function of the epochs with (b) Lucchi++ and (d) Kasthuri++ as target domains. The magenta lines represent the maximum IoU value obtained by the fully supervised baseline models, while the blue and orange lines represent the IoU values obtained by the baseline methods applied without adaptation and after histogram matching to the target datasets, respectively.}
\end{figure}

\begin{figure}[ht]
    \begin{minipage}{.5\linewidth}
        \centering
        \subfloat[]{\includegraphics[scale=.39]{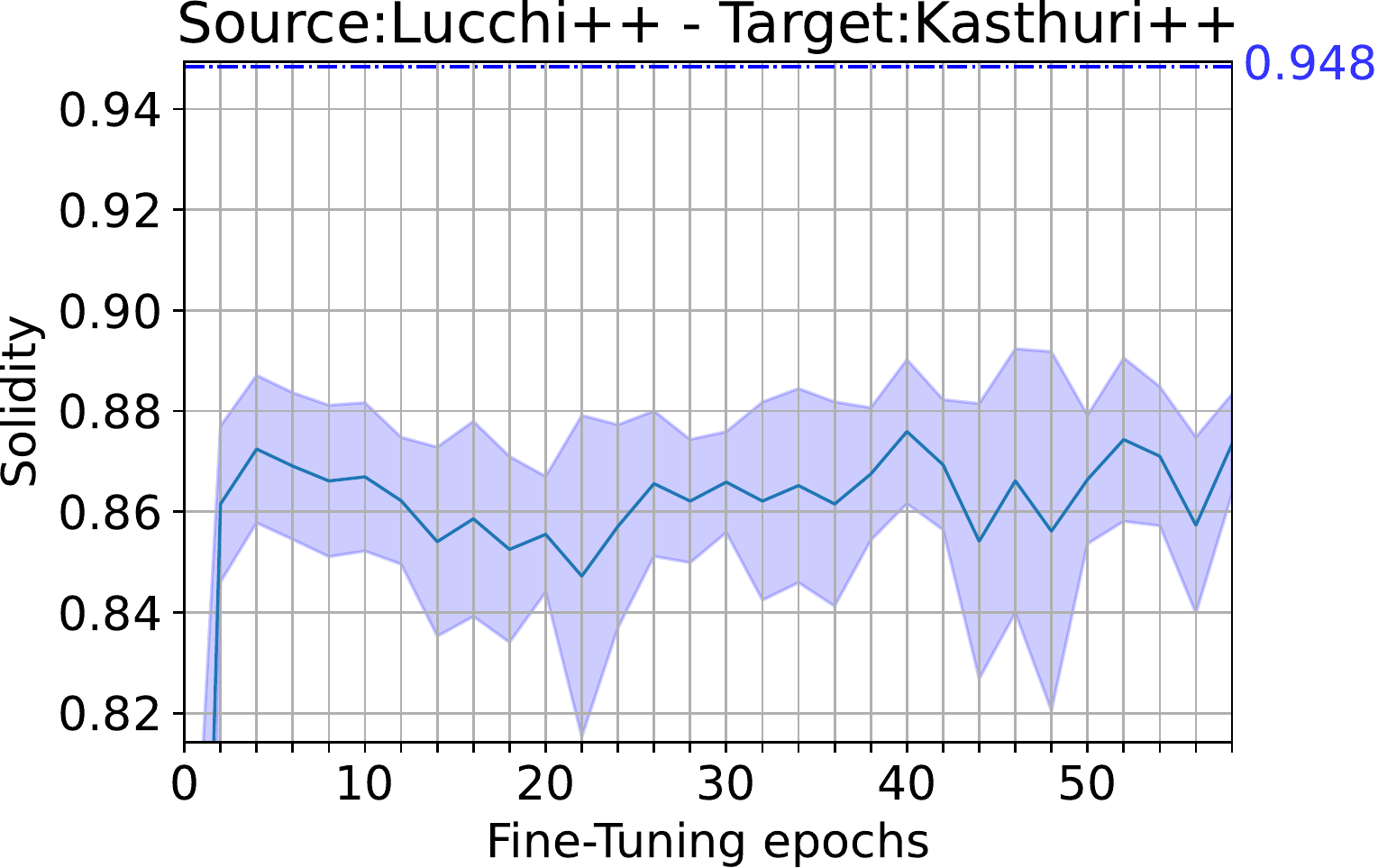}}
    \end{minipage}
    \begin{minipage}{.5\linewidth}
        \centering
        \subfloat[]{\includegraphics[scale=.39]{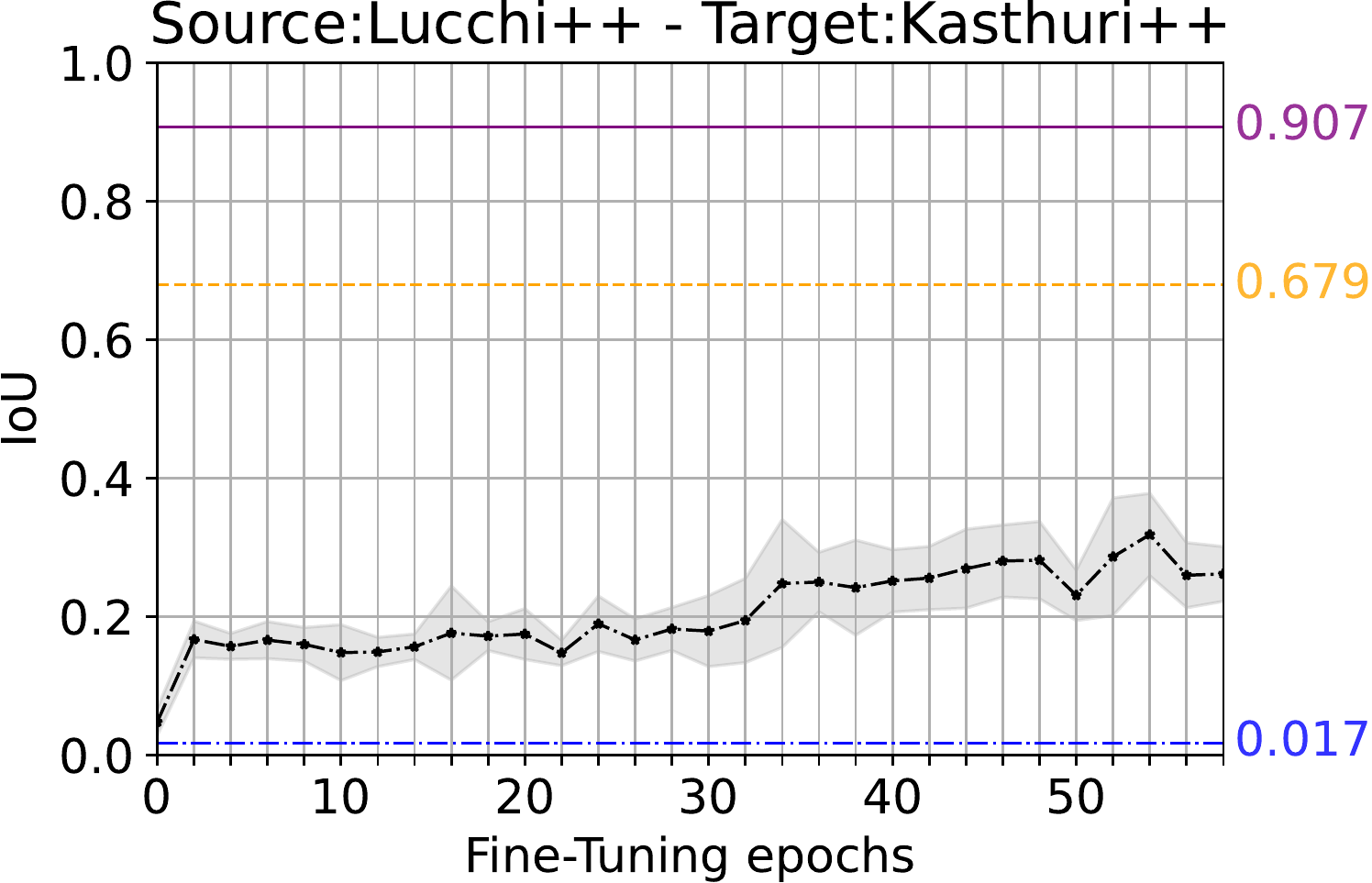}}
    \end{minipage}\par\medskip
    \begin{minipage}{.5\linewidth}
        \centering
        \subfloat[]{\includegraphics[scale=.39]{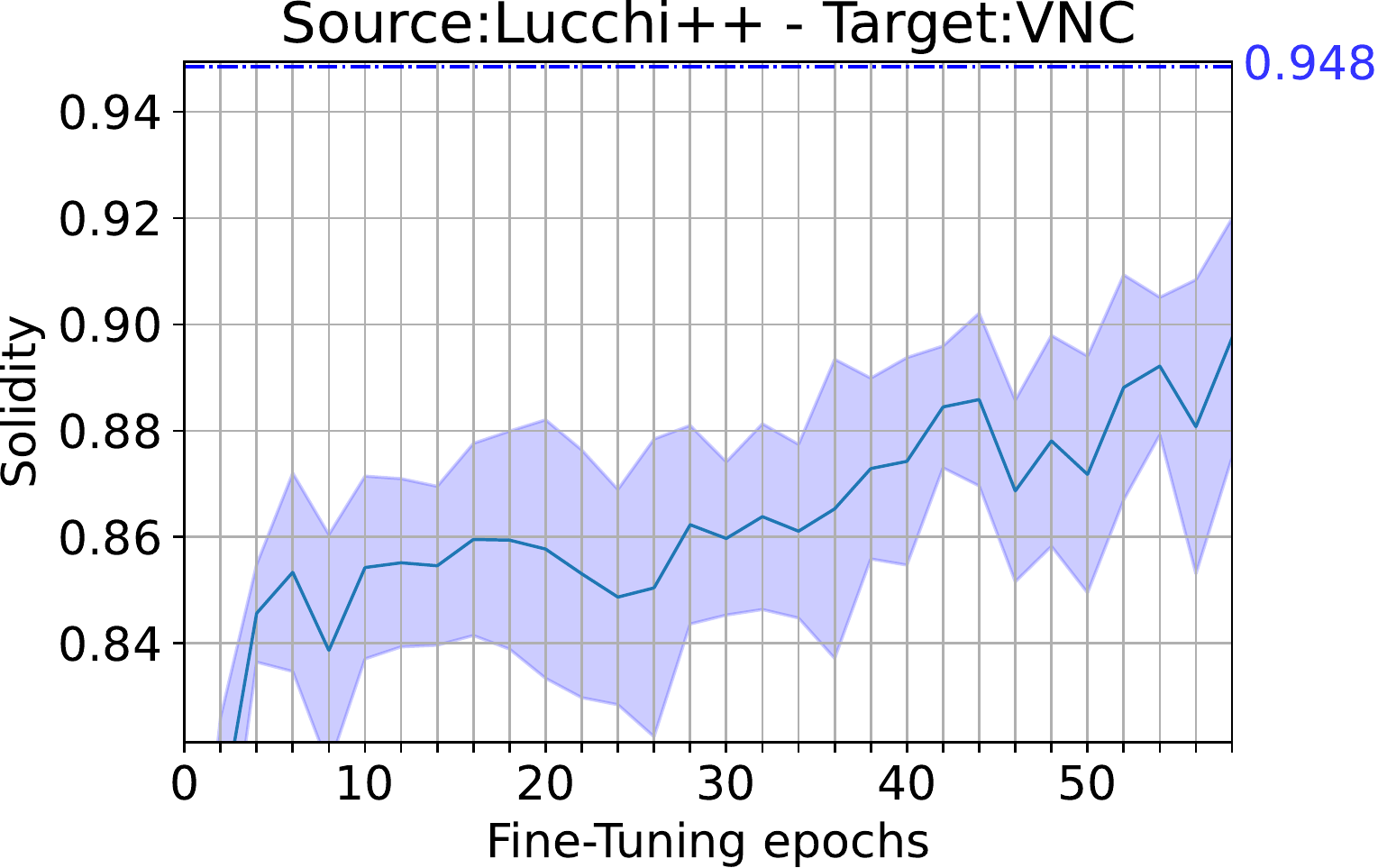}}
    \end{minipage}
    \begin{minipage}{.5\linewidth}
        \centering
        \subfloat[]{\includegraphics[scale=.39]{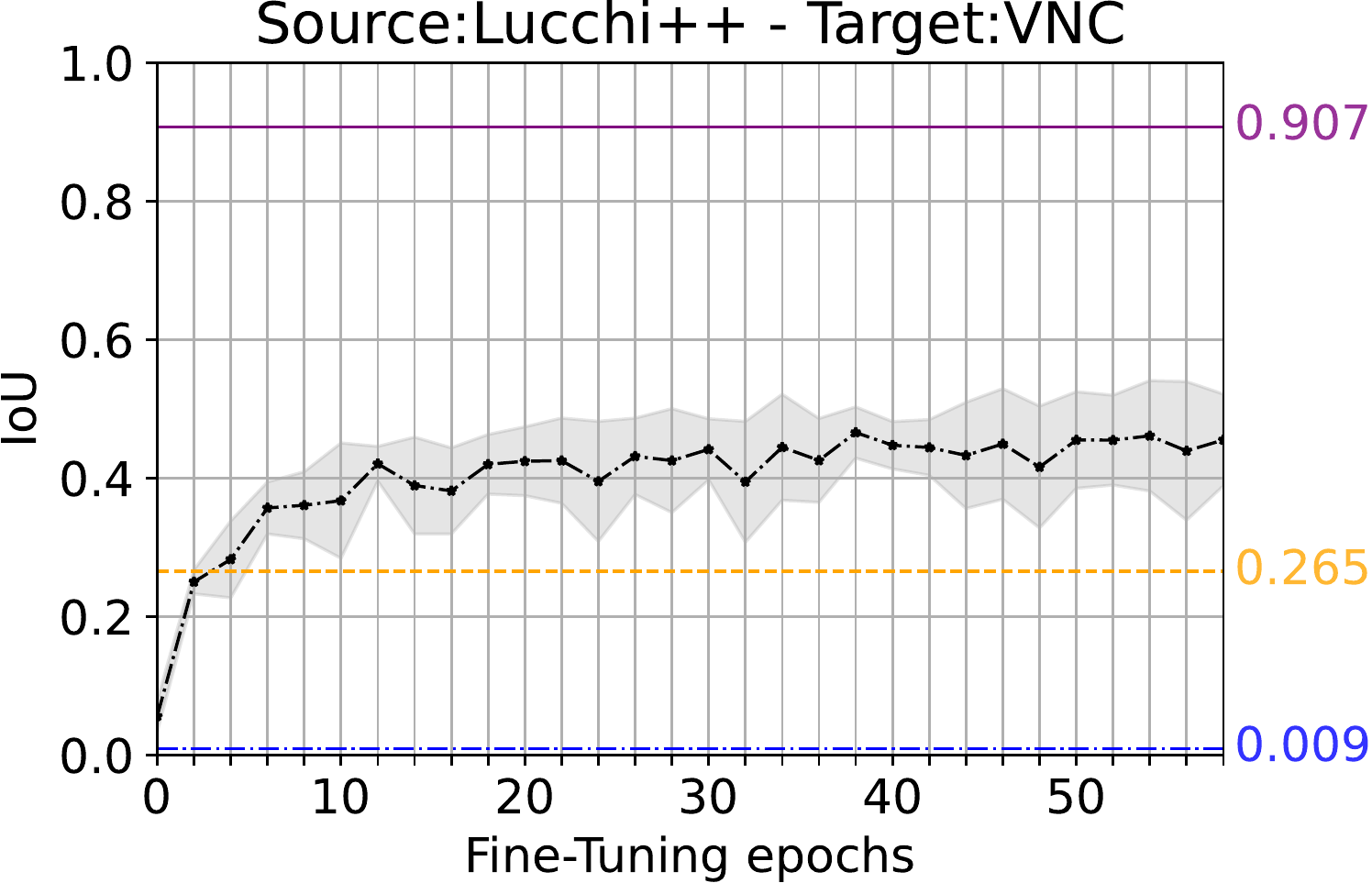}}
    \end{minipage}\par\medskip
    
    \caption{Relation between solidity and IoU in the DAMT-Net approach with Lucchi++ as source domain. On the left, the evolution of the solidity value (averaged for ten executions) as a function of the stylization epochs with (a) Kasthuri++ and (c) VNC as target domains (dashed lines represent the source solidity value). On the right, the evolution of the test IoU (also averaged over ten executions) as a function of the epochs with (b) Kasthuri++ and (d) VNC as target domains. The magenta lines represent the maximum IoU value obtained by the fully supervised baseline models. In contrast, the blue and orange lines represent the IoU values obtained by the baseline methods applied without adaptation and after histogram matching to the target datasets, respectively.}
\end{figure}

\begin{figure}[ht]
    \begin{minipage}{.5\linewidth}
        \centering
        \subfloat[]{\includegraphics[scale=.39]{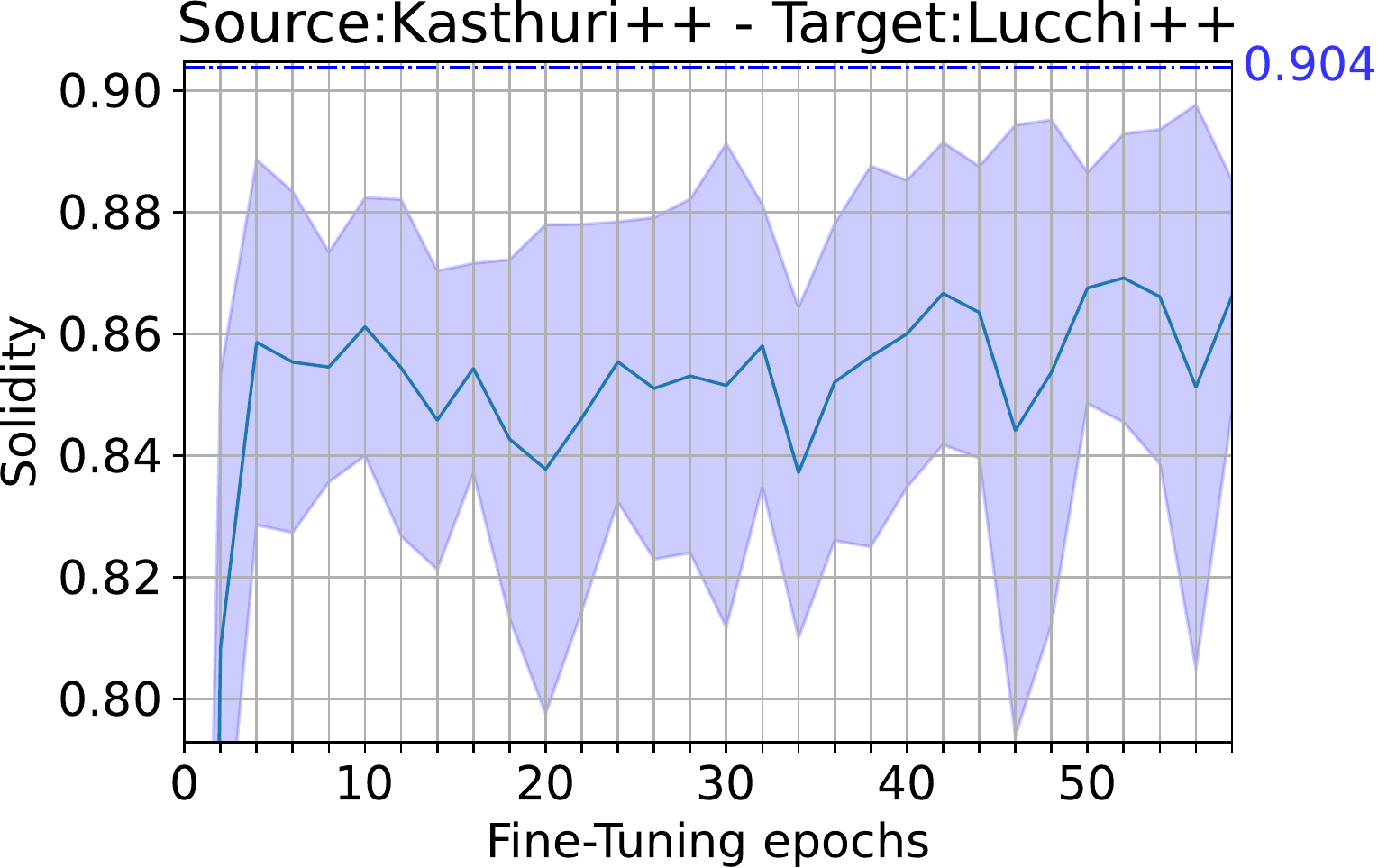}}
    \end{minipage}
    \begin{minipage}{.5\linewidth}
        \centering
        \subfloat[]{\includegraphics[scale=.39]{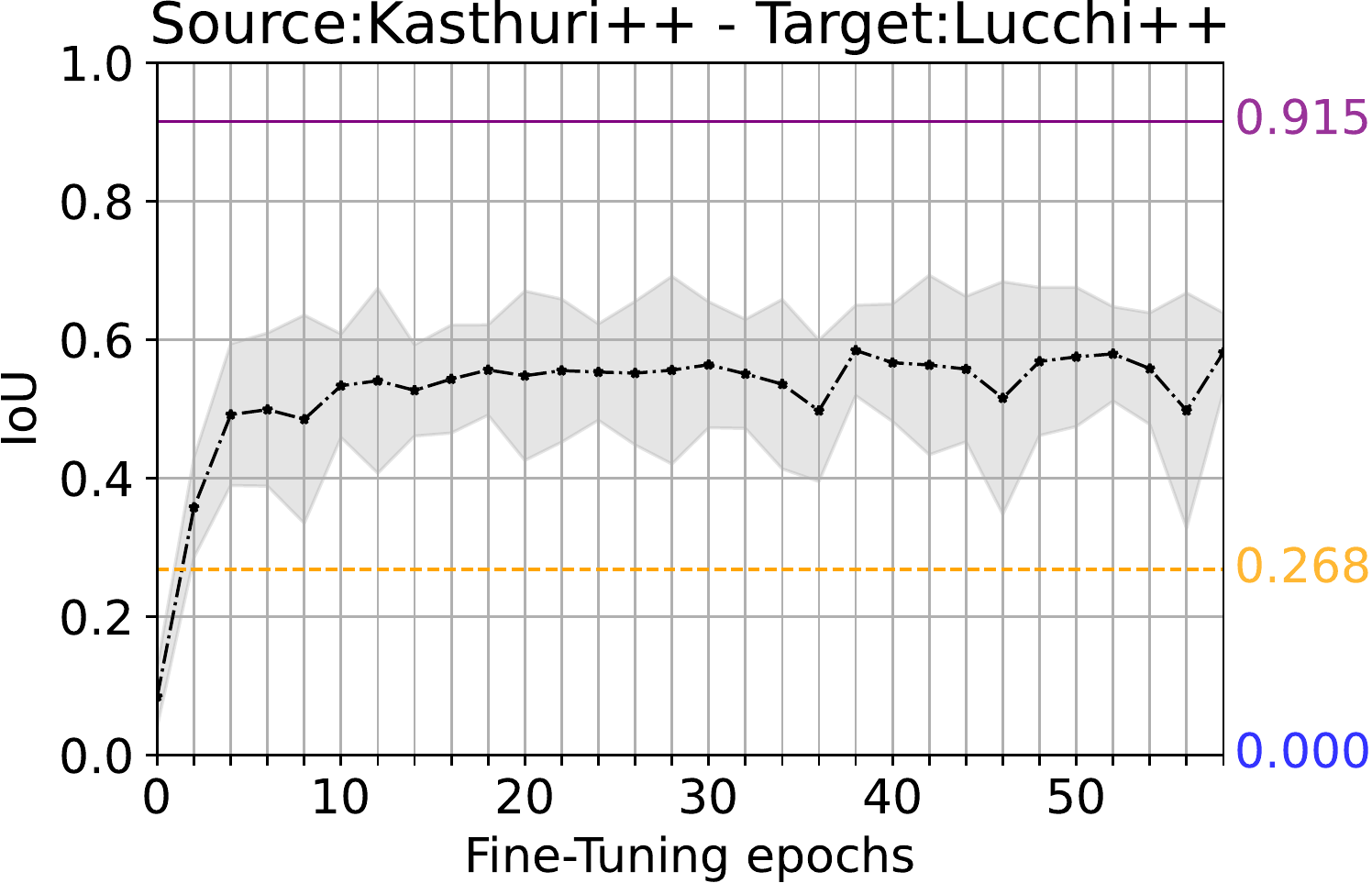}}
    \end{minipage}\par\medskip
    \begin{minipage}{.5\linewidth}
        \centering
        \subfloat[]{\includegraphics[scale=.39]{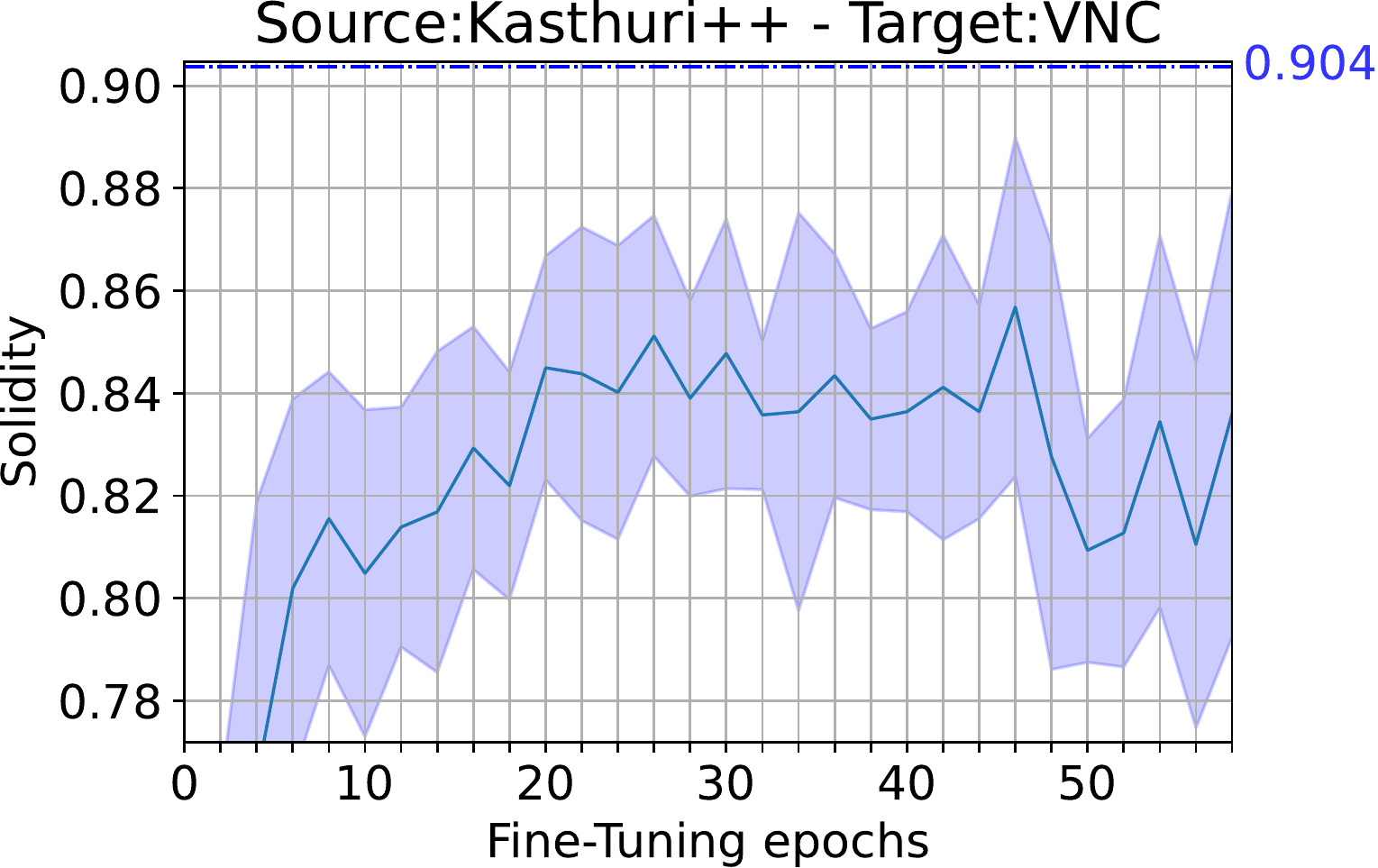}}
    \end{minipage}
    \begin{minipage}{.5\linewidth}
        \centering
        \subfloat[]{\includegraphics[scale=.39]{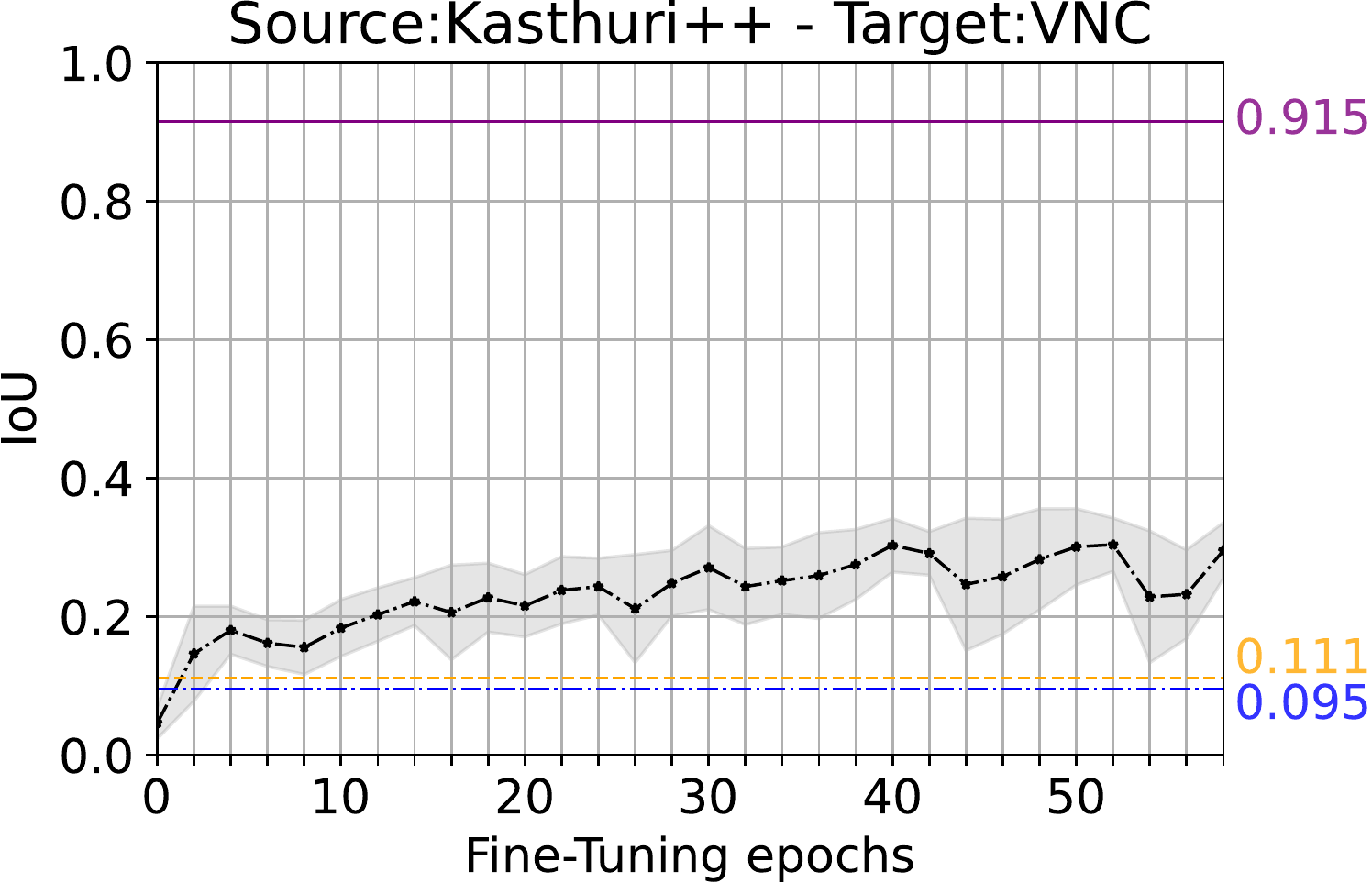}}
    \end{minipage}\par\medskip
    
    \caption{Relation between solidity and IoU in the DAMT-Net approach with Kasthuri++ as source domain. On the left, the evolution of the solidity value (averaged for ten executions) as a function of the stylization epochs with (a) Lucchi++ and (c) VNC as target domains (dashed lines represent the source solidity value). On the right, the evolution of the test IoU (also averaged over ten executions) as a function of the epochs with (b) Lucchi++ and (d) VNC as target domains. The magenta lines represent the maximum IoU value obtained by the fully supervised baseline models. In contrast, the blue and orange lines represent the IoU values obtained by the baseline methods applied without adaptation and after histogram matching to the target datasets, respectively.}
\end{figure}

\begin{figure}[ht]
    \begin{minipage}{.5\linewidth}
        \centering
        \subfloat[]{\includegraphics[scale=.39]{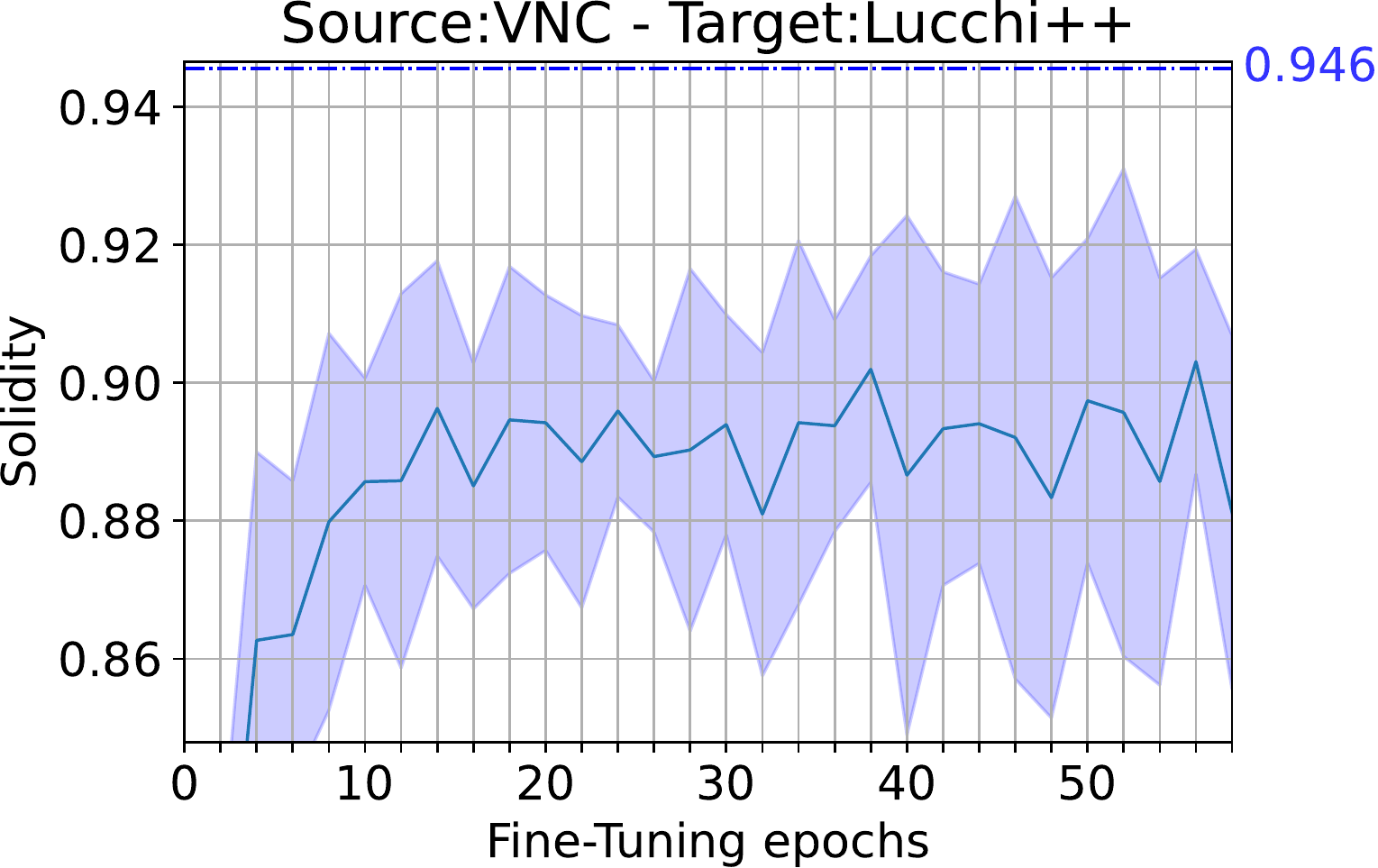}}
    \end{minipage}
    \begin{minipage}{.5\linewidth}
        \centering
        \subfloat[]{\includegraphics[scale=.39]{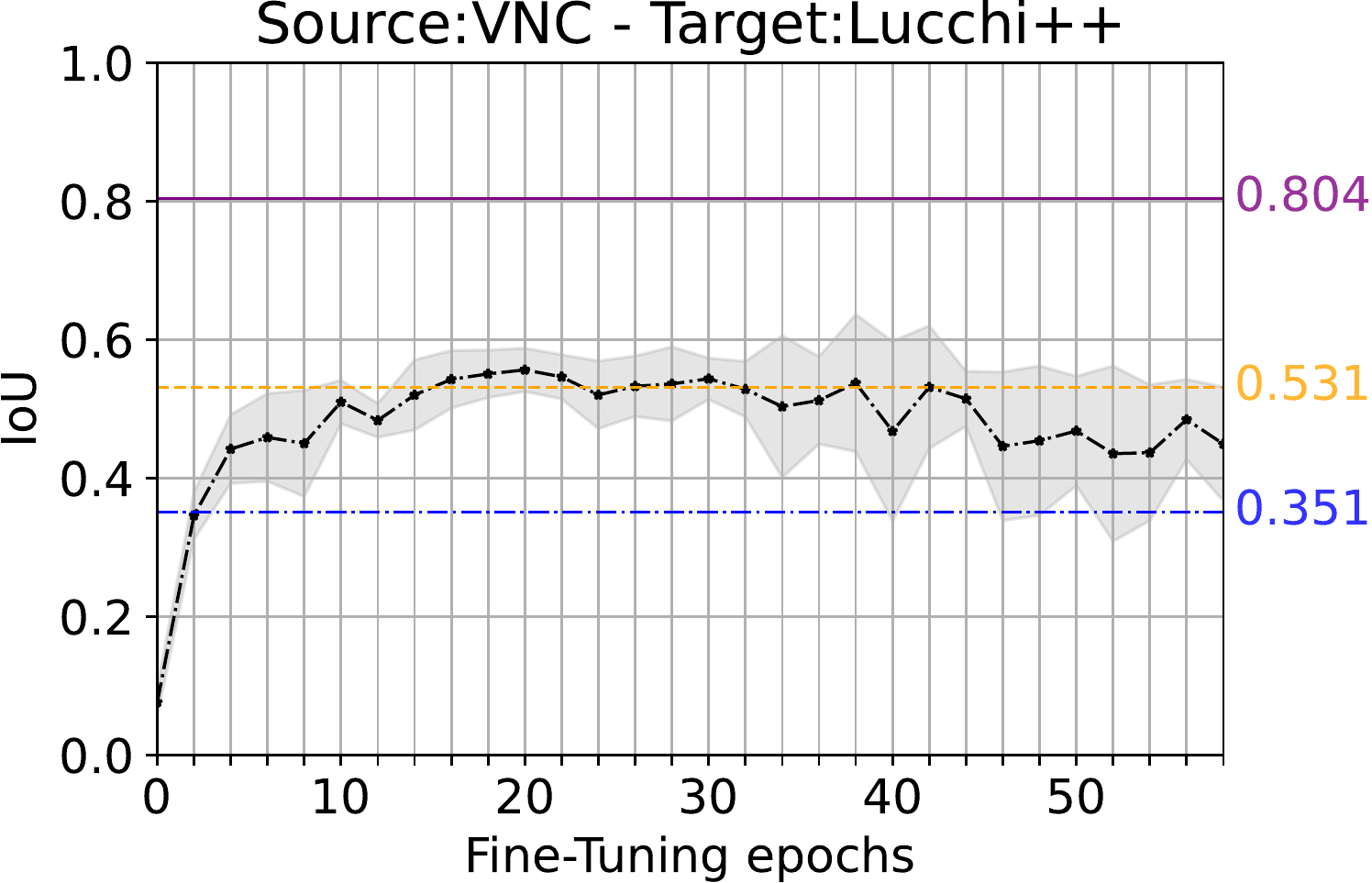}}
    \end{minipage}\par\medskip
    \begin{minipage}{.5\linewidth}
        \centering
        \subfloat[]{\includegraphics[scale=.39]{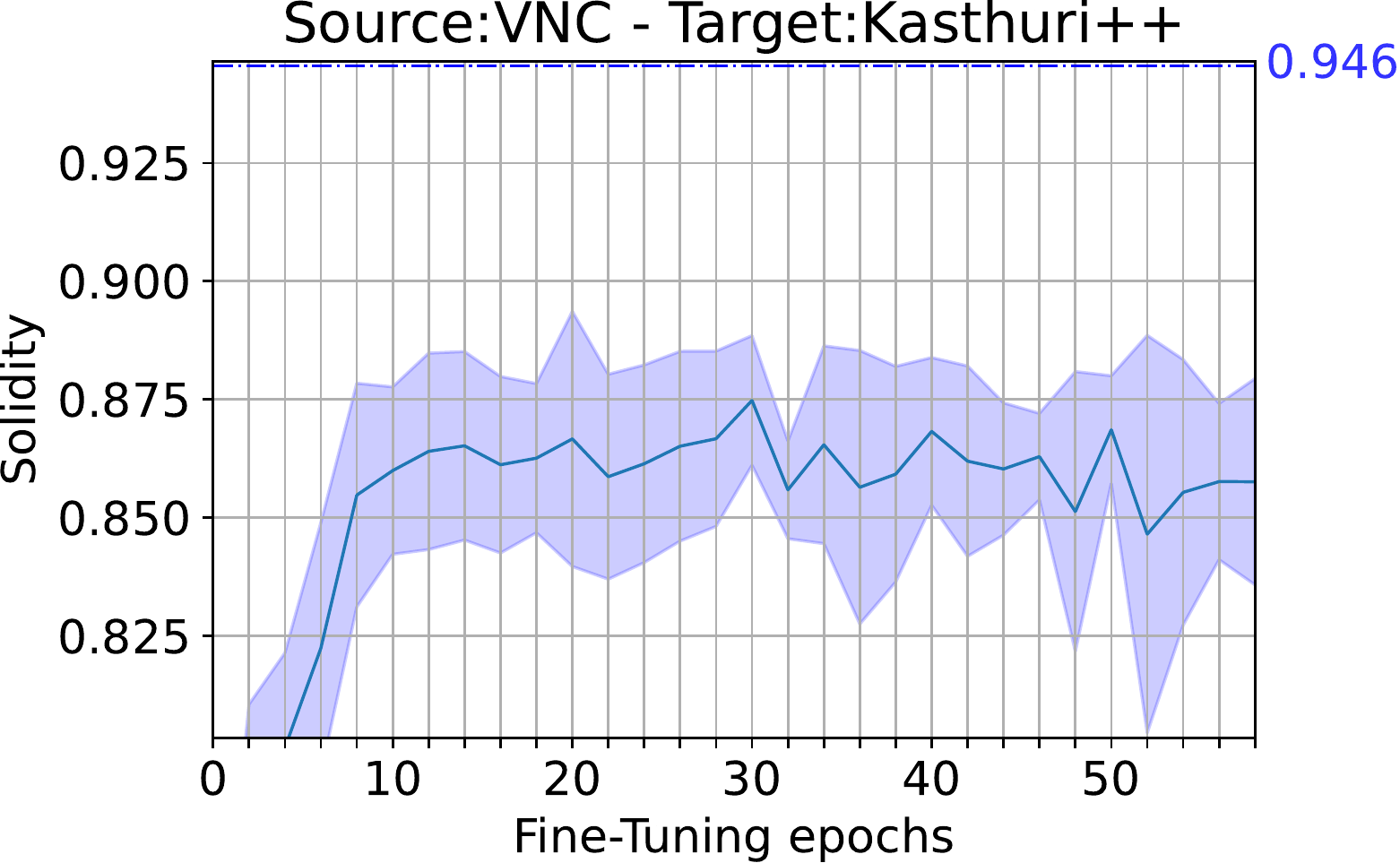}}
    \end{minipage}
    \begin{minipage}{.5\linewidth}
        \centering
        \subfloat[]{\includegraphics[scale=.39]{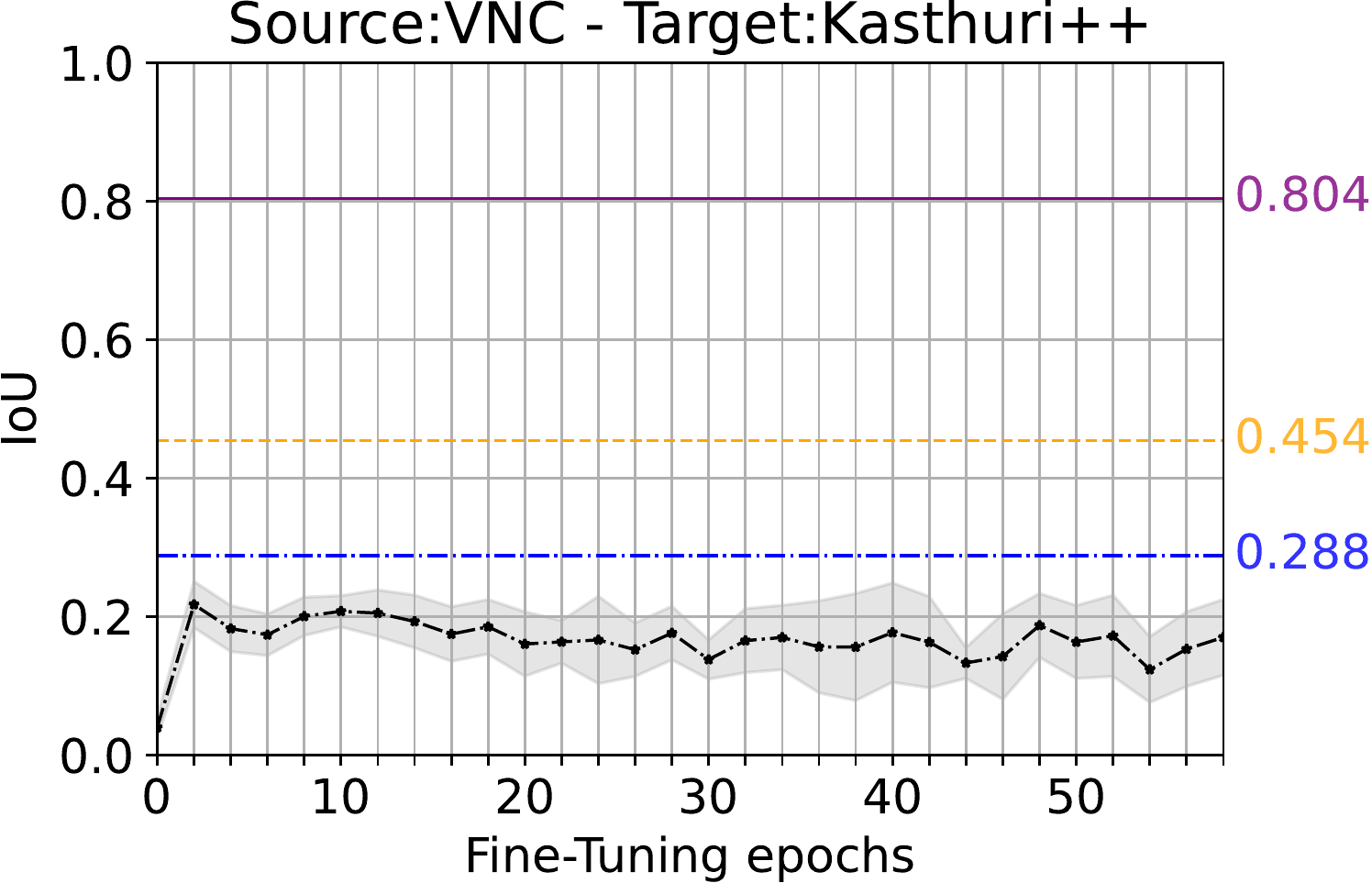}}
    \end{minipage}\par\medskip
    
    \caption{Relation between solidity and IoU in the DAMT-Net approach with VNC as source domain. On the left, the evolution of the solidity value (averaged for ten executions) as a function of the stylization epochs with (a) Lucchi++ and (c) Kasthuri++ as target domains (dashed lines represent the source solidity value). On the right, the test IoU evolution (averaged over ten executions) as a function of the epochs with (b) Lucchi++ and (d) Kasthuri++ as target domains. The magenta lines represent the maximum IoU value obtained by the fully supervised baseline models. In contrast, the blue and orange lines represent the IoU values obtained by the baseline methods applied without adaptation and after histogram matching to the target datasets, respectively.}
\end{figure}
        
\clearpage
\newpage

\section{Hyperparameter search}\label{hyperparameter_search}
\setcounter{figure}{0}

This section describes in detail the search we performed for the optimal training configuration and set of hyperparameters in all our proposed approaches. The corresponding search space and best values are summarized in the tables below using the following notation:
\begin{itemize}
    \item $[a, b]$: Range between two possible values. E.g. \textit{zoom([0.75,1.25])} corresponds to a random zoom value between $0.75$ and $1.25$.
    \item $[a, b, c]$: All values from $a$ to $b$ with $c$ step. E.g. $[10, 300, 10]$ corresponds to $10, 20, 30, 40, ..., 300$. 
    \item $(a, b, c)$: All values set, e.g., dropout(0.1,0.2,0.3) in a 3-depth level network indicate that 0.1 dropout value has been set in the first level, 0.2 dropout in the second level and 0.3 in the third level.  
    \item $choice[a, b, ...]$: One value between $a$, $b$ and so on. E.g. \textit{[10, 15, 20, 30, 60]} possible values are: $10$ or $15$ or $20$ or $30$ or $60$ (but only one). 
    \item $a, b, c, ...$: All tested values, e.g., flips, rotations.
    \item $N(\mu,\sigma)$: normal distribution with mean $\mu$ and standard deviation $\sigma$. 
\end{itemize}

There are also acronyms used in tables:

\begin{itemize}
    \item MAE: Mean absolute error.
    \item MSE: Mean squared error.
    \item BCE: Binary cross entropy.
    \item SGD: Stochastic gradient descent.
    \item Reduce on plateau: A learning rate policy to reduce the value of the learning rate when the monitored metric has stopped improving (\url{https://keras.io/api/callbacks/reduce_lr_on_plateau/}).
    \item OneCycle: One-cycle learning rate policy for super-convergence~\cite{smith2019super}.
    \item ELU: Exponential linear unit activation function. 
    \item he\_normal: \textit{He normal}~\cite{he2015delving} as kernel initialization.
\end{itemize}

\clearpage
\newpage

\subsection{Self-supervised approach}
\label{subsec:apendix:hyperparam_search_SSL}
\begin{table}[]
\hspace{-1.4cm}
\begin{adjustbox}{max width=1.22\textwidth}
\begin{tabular}{rcrcr}
\hline
 \textbf{Hyperparameter}                                                                                                                                       &  \phantom{abc} &  \textbf{Search space}                                                                                                                      &  \phantom{abc} &  \textbf{Best assignment}                                                                                                                   \\ \hline
 \textbf{Data}                                                                                                                                                 &  \phantom{abc} &  \phantom{abc}                                                                                                                              &  \phantom{abc} &  \phantom{abc}                                                                                                                              \\ \hline
Validation                                                                                                                                                                     &                                & True                                                                                                                                                        &                                & True                                                                                                                                                        \\
Random validation                                                                                                                                                              &                                & True                                                                                                                                                        &                                & True                                                                                                                                                        \\
\% of train as validation                                                                                                                                                      &                                & 10\%                                                                                                                                                        &                                & 10\%                                                                                                                                                        \\
Patches                                                                                                                                                                        &                                & \begin{tabular}[c]{@{}r@{}} Maximum number that fitted \\ into the slice dimensions  \end{tabular}                     &                                & \begin{tabular}[c]{@{}r@{}} Maximum number that fitted \\ into the slice dimensions  \end{tabular}                     \\
Patch size                                                                                                                                                                     &                                & $256\times256$                                                                                                                                              &                                & $256\times256$                                                                                                                                              \\
  \begin{tabular}[c]{@{}r@{}}Discard patches with less than  \\ a \% of the foreground class \end{tabular} &                                &  \textit{choice}{[}True(2\%),True(5\%), True(10\%), False{]}                                                                      &                                & True(2\%)                                                                                                                                                   \\
Shuffle train on each epoch                                                                                                                                                    &                                & True                                                                                                                                                        &                                & True                                                                                                                                                        \\
Data augmentation                                                                                                                                                              &                                & flips, rotation\_range({[}-180,180,90{]})                                                                                                                   &                                & flips, rotation\_range({[}-180,180,90{]})                                                                                                                   \\
Number of epochs                                                                                                                                                               &                                &  \textit{choice}{[}100,200,300{]}                                                                                                           &                                & 200                                                                                                                                                         \\
Batch size                                                                                                                                                                     &                                &  \textit{choice}{[}1, 2, 5{]}                                                                                                                &                                & 1                                                                                                                                                           \\
Loss type                                                                                                                                                                      &                                &  \textit{choice}{[}MAE,MSE{]}                                                                                                               &                                & MAE                                                                                                                                                         \\
Optimizer                                                                                                                                                                      &                                &  \textit{choice}{[}SGD, Adam{]}                                                                                                             &                                & Adam                                                                                                                                                        \\
Learning rate                                                                                                                                                                  &                                &  \textit{choice}{[}1e-3, 2e-3,1e-4,5e-4{]}                                                                                                       &                                & 5e-4                                                                                                                                                        \\
Scheduler                                                                                                                                                                      &                                &  \textit{choice}{[}Reduce on Plateau, OneCycle{]}                                                                                           &                                & OneCycle                                                                                                                                                    \\
Patience                                                                                                                                                                       &                                &  \textit{choice}{[}None,3, 5, 7, 10, 50{]}                                                                                                       &                                & None                                                                                                                                                           \\ \hline
 \textbf{Architecture}                                                                                                                                         &  \phantom{abc} &  \phantom{abc}                                                                                                                              &  \phantom{abc} &                                                                                                                                                             \\ \hline
Number of feature maps to start with                                                                                                                                           &                                & 32                                                                                                                                                          &                                & 32                                                                                                                                                          \\
Dropout type                                                                                                                                                                   &                                &   \begin{tabular}[c]{@{}r@{}}Spatial dropout  (0.1, 0.1, 0.2, 0.2, 0.3) \end{tabular} &                                &   \begin{tabular}[c]{@{}r@{}}Spatial dropout (0.1, 0.1, 0.2, 0.2, 0.3) \end{tabular} \\
Pooling type                                                                                                                                                                   &                                & Max-pooling                                                                                                                                                 &                                & Max-pooling                                                                                                                                                 \\
Kernel initializer                                                                                                                                                             &                                & he\_normal                                                                                                                                                    &                                & he\_normal                                                                                                                                                  \\
Activation                                                                                                                                                                     &                                & ELU                                                                                                                                                         &                                & ELU                                                                                                                                                         \\ \hline
\end{tabular}
\end{adjustbox}
\vspace{0.2cm}
\caption{ Hyperparameter search space for the proposed self-supervised learning method.}

\label{tab:SSLHyper}
\end{table}

\begin{table}[]
\hspace{-1.4cm}
\begin{adjustbox}{max width=1.22\textwidth}
 \begin{tabular}{rcrcr}
\hline
 \textbf{Hyperparameter}                                                                                                                                       &  \phantom{abc} &  \textbf{Search space}                                                                                                                      &  \phantom{abc} &  \textbf{Best assignment}                                                                                                                   \\ \hline
 \textbf{Data}                                                                                                                                                 &  \phantom{abc} &  \phantom{abc}                                                                                                                              &  \phantom{abc} &  \phantom{abc}                                                                                                                              \\ \hline
Validation                                                                                                                                                                     &                                & True                                                                                                                                                        &                                & True                                                                                                                                                        \\
Random validation                                                                                                                                                              &                                & True                                                                                                                                                        &                                & True                                                                                                                                                        \\
\% of train as validation                                                                                                                                                      &                                & 10\%                                                                                                                                                        &                                & 10\%                                                                                                                                                        \\
Patches                                                                                                                                                                        &                                &\begin{tabular}[c]{@{}r@{}} Maximum number that fitted \\ into the slice dimensions  \end{tabular}                    &                                &\begin{tabular}[c]{@{}r@{}} Maximum number that fitted \\ into the slice dimensions  \end{tabular}                     \\
Patch size                                                                                                                                                                     &                                & $256\times256$                                                                                                                                              &                                & $256\times256$                                                                                                                                              \\
  \begin{tabular}[c]{@{}r@{}}Discard patches with less than  \\ a \% of the foreground class \end{tabular} &                                &  \textit{choice}{[}True(2\%),True(5\%), True(10\%), False{]}                                                                      &                                & True(2\%)                                                                                                                                                   \\
Shuffle train on each epoch                                                                                                                                                    &                                & True                                                                                                                                                        &                                & True                                                                                                                                                        \\
Data augmentation                                                                                                                                                              &                                & flips, rotation\_range({[}-180,180,90{]})                                                                                                                   &                                & flips, rotation\_range({[}-180,180,90{]})                                                                                                                   \\
Number of epochs                                                                                                                                                               &                                & \textit{choice}{[}10,200,10{]}                                                                                                                                             &                                & 60                                                                                                                                                          \\
Batch size                                                                                                                                                                     &                                & \textit{choice}{[}1, 2, 5{]}                                                                                                                &                                & 1                                                                                                                                                           \\
Loss type                                                                                                                                                                      &                                & BCE                                                                                                                                                         &                                & BCE                                                                                                                                                         \\
Optimizer                                                                                                                                                                      &                                & \textit{choice}{[}SGD, Adam{]}                                                                                                             &                                & Adam                                                                                                                                                        \\
Learning rate                                                                                                                                                                  &                                & \textit{choice}{[}1e-3, 2e-3,1e-4{]}                                                                                                       &                                & 1e-4                                                                                                                                                        \\
Scheduler                                                                                                                                                                      &                                & \textit{choice}{[}Reduce on Plateau, OneCycle{]}                                                                                           &                                & OneCycle                                                                                                                                                    \\
Patience                                                                                                                                                                       &                                & \textit{choice}{[}None,3, 5, 7, 10, 50{]}                                                                                                       &                                & None                                                                                                                                                          \\ \hline
 \textbf{Architecture}                                                                                                                                         &  \phantom{abc} &  \phantom{abc}                                                                                                                              &  \phantom{abc} &                                                                                                                                                             \\ \hline
Number of feature maps to start with                                                                                                                                           &                                & 32                                                                                                                                                          &                                & 32                                                                                                                                                          \\
Dropout type                                                                                                                                                                   &                                &   \begin{tabular}[c]{@{}r@{}}Spatial dropout  (0.1, 0.1, 0.2, 0.2, 0.3) \end{tabular} &                                &   \begin{tabular}[c]{@{}r@{}}Spatial dropout (0.1, 0.1, 0.2, 0.2, 0.3) \end{tabular} \\
Pooling type                                                                                                                                                                   &                                & Max-pooling                                                                                                                                                 &                                & Max-pooling                                                                                                                                                 \\
Kernel initializer                                                                                                                                                             &                                & he\_normal                                                                                                                                                   &                                & he\_normal                                                                                                                                                    \\
Activation                                                                                                                                                                     &                                & ELU                                                                                                                                                         &                                & ELU                                                                                                                                                         \\ \hline
\end{tabular}
\end{adjustbox}
\vspace{0.2cm}
\caption{ Hyperparameter search space for the proposed self-supervised learning training step
with the Attention U-Net.}

\label{tab:SSLHyper_train}
\end{table}

\clearpage
\newpage

\subsection{Attention Y-Net}
\label{subsec:apendix:hyperparam_search_attYnet}


\begin{table}[]
\hspace{-1.4cm}
\begin{adjustbox}{max width=1.22\textwidth}
\begin{tabular}{rcrcr}
\hline
\textbf{Hyperparameter}                                                                                & \phantom{abc}  & \textbf{Search space}                  & \phantom{abc} & \textbf{Best assignment}         \\ \hline
\textbf{Data}                                                                                          & \phantom{abc}  & \phantom{abc}                          & \phantom{abc} & \phantom{abc}                    \\ \hline
Validation                                                                                             &                & True                                   &               & True                             \\
Random validation                                                                                      &                & True                                   &               & True                             \\
\% of train as validation                                                                              &                & 10\%                                   &               & 10\%                             \\
Patches                                                                                                &                & \textit{choice}[Sequential, Random(1000)]           &               & Random(1000)                     \\
Patch size                                                                                             &                & $256\times256$                         &               & $256\times256$                   \\
\begin{tabular}[c]{@{}r@{}}Discard patches with less than \\ a \% of the foreground class\end{tabular} &                & \textit{choice}[True(5\%), True(10\%), False]       &               & False                            \\
\begin{tabular}[c]{@{}r@{}}Discard patches with more than \\ a \% of zeros in the image\end{tabular}   &                & \textit{choice}[True(50\%), True(80\%), False]      &               & True(50\%)                       \\
Shuffle train on each epoch                                                                            &                & True                                   &               & True                             \\
Data augmentation                                                                                      &                & flips, rotation\_range(180)  &               & flips, rotation\_range(180)\\
Number of epochs                                                                                       &                & \textit{choice}[20, 30, 50, 360]                       &               & 50                               \\
Batch size                                                                                             &                & \textit{choice}[1, 2]                                &               & 1                                \\
Loss type                                                                                              &                &  $\alpha MSE + (1-\alpha)BCE$            &               & $\alpha MSE + (1-\alpha)BCE$       \\
loss weight $\alpha$                                                                                   &                & \textit{choice}[0.9, 0.98]                          &               & 0.98                             \\
Optimizer                                                                                              &                & \textit{choice}[SGD, Adam]                           &               & SGD                              \\
Learning rate                                                                                          &                & \textit{choice}[1e-3, 2e-3]                         &               & 1e-3                             \\
Scheduler                                                                                              &                & \textit{choice}[Reduce on Plateau, OneCycle]        &               & Reduce on Plateau                \\
Patience                                                                                               &                & \textit{choice}[3, 5, 7, 10, 50]                        &               & 7                                \\ \hline
\textbf{Architecture}                                                                                  & \phantom{abc}  & \phantom{abc}                          & \phantom{abc} &                                  \\ \hline
Number of feature maps to start with                                                                   &                & \textit{choice}[32, 64]                              &               & 32                               \\
Dropout type                                                                                           &                & \begin{tabular}[c]{@{}r@{}}Spatial dropout\\ (0.1, 0.1, 0.2, 0.2, 0.3)\end{tabular}&               & \begin{tabular}[c]{@{}r@{}}Spatial dropout\\ (0.1, 0.1, 0.2, 0.2, 0.3)\end{tabular} \\
Pooling type                                                                                           &                & Max-pooling                            &               & Max-pooling                      \\
Kernel initializer                                                                                     &                & he\_normal                              &               & he\_normal                        \\
Activation                                                                                             &                & ELU                                    &               & ELU                              \\ \hline
\end{tabular}
\end{adjustbox}
\vspace{0.2cm}
\caption{Hyperparameter search space for the proposed Attention Y-Net, first multitask step.}
\end{table}


\begin{table}[]
\hspace{-1.4cm}
\begin{adjustbox}{max width=1.22\textwidth}
\begin{tabular}{rcrcr}
\hline
\textbf{Hyperparameter}                                                                                & \phantom{abc}  & \textbf{Search space}                  & \phantom{abc} & \textbf{Best assignment}         \\ \hline
\textbf{Data}                                                                                          & \phantom{abc}  & \phantom{abc}                          & \phantom{abc} & \phantom{abc}                    \\ \hline
Validation                                                                                             &                & True                                   &               & True                             \\
Random validation                                                                                      &                & True                                   &               & True                             \\
\% of train as validation                                                                              &                & 10\%                                   &               & 10\%                             \\
Patches                                                                                                &                & \textit{choice}[Sequential, Random(1000)]           &               & Random(1000)                     \\
Patch size                                                                                             &                & $256\times256$                         &               & $256\times256$                   \\
\begin{tabular}[c]{@{}r@{}}Discard patches with less than \\ a \% of the foreground class\end{tabular} &                & \textit{choice}[True(5\%), True(10\%), False]       &               & False                            \\
\begin{tabular}[c]{@{}r@{}}Discard patches with more than \\ a \% of zeros in the image\end{tabular}   &                & \textit{choice}[True(50\%), True(80\%), False]      &               & True(50\%)                       \\
Shuffle train on each epoch                                                                            &                & True                                   &               & True                             \\
Data augmentation                                                                                      &                & \begin{tabular}[c]{@{}r@{}}flips, rotation\_range(180),\\histogram\_matching(0\%, 25\%, 50\%, 100\%), \\ CLAHE(50\%) \end{tabular}  &               & \begin{tabular}[c]{@{}r@{}}flips, rotation\_range(180),\\ histogram\_matching(50\%) \end{tabular}\\
Number of epochs                                                                                       &                & \textit{choice}[15, 20, 30, 40]                        &               & 40                               \\
Batch size                                                                                             &                & 1                                      &               & 1                                \\
Loss type                                                                                              &                & MSE                                    &               & MSE                              \\
Optimizer                                                                                              &                & \textit{choice}[SGD, Adam]                           &               & Adam                              \\
Learning rate                                                                                          &                & \textit{choice}[2e-3, 2e-4]                         &               & 2e-4                             \\
Scheduler                                                                                              &                & \textit{choice}[Reduce on Plateau, OneCycle]        &               & Reduce on Plateau                \\
Patience                                                                                               &                & \textit{choice}[5, 6, 7, 10, 20]                        &               & 6                                \\ \hline
\textbf{Architecture}                                                                                  & \phantom{abc}  & \phantom{abc}                          & \phantom{abc} &                                  \\ \hline
Number of feature maps to start with                                                                   &                & 32                                     &               & 32                               \\
Dropout type                                                                                           &                & \begin{tabular}[c]{@{}r@{}}Spatial dropout\\ (0.1, 0.1, 0.2, 0.2, 0.3)\end{tabular}&               & \begin{tabular}[c]{@{}r@{}}Spatial dropout\\ (0.1, 0.1, 0.2, 0.2, 0.3)\end{tabular} \\
Pooling type                                                                                           &                & Max-pooling                            &               & Max-pooling                      \\
Kernel initializer                                                                                     &                & he\_normal                              &               & he\_normal                        \\
Activation                                                                                             &                & ELU                                    &               & ELU                              \\ \hline
\end{tabular}
\end{adjustbox}
\vspace{0.2cm}
\caption{Hyperparameter search space for the proposed Attention Y-Net, second step focused in reconstruction.}
\end{table}


\begin{table}[]
\hspace{-1.4cm}
\begin{adjustbox}{max width=1.22\textwidth}
\begin{tabular}{rcrcr}
\hline
\textbf{Hyperparameter}                                                                                & \phantom{abc}  & \textbf{Search space}                  & \phantom{abc} & \textbf{Best assignment}         \\ \hline
\textbf{Data}                                                                                          & \phantom{abc}  & \phantom{abc}                          & \phantom{abc} & \phantom{abc}                    \\ \hline
Validation                                                                                             &                & True                                   &               & True                             \\
Random validation                                                                                      &                & True                                   &               & True                             \\
\% of train as validation                                                                              &                & 10\%                                   &               & 10\%                             \\
Patches                                                                                                &                & \begin{tabular}[c]{@{}r@{}}\textit{choice}[Sequential, Random(160,\\ 500, 1000, 2000)]\end{tabular}           &               & Random(1000)                     \\
Patch size                                                                                             &                & $256\times256$                         &               & $256\times256$                   \\
\begin{tabular}[c]{@{}r@{}}Discard patches with less than \\ a \% of the foreground class\end{tabular} &                & \textit{choice}[True(5\%), True(10\%), False]       &               & False                            \\
\begin{tabular}[c]{@{}r@{}}Discard patches with more than \\ a \% of zeros in the image\end{tabular}   &                & \textit{choice}[True(50\%), True(80\%), False]      &               & True(50\%)                       \\
Shuffle train on each epoch                                                                            &                & True                                   &               & True                             \\
Data augmentation                                                                                      &                & \begin{tabular}[c]{@{}r@{}}flips, rotation\_range(180),\\histogram\_matching(0\%, 25\%, 50\%, 100\%), \\ CLAHE(50\%) \end{tabular}  &               & \begin{tabular}[c]{@{}r@{}}flips, rotation\_range(180),\\ CLAHE(50\%) \end{tabular}\\
Number of epochs                                                                                       &                & \textit{choice}[5, 10, 20, 100]                        &               & 100                               \\
Batch size                                                                                             &                & 1                                      &               & 1                                \\
Loss type                                                                                              &                & BCE                                    &               & BCE                              \\
Optimizer                                                                                              &                & \textit{choice}[SGD, Adam]                           &               & Adam                              \\
Learning rate                                                                                          &                & \textit{choice}[2e-3, 2e-4]                         &               & 2e-4                             \\
Scheduler                                                                                              &                & \textit{choice}[Reduce on Plateau, OneCycle, None]        &               & oneCycle                \\
Patience                                                                                               &                & \textit{choice}[10, 15, 20, 150]                       &               & 15                                \\ \hline
\textbf{Architecture}                                                                                  & \phantom{abc}  & \phantom{abc}                          & \phantom{abc} &                                  \\ \hline
Number of feature maps to start with                                                                   &                & 32                                     &               & 32                               \\
Dropout type                                                                                           &                & \begin{tabular}[c]{@{}r@{}}Spatial dropout\\ \{0.1, 0.1, 0.2, 0.2, 0.3\}\end{tabular}&               & \begin{tabular}[c]{@{}r@{}}Spatial dropout\\ \{0.1, 0.1, 0.2, 0.2, 0.3\}\end{tabular} \\
Pooling type                                                                                           &                & Max-pooling                            &               & Max-pooling                      \\
Kernel initializer                                                                                     &                & he\_normal                              &               & he\_normal                        \\
Activation                                                                                             &                & ELU                                    &               & ELU                              \\ \hline
\end{tabular}
\end{adjustbox}
\vspace{0.2cm}
\caption{Hyperparameter search space for the proposed Attention Y-Net, third and last training step, focused in segmentation.}
\end{table}

\clearpage
\newpage
\subsection{DAMT-Net}
To execute DAMT-Net, we follow the publicly available implementation provided by its authors~\cite{peng2020unsupervised}. Since they use two images in each training step, we interpret the batch size as $2$. Bearing this in mind, we define an epoch as follows: 
\begin{equation}
    Epoch = \frac{|X_{train}|}{batch\_size}
\end{equation}
where $|X_{train}|$ is the cardinality of the training set. Taking this into account, we explored a few hyperparameters:
\begin{itemize}
    \item Patch size: \textbf{512}$\times$\textbf{512} pixels and $256\times256$ pixels
    \item Epochs: 30, \textbf{60}, 100
    \item Save checkpoint every \textbf{2} epochs.
\end{itemize}
Among the different options, the best assignment has been highlighted in bold. The rest of the parameters are those proposed by default in the original publication~\cite{peng2020unsupervised}.

\newpage
\bibliographystyle{IEEEtran}
\bibliography{supplement}